\documentclass[letterpaper]{article} 
\usepackage{aaai24}  
\usepackage{times}  
\usepackage{helvet}  
\usepackage{courier}  
\usepackage[hyphens]{url}  
\usepackage{graphicx} 
\usepackage{natbib}  
\usepackage{caption} 
\usepackage{algorithm}
\usepackage{algorithmic}

\usepackage{multirow}
\usepackage{subfig}
\usepackage{float}
\usepackage{multicol}

\usepackage{amsmath}
\usepackage{amsthm}
\usepackage{booktabs}
\usepackage{algorithm}
\usepackage{algorithmic}
\usepackage[switch]{lineno}
\usepackage{multirow}
\usepackage{graphicx}
\usepackage{subfig}
\usepackage{xcolor}

\usepackage{amsfonts}       
\usepackage{nicefrac}       
\usepackage{microtype}      
\usepackage{xcolor}         
\usepackage{graphicx}
\usepackage{tcolorbox}
\usepackage{multirow}    
\usepackage{amsmath,amssymb,amsfonts}
\usepackage{subfig}
\usepackage{pifont}

\usepackage{overpic}
\usepackage{array} 
\usepackage{tabularx}
\usepackage{ltablex}
\newcolumntype{b}{X}
\newcolumntype{s}{>{\hsize=2.2\hsize}X}

\usepackage{newfloat}
\usepackage{listings}
\DeclareCaptionStyle{ruled}{labelfont=normalfont,labelsep=colon,strut=off} 
\floatstyle{ruled}
\newfloat{listing}{tb}{lst}{}
\floatname{listing}{Listing}
\title{Sparks of Artificial General Intelligence(AGI) in Semiconductor Material Science: Early Explorations into the Next Frontier of Generative AI-Assisted Electron Micrograph Analysis}
\author{
    Sakhinana Sagar Srinivas\textsuperscript{\rm 1}\thanks{Designed, programmed the software, and drafted manuscript.},
    \{Geethan Sannidhi\textsuperscript{\rm 2}, Sreeja Gangasani\textsuperscript{\rm 3}, Chidaksh Ravuru\textsuperscript{\rm 4}\}\thanks{Conducted experiments and analyzed visual results}, 
    Venkataramana Runkana\textsuperscript{\rm 1}\\
}
\affiliations{
    \textsuperscript{\rm 1}TCS Research,
    \textsuperscript{\rm 2}IIIT Pune,
    \textsuperscript{\rm 3}IIT Palakkad,
    \textsuperscript{\rm 4}IIT Dharwad \\

    \texttt{sagar.sakhinana@tcs.com}, \texttt{geethan.iiitp.ac.in}, \texttt{111901023@smail.iitpkd.ac.in}, \texttt{200010046@iitdh.ac.in}, \texttt{venkat.runkana@tcs.com}\\
}

\usepackage{bibentry}

\begin{document}

\maketitle
\vspace{-3mm}
\begin{abstract}
\vspace{-2mm}
Characterizing materials with electron micrographs poses significant challenges for automated labeling due to the complex nature of nanomaterial structures. To address this, we introduce a fully automated, end-to-end pipeline that leverages recent advances in Generative AI. It is designed for analyzing and understanding the microstructures of semiconductor materials with effectiveness comparable to that of human experts, contributing to the pursuit of Artificial General Intelligence (AGI) in nanomaterial identification. Our approach utilizes Large MultiModal Models (LMMs) such as GPT-4V, alongside text-to-image models like DALL·E-3. We integrate a GPT-4 guided Visual Question Answering (VQA) method to analyze nanomaterial images, generate synthetic nanomaterial images via DALL·E-3, and employ in-context learning with few-shot prompting in GPT-4V for accurate nanomaterial identification. Our method surpasses traditional techniques by enhancing the precision of nanomaterial identification and optimizing the process for high-throughput screening.
\vspace{-2mm}
\end{abstract}

\vspace{-5mm}
\section{Introduction}
\begin{figure}[ht!]
     \centering
     \subfloat[\textit{MEMS} devices exhibit high intra-class dissimilarity, indicating that they can appear distinct even if they are from the same category.]{
     \includegraphics[height=0.085\textwidth]{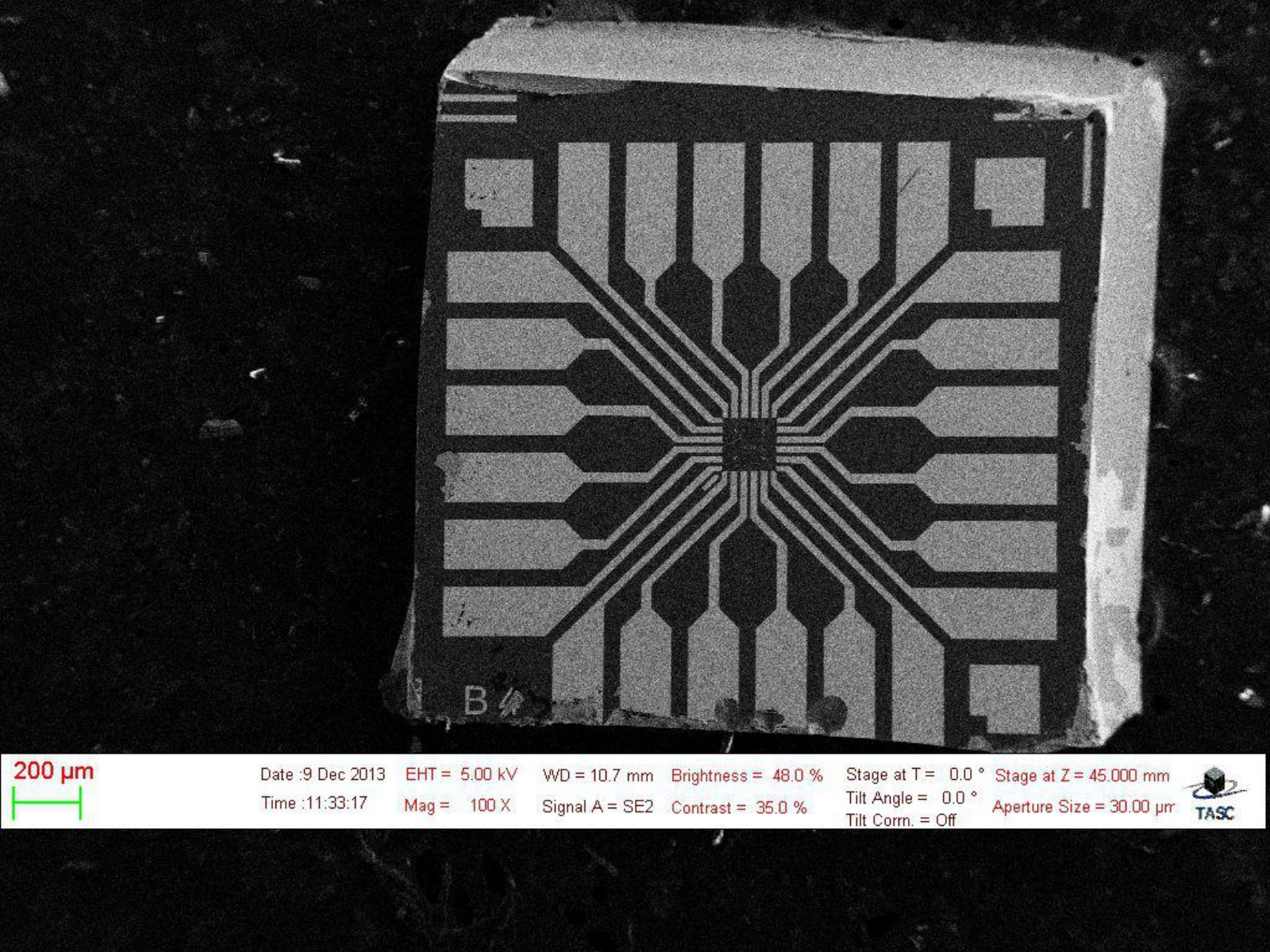}
     \includegraphics[height=0.085\textwidth]{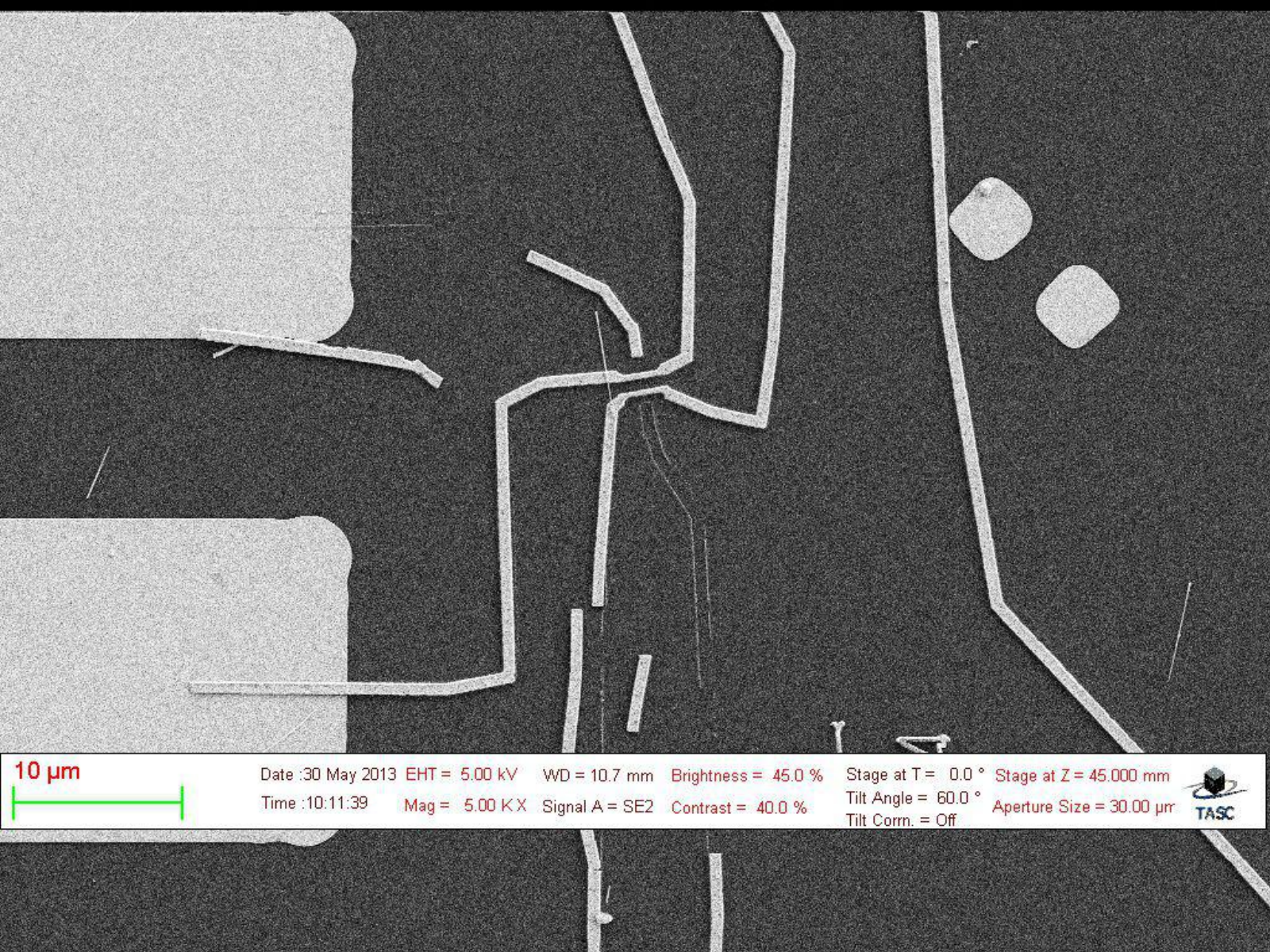}
     \includegraphics[height=0.085\textwidth]{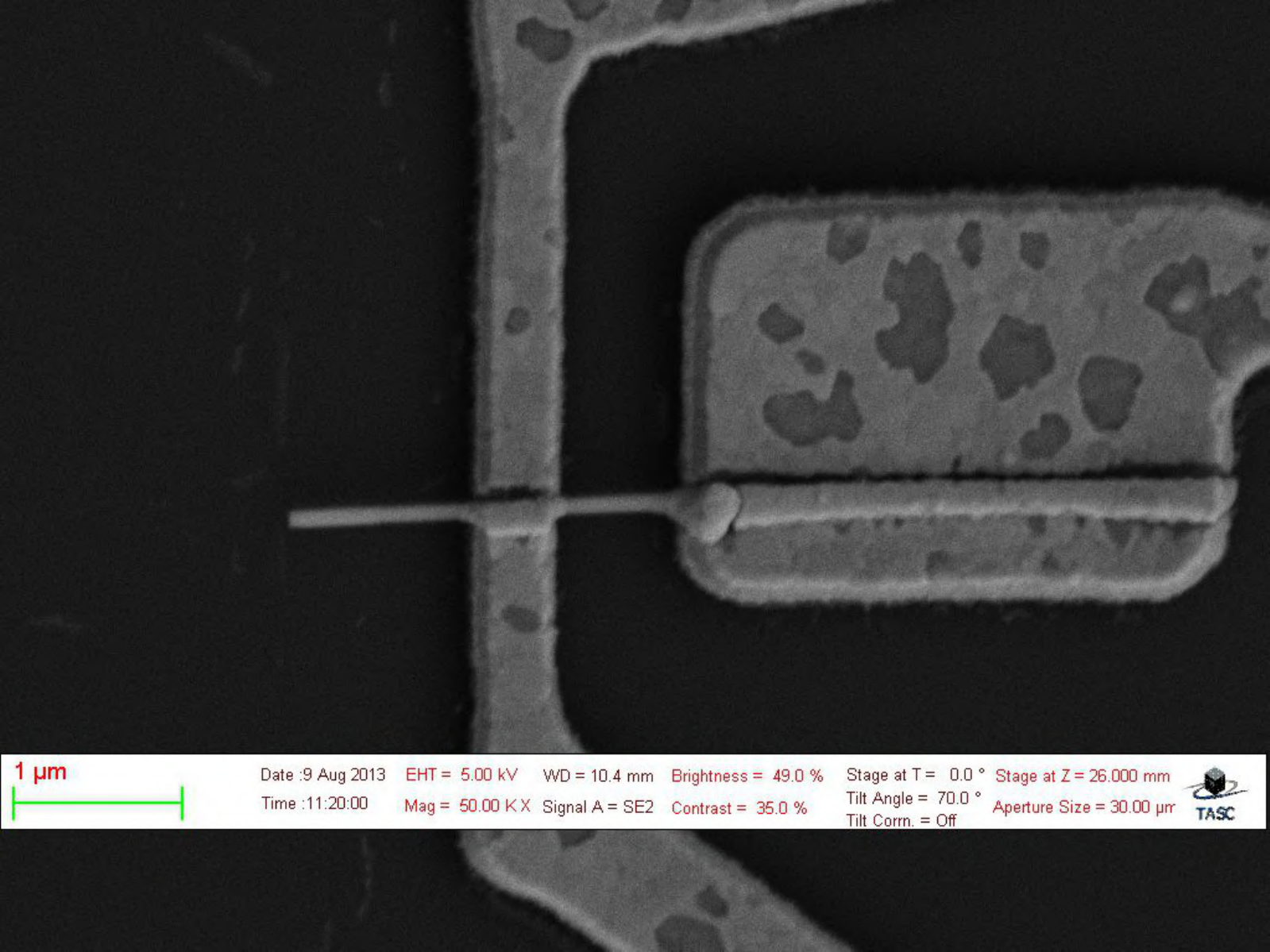}
     \label{fig:subfig1}
     }
     \vspace{-0.5mm}
     \qquad
     \subfloat[Nanomaterials from different categories (\textit{listed from left to right as films and coated surfaces, porous sponges and powders}), have a high degree of inter-class similarity as they look similar or even identical.]{
     \includegraphics[height=0.085\textwidth]{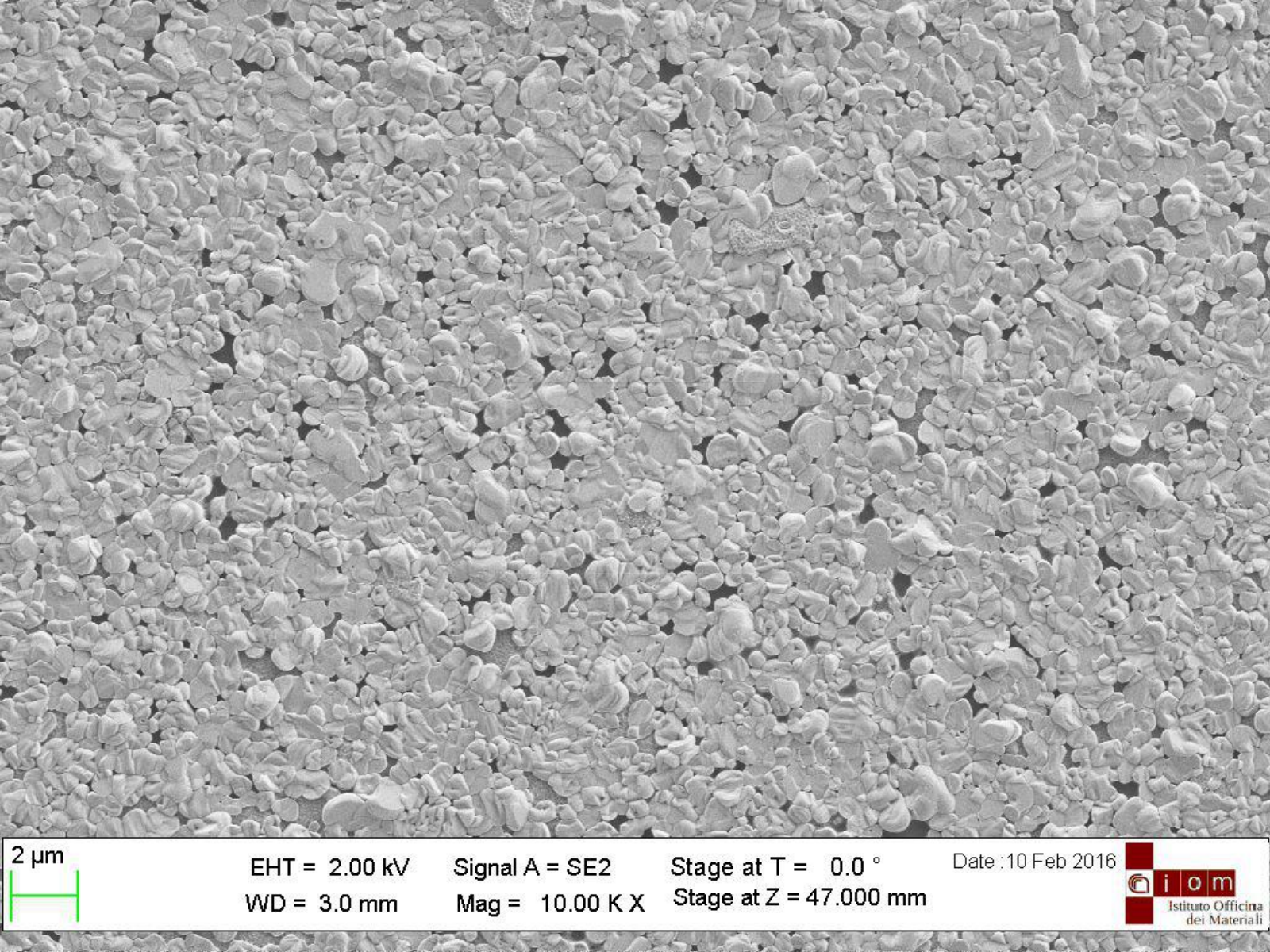}
     \includegraphics[height=0.085\textwidth]{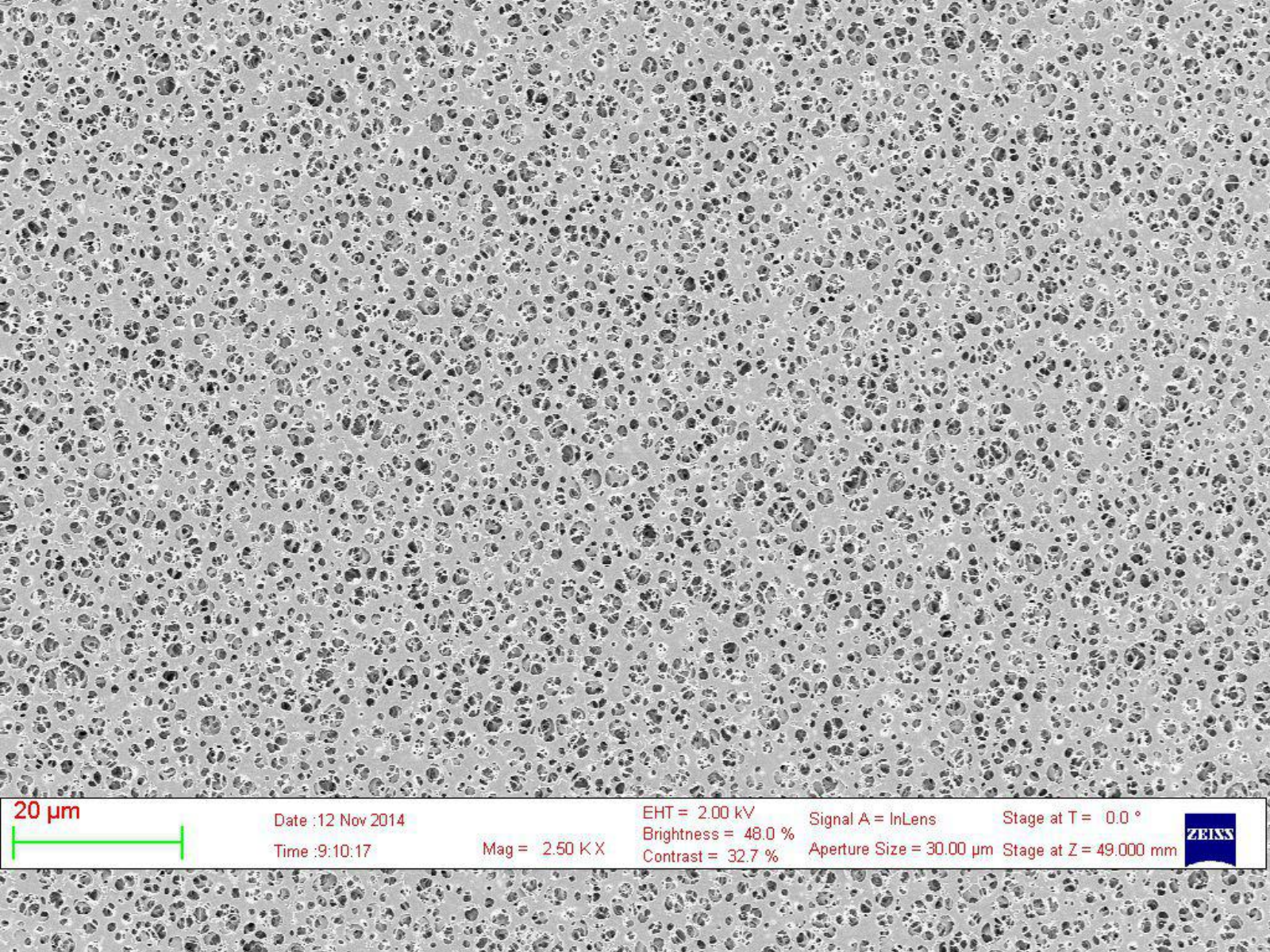}
     \includegraphics[height=0.085\textwidth]{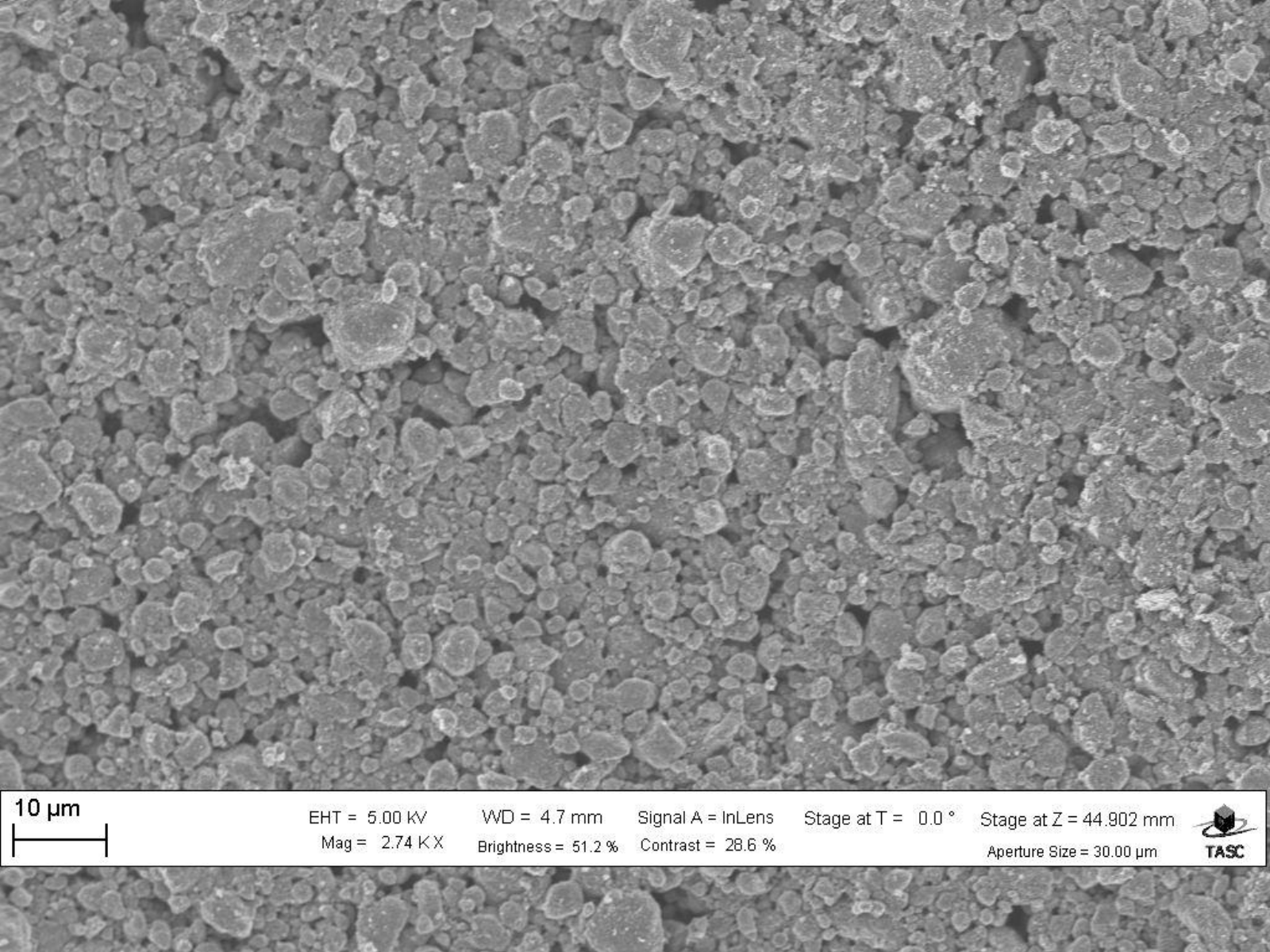}
     \label{fig:subfig2}
     }
     \vspace{-0.5mm}
     \qquad
     \subfloat[The spatial heterogeneity of \textit{nanoparticles} can be observed with different patterns appearing at different magnifications.]{
     \includegraphics[height=0.085\textwidth]{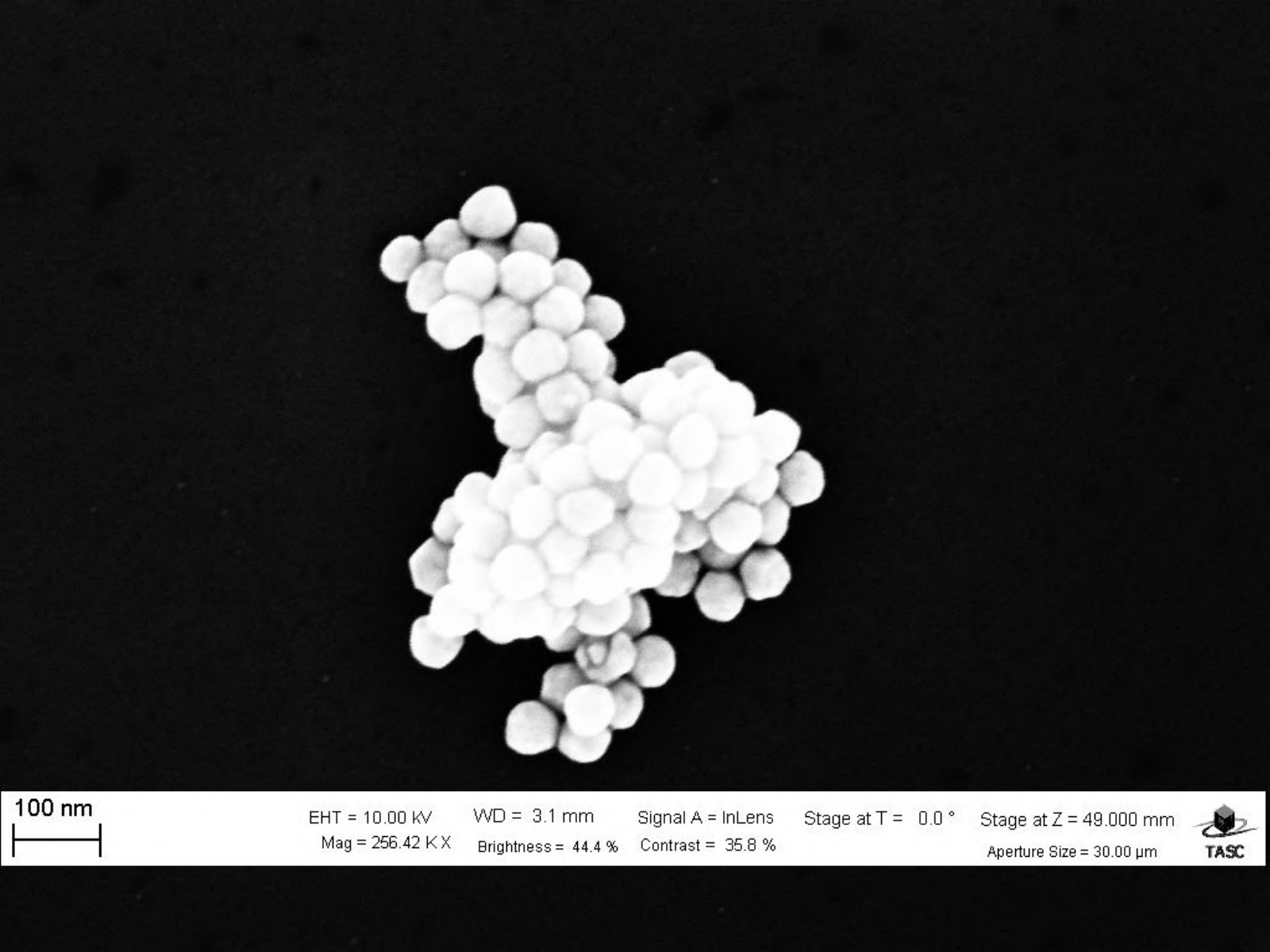}
     \includegraphics[height=0.085\textwidth]{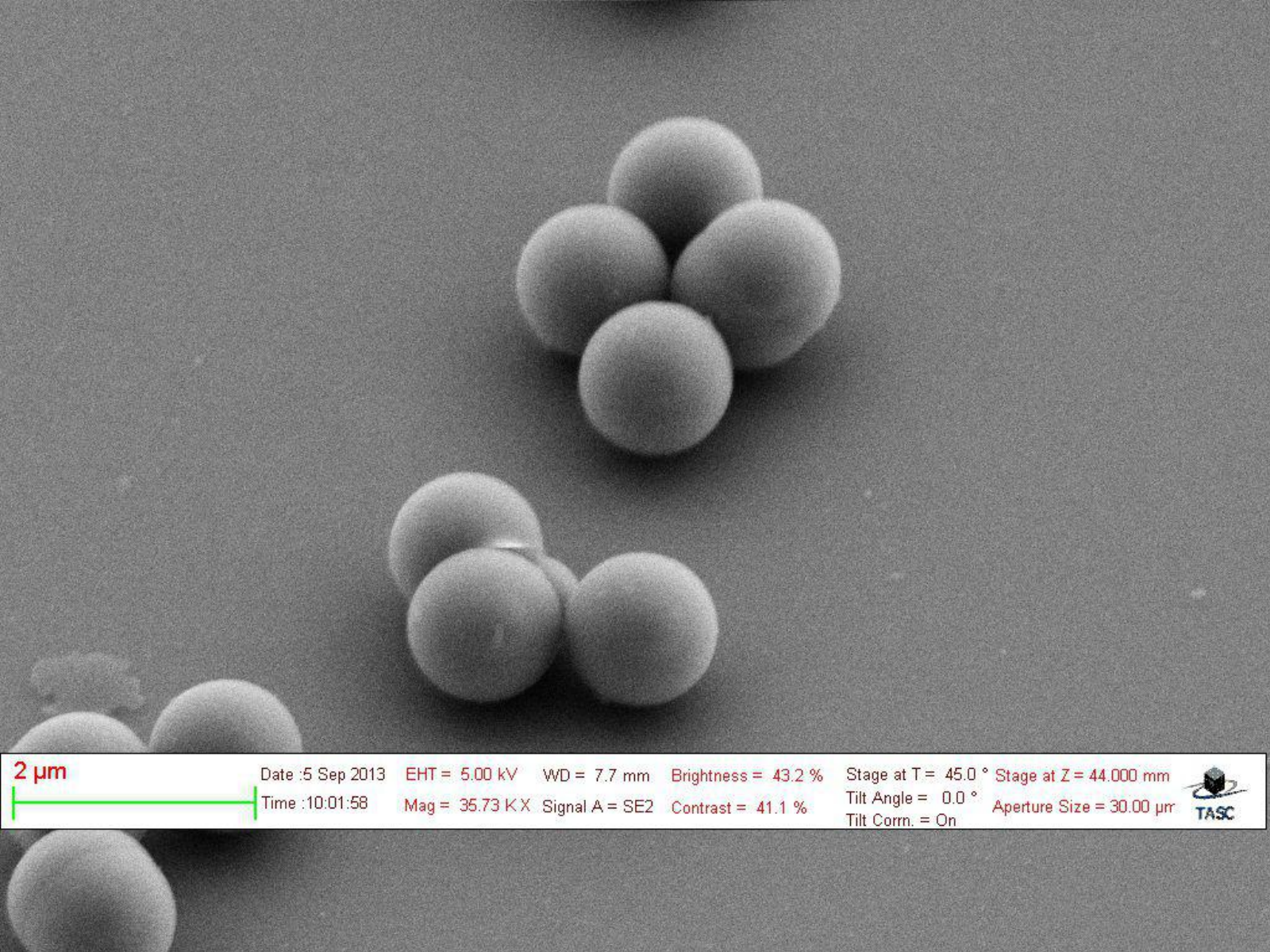}
     \includegraphics[height=0.085\textwidth]{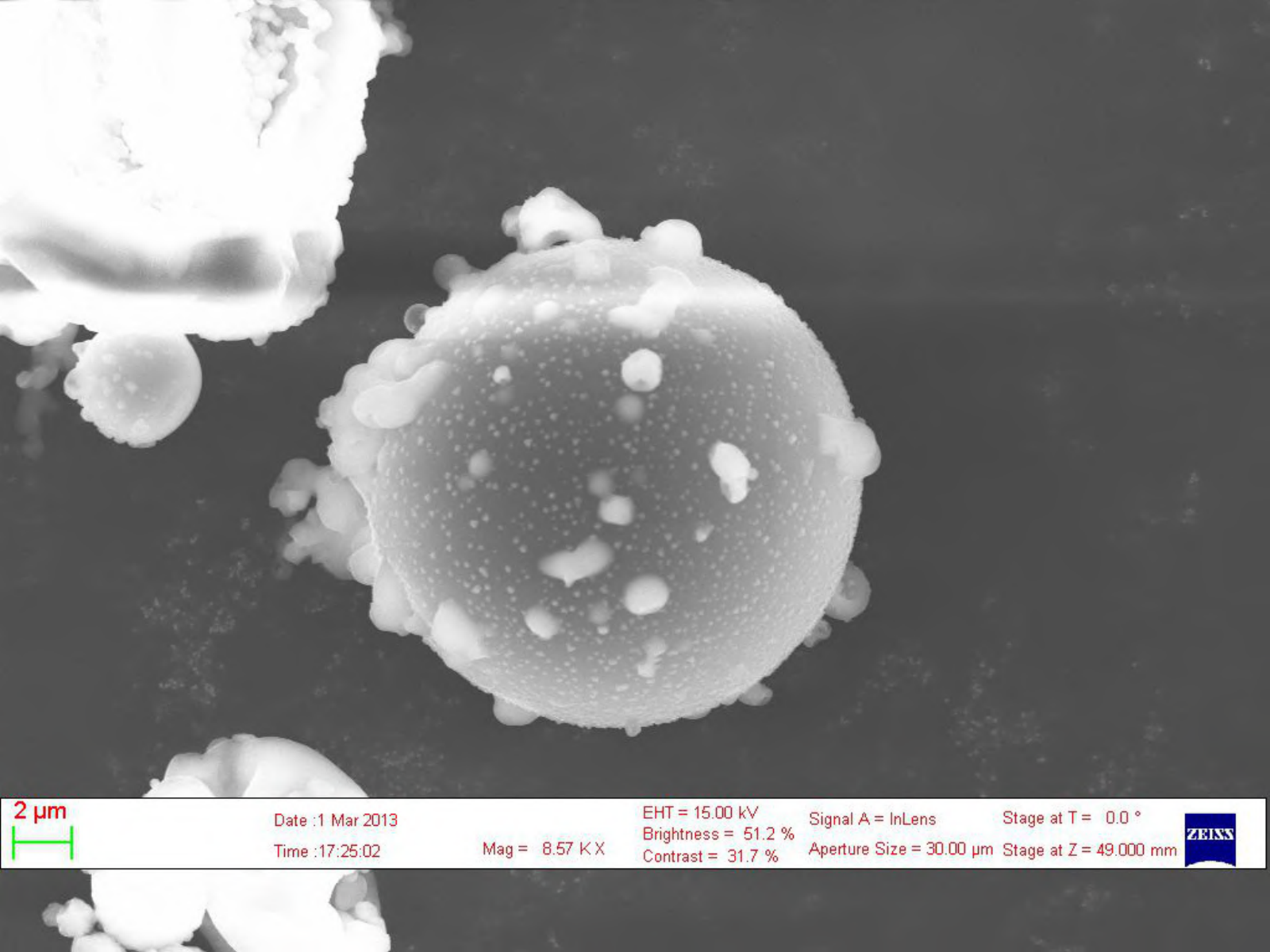}
     \label{fig:subfig3}
     }
     \vspace{-3mm}
     \caption{The figure highlights the complexity of classifying micrographs in the SEM dataset(\cite{aversa2018first}).}
     \vspace{-4mm}
     \label{fig:figure1}
\end{figure}

\vspace{-1mm}
The multifaceted journey of semiconductor production involves several stakeholders. Fabless firms such as NVIDIA, Qualcomm, and AMD focus on designing and developing semiconductor chips, yet they do not own fabrication facilities. Instead, they utilize Electronic Design Automation (EDA) tools for designing, simulating circuits, and verifying semiconductor devices. Following this phase, specialized foundries like Taiwan Semiconductor Manufacturing Company (TSMC) and Samsung Electronics fabricate the designs provided by the fabless companies onto silicon wafers. These foundries employ advanced sub-14 nm technology to etch precise geometries essential for modern high-performance chips. After fabrication, companies like Advantest and Teradyne employ specialized semiconductor test equipment to subject the chips to a rigorous evaluation phase, ensuring they meet performance and reliability standards. Post-testing, packaging, and assembly companies such as ASE Technology Holding and Amkor Technology prepare the semiconductor devices for integration into larger electronic systems. In contrast to these individual stakeholders, Integrated Device Manufacturers (IDMs) like Intel and Texas Instruments oversee nearly all aspects of the semiconductor production process, from design to packaging. As the semiconductor industry continues to strive toward miniaturization, aiming for more powerful and energy-efficient chips, it faces challenges such as manufacturing errors and quantum tunneling. Addressing these challenges requires advanced imaging and analysis, as well as innovative engineering approaches, all of which are crucial for maintaining the rapid evolution of semiconductor technology in today's digital age. One of the key advancements in the industry, particularly in sub-7 nm technology, hinges on achieving micro and nanoscale precision. Tools like Scanning Electron Microscopy (SEM) and Transmission Electron Microscopy (TEM) are at the forefront of this effort. These electron beam tools provide detailed micrographs, or nano images, of semiconductor materials and structures. The advanced imaging techniques play a vital role in manufacturing analysis, enabling clear visualization and analysis of microstructures. This makes these tools indispensable for quality control, process monitoring, and failure analysis to ensure that semiconductors adhere to design parameters and identify defects. Materials characterization at the micro and nanoscale is imperative for continued technological advancement. However, automated labeling of electron micrographs faces challenges due to the high similarity between different nanomaterial categories (high inter-similarity), wide appearance variance within a single category (high intra-dissimilarity), and spatial heterogeneity of patterns of nanomaterials across different length scales in electron micrographs. The manifold complexities of automated nanomaterial identification tasks are illustrated in Figure \ref{fig:figure1}. Advancements in machine learning and image recognition technologies are crucial for the accurate labeling and analysis of electron micrographs, thereby improving quality control and performance in the semiconductor industry and aiding its further progression. In the realm of AI, Large Language Models (LLMs) such as GPT-4 (language-only), which empower conversational agents like ChatGPT to generate human-like dialogue as responses to user inputs, have recently gained prominence and showcased unprecedented capabilities in human-AI interaction. These large-scale models leverage an autoregressive, decoder-only architecture and undergo pre-training in a self-supervised learning paradigm on vast amounts of unlabeled text corpora. At their core, they operate by predicting the next token in a sequence based on the context provided by preceding tokens—a foundational principle of language modeling. Additionally, to refine their outputs and better align with human preferences, they are fine-tuned using reinforcement learning from human feedback (RLHF). The foundational LLMs have revolutionized natural language processing(NLP) with their advanced text comprehension and sophisticated logical reasoning, leading to remarkable performance across various NLP tasks. A key feature of these large-scale models is their ``prompt and predict'' paradigm, which allows users to instruct LLMs using natural language prompts to set the context and task-specific instructions to generate the text-based response. The term ``prompting'' refers to the method of conditioning the language model to respond to the instructions based solely on the patterns and knowledge acquired during the training phase. General-purpose language models, like GPT-4, can be steered to generate desired outputs using various prompt engineering strategies. One of these strategies is zero-shot learning, where the language model generates an output based solely on its pre-trained knowledge, without any task-specific demonstrations (input-output mappings). In contrast, few-shot learning provides the language model with a limited number of demonstrations to guide its output. In essence, prompts that include both explicit conditioning based on task-specific instructions and a few demonstrations are termed few-shot prompts, while those that rely solely on task-specific instructions are referred to as zero-shot prompts. Chain-of-thought (CoT) and tree-of-thought (ToT) prompting techniques assist LLMs in explaining their reasoning step-by-step and in exploring multiple possible thought paths simultaneously, thus enhancing performance on tasks involving reasoning, logic, and more. The choice between these strategies typically depends on the context and specific objectives of the request, with each designed to optimize the language model’s performance. Proprietary LLMs, such as GPT-4\cite{gpt4}, demonstrate advanced language comprehension. However, their `black-box' nature can pose challenges to interpretability and explainability, especially given the lack of direct access to internal state representations like logits or token embeddings. Furthermore, while general-purpose LLMs are designed to handle a broad range of tasks, adapting them for niche tasks can be highly resource-intensive due to their high model complexity and size, and their performance might not always be optimized for specialized applications. In contrast, open-source small-scale models like BERT\cite{devlin2018bert}, following a ``pre-training and fine-tuning" approach, can be more cost-efficient for task-specific customization. These smaller language models also provide better interpretability because they allow access to internal state representations like logits or token embeddings, thanks to their open nature. However, they might not match the reasoning and generalization capabilities of proprietary LLMs, sometimes producing less coherent and contextually apt outputs. In recent times, the trend has shifted towards exploring and expanding the capabilities of foundational language-only LLMs in multimodal settings to enhance their performance and applicability across a wider range of tasks. GPT-4 with Vision (GPT-4V\cite{gpt4v}) represents a significant advancement over the earlier, text-focused OpenAI GPT-4, which was limited to language processing. Large multimodal models (LMMs) such as GPT-4V are instruction-following, language-based human-AI interaction systems capable of analyzing image inputs by interpreting and responding to text prompts, which enables them to generate text-only outputs conditioned on the provided visual context. GPT-4V integrates visual capabilities with its existing language processing abilities, enabling it to perform tasks such as analyzing and describing images based on textual prompts, transcribing text from images, and deciphering data, among others, thereby broadening the horizons for real-world applications. Similarly, DALL·E-3\cite{dalle3, dalle3latest}, an advanced version of OpenAI's DALL-E\cite{ramesh2022hierarchical}, excels in text-to-image synthesis, designed to generate accurate images from textual prompts. A significant advancement over previous models in the realm of AI image generation, DALL·E-3 not only generates high-quality images that accurately reflect the intended textual descriptions but also has the new ability to modify (edit) and transform existing images based on textual inputs

\vspace{-3mm}

\begin{figure}[ht!]
\centering
\resizebox{1.05\linewidth}{!}{ 
\hspace*{-4mm}\includegraphics[keepaspectratio,height=4.5cm,trim=0.0cm 3.0cm 0cm 2.5cm,clip]{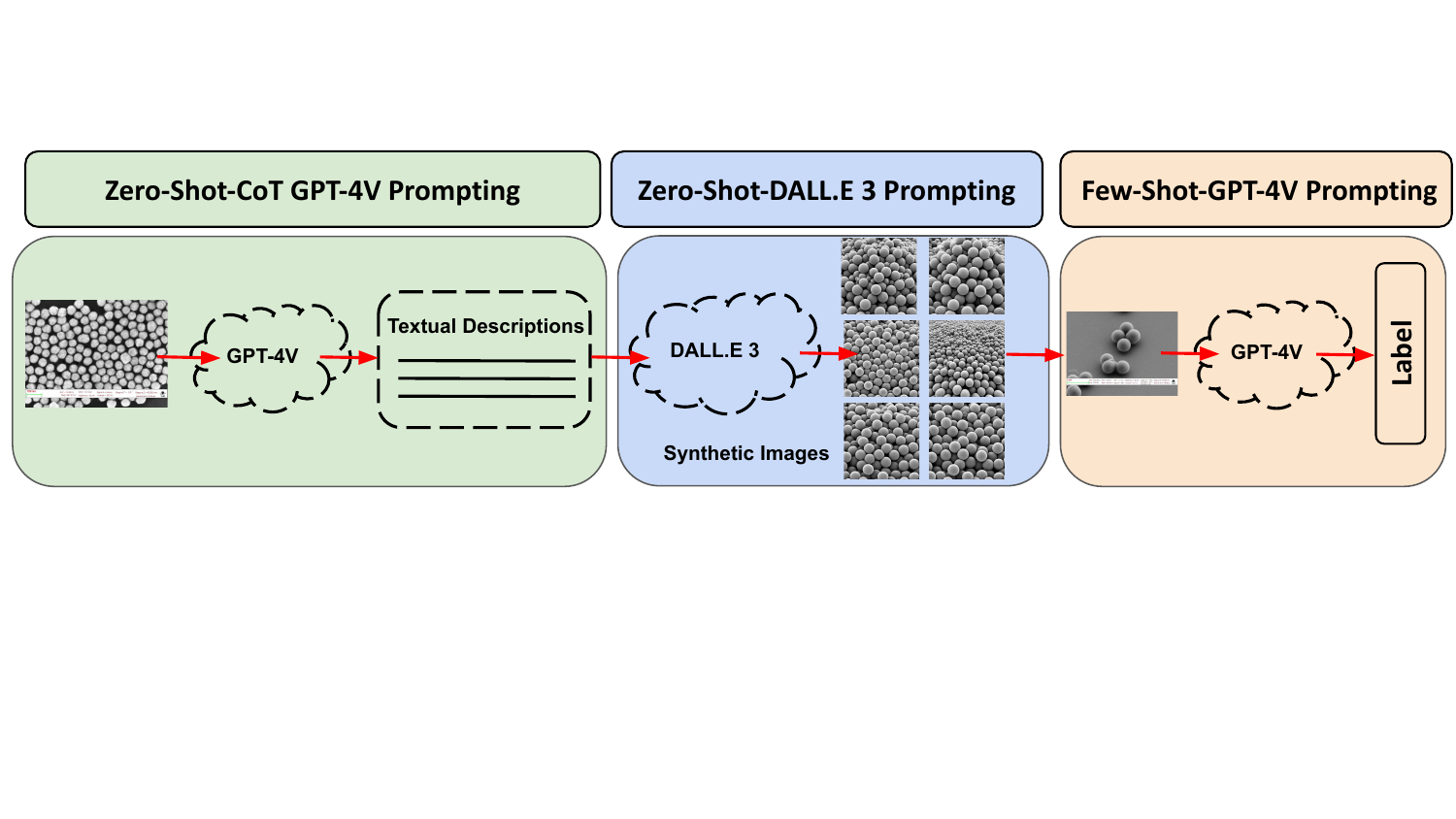} 
}
\vspace{-16mm}
\caption{Our framework comprises three sequentially operating methods: (a) \textbf{GPT-4 Guided VQA for Nanomaterial Image Analysis}: GPT-4 actively formulates questions to analyze nanomaterial images, which, in conjunction with GPT-4V, extract detailed insights from the images, yielding comprehensive textual descriptions of the nanomaterial structures and patterns. These Q$\&$A pairs subsequently guide DALL-E 3 in generating synthetic nanomaterial images. (b) \textbf{Zero-Shot Prompting with DALL-E 3}: DALL-E 3 uses zero-shot prompting to transform the Q$\&$A pairs into visual representations, demonstrating its ability to generate synthetic images of nanomaterials without prior examples. This addresses nanomaterial identification and mitigates data scarcity challenges. (c) \textbf{In-Context Learning with Few-Shot Prompting with GPT-4V}: This method employs few-shot prompting with GPT-4V for nanomaterial classification, sidestepping traditional fine-tuning through analogy-based learning. Overall, the framework operates autonomously, eliminating the need for human intervention (human-out-of-the-loop) and forgoing parameter updates (parameter-free), thus highlighting its ease of use and efficient data processing.}
\label{fig:figure2}
\vspace{-5mm}
\end{figure}

\vspace{0mm}
In the semiconductor manufacturing sector, traditional vision-based frameworks are becoming increasingly limited, especially in comparison to recent advancements in Generative Deep Learning. The lack of an integrated approach in existing architectures, which can process both visual and linguistic data simultaneously, diminishes their robustness and precision. This significant gap has the potential to hinder future innovation in the semiconductor industry. Our study introduces a novel approach to the automatic nanomaterial identification task by harnessing the strengths of GPT-4V and DALL-E-3. This endeavor represents a pioneering step towards addressing this challenge. The workflow of the proposed approach, Generative Deep Learning for Nanomaterial Identification (\texttt{GDL-NMID}), is illustrated in Figure \ref{fig:figure2}. Our novel approach to nanomaterial identification is both autonomous, reducing the need for constant human oversight, and versatile, requiring minimal manual configuration or adjustment to work effectively. The main contributions of our work can be summarized as follows:

\begin{itemize}
\item \textbf{GPT-4V-Guided Visual Question Answering(VQA) for Nanomaterial Image Analysis:} Utilizing the language model GPT-4, we generate natural language questions tailored for analyzing nanomaterial images. These open-ended questions serve as detailed instructions to unearth insights about the material's structure, properties, and potential applications. Combined with visual data such as nanomaterial images, these instructions become prompts for VQA, facilitated by the multimodal capabilities of GPT-4V. We employ the Zero-shot Chain of Thought (Zero-shot-CoT) prompting technique with LMMs like GPT-4V to delve deeper into nanomaterial images, leveraging the model's pre-trained knowledge to generate the technical descriptions conditioned solely on the multimodal prompt. Unlike traditional language-centric CoT, our multimodal CoT approach combines textual queries with visual inputs within its prompts. GPT-4V can thus produce context-rich text responses that detail the visual intricacies of nanomaterial images. The structured CoT prompts ensure comprehensive exploration of the nanomaterial's characteristics. Additionally, the textual descriptions guide the generation of synthetic nanomaterial images with DALL-E 3, translating text into precise visual representations.
\vspace{-1mm}
\item \textbf{Zero-Shot Prompting with DALL-E 3 for Synthetic Image Generation:} We utilize DALL-E 3 ability to convert textual descriptions, referred to as `prompts', into high-quality nanomaterial images without task-specific fine-tuning. This text-to-image model leverages its prior knowledge acquired during training in a manner similar to zero-shot prompting in language models. DALL-E 3 generates images based on text inputs, especially from the Q$\&$A pairs provided by GPT-4V. Our research highlights the zero-shot prompting capability of DALL-E 3, which interprets Q$\&$A pairs and visually translates them into synthetic nanomaterial images. Data augmentation using synthetic images enhances nanomaterial identification in electron micrographs. This approach addresses data scarcity, boosts the diversity of training datasets, and improves the robustness of classification models. By generating images that simulate rare scenarios, it offers a cost-efficient alternative to extensive data collection.
\vspace{-1mm}
\item \textbf{In-Context Learning for Nanomaterial Identification with Few-Shot Prompting with Multimodal Models(GPT-4V):} Our work investigates in-context learning using few-shot prompting with Language Model Multimodals (LMMs), such as GPT-4V, to eliminate the need for traditional gradient-based fine-tuning when classifying various nanomaterials in microscopy images. These LMMs utilize minimal examples(demonstrations) based on few-shot prompting---without any updates to the model parameters---to leverage analogy-based learning from prior knowledge for nanomaterial identification.
\end{itemize}

\vspace{-5mm}
\section{Problem Statment}
\label{pm}
\vspace{-1mm}
Our study focuses on the classification of electron micrographs using few-shot learning in large multimodal models (LMMs), such as GPT-4V. This approach involves leveraging a small set of relevant demonstrations(image-label pairs) to make predictions on new data (query images) without further fine-tuning of the model parameters. A common scenario is where the model samples image-label pairs from a training dataset \( \mathcal{D} \) as demonstrations, and then predicts the label of a query image from the test dataset based on these demonstrations. Consider a training dataset \( \mathcal{D} \) consisting of image-label pairs \( \{(I_i, y_i)\}_{i=1}^N \). Additionally, let \( I_q \) denote a query image. The task is to predict the label \( y_q \) of the query image \( I_q \) based on \( \mathcal{D} \), without model parameters update. In this scenario, using GPT-4V, the task can be framed as a probabilistic inference problem where the objective is to estimate the conditional probability distribution \( P(y_q | I_q, \mathcal{D}) \), representing the probability of the label \( y_q \) given the query image \( I_q \) and the training dataset \( \mathcal{D} \). Through this formulation, the few-shot learning task aims to sample the relevant demonstrations in the dataset \( \mathcal{D} \) to make an informed prediction for the label \( y_q \) of the query image \( I_q \), without requiring additional training of the model parameters.

\vspace{-4mm}

\vspace{-3mm}
\section{Proposed Method}
\label{pm}

\vspace{-1mm}
\paragraph{Electron Micrograph Encoder:} 
Let's consider an input image, denoted as $\mathbf{I}$ expressed as a 3D tensor with dimensions $H \times W \times C$, where $H$ represents the image's pixel height, $W$ its pixel width, and $C$ the number of channels associated with each pixel in the image. To process this input image, it is divided into smaller, non-overlapping patches, with each patch treated as a token having fixed-size spatial dimensions of $P \times P \times C$, where $P$ represents the patch size. The tokenization of the image results in a total number of patches given by $n = \left(\frac{HW}{P^{2}}\right)$. These patches are linearly encoded into 1D vectors, forming a sequence of tokens represented as $\mathbf{I'} \in \mathbb{R}^{n \times d}$ where $d$ is the patch embedding dimension. To maintain the spatial information of patches from the original image, differentiable positional embeddings representing patch positions are added element-wise to the patch embeddings. This process allows the framework to effectively analyze and understand the complex visual and spatial context of image patches. We also append a classification token $<\hspace{-1mm}\textit{cls}\hspace{-1mm}>$ to the token sequence. This token aggregates information from all patches, creating a global representation that helps the framework gain a coherent understanding of the holistic visual context of the image. We input this augmented token sequence into ViT\cite{dosovitskiy2020image}, which is composed of multiple stacked transformer encoder layers. Each encoder layer processes the patch embeddings hierarchically using a higher-order attention mechanism instead of the standard multi-head self-attention (MHSA), iteratively updating patch representations at different levels of abstraction. The hierarchical attention mechanism allows the framework to grasp visual information comprehensively at different levels of detail, from fine-grained features to high-level context. This process operates in two stages: local attention, which focuses on patch-level relationships to capture the interactions between patches and their immediate context within the image, and global attention, which aggregates global information by incorporating the classification token, aiding the framework in achieving an overarching understanding of the visual context throughout the entire image. After passing through the transformer layers, we consider only the output embedding $h_{\textit{cls}}$ corresponding to the $<\hspace{-1mm}\textit{cls}\hspace{-1mm}>$ token as the unified, holistic representation of the entire image, aggregating information from all patches by distilling the diverse and distributed information from the smaller, localized regions of the image. In summary, the framework processes input images by dividing them into patches, encoding them into tokens, incorporating a classification token $<\hspace{-1mm}\textit{cls}\hspace{-1mm}>$, and using a hierarchical attention mechanism to create a holistic image representation, $h_{\textit{cls}}$, that embodies both local and global context. For few-shot prompting of LMMs such as GPT-4V, we provide a small number of demonstrations(image-label pairs as input-output mappings) for nanomaterial identification in the query image. This is accomplished using an electron micrograph encoder that selects relevant images from the training set resembling or matching the query image.

\vspace{-2mm}
\paragraph{Zero-Shot Chain-of-Thought (CoT) GPT-4V Prompting:}
The GPT-4V API, accessible through Multimodal Modeling as a Service (MMaaS)—a cloud-based platform that accepts both image and text inputs to generate output—is not yet fully available to the public. While still in beta phase, GPT-4V can be accessed by ChatGPT Plus subscribers at chat.openai.com, but usage is subject to a cap. Our work on nanomaterial image interpretation begins with using GPT-4 to generate natural language questions that serve as task-specific instructions. These textual prompts, combined with visual (image) inputs, are employed to construct multimodal prompts that guide GPT-4V in Visual Question Answering (VQA) tasks for analyzing nanomaterial images. Consequently, GPT-4V provides contextually rich textual responses that encapsulate the information within the visual inputs. The task instructions created by GPT-4 (language-only) are crucial for directing GPT-4V's VQA performance on nanomaterial images. By utilizing a zero-shot CoT prompt template with these instructions and the query image, LMMs like GPT-4V can generate detailed descriptions of nanomaterial images. This approach takes advantage of the multimodal model’s intrinsic domain-specific knowledge acquired during training to provide comprehensive insights into the images. Essentially, GPT-4 formulates general questions about nanomaterial images, which are then converted into structured CoT prompts guiding GPT-4V in its detailed visual analysis that explores the image's structure, patterns, imaging techniques, and context—be it experimental, real-world, or theoretical.  In guiding GPT-4V's analysis of nanomaterial images, we focus on the following key areas: (a) Basics: Identify the type and scale of the nanomaterial. (b) Morphology and Structure: Describe the shape, layers, domains, and uniformity. (c) Size and Distribution: Determine size, distribution pattern, and signs of aggregation. (d) Surface Characteristics: Observe texture, defects, or impurities. (e) Composition and Elements: Identify compositional variations and specific elements. (f) Interactions and Boundaries: Examine nanostructure interactions and boundaries. (g) External Environment: Observe interactions with surroundings and identify non-nanomaterial structures. (h) Image Technique and Modifications: Identify the imaging technique and any post-processing. (i) Functional Features: Look for functional features and assess if dynamic processes are captured. (j) Context and Application: Understand the sample's intended use and its status as real, experimental, or theoretical. The CoT prompt format is as follows:

\vspace{-5mm}
\begin{tcolorbox}[colback=white!5!white,colframe=black!75!black]
\vspace{-2mm}
\textbf{Prompt 1:} **Basics** - What type of nanomaterial is depicted in the image? - What is the scale of the image? (e.g., what does one unit of measurement represent?). \textbf{Prompt 2:} **Morphology and Structure** - What is the general shape or morphology of the nanomaterials in the image? - Are there distinct layers, phases, or domains visible?
- Do the nanomaterials appear uniform in size and shape or are they varied?. \textbf{Prompt 3:} **Size and Distribution** - What is the approximate size or size range of the individual nanostructures?  - How are the nanomaterials distributed throughout the image? (e.g., evenly spaced, clustered, random) - Is there any evidence of aggregation or bundling?. \textbf{Prompt 4:} **Surface Characteristics** - Does the nanomaterial appear smooth, rough, or have any specific textures? - Are there any visible defects, pores, or impurities on the surface?. \textbf{Prompt 5:} **Composition and Elements** - Is there evidence of compositional variations in the image (e.g., different colors, brightness, or contrasts)? - Are there any labels or markers indicating specific elements or compounds present?. \textbf{Prompt 6:} **Interactions and Boundaries** - How do individual nanostructures interact with one another? (e.g., are they touching, fused, or separate?) - Are there clear boundaries between different structures or phases?. \textbf{Prompt 7:} 
\vspace{-3mm}
\end{tcolorbox}

\vspace{-3mm}
\begin{tcolorbox}[colback=white!5!white,colframe=black!75!black]
\vspace{-2mm}
**External Environment** - Is there any evidence of the nanomaterial interacting with its surrounding environment or matrix (e.g., solvents, polymers, or other materials)? - Are there other structures or objects in the image that are not nanomaterials? If so, what are they?. \textbf{Prompt 8:} **Image Technique and Modifications** - What imaging technique was used to capture this image? (e.g., SEM, TEM) - Were there any post-processing or modifications made to the image (e.g., false coloring, 3D rendering)?. \textbf{Prompt 9:} **Functional Features** - If applicable, are there any functional features visible (e.g., active sites, regions with distinct properties)? - Are there dynamic processes captured in the image or is it a static representation?. \textbf{Prompt 10:}  **Context and Application** - What is the intended application or use of the nanomaterial being depicted? - Is this a experimental sample, or a theoretical or simulation-based representation?
\vspace{-2mm}
\end{tcolorbox}

\vspace{-1mm}
The structured prompts are designed to facilitate a comprehensive, in-depth exploration of various facets, ranging from fundamental aspects like size and distribution, to morphology and structure, to practical applications associated with these nanomaterials. Zero-shot CoT prompting in LMMs such as GPT-4V generates text that responds to and elaborates on the specific aspects mentioned in each prompt.

\vspace{-1mm}
\begin{tcolorbox}[colback=white!5!white,colframe=black!75!black]
\vspace{-1.5mm}
\centering
(\textbf{Chatbot's Response}) [Generated Text]
\vspace{-1.5mm}
\end{tcolorbox}

\vspace{-1mm}
In the following section, we will outline our approach to integrating these generated technical descriptions, which will serve as input for creating synthetic nanomaterial images using DALL-E 3, an advanced text-to-image generation model, capable of translating textual descriptions into highly accurate images that adhere closely to the provided text prompts.
Table \ref{tab:tab9} shows the Q$\&$A pairs for patterned surface nanomaterials using Zero-Shot CoT prompting of GPT-4V.

\vspace{-2mm}
\paragraph{Zero-Shot DALL-E 3 Prompting:}
The DALL-E 3 API, available as a cloud service, is engineered to transform text inputs into high-quality images. Public access to the API, however, is limited. ChatGPT Plus subscribers can utilize DALL-E 3 on chat.openai.com, subject to usage limits, which enables the generation of realistic nanomaterial images based on textual descriptions. These technical descriptions are provided in the form of question-answer pairs by GPT-4V and contain in-depth information about the nanomaterials depicted in the images. DALL-E 3 is designed to understand textual prompts and create visually accurate representations based on those prompts. The zero-shot prompting capability of DALL-E 3 emphasizes its ability to accurately convert text into images without requiring additional prompt engineering or task-specific tuning. This capability is achieved by leveraging its pre-existing knowledge, akin to zero-shot prompting with language models, which respond to tasks without having been exposed to specific examples during training. The zero-shot prompt format is as follows,

\vspace{-2mm}
\begin{tcolorbox}[colback=white!5!white,colframe=black!75!black]
\vspace{-2mm}
Please generate multiple synthetic images based on the textual information provided below in the form of question-answer pairs for a given nanomaterial.
\vspace{-2mm}
\end{tcolorbox}

\vspace{-2mm}
Table \ref{tab:tab10} displays synthetic images of patterned surface nanomaterials, created by DALL-E 3 through Zero-Shot prompting, using textual descriptions generated by GPT-4V.

\vspace{-2mm}
\paragraph{Few-Shot GPT-4V Prompting in Nanomaterial Identification:}
Few-shot prompting is a technique that enables in-context learning in language-and-vision multimodal models (LMMs) such as GPT-4V, guiding these large-scale models to better performance on complex, unseen tasks. With this technique, LMMs can tackle new tasks without the need for traditional gradient-based fine-tuning on labeled data for domain-specific task adaptation. Instead, the multimodal model uses a minimal set of task-specific input-output pairs as demonstrations to apply analogy-based learning, using the implicit prior knowledge acquired during pre-training to handle new tasks. Context-augmented prompting enhances the emerging few-shot learning abilities of LMMs by including both task-specific instructions and demonstrations in the prompt, aiding LMMs to better adapt and perform on unseen tasks, thereby improving their generalization capabilities. In the realm of nanomaterial identification, few-shot prompting employs a small number of image-label pairs, represented as \((\mathcal{I}_i, \mathcal{Y}_i)\), sampled from the training set relevant to the query image, which serves as guiding demonstrations. Task-specific instructions involve a natural language question to instruct GPT-4V to predict the labels of query images. At inference time, for test images denoted as \( \mathcal{I}_{\text{test}} \), few-shot prompting determines the output label using the conditional probability distribution, articulated as  \( \mathbf{P}(\mathcal{Y}_{\text{test}} \mid ((\mathcal{I}_{\text{train}}, \mathcal{Y}_{\text{train}}), \mathcal{I}_{\text{test}})) \). This showcases a data-efficient learning paradigm that enhances the multimodal models' adaptability and generalization capabilities for unseen or novel tasks, crucial in scenarios with limited labeled data. Building upon the foundation of few-shot prompting in the context of nanomaterial identification, we delve into the influence of both the quality and quantity of demonstrations on task performance. Specifically, we evaluate two contrasting sampling strategies for selecting these demonstrations: random and similarity-driven sampling. The random method offers a naive approach by arbitrarily selecting demonstrations (image-label pairs) from the training data, without adhering to any specific criteria or systematic approach, thereby serving as a baseline for our evaluations. On the other hand, similarity-driven sampling utilizes the cosine similarity method to find the most similar images in the training data to the query image. Underlying this strategy is the hypothesis that demonstrations closely mirroring the data distribution of the query image can potentially boost the adaptability and precision of the multimodal model used for nanomaterial identification. By employing diverse strategies to sample demonstrations when constructing multimodal prompts, we aim to provide a thorough analysis of how different demonstration sampling methods affect the efficacy of few-shot learning of LMMs in nanomaterial identification tasks. Furthermore, the effectiveness of these demonstrations is directly linked to the sampling methods used to retrieve the top-K images that closely align with a new or unseen query image. To delve deeper into the impact of the number of demonstrations (\( K \)) on performance, we adjust \( K \) for each query image. Utilizing the electron micrograph encoder, we process an image dataset to extract the holistic representation 

\clearpage
\onecolumn

\begin{tcolorbox}[colback=white!2!white,colframe=black!75!black]
\vspace{-5mm}
\begin{tabularx}{1.2\textwidth}{bss}
\caption{The table presents the question-answer pairs generated by the LMM(GPT-4V) in response to natural language questions about the input image of the patterned surface nanomaterial category. In summary, the question-answer pairs (generated text) provide in-depth information about an image related to patterned surface nanomaterials.}\label{tab:tab9}\\ [-0.75em]
\toprule 
 \\ [-0.75em]
\begin{overpic}[width=0.175\columnwidth]{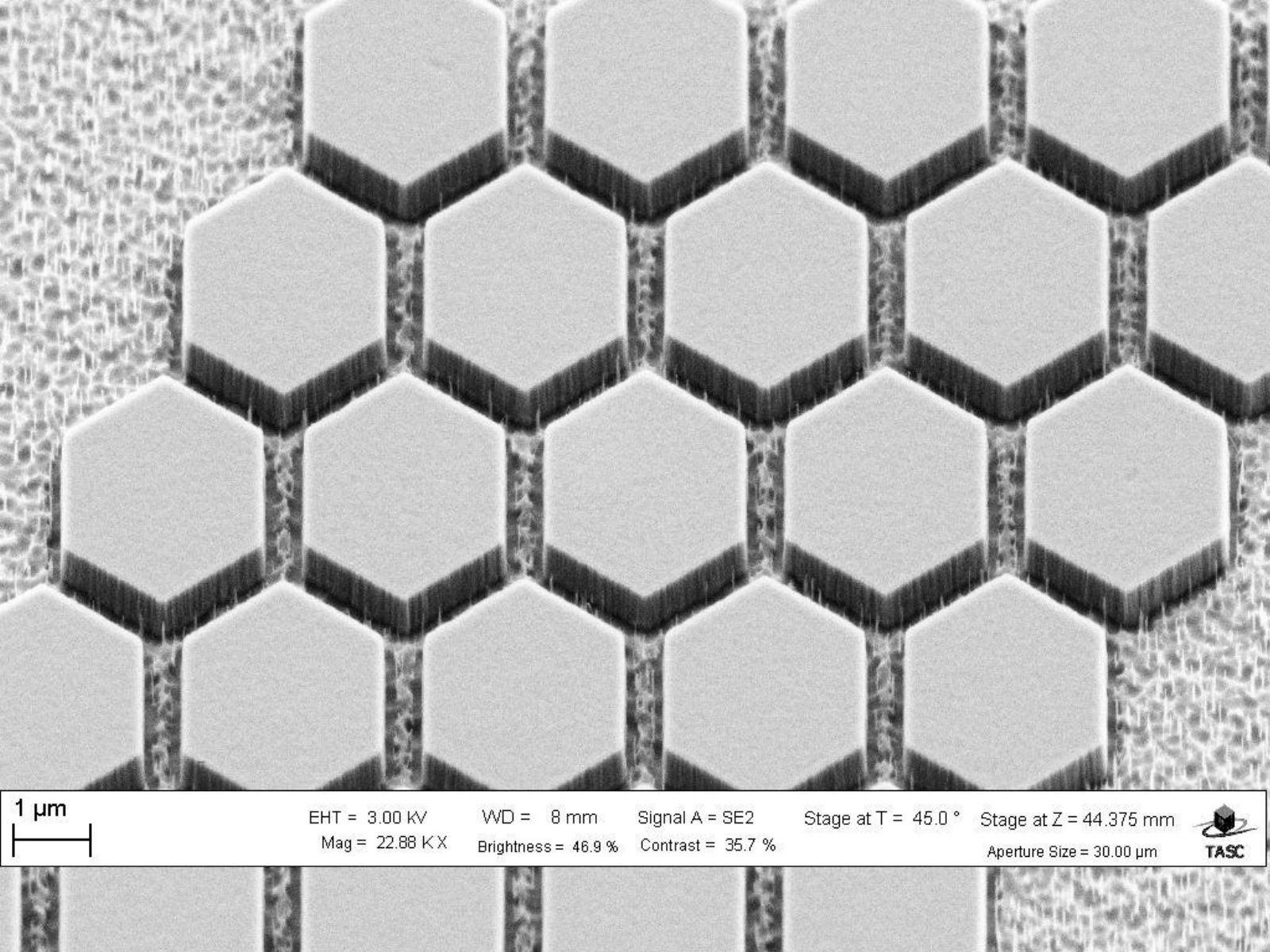}
  \put(400,40){\makebox[0pt][c]{%
    \parbox{0.5\columnwidth}{%
      \centering
      \textbf{Please answer the following questions based on the provided input image belonging to the patterned surface nanomaterial category.} 
    }%
  }}  
\end{overpic}  \\ [-0.25em]

\midrule
\textbf{Query prompt}  & \textbf{GPT-4 Vision output} \\
\midrule
\endhead
\textcolor{black}{**Basics**} & The image depicts a patterned surface nanomaterial. The scale of the image is indicated by the bar at the bottom-left corner, which represents 1 $\mu$m.
\\

\midrule
\textcolor{black}{**Morphology and Structure**} & The general shape or morphology of the nanomaterials in the image is hexagonal. The image seems to show one distinct layer of hexagonal nanostructures on a textured background. The nanomaterials appear uniform in size and shape.
\\

\midrule
\textcolor{black}{**Size and Distribution**} & The approximate size of the individual nanostructures is slightly less than 1 $\mu$m given the provided scale. The nanomaterials are evenly spaced and arranged in a hexagonal grid pattern. There is no visible evidence of aggregation or bundling.
\\

\midrule
\textcolor{black}{**Surface Characteristics**} & The hexagonal nanomaterials appear smooth, while the background has a textured pattern. There are no obvious defects, pores, or impurities on the hexagonal structures.
\\

\midrule
\textcolor{black}{**Composition and Elements**} & The image is grayscale, so it's difficult to determine compositional variations based on colors. However, there is contrast between the hexagonal structures and the background. There are no visible labels or markers indicating specific elements or compounds.
\\
 
\midrule
\textcolor{black}{**Interactions and Boundaries**} & The individual hexagonal nanostructures seem to be separate from one another with clear gaps in between. There are clear boundaries between the hexagonal structures and the background.
\\
 
\midrule
\textcolor{black}{**External Environment**} & There isn't direct evidence from the image to indicate interactions between the nanomaterial and its surrounding environment. The background texture appears distinct from the hexagonal nanostructures but is not labeled, making its composition or identity unclear.

\\
 
\midrule
\textcolor{black}{**Image Technique and Modifications**} & The image appears to be taken using Scanning Electron Microscopy (SEM) based on the details provided in the image. There is no indication in the image about post-processing or modifications.
\\
 
\midrule
\textcolor{black}{**Functional Features**} & It's unclear from the image alone if there are any functional features visible. The image represents a static view of the nanostructures.
\\

\midrule
\textcolor{black}{**Context and Application**} & The intended application or use of the nanomaterial is not provided in the image. Given the detailed nature of the image and the presence of measurement scales and settings, it appears to be a real, experimental sample.
\\

\bottomrule
\end{tabularx}
\end{tcolorbox}

\vspace{-8mm}

\begin{tcolorbox}[colback=white!2!white,colframe=black!75!black]
\vspace{-5mm}
\begin{tabularx}{1.35\textwidth}{bsssss}
\\ [-0.75em]
\hspace*{-3mm}\includegraphics[width=0.15\textwidth]{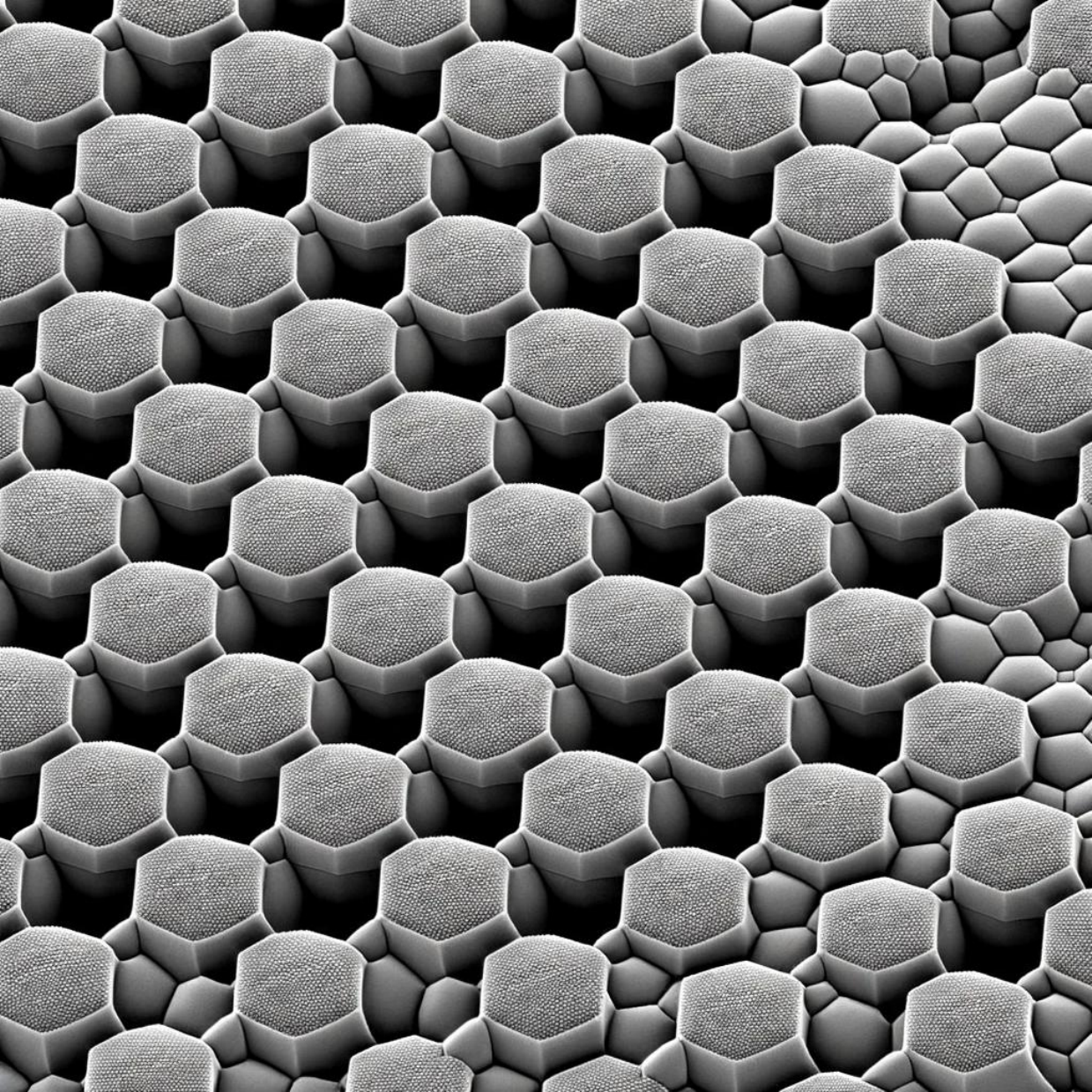} & 
\includegraphics[width=0.15\textwidth]{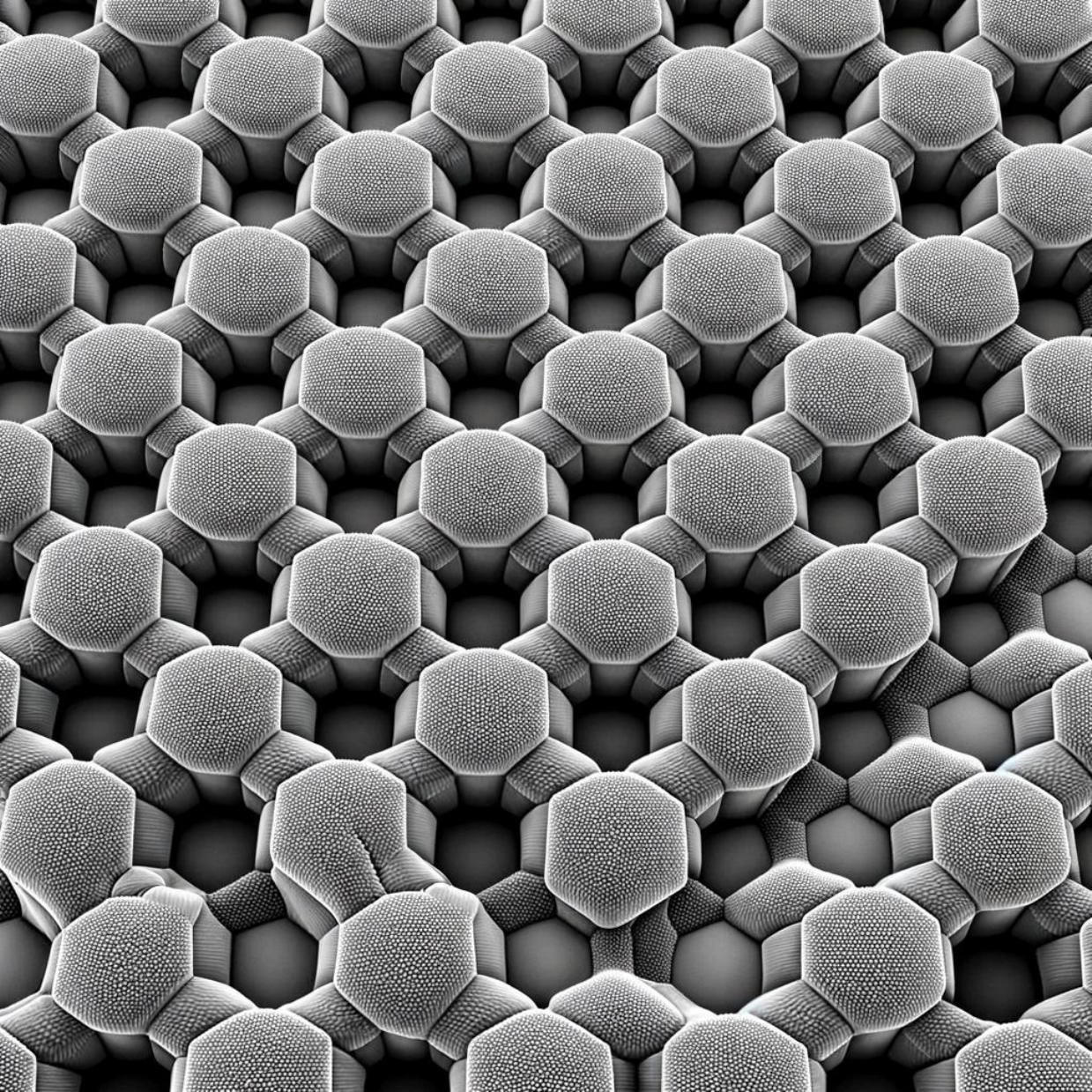} & 
\includegraphics[width=0.15\textwidth]{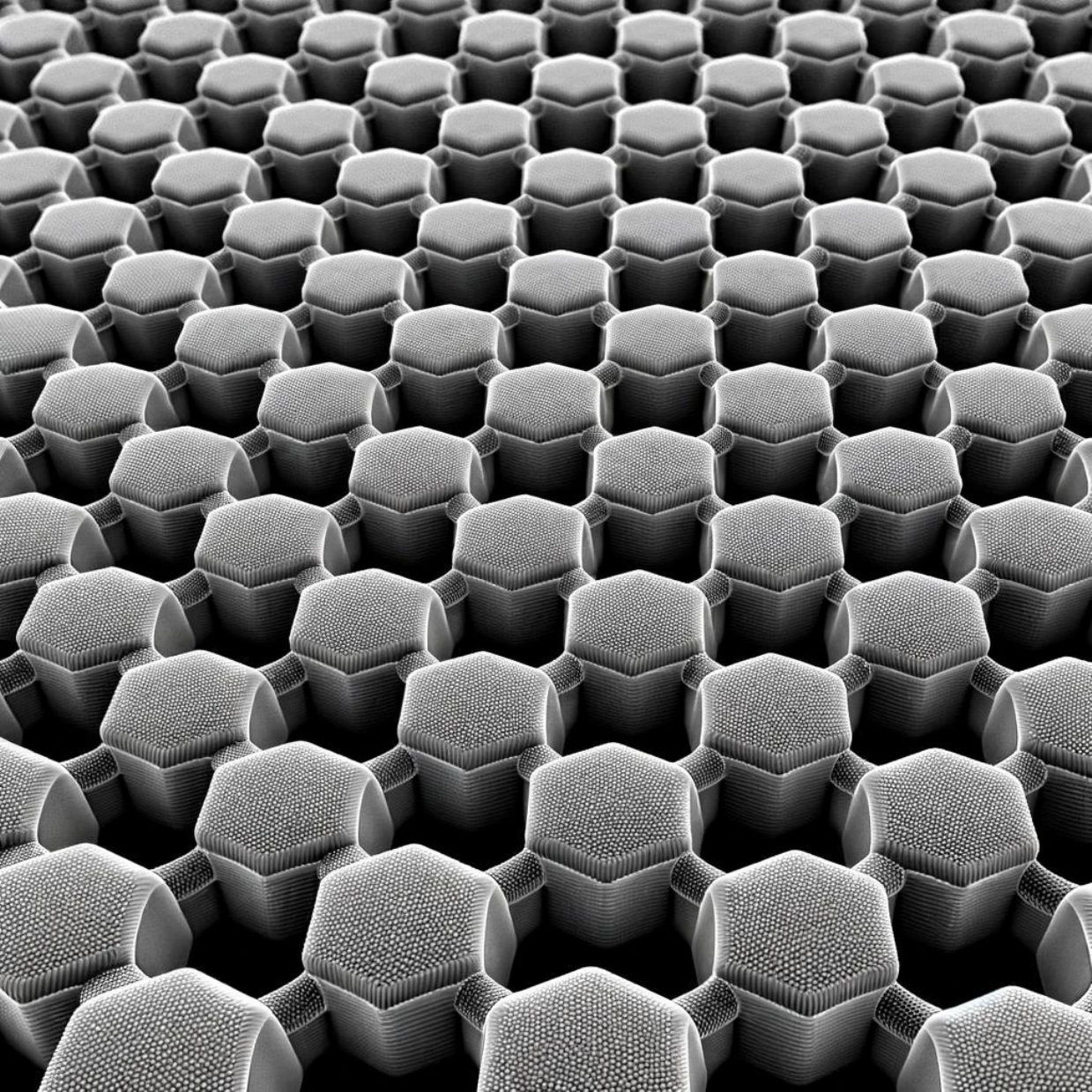} &
\includegraphics[width=0.15\textwidth]{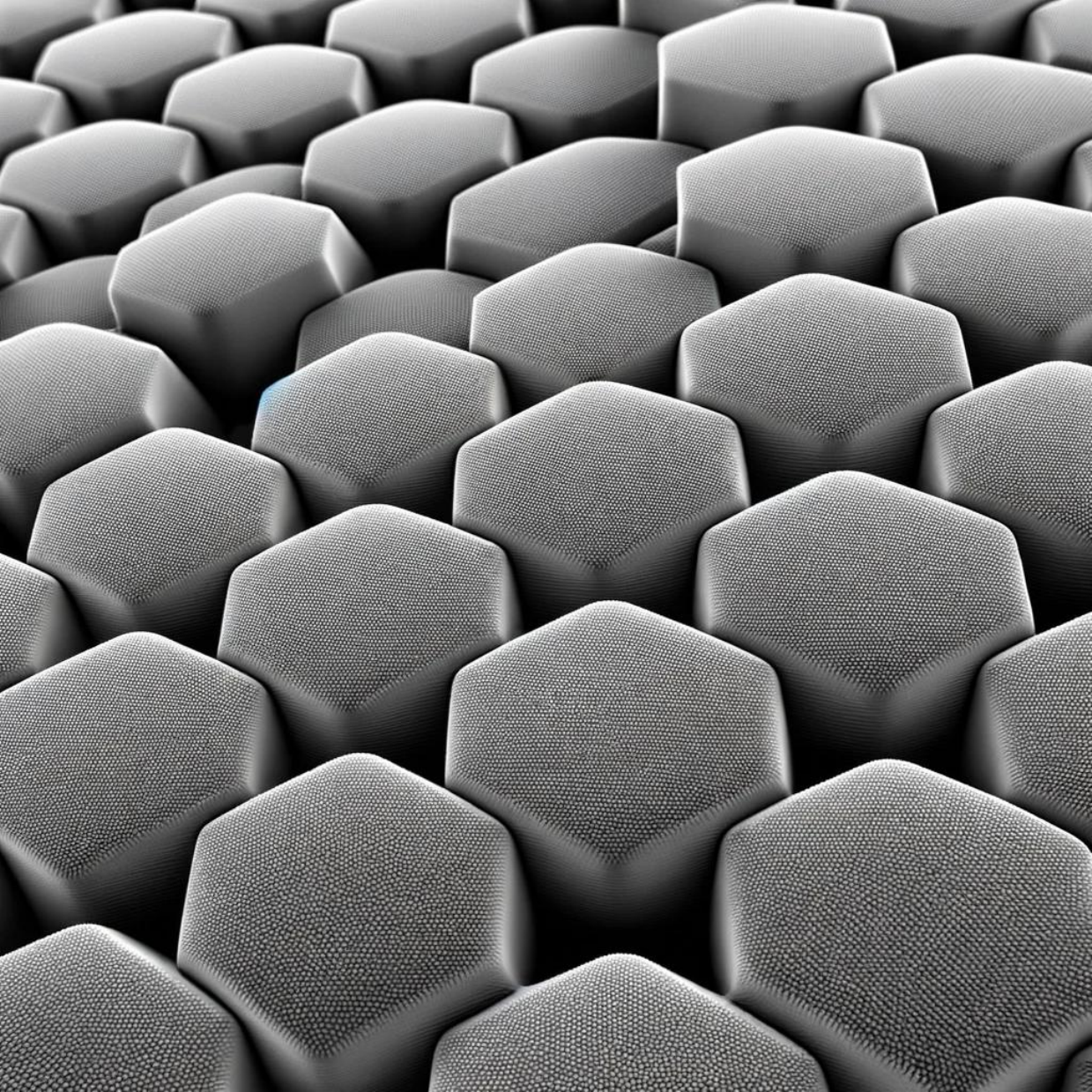} & 
\includegraphics[width=0.15\textwidth]{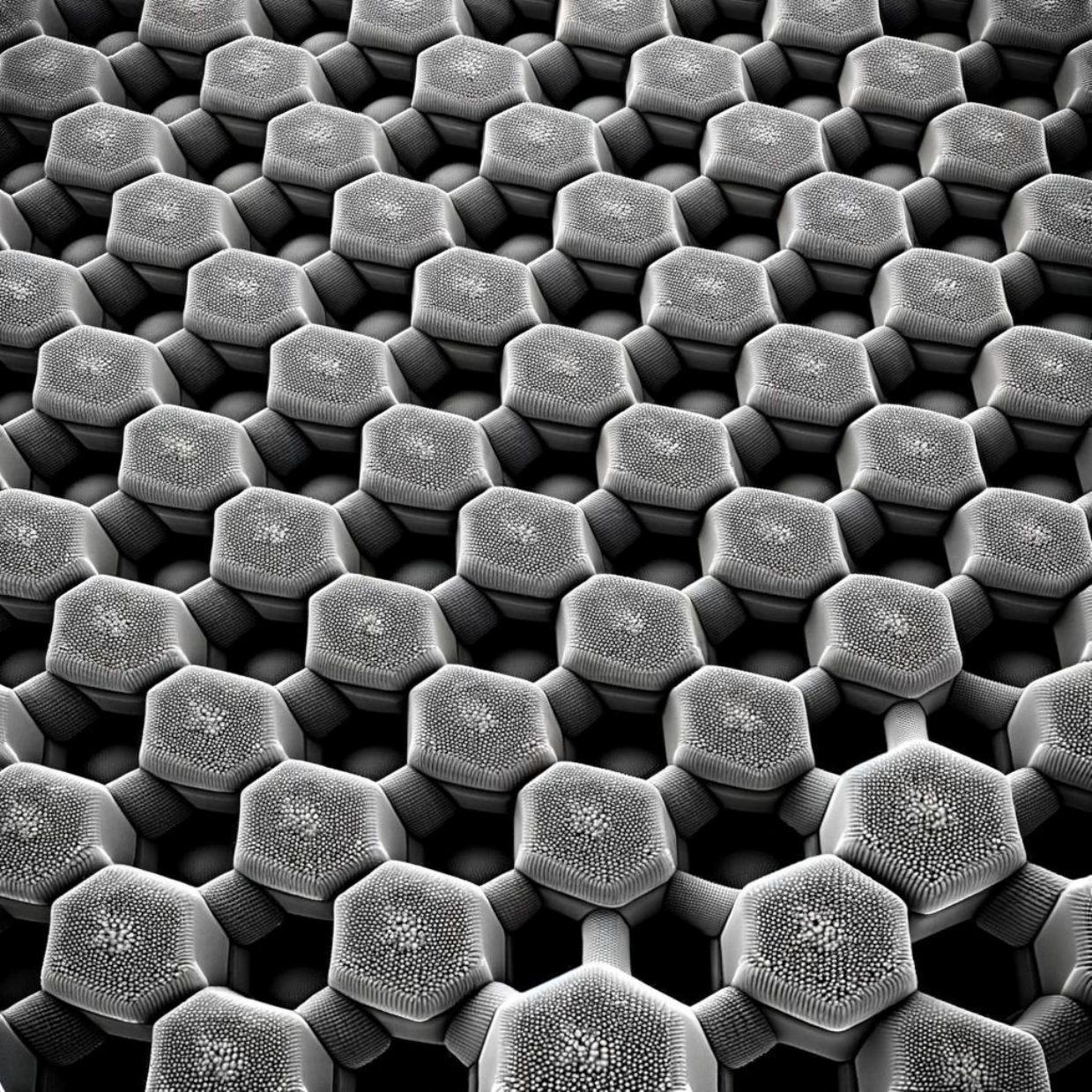} & 
\includegraphics[width=0.15\textwidth]{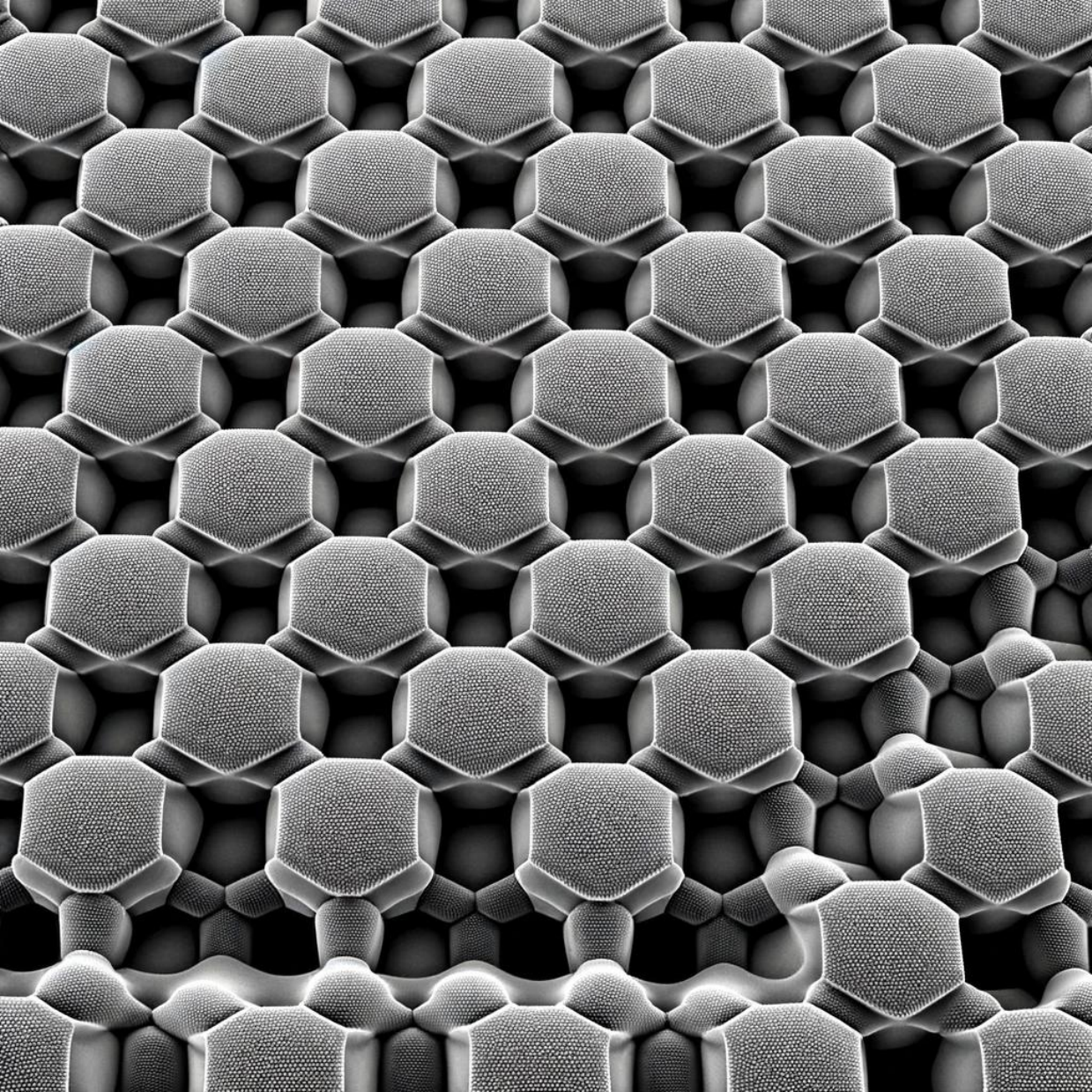} \\ [-0.65em] 
\caption{The table shows sample synthetic images generated by DALLE-3 from textual descriptions provided in a series of question-answer pairs, as shown in Table \ref{tab:tab9}.}
\label{tab:tab10}
\end{tabularx}
\vspace{-2mm}
\end{tcolorbox}

\clearpage
\twocolumn

\( h_{\text{cls}} \) for each image. We then compute similarity scores between the query image and the images in the dataset using metrics such as cosine similarity or Euclidean distance.  By ranking the images based on these similarity scores, we select the top-K most similar images. These selected images serve as the demonstration set for few-shot prompting, aiding the model in making accurate predictions for the query image. In brief, our objective is to explore the promising, few-shot learning abilities of LMMs via prompting on nanomaterial identification task. A multimodal prompt consists of selected few image-label pairs from the training data, accompanied by task-specific instructions that guide the LMMs in predicting the nanomaterial category of the query image. This evaluation examines the LMMs' capability to predict nanomaterial categories based solely on the contextual prompt, without any parameter updates or access to external knowledge, distinguishing it from traditional supervised learning where models are fine-tuned on labeled data.

\vspace{-2mm}
\begin{tcolorbox}[colback=white!5!white,colframe=black!75!black]
\centering
\vspace{-2mm}
Below are the provided image-label pairs for the nanomaterial identification task. Based on these pairs, predict the nanomaterial category for the given query image.
\vspace{-2mm}
\end{tcolorbox}

\vspace{-1mm}
In summary, few-shot prompting enables models like GPT-4V to predict nanomaterial categories without fine-tuning by utilizing select demonstrations, task-specific instructions, and the prior knowledge acquired from training on diverse multimodal datasets.

\vspace{-4mm} 
\section{Experiments And Results}
\label{expresults}

\vspace{-12.5mm}
\begin{figure}[htbp]
\centering
     \subfloat{\hspace{-0mm}\includegraphics[width=0.13\textwidth]{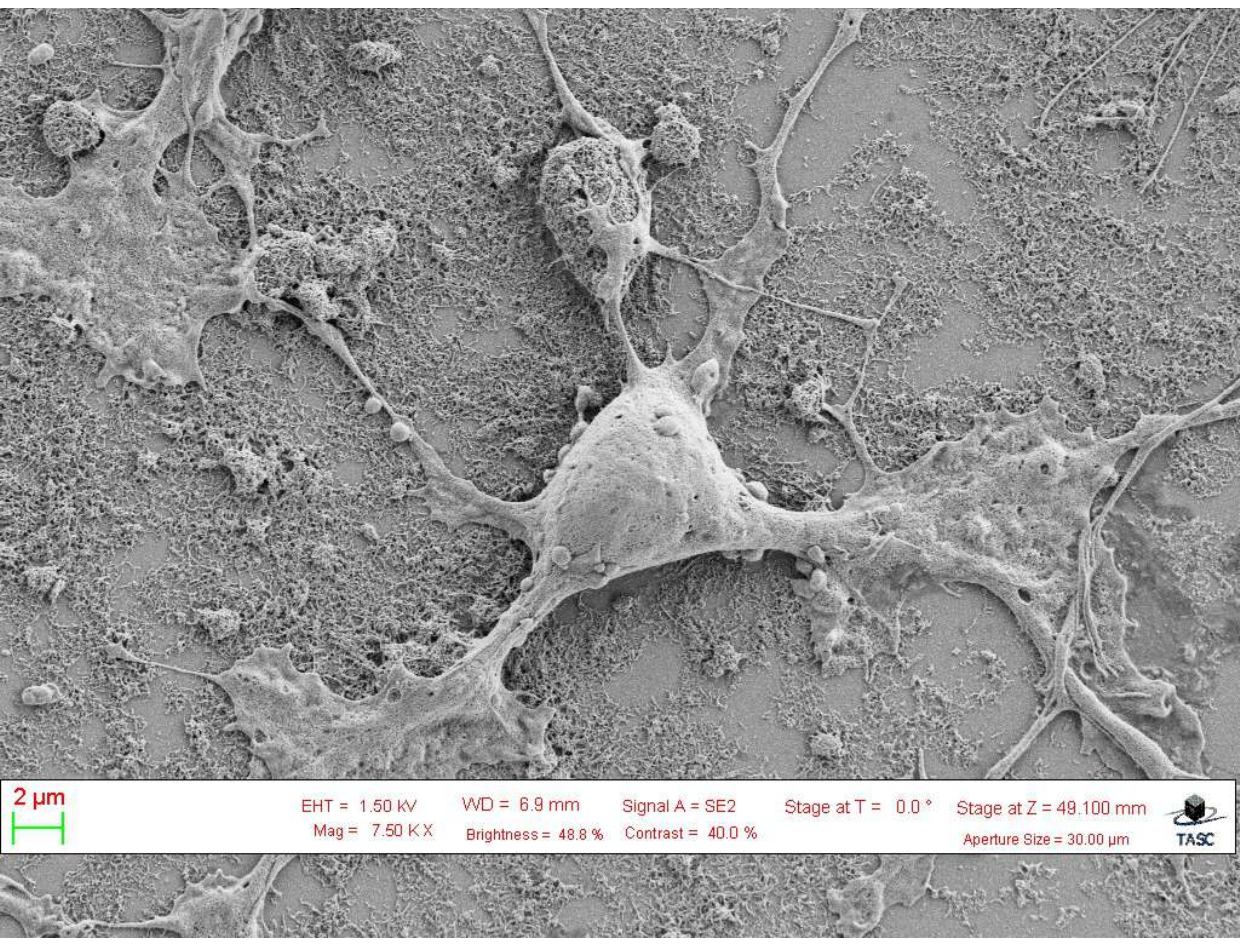}
     \includegraphics[width=0.13\textwidth]{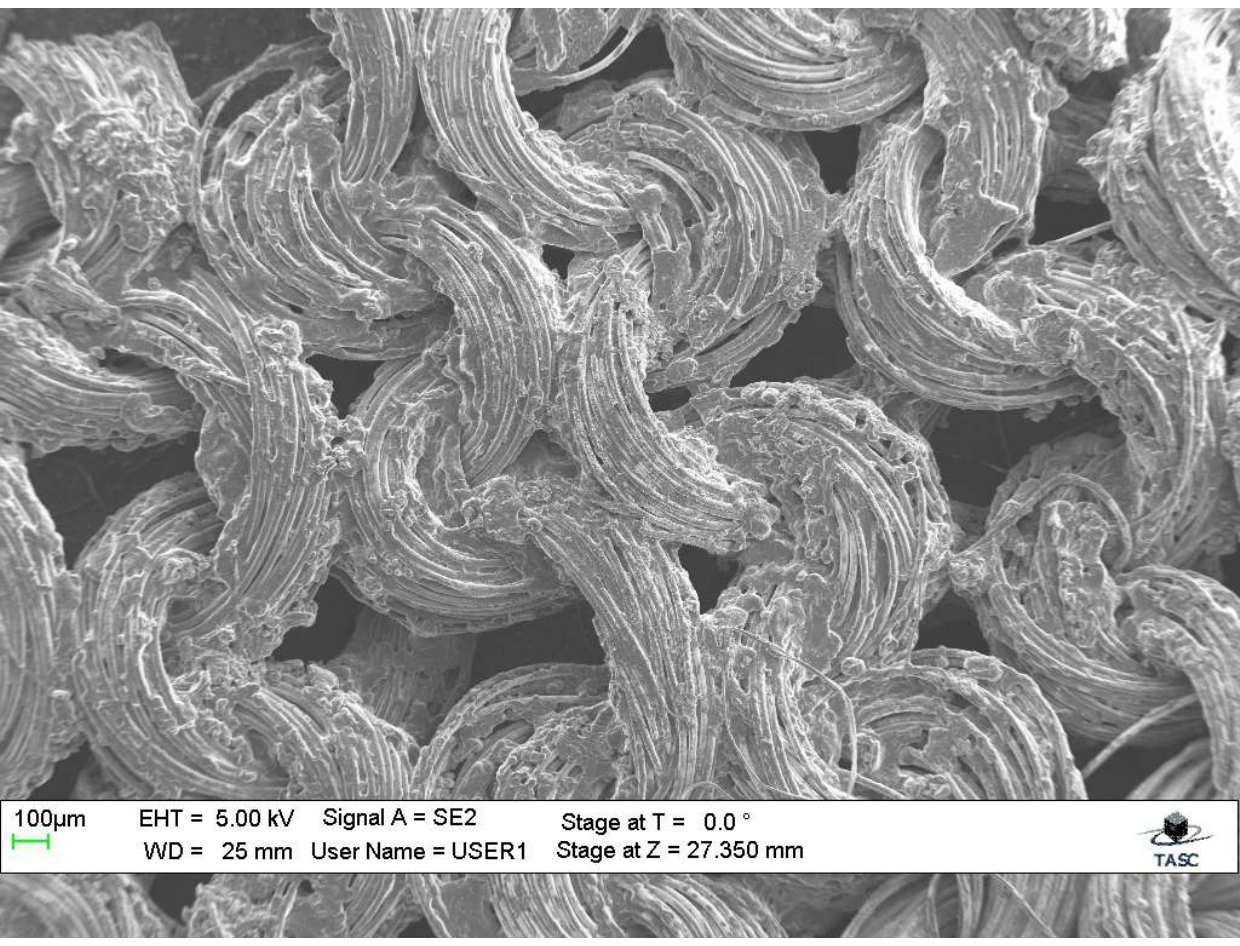}
     \includegraphics[width=0.13\textwidth]{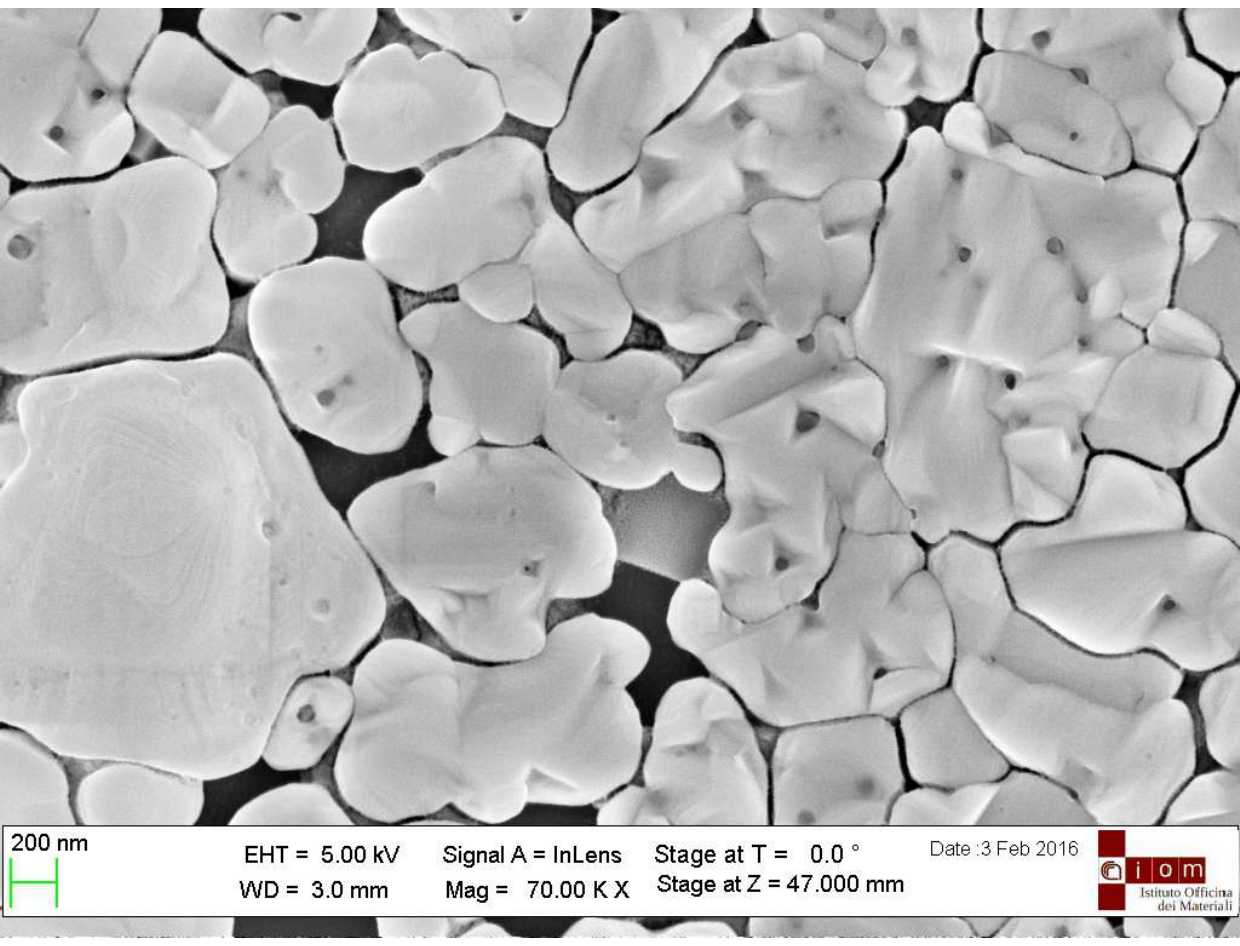}
     }
     \vspace{-15mm}
     \qquad
     \subfloat{\hspace{-0mm}\includegraphics[width=0.13\textwidth]{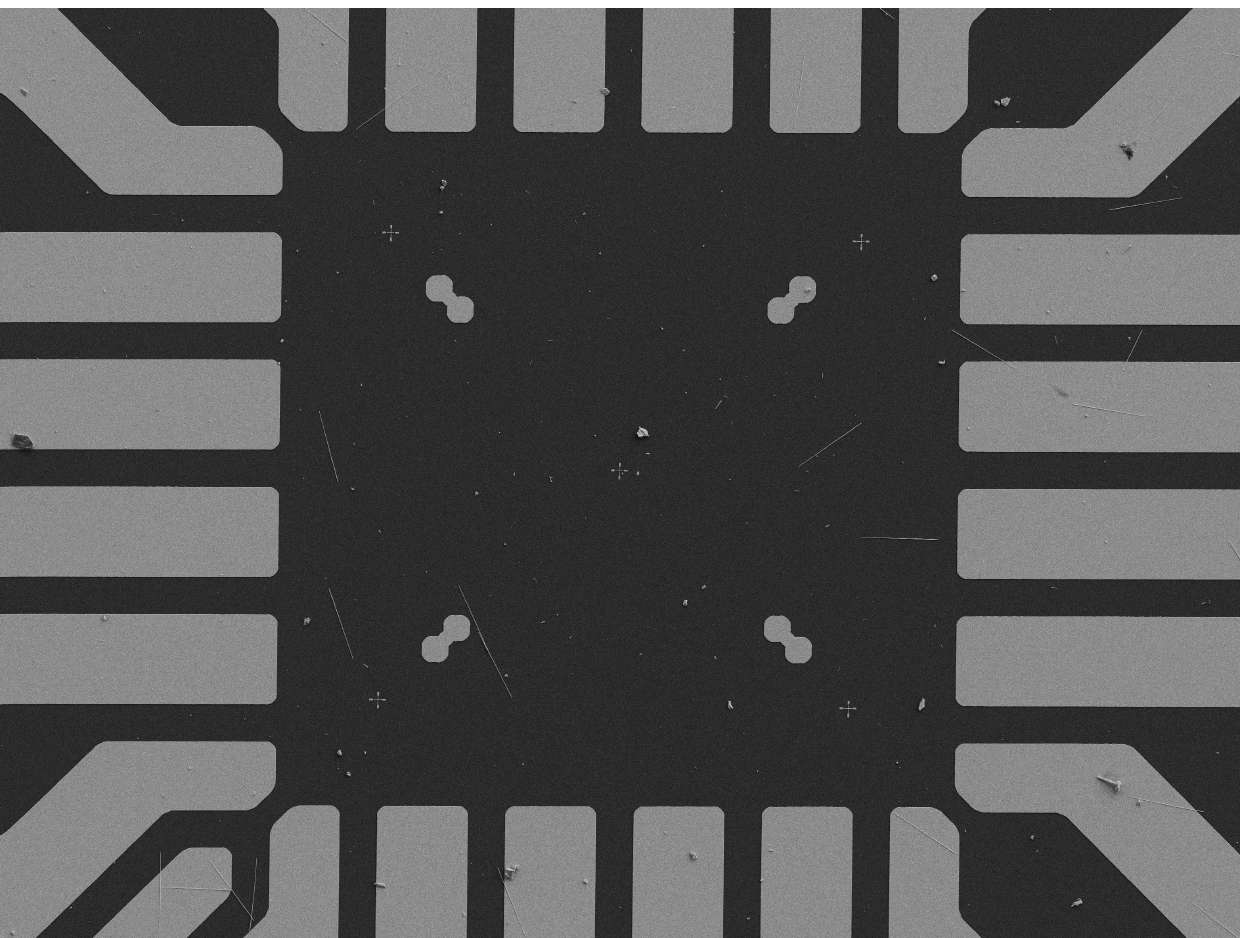}
     \includegraphics[width=0.13\textwidth]{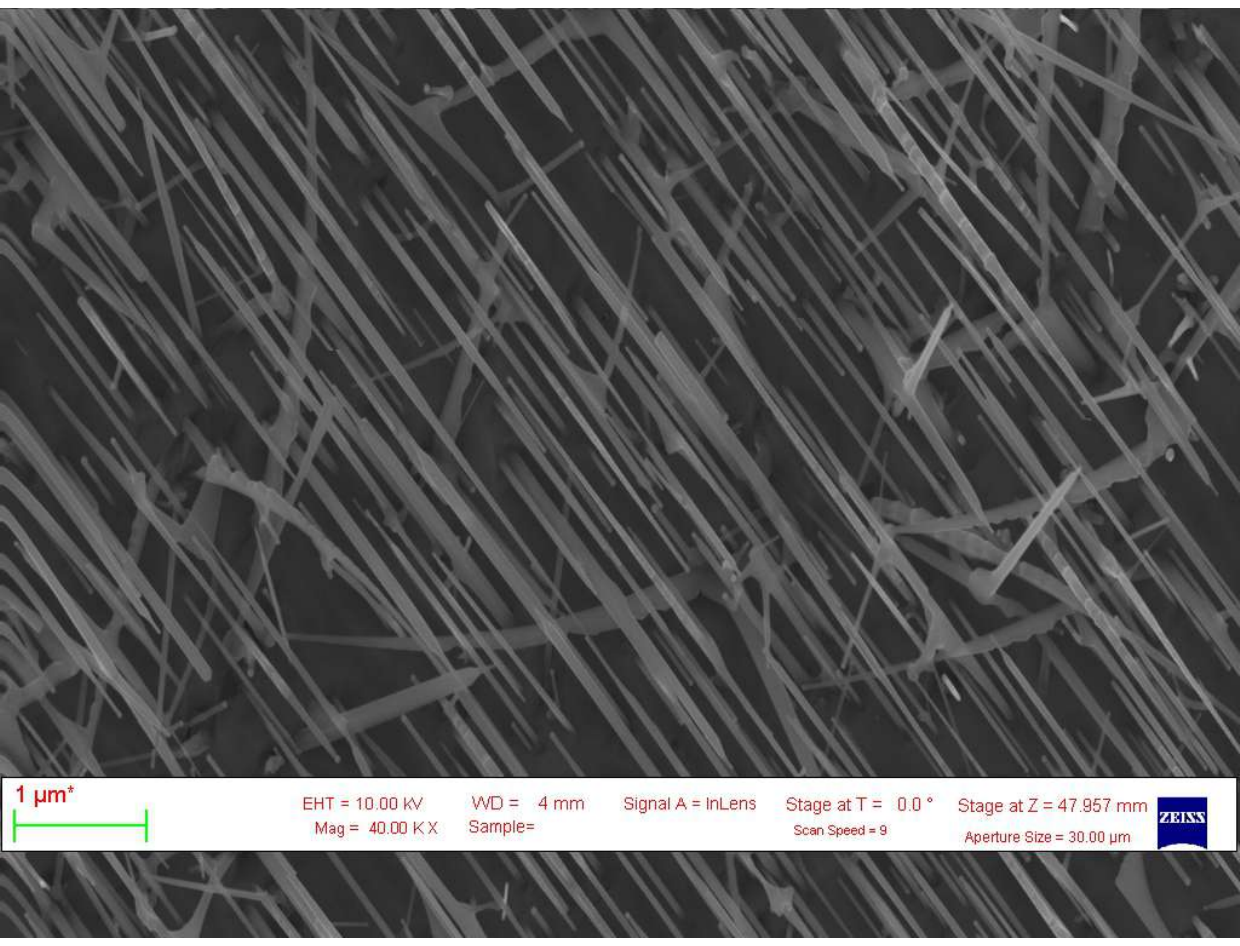}
     \includegraphics[width=0.13\textwidth]{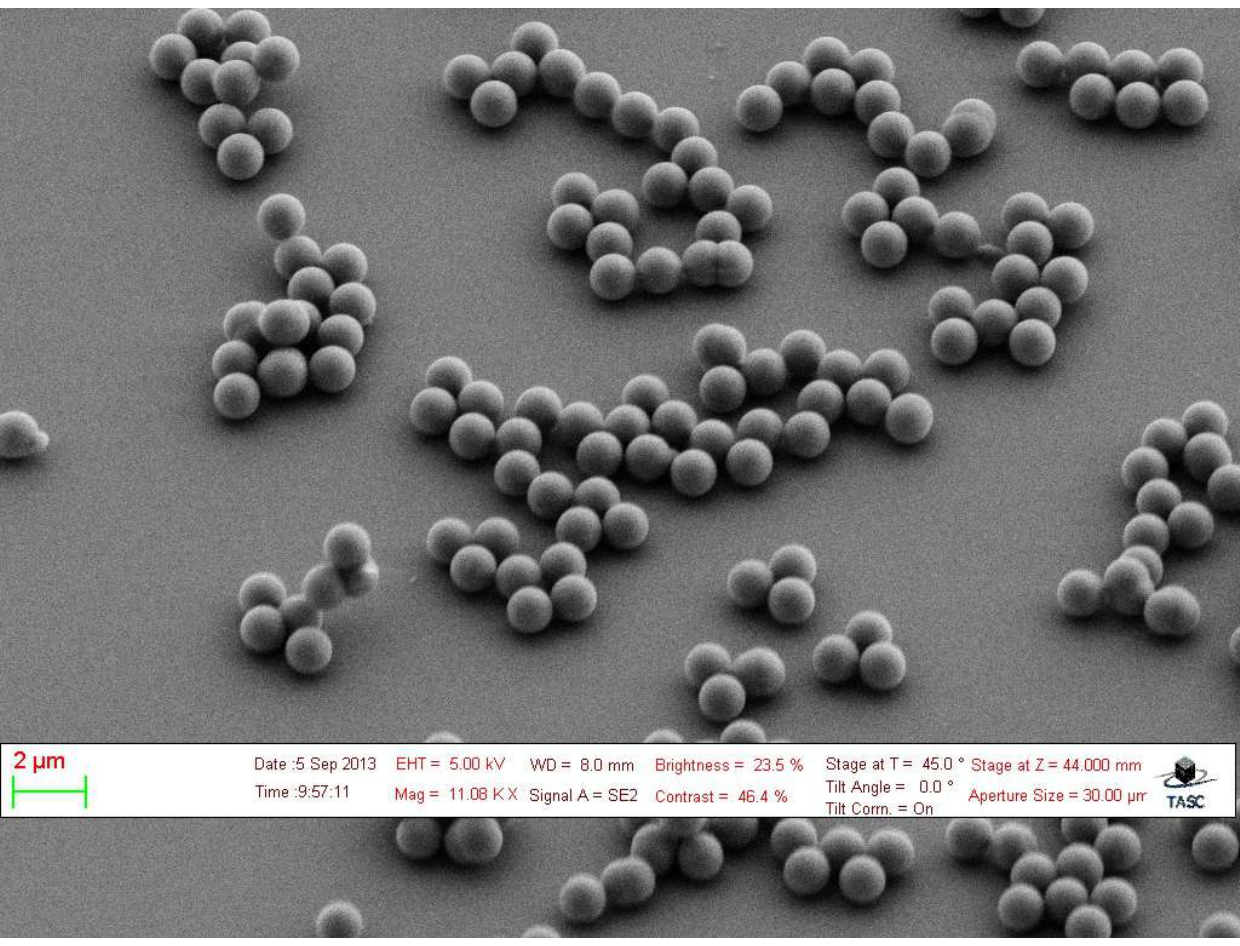}
     }
     \vspace{-15mm}
     \qquad    
     \subfloat{\hspace{-0mm}\includegraphics[width=0.13\textwidth]{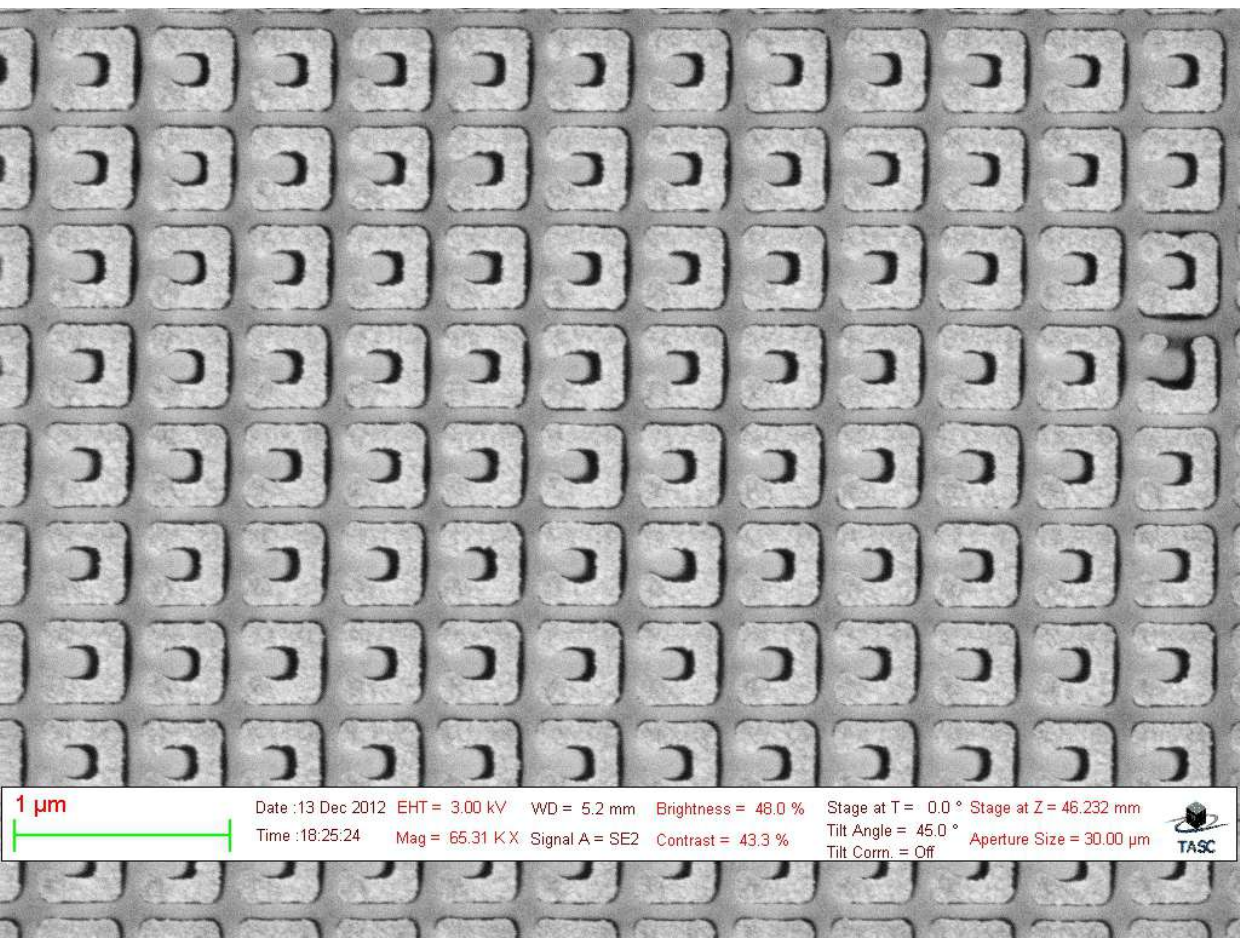}
     \includegraphics[width=0.13\textwidth]{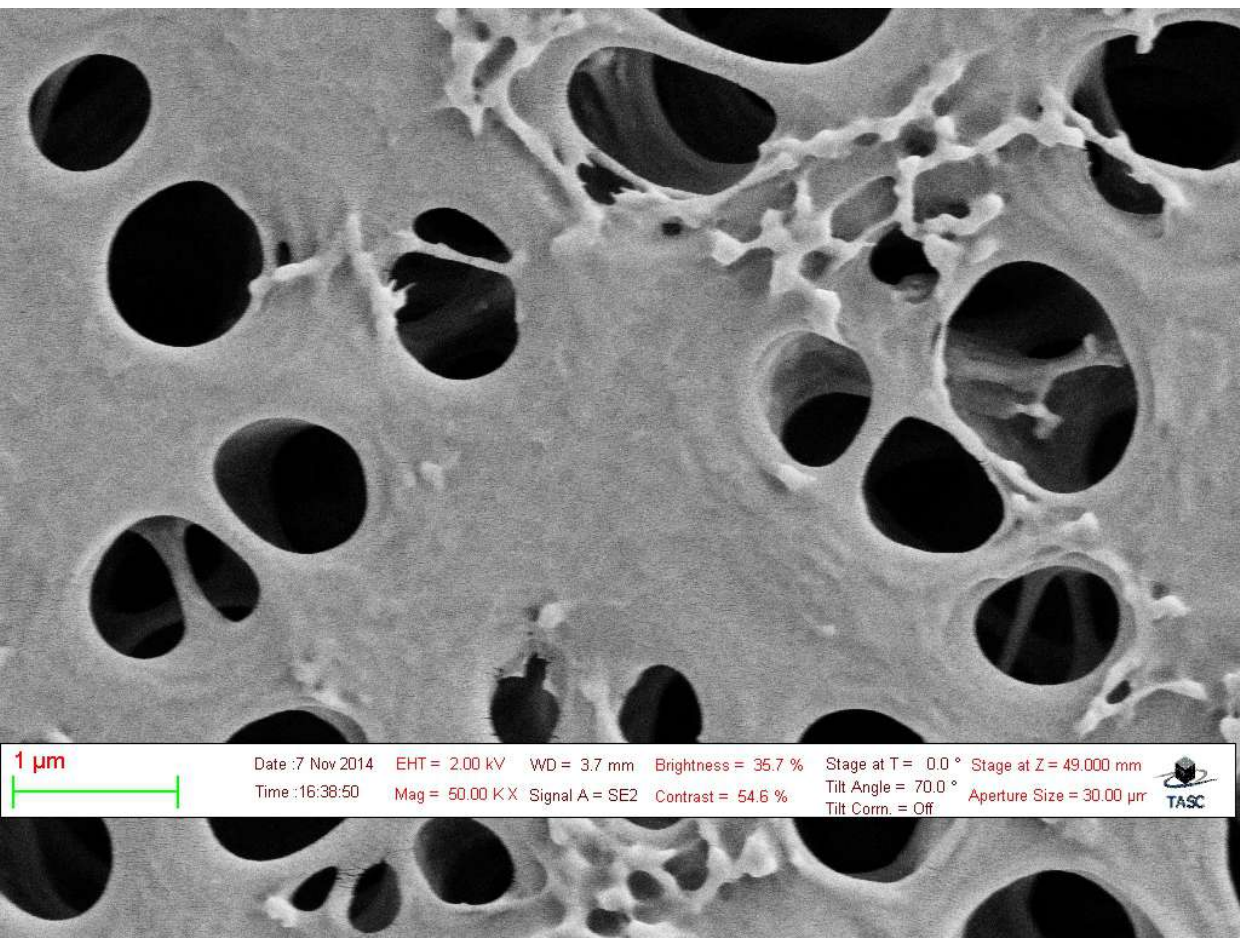}
     \includegraphics[width=0.13\textwidth]{Images/ShortlistedImages/Powder.pdf}
     }
     \vspace{-15mm}
     \qquad
     \subfloat{\hspace{-0mm}\includegraphics[width=0.13\textwidth]{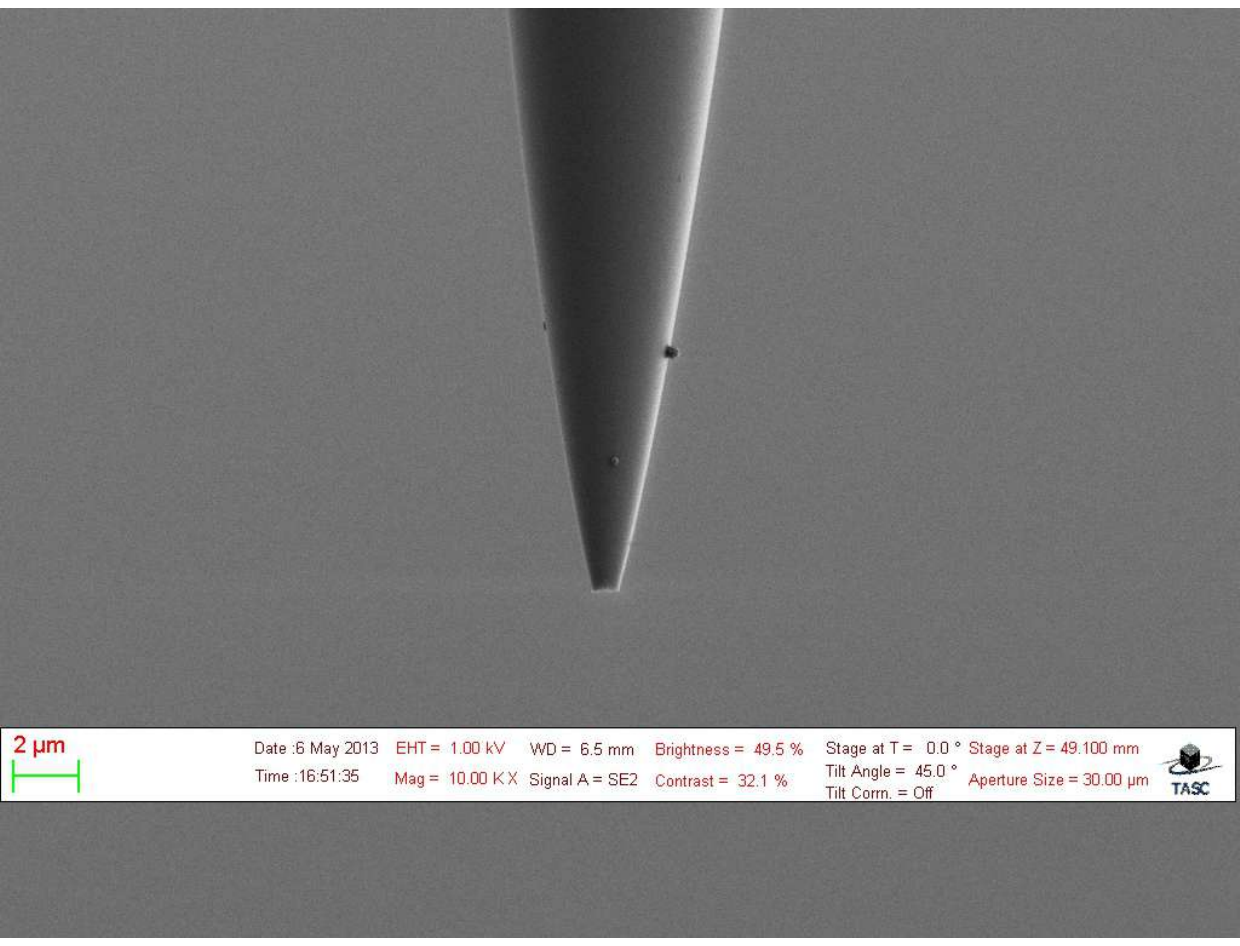}
     }
     \vspace{-10mm}
     \caption{The figure shows sample microscopic images of nanomaterials with different structures and morphologies found in the SEM dataset \cite{aversa2018first}. From left to right in the first row: \textit{biological, fibers, films}; in the second row: \textit{MEMS, nanowires, particles}; in the third row: \textit{patterned surface, porous sponges, powder}; and in the last row: \textit{tips}.}
      \vspace{-5mm}
     \label{fig:illustrationpics}
\end{figure}

\vspace{-1mm} 
\paragraph{Datasets:} The primary focus of our research was to automate the identification of nanomaterials using the SEM dataset \cite{aversa2018first}. This benchmark dataset, annotated by human experts, encompasses 10 unique categories that reflect a wide variety of nanomaterials, including \textit{particles, nanowires, and patterned surfaces}. It contains around 21,283 electron micrographs in total. A visual depiction of the nanomaterial categories within the SEM dataset is provided in Figure \ref{fig:illustrationpics}. Although the first experimental findings \cite{modarres2017neural} explored a subset of this dataset, our work leveraged the entire dataset as the subset was not publicly available. The curators of the original dataset \cite{aversa2018first} did not specify predefined splits for training, validation, and test datasets, prompting us to employ a custom approach to evaluate the performance of our framework. This approach enabled a balanced comparison with widely-accepted baseline models in a competitive benchmark scenario.

\vspace{-3mm}
\subsection{Data Prepration : Identifying Hard-to-Classify Micrographs: A Train/Test Approach}
\label{hardtoclassify}
\vspace{-1mm}
The SEM dataset\cite{aversa2018first}, comprises images with original dimensions of $1024\times 768\times 3$ pixels, which were downscaled to $224\times 224\times 3$ pixels to facilitate our analysis. We standardized the images using z-score normalization to ensure a mean of zero and a variance of one, and then flattened the images into 1D-vectors. Subsequently, we employed Principal component analysis (PCA) to reduce the dimensionality of the image data, which involved computing the eigenvectors and eigenvalues of the data covariance matrix. We selected the top-$N$ eigenvectors, where $N$ represents the desired reduced dimensionality, and projected the original data onto the lower-dimensional subspace spanned by these eigenvectors. After dimensionality reduction with PCA, we applied K-Means clustering to segment the images into distinct groups based on inherent patterns and similarities, enabling a more structured analysis of the electron micrographs to identify and understand underlying structures and variations. For our analysis, we set the initial number of clusters at K=10, in line with the predefined number of nanomaterial categories in the SEM dataset. K-Means clustering iteratively works by randomly initializing centroids, assigning each image to the nearest centroid, recalculating the centroids as the mean of the images in each cluster, and repeating this process until the centroids no longer change significantly. After clustering, the most difficult images to classify can be identified by calculating the distance of each image from its assigned centroid, where larger distances suggest greater classification difficulty. Images in smaller or high-variance clusters may also indicate a more challenging classification task. Additionally, calculating the silhouette score for each image, with lower scores indicating a possible better fit with neighboring clusters, further highlights classification challenges. Evaluating the clustering and pinpointing hard-to-classify images through comparison with available ground truth labels enables a thorough analysis and deeper understanding of the image data. We sampled 10$\%$ of the hard-to-classify images from the SEM dataset to create a fixed test dataset, and used the remaining images as the training dataset. We then evaluated our proposed framework and the baseline algorithms on these datasets. Incorporating hard-to-classify images into the test set is essential for a thorough evaluation of classification algorithms. This approach challenges the algorithms, providing a rigorous assessment that prevents overestimation of performance based on simpler examples. Moreover,

\vspace{-5mm}
\begin{table*}[ht!]
\footnotesize
\centering
\setlength{\tabcolsep}{5pt}
\caption{The table compares our method to baseline algorithms, such as vision-based supervised convolutional neural networks (ConvNets), vision transformers (ViTs), and self-supervised learning (VSL) algorithms.}
\vspace{-2mm}
\label{tab:table2}
\begin{tabular}{c|c|c|c|c|c|c|c}
\toprule
\multicolumn{2}{c|}{\textbf{Algorithms}} & \textbf{Top-1} & \textbf{Top-2} & \textbf{Top-3} & \textbf{Top-5} \\ 
\midrule
\multirow{6}{*}{\rotatebox[origin=c]{90}{\textbf{ConvNets}}} & AlexNet(\cite{krizhevsky2017imagenet}) & 0.528 & 0.551 & 0.700 & 0.827 \\
& DenseNet(\cite{huang2017densely}) & 0.569 & 0.799 & 0.876 & 0.929 \\
& ResNet(\cite{he2016deep}) & 0.485 & 0.729 & 0.847 & 0.897 \\
& VGG(\cite{simonyan2014very}) & 0.538 & 0.669 & 0.756 & 0.808 \\
& GoogleNet(\cite{szegedy2015going}) & 0.609 & 0.883 & 0.922 & 0.969 \\
& SqueezeNet(\cite{iandola2016squeezenet}) & 0.404 & 0.496 & 0.646 & 0.698 \\
\midrule
\multirow{6}{*}{\rotatebox[origin=c]{90}{\textbf{VSL}}} & Barlowtwins\cite{zbontar2021barlow} & 0.148 & 0.218 & 0.305 & 0.410 \\
& SimCLR\cite{chen2020simple} & 0.130 & 0.218 & 0.266 & 0.379 \\
& byol\cite{grill2020bootstrap} & 0.143 & 0.248 & 0.318 & 0.453 \\
& moco\cite{he2020momentum} & 0.169 & 0.201 & 0.280 & 0.472 \\
& nnclr\cite{dwibedi2021little} & 0.158 & 0.278 & 0.331 & 0.563 \\
& simsiam\cite{chen2021exploring} & 0.188 & 0.283 & 0.419 & 0.535 \\
\midrule
\multirow{24}{*}{\rotatebox[origin=c]{90}{\textbf{Vision Transformers(ViTs)}}} & CCT\cite{hassani2021escaping} & 0.570 & 0.802 & 0.906 & 0.981 \\
& CVT\cite{CVT} & 0.577 & 0.802 & 0.867 & 0.930 \\
& ConViT\cite{ConViT} & 0.609 & 0.764 & 0.863 & 0.957 \\
& ConvVT\cite{CVT} & 0.319 & 0.598 & 0.781 & 0.921 \\
& CrossViT\cite{Crossvit} & 0.442 & 0.692 & 0.805 & 0.915 \\
& PVTC\cite{PVT} & 0.596 & 0.812 & 0.856 & 0.964 \\
& SwinT\cite{SwinT} & 0.707 & 0.804 & 0.940 & 0.993 \\
& VanillaViT\cite{dosovitskiy2020image} & 0.655 & 0.870 & 0.891 & 0.970 \\
& Visformer\cite{visformer} & 0.398 & 0.609 & 0.679 & 0.856 \\
& ATS\cite{fayyaz2021ats} & 0.540 & 0.744 & 0.861 & 0.973 \\
& CaiT\cite{CaiT} & 0.657 & 0.799 & 0.974 & 0.989 \\
& DeepViT\cite{Deepvit} & 0.546 & 0.746 & 0.919 & 0.988 \\
& Dino\cite{Dino} & 0.049 & 0.230 & 0.396 & 0.437 \\
& Distillation\cite{Distillation} & 0.533 & 0.751 & 0.885 & 0.955 \\
& LeViT\cite{Levit} & 0.624 & 0.841 & 0.903 & 0.970 \\
& MA\cite{MA} & 0.202 & 0.311 & 0.362 & 0.491 \\
& NesT\cite{Nest} & 0.660 & 0.863 & 0.922 & 0.985 \\
& PatchMerger\cite{PatchMerger} & 0.578 & 0.756 & 0.913 & 0.975 \\
& PiT\cite{PiT} & 0.555 & 0.742 & 0.863 & 0.979 \\
& RegionViT\cite{Regionvit} & 0.606 & 0.827 & 0.883 & 0.948 \\
& SMIM\cite{SMIM} & 0.171 & 0.319 & 0.478 & 0.646 \\
& T2TViT\cite{T2TViT} & 0.749 & 0.918 & 0.978 & 0.992 \\
& ViT-SD\cite{ViT-SD} & 0.597 & 0.802 & 0.940 & 0.973 \\
\midrule
& \texttt{GDL-NMID} & \textbf{0.962} & \textbf{0.973} & \textbf{0.989} & \textbf{0.999} \\
\bottomrule
\end{tabular}
\end{table*}

\vspace{5mm}
it gauges the algorithm's ability to generalize to complex, ambiguous data and helps prevent the model from developing a bias towards easier, more common cases during training.

\vspace{-2mm}
\paragraph{Results:} We conducted a comprehensive evaluation of the efficacy of our proposed framework, comparing it with well-established computer vision standard baseline models. Our method was juxtaposed with both supervised learning approaches, such as Convolutional Neural Networks (ConvNets) and Vision Transformers (ViTs, \cite{philvformer, neelayvformer}), and self-supervised techniques like Vision Contrastive Learning (VCL, \cite{susmelj2020lightly}). The results of this evaluation are summarized in Table \ref{tab:table2}. To maintain an unbiased and thorough evaluation, we ensured uniform experimental settings across all baseline algorithms, utilizing the Top-\text{N} accuracy as the evaluation metric and focusing on values of \text{N} within $\{1, 2, 3, 5\}$. Our proposed framework demonstrates superior performance, with a marginal rise of 28.43$\%$ in the Top-1 accuracy and a slight increment of 0.70$\%$ in the Top-5 accuracy when compared with the second-best baseline algorithm, T2TViT (\cite{T2TViT}). Table \ref{tab:table3} shows the experimental findings contrasting our framework's performance with multiple supervised learning-centric standard models, encompassing different Graph Neural Networks (GNNs) variants (\cite{rozemberczki2021pytorch, Fey/Lenssen/2019}). For further comparison, we incorporated Graph Contrastive Learning (GCL, \cite{Zhu:2021tu}) methods. Our proposed framework achieves state-of-the-art results on the benchmark dataset \cite{aversa2018first} when contrasted with the baselines. 

\vspace{-4mm}
\section{Conclusion}
\vspace{-1mm}
In this work, we introduce an autonomous framework that innovatively applies advanced generative AI for identifying nanomaterials in electron micrographs. Our framework synergizes the sophisticated capabilities of large multimodal models like GPT-4V with the generative prowess of text-to-image models such as DALL·E 3 to substantially enhance nanomaterial classification accuracy. It employs GPT-4V’s Visual Question Answering (VQA) for in-depth analysis of nanomaterial images, utilizes DALL·E 3 for creating synthetic images from question-and-answer pairs generated by GPT-4V, and leverages few-shot prompting of GPT-4V's for in-context learning, enabling more efficient classification. The method marks a significant advance over conventional techniques, offering a streamlined process for high-throughput screening within the semiconductor industry.

\bibliography{aaai24}

\newpage
\pagebreak

\vspace{-2mm}
\begin{figure*}[ht!]
\centering
\resizebox{0.8\linewidth}{!}{ 
\hspace*{5mm}\includegraphics[keepaspectratio,height=4.5cm,trim=0.0cm 7.0cm 0cm 7.0cm,clip]{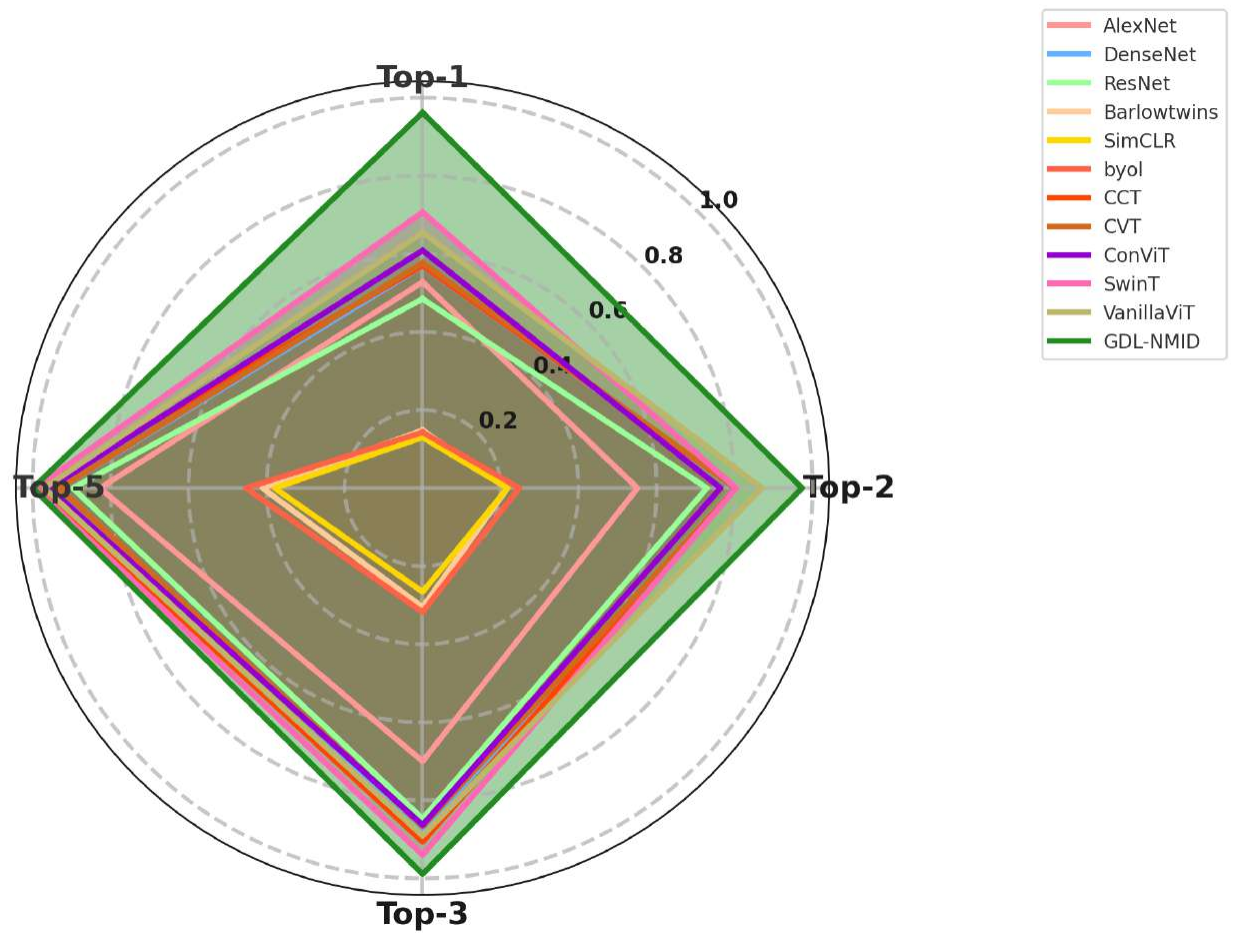} 
}
\vspace{-1mm}
\caption{The figure shows the extended comparison of the proposed framework with vision-based supervised convolutional neural networks (ConvNets), vision transformers (ViTs), and self-supervised learning (VSL) algorithms on the SEM dataset \cite{aversa2018first}.}
\label{fig:figure6}
\vspace{-2mm}
\end{figure*}

\vspace{-1mm}
\section{Technical Appendix}

\begin{table*}[ht!]
\footnotesize
\centering
\setlength{\tabcolsep}{4pt}
\caption{The table compares our proposed method to supervised-learning based GNNs and self-supervised graph contrastive learning (GCL) algorithms on the SEM dataset \cite{aversa2018first}.}
\label{tab:table3}
\vspace{-2mm}
\begin{tabular}{cc|c|c|c|c|c}
\hline
\multicolumn{2}{c|}{\textbf{Algorithms}} & \textbf{Top-1} & \textbf{Top-2} & \textbf{Top-3} & \textbf{Top-5}  \\ \hline
\multicolumn{1}{c|}{\multirow{4}{*}{\rotatebox[origin=c]{90}{\textbf{GCL}}}} & GBT\cite{bielak2021graph} & 0.547 & 0.577 & 0.646 & 0.706 \\
\multicolumn{1}{c|}{} & GRACE\cite{zhu2020deep} & 0.598 & 0.617 & 0.680 & 0.750 \\
\multicolumn{1}{c|}{} & BGRL\cite{thakoor2021bootstrapped} & 0.556 & 0.605 & 0.649 & 0.696 \\
\multicolumn{1}{c|}{} & InfoGraph\cite{sun2019infograph} & 0.526 & 0.601 & 0.651 & 0.702 \\
\hline
\multicolumn{1}{c|}{\multirow{15}{*}{\rotatebox[origin=c]{90}{\textbf{Graph Neural Networks}}}} & APPNP\cite{klicpera2018predict} & 0.632 & 0.699 & 0.742 & 0.786 \\
\multicolumn{1}{c|}{} & AGNN\cite{thekumparampil2018attention} & 0.538 & 0.760 & 0.819 & 0.894 \\
\multicolumn{1}{c|}{} & ARMA\cite{bianchi2021graph} & 0.582 & 0.800 & 0.907 & 0.987 \\
\multicolumn{1}{c|}{} & DNA\cite{fey2019just} & 0.622 & 0.634 & 0.853 & 0.916 \\
\multicolumn{1}{c|}{} & GAT\cite{velivckovic2017graph} & 0.491 & 0.761 & 0.849 & 0.985 \\
\multicolumn{1}{c|}{} & GGConv\cite{li2015gated} & 0.563 & 0.834 & 0.907 & 0.992 \\
\multicolumn{1}{c|}{} & GraphConv\cite{morris2019weisfeiler} & 0.658 & 0.822 & 0.924 & 0.996 \\
\multicolumn{1}{c|}{} & GCN2Conv\cite{chen} & 0.732 & 0.869 & 0.929 & 0.998 \\
\multicolumn{1}{c|}{} & ChebConv\cite{defferrard2016convolutional} & 0.504 & 0.805 & 0.875 & 0.951 \\ 
\multicolumn{1}{c|}{} & GraphConv\cite{morris2019weisfeiler} & 0.509 & 0.694 & 0.895 & 0.993 \\
\multicolumn{1}{c|}{} & GraphUNet\cite{gao2019graph} & 0.657 & 0.680 & 0.930 & 0.978 \\
\multicolumn{1}{c|}{} & MPNN\cite{gilmer2017neural} & 0.603 & 0.822 & 0.939 & 0.999 \\
\multicolumn{1}{c|}{} & RGGConv\cite{bresson2017residual} & 0.618 & 0.692 & 0.951 & 0.961 \\
\multicolumn{1}{c|}{} & SuperGAT\cite{kim2022find} & 0.598 & 0.627 & 0.920 & 0.985 \\
\multicolumn{1}{c|}{} & TAGConv\cite{du2017topology} & 0.598 & 0.718 & 0.841 & 0.999 \\
\hline
\multicolumn{1}{c|}{} & \texttt{GDL-NMID} & \textbf{0.962} & \textbf{0.973} & \textbf{0.989} & \textbf{0.999} \\  \bottomrule
\end{tabular}
\vspace{-2mm}
\end{table*}

\vspace{-2mm}
\begin{figure*}[ht!]
\centering
\resizebox{0.8\linewidth}{!}{ 
\hspace*{5mm}\includegraphics[keepaspectratio,height=4.5cm,trim=0.0cm 7.0cm 0cm 7.0cm,clip]{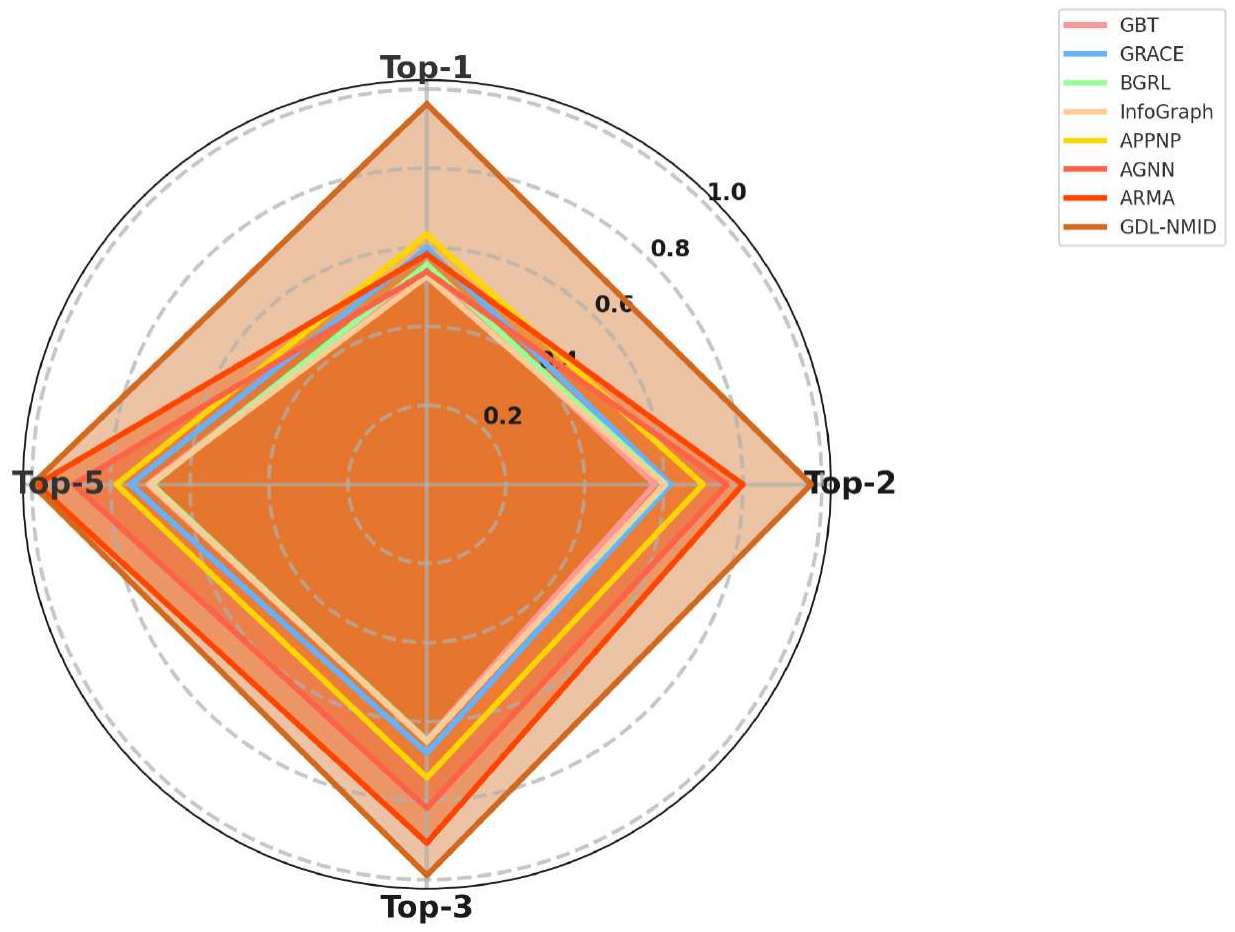} 
}
\vspace{-1mm}
\caption{The figure shows the extended comparison of the proposed framework with supervised-learning based GNNs and self-supervised graph contrastive learning (GCL) algorithms on the SEM dataset \cite{aversa2018first}.}
\label{fig:figure5}
\vspace{-3mm}
\end{figure*}

\vspace{3mm}
\subsection{Experimental Setup}
\vspace{0mm}
In our experiments, we specifically designed an electron micrograph encoder to process electron micrographs and generate a comprehensive image representation. The ultimate goal is to leverage this encoder for few-shot prompting of Large Multimodal Models (LMMs), such as GPT-4V, to identify the nanomaterial category for a given query image. In this few-shot prompting approach, the encoder computes image embeddings and then identifies a select number of analogous or identical images from the training set, relevant to the query image, through a similarity learning technique. By presenting these selected demonstrations (sampled image-label pairs) to the LMMs, they can effectively predict the nanomaterial category of the query image, even with minimal demonstrations. Unsupervised image representation learning is essential in this context for several reasons. First, it provides the foundation for few-shot prompting with LMMs like GPT-4V, enabling the electron micrograph encoder to capture comprehensive image representations that are critical for effectively identifying relevant demonstrations (input-output pairs). Unsupervised learning may lead to more generalized image representations since the encoder, not limited by predefined labels, can capture a wider range of features potentially relevant to the identification of nanomaterials—features that supervised training sets might not include. Moreover, the encoder's ability to identify similar images affords a nuanced understanding of the data, uncovering relationships and structures within the electron micrographs that could elude human observers or be too complex for supervised models to discern without extensive labeled data. In essence, this approach is a calculated strategy that utilizes the abundance of data to set the stage for proficient few-shot prompting with LMMs. We describe the training of the electron micrograph encoder in unsupervised learning settings as follows: We utilized the SEM dataset\cite{aversa2018first}, which is a compilation of electron micrographs of various nanomaterials with dimensions of $1024 \times 768 \times 3$ pixels. For our analysis, we resized these images to $224 \times 224 \times 3$ pixels and standardized them to maintain a constant mean and covariance of 0.5 across channels. This data preprocessing ensures that image values span between -1 and 1. Subsequently, we split the downsized images into distinct patches, representing the micrographs as patch sequences. We obtained patch sequences with a resolution of 32 pixels each. The patch dimension ($d_\text{pos}$) and the position embedding dimension ($d$) were both set to 128. The encoder was trained for 50 epochs with an initial learning rate of $1 \times 10^{-3}$ and a batch size of 48. Additionally, we configured a few hyperparameters for the attention layer: the number of attention heads (H) was set to 4, and the dimensionality of Key/Query/Value ($d_{h}$) was set to 32. To enhance the performance of the electron micrograph encoder, we employed two key strategies: (a) early stopping on the validation set, which halts training when the encoder's performance on the validation data plateaus, thereby preventing overfitting; and (b) a learning rate scheduler that systematically reduces the learning rate by half if the validation loss does not improve for five consecutive epochs. This reduction in the learning rate aids the encoder in converging to a better solution and mitigates overfitting. Moreover, we utilized the Adam optimization algorithm \cite{kingma2014adam} to update the encoder's trainable parameters.
Training the electron micrograph encoder for unsupervised image representation learning involves optimizing a similarity measure between the representations of different views of the same image while minimizing similarity between views of different images. The Normalized Temperature-Scaled Cross Entropy Loss (NT-Xent Loss\cite{sohn2016multiclass, chen2020simple}) is a commonly employed loss function for this task. Given a batch of images, we first generate two augmented views of each image. The micrograph encoder is then used to obtain representations \( h^{k}_{\textit{cls}} \) and \( h^{k^+}_{\textit{cls}} \) of the two views of each image. The NT-Xent loss is defined as follows:

\vspace{0mm}
\resizebox{0.925\linewidth}{!}{
\hspace{0mm}\begin{minipage}{\linewidth}
\centering
\begin{equation}
\mathcal{L}_{\text{NT-Xent}} = -\frac{1}{2N} \sum_{k=1}^{2N} \log \left( \frac{\exp(\text{sim}(h^{k}_{\textit{cls}}, h^{k^+}_{\textit{cls}}) / \tau)}{\sum_{l=1, l\neq k, l\neq k^+}^{2N} \exp(\text{sim}(h^{k}_{\textit{cls}}, h^{l}_{\textit{cls}}) / \tau)} \right) \nonumber
\end{equation}
\end{minipage}
} 

\begin{figure}[ht!]
\centering
\resizebox{1.25\linewidth}{!}{ 
\hspace*{-6mm}\includegraphics[keepaspectratio,height=4.5cm,trim=0.0cm 0.0cm 0cm 0.0cm,clip]{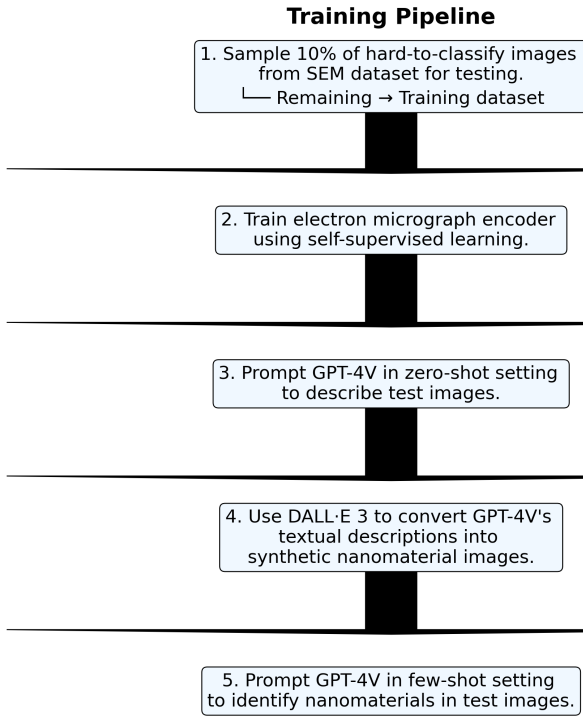} 
}
\vspace{-5mm}
\caption{A visual representation of the training pipeline details the step-by-step process, which begins by sampling hard-to-classify images from the SEM dataset and ends with employing few-shot prompting to instruct GPT-4V for nanomaterial identification. The pipeline illustrates the integration of a self-supervised learning micrograph encoder, the zero-shot prompting of GPT-4V, and the utilization of DALL·E 3 to generate synthetic images from textual descriptions. Note: The micrograph encoder model parameters are updated through unsupervised learning on the training dataset.}
\label{fig:figure3}
\vspace{-3mm}
\end{figure}

\vspace{2mm}
where \( N \) is the number of images in the batch, \( \text{sim}(z_k, z_l) = \frac{z_k^\top z_l}{\|z_k\| \|z_l\|} \) is the cosine similarity between representations \( z_k \) and \( z_l \), \( k^+ \) is the index of the positive pair for \( z_k \), and \( \tau \) is the temperature parameter.  The objective is to minimize \( \mathcal{L}_{\text{NT-Xent}} \) with respect to the parameters of the micrograph encoder, typically using gradient-based optimization algorithms to learn a representation space where similar images are mapped close together and dissimilar images are mapped far apart, thus maximizing similarity between like images. Once the micrograph encoder has been trained to represent images, it can be used to sample related images from the entire training dataset for few-shot prompting of GPT-4V. This is achieved by using the unsupervised image embeddings computed by the micrograph encoder to determine the similarity between different images. The images most similar to a given  query image are then selected. The corresponding image-label pairs (demonstrations), along with the task-specific instruction to predict the nanomaterial category of the query image, are provided to GPT-4V, which then outputs the predicted nanomaterial category. The experiments were carefully designed to demonstrate the effectiveness of the proposed fusion framework, Generative Deep Learning for Nanomaterial Identification (\texttt{GDL-NMID}) leveraging the strengths of both GPT-4V and DALL·E 3, in comparison to the baselines. Note: API access for GPT-4V and DALL·E 3 has been restricted from public use but may become accessible starting in mid-November 2023. ChatGPT Plus subscribers can access GPT-4V and DALL·E 3 through the OpenAI ChatGPT web interface. To optimize computational resource usage, the system is trained on two V100 GPUs, each equipped with 8 GB of GPU memory, utilizing the PyTorch framework. This configuration ensures that the training process is completed within a reasonable timeframe. We conducted two individual experiments and reported the averaged results. Figure \ref{fig:figure3} illustrates the end-to-end pipeline of the framework. In our work, we explore Large Multimodal Models (LLMs) such as GPT-4V, which can process both input text and images to generate text responses, and text-to-image diffusion generative models like DALL·E 3. These large-scale general-purpose models build upon the capabilities of Large Language Models (LLMs) like GPT-4 (text only), integrating language understanding with visual data interpretation. While GPT-4V exhibits impressive skills, such as describing image contents in detail, and DALL·E 3 generates high-quality synthetic images from textual descriptions, they sometimes misinterpret images or textual descriptions. This challenge is known as `hallucination', and it is a recognized issue in the current development of multi-purpose large-scale models. In our work, we manually discard both the textual descriptions generated by GPT-4V and the corresponding synthetic images generated by DALL·E 3 from these textual descriptions if they are misaligned with the ground-truth image.

\vspace{-2mm}
\subsection{A Multi-Metric Evaluation of Framework Performance in the Classification of Nanomaterials Using Electron Micrographs}
\begin{figure*}[ht!]
\centering
\resizebox{0.75\linewidth}{!}{ 
\hspace*{-0mm}\includegraphics[keepaspectratio,height=4.5cm,trim=0.0cm 0.0cm 0cm 0.0cm,clip]{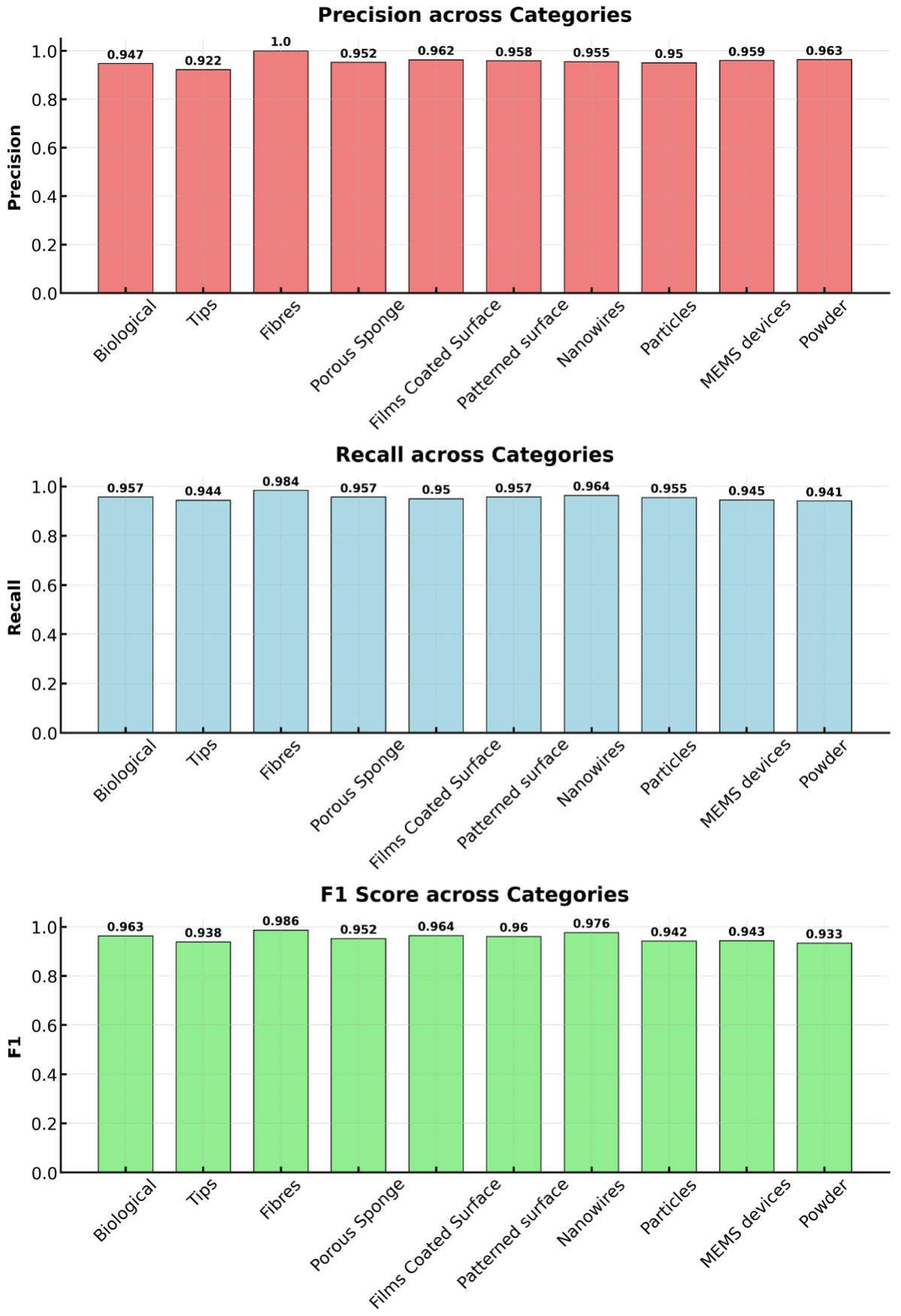} 
}
\vspace{-3mm}
\caption{The figure shows the bar charts displaying the evaluation metrics for nanomaterial categorization through electron micrographs within the SEM dataset. The top chart represents Precision values across categories, the middle chart displays Recall values, and the bottom chart showcases the F1 score values. Each category is represented by distinct bars, and their respective values are labeled on top of each bar.}
\label{fig:figure4}
\vspace{-3mm}
\end{figure*}

\vspace{-1mm}
We conducted systematic experimentation to evaluate the capabilities of our proposed framework in classifying electron micrographs of diverse nanomaterials, spanning from elementary to sophisticated patterns. Nanomaterials exhibit a wide spectrum of patterns due to variations in attributes such as composition, morphology, and crystalline nature. Consequently, electron micrographs offer invaluable insights into the inherent characteristics of these nanomaterials, making their precise classification essential for applications in materials science. In the classification of nanomaterials using electron micrographs, several critical metrics gauge the accuracy and precision of the framework. We employ a detailed multi-metric evaluation to compare the performance of our framework with baseline models, with a primary focus on classifying electron micrographs across various nanomaterial categories.
The evaluation focuses on a confusion matrix that captures key metrics: True Positives (TP) represent correctly classified micrographs for a specific category; False Negatives (FN) are cases where micrographs belonging to a specific category were incorrectly overlooked or misclassified. True Negatives (TN) indicate accurate identifications of micrographs that do not belong to a particular category, whereas False Positives (FP) represent micrographs that have been incorrectly assigned to a category. Precision (TP/(TP + FP)) evaluates the proportion of correctly classified micrographs among all those classified for a category, with an emphasis on minimizing false positives. Recall (TP/(TP + FN)) measures how effectively the framework identifies actual micrographs of a category, prioritizing the reduction of false negatives. The F1-score seamlessly combines precision and recall into a unified metric, offering a comprehensive assessment of the framework's performance in classifying electron micrographs across nanomaterial categories. In the intricate domain of nanomaterial identification via electron micrographs, these metrics are indispensable tools, enabling a comprehensive and nuanced evaluation of the effectiveness and reliability of the classification framework. Our results, highlighted in Figure \ref{fig:figure4}, show the bar chart overview of the metrics for different nanomaterial categories and validate the framework's robustness using multiple metrics on the SEM dataset\cite{aversa2018first}. Incorporating these metrics into our analysis provides deeper insight into our model's effectiveness in categorizing electron micrographs across diverse nanomaterial categories. It's important to note that the SEM dataset exhibits significant class imbalance. Notably, our framework demonstrates higher classification scores for nanomaterial categories with a substantial number of labeled instances, outperforming those with fewer instances. This remarkable success in classifying categories with fewer labeled instances can be attributed to our proposed framework's reduced reliance on nanomaterial-specific relational inductive biases, setting it apart from conventional methods. In summary, our extended experiments have significantly bolstered our confidence in the framework's ability to generalize and accurately categorize various nanomaterials using electron micrographs. We anticipate that these advancements will have a substantial impact on the broader scientific community, facilitating the acceleration of materials characterization and related research. 

\vspace{-2mm}

\vspace{-3mm}
\subsection{Baseline Algorithms}
\vspace{0mm}
Our baseline methods are categorized into four primary groups. First, we leverage Graph Neural Networks (GNNs) for supervised multi-class classification of vision graphs \cite{rozemberczki2021pytorch, Fey/Lenssen/2019}. In this approach, we construct vision graphs from electron micrographs by employing the Top-K nearest neighbor search method. Here, patches are used as nodes, and edges connect semantically similar neighboring nodes. We opt for a 32-pixel patch size and set K to 5 for simplicity, thus avoiding the complexity of multi-scale vision graphs with varying patch resolutions. Second, we employ Graph Contrastive Learning (GCL) techniques \cite{Zhu:2021tu}, designed to generate multiple correlated graph views for graph data augmentation and then learn representations consistent across these views. These methods using diverse contrastive loss functions, aim to maximize the similarity between positive pairs and minimize it between negative pairs from different graphs. Typically, the Graph Attention Network (GAT) \cite{velivckovic2017graph} serves as a node-level graph encoder to compute unsupervised node embeddings. Graph-level embeddings are then obtained by sum-pooling of node-level embeddings. During inference, Random Forest (RF) algorithms use these unsupervised graph-level embeddings to predict nanomaterial categories. We evaluate the effectiveness of these unsupervised embeddings based on the RF model's accuracy with holdout data. Third, for the for supervised classification of electron micrographs, we use Convolutional Neural Networks (ConvNets) \cite{philvformer, neelayvformer} to operate on electron micrographs grids and also utilize Vision Transformers (ViTs) \cite{philvformer, neelayvformer} by evaluating patch sequences within electron micrographs for nanomaterial identification. In addition, Vision Contrastive Learning (VCL) techniques \cite{susmelj2020lightly} are applied for self-supervised learning in computer vision, utilizing the ResNet architecture for feature extraction.

\vspace{-1mm}
\subsection{Background}
\vspace{0mm}
Text-to-image generation models are technologies that create visual representations from textual descriptions. In the field of artificial intelligence and deep learning, numerous open-source models have emerged, translating text into images. The evolution of text-to-image models has been rapid. Early models like the Generative Adversarial Network (GAN) \cite{goodfellow2020generative} laid the foundation. Subsequent models such as BigGAN \cite{brock2018large} improved resolution and fidelity. DALL-E \cite{ramesh2021zero}, introduced by OpenAI, showcased remarkable capability in generating complex images from simple textual prompts. Its successor, DALL-E 2 \cite{ramesh2022hierarchical}, highlights the ongoing developments in this field. More recently, DALL-E 3, unveiled by OpenAI \cite{dalle3}, is an advanced text-to-image generation model that translates nuanced requests into highly detailed and accurate images. Integrated with the AI chatbot ChatGPT, it allows users to refine image prompts interactively. Another significant model is Stable Diffusion \cite{rombach2022high, podell2023sdxl}, an open-source AI-based image generation model that can generate detailed and coherent images from textual descriptions. It is utilized in popular applications such as Wombo\footnote{For more information, refer to \url{https://www.wombo.ai/}.} and Lensa\footnote{For more information, refer to \url{https://prisma-ai.com/lensa}.}. The model operates by gradually transforming a pattern of random noise into an image that aligns with the provided text prompt. Furthermore, the Grounded-Language-to-Image Generation (GLIGEN) \cite{li2023gligen} model proposes an extended approach to traditional text-to-image diffusion models by allowing them to use additional grounding inputs such as bounding boxes and reference images. This approach improves image realism and controllability by combining these inputs with pre-trained model knowledge to generate more accurate and contextually appropriate images. Google's text-to-image neural network, Imagen \cite{saharia2022photorealistic}, generates high-quality images by understanding and interpreting text inputs with a high degree of fidelity. Additionally, the integration of Dreambooth with Stable Diffusion \cite{ruiz2023dreambooth} brings Dreambooth's personalization capabilities into the Stable Diffusion text-to-image model, enabling the creation of custom images that reflect specific subjects or styles from a user's text descriptions. These developments collectively demonstrate collaborative advancements in text-to-image and text-to-video generation, respectively. On the other hand, SDXL\footnote{For more information, refer to \url{https://docs.sdxl.ai/}.} from Stability AI is touted for its significant improvements over previous diffusion models, such as DALL-E 2 and Imagen, in terms of image quality, diversity, and efficiency, delivering more realistic image generation with improved composition and text interpretation. OpenJourney\footnote{For more information, refer to \url{https://openjourney.art/}.}, a fine-tuned version of the Stable Diffusion XL (SDXL) text-to-image diffusion model, creates AI art in the style known as `Midjourney', crafting images that are reminiscent of the aesthetic associated with Midjourney\footnote{For more information, refer to \url{https://docs.midjourney.com/}.}. Furthermore, Deep Daze, a simple command-line tool for text-to-image generation using OpenAI's CLIP and Siren, enriches the ecosystem of open-source tools for text-to-image synthesis \cite{radford2021learning, sitzmann2020implicit}. DeepFloyd IF\cite{deepfloydIF}, also from Stability AI, is a modular, cascaded pixel diffusion model capable of generating high-resolution images, its design adeptly intertwining realistic visuals with language comprehension. Meanwhile, DreamShaper, another model in this field, elevates photorealism and anime-style generation with its diffusion model architecture, seamlessly aligning images with input text. Additionally, Waifu Diffusion\cite{waifudiffusion}, a descendant of Stable Diffusion, garners acclaim for its ability to generate high-quality anime images from text prompts, even those that are complex or abstract. These open-source models, each boasting distinctive flair and technological underpinnings, are propelling the text-to-image generation domain toward new horizons with applications that sprawl across content creation, data visualization, and beyond. The convergence of language and vision has ushered in a transformative paradigm in artificial intelligence, culminating in the development of Large Multimodal Models (LMMs). State-of-the-art multi-modal language models such as GPT-4(V)ision and LLaVA-1.5 exemplify this advancement, showcasing unprecedented levels of image understanding and reasoning. OpenAI's GPT-4V is a groundbreaking general-purpose LMM capable of processing and interrelating text and image data. It is designed to understand and generate language based on textual and visual contexts. Built on a transformer-based design and fine-tuned with reinforcement learning from human feedback, GPT-4V can handle both text and image inputs. This breakthrough in multimodal learning unlocks a myriad of new possibilities, including generating text descriptions from images, translating images into different languages, or crafting creative content based on visual prompts. Additionally, GPT-4V has been conscientiously developed to be safe and ethical, with significant efforts to mitigate potential misuse or harm. Overall, GPT-4V represents a major milestone at the forefront of multimodal AI chatbots, integrating language and vision capabilities and signifying a major milestone in multimodal learning. In the evolving domain of multimodal learning, several models have emerged as noteworthy counterparts to OpenAI's GPT-4V, fostering the fusion of visual and textual data processing to generate descriptive textual output from image inputs. LLaVA-1.5\cite{liu2023improved, liu2023visual}, which embodies an end-to-end trained large multimodal model, is an auto-regressive language model built on the transformer architecture and was fine-tuned using LLaMA/Vicuna based on GPT-generated multimodal instruction-following data. For its visual understanding capabilities, LLaVA-1.5 uses a CLIP (Contrastive Language–Image Pre-training) model as its visual encoder. LLaVA, although not compared on the same benchmarks as GPT-4, shows promising results in understanding visual content and responding to queries, performing well even on out-of-domain images. However, in certain aspects of detailed analysis, GPT-4V may demonstrate superior performance compared to LLaVA. On a similar trajectory, Alibaba Cloud's Qwen-VL\cite{bai2023qwen} aims to harmonize vision and language processing, albeit with fewer documented specifics regarding its capabilities. Lastly, the Google PaLI-X model\cite{chen2023pali} enhances the synergy between vision and language processing by scaling up both the component size and the training task mixture, achieving improved performance across a broad spectrum of tasks such as image-based captioning, question answering, and object detection. The advent of these models underscores the burgeoning exploration and achievements in multimodal learning, delineating a promising trajectory for more intuitive and capable AI applications. However, LMMs are susceptible to vulnerabilities such as language hallucination and visual illusion, caused by the imbalance between their language and vision modules. Language hallucination leads LMMs to generate text descriptions for images that do not exist, while visual illusion results in erroneous visual interpretations. There is a need for new methods to address these challenges, such as developing more robust vision modules and new training methods that explicitly teach LMMs to avoid these pitfalls. Overall, the convergence of language and vision is a promising new direction in artificial intelligence, with LMMs having the potential to revolutionize our interactions with intelligent machines and the world around us.

\vspace{-2mm}

\clearpage
\onecolumn

\begin{tcolorbox}[colback=white!2!white,colframe=black!75!black]
\vspace{-5mm}
\begin{tabularx}{1.2\textwidth}{bss}
\caption{The table presents the question-answer pairs generated by the LMM(GPT-4V) in response to natural language questions about the input image of the biological nanomaterial category. In summary, the question-answer pairs (generated text) provide in-depth information about an image related to biological nanomaterials.}\label{tab:tab1}\\ [-0.75em]
\toprule 
 \\ [-0.95em]
\begin{overpic}[width=0.155\columnwidth]{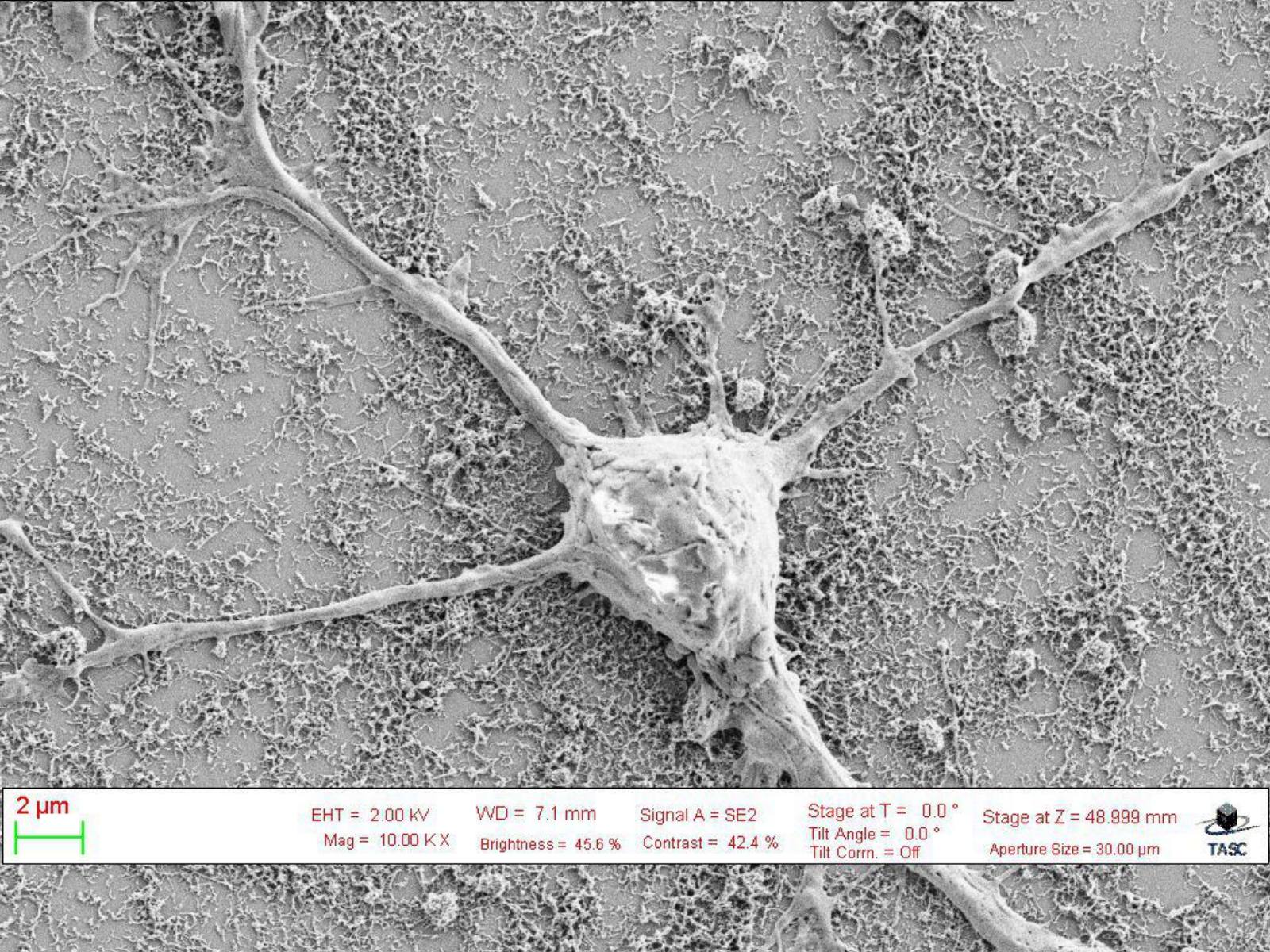}
  \put(400,35){\makebox[0pt][c]{%
    \parbox{0.4\columnwidth}{%
      \centering
      \textbf{Please answer the following questions based on the provided input image belonging to the biological nanomaterial category.} 
    }%
  }}  
\end{overpic}  \\ [-0.25em]

\midrule
\textbf{Query prompt}  & \textbf{GPT-4 Vision output} \\
\midrule
\endhead
\textcolor{black}{**Basics**} & The image depicts a biological nanomaterial, possibly a cellular structure or related biological entities. The scale of the image is 2 micrometers (µm) as indicated by the scale bar.
\\

\midrule
\textcolor{black}{**Morphology and Structure**} & The nanomaterials in the image display a dendritic or branched morphology. There are no distinct layers or phases, but there are branching structures and interconnected networks. The nanomaterials appear varied in size, with some structures appearing larger and more prominent than others.  
\\

\midrule
\textcolor{black}{**Size and Distribution**} & The size of individual nanostructures varies, but most structures seem to be within the sub-micrometer range, given the 2 µm scale bar. The nanomaterials are distributed in a web-like or interconnected manner throughout the image. There's evidence of branching or interconnection, but not necessarily aggregation or bundling.
\\

\midrule
\textcolor{black}{**Surface Characteristics**} & The surface of the nanomaterial appears rough and textured. There are no clear visible defects, but there are intricate details suggesting complex surface features.
\\

\midrule
\textcolor{black}{**Composition and Elements**} & The grayscale image doesn't indicate compositional variations based on color, but brightness contrasts suggest variations in material density or composition. No specific labels or markers indicating elements or compounds are visible.
\\
 
\midrule
\textcolor{black}{**Interactions and Boundaries**} & The individual nanostructures appear interconnected, forming a network. There are clear boundaries between the larger structures, but the smaller branching structures are closely intertwined.
\\
 
\midrule
\textcolor{black}{**External Environment**} & There doesn't appear to be an interaction between the nanomaterial and a surrounding matrix or solvent in the image. Most structures in the image seem to be related to the biological nanomaterial. The background texture might be a substrate or another type of material on which these structures are grown or attached.
\\
 
\midrule
\textcolor{black}{**Image Technique and Modifications**} & The imaging technique used is Scanning Electron Microscopy (SEM) as indicated by the parameters given (like EHT, Signal A, and WD). There is no clear evidence of post-processing or modifications from the provided image.
\\
 
\midrule
\textcolor{black}{**Functional Features**} & It's challenging to identify specific functional features from this image without additional context. However, the branched structures might have distinct properties related to their morphology. The image appears to be a static representation.
\\

\midrule
\textcolor{black}{**Context and Application**} & The intended application or use of the depicted nanomaterial isn't provided in the image or accompanying information. This appears to be a real, experimental sample given the details and the parameters provided in the SEM image.
\\ [-0.05em]

\bottomrule
\end{tabularx}
\end{tcolorbox}

\vspace{-7mm}

\begin{tcolorbox}[colback=white!2!white,colframe=black!75!black]
\vspace{-5mm}
\begin{tabularx}{1.2\textwidth}{bsssss}
\\ [-1.05em]
\includegraphics[width=0.14\textwidth]{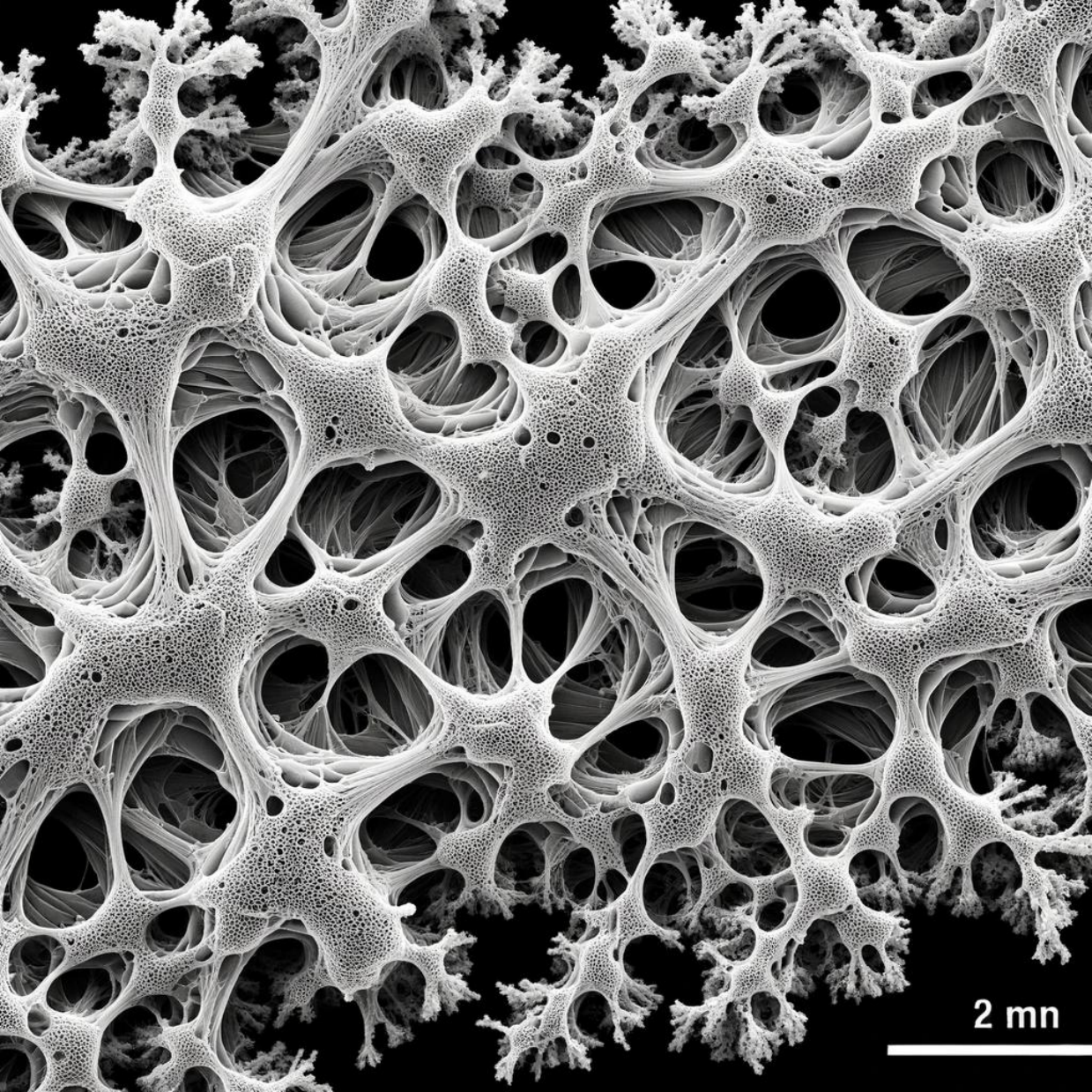} & 
\includegraphics[width=0.14\textwidth]{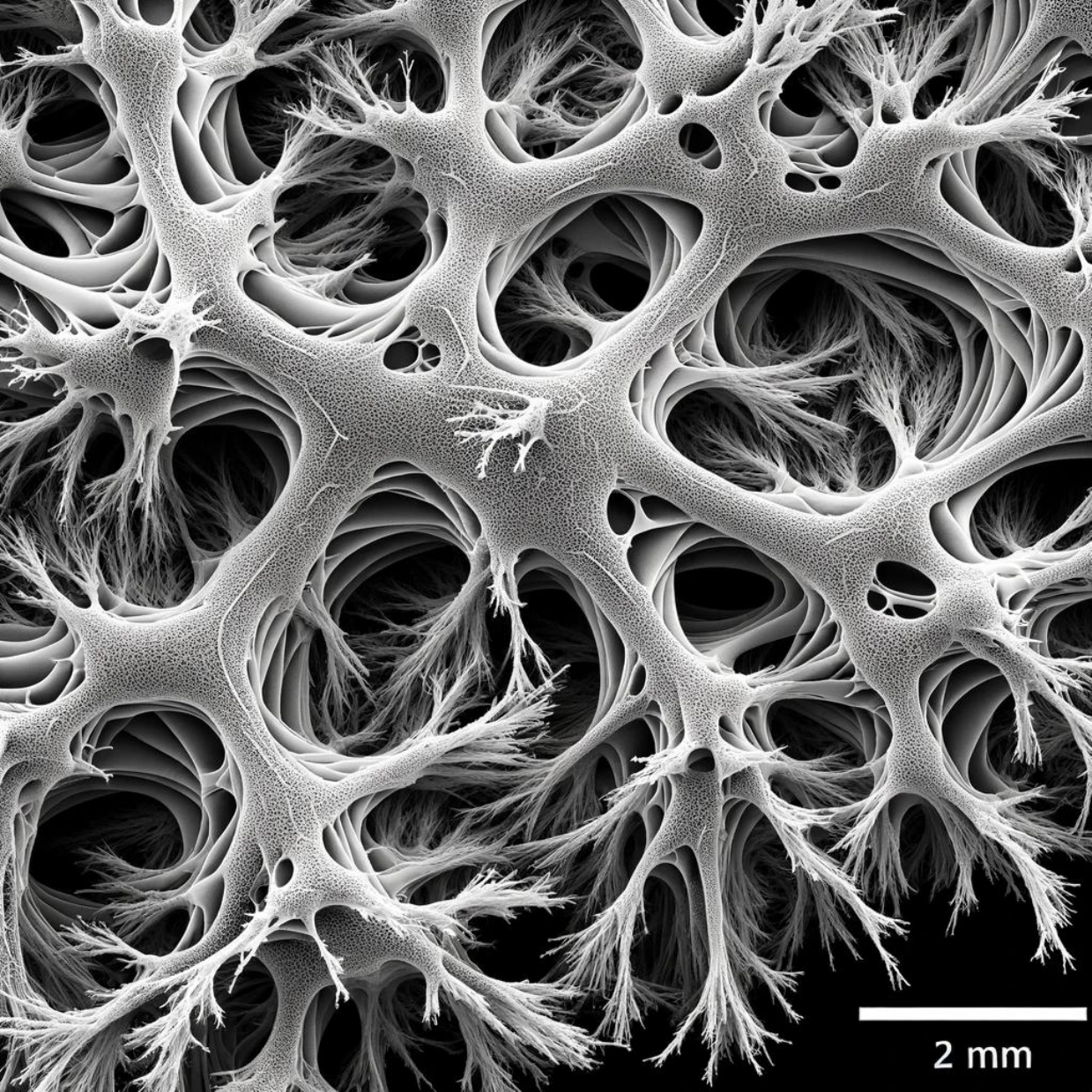} & 
\includegraphics[width=0.14\textwidth]{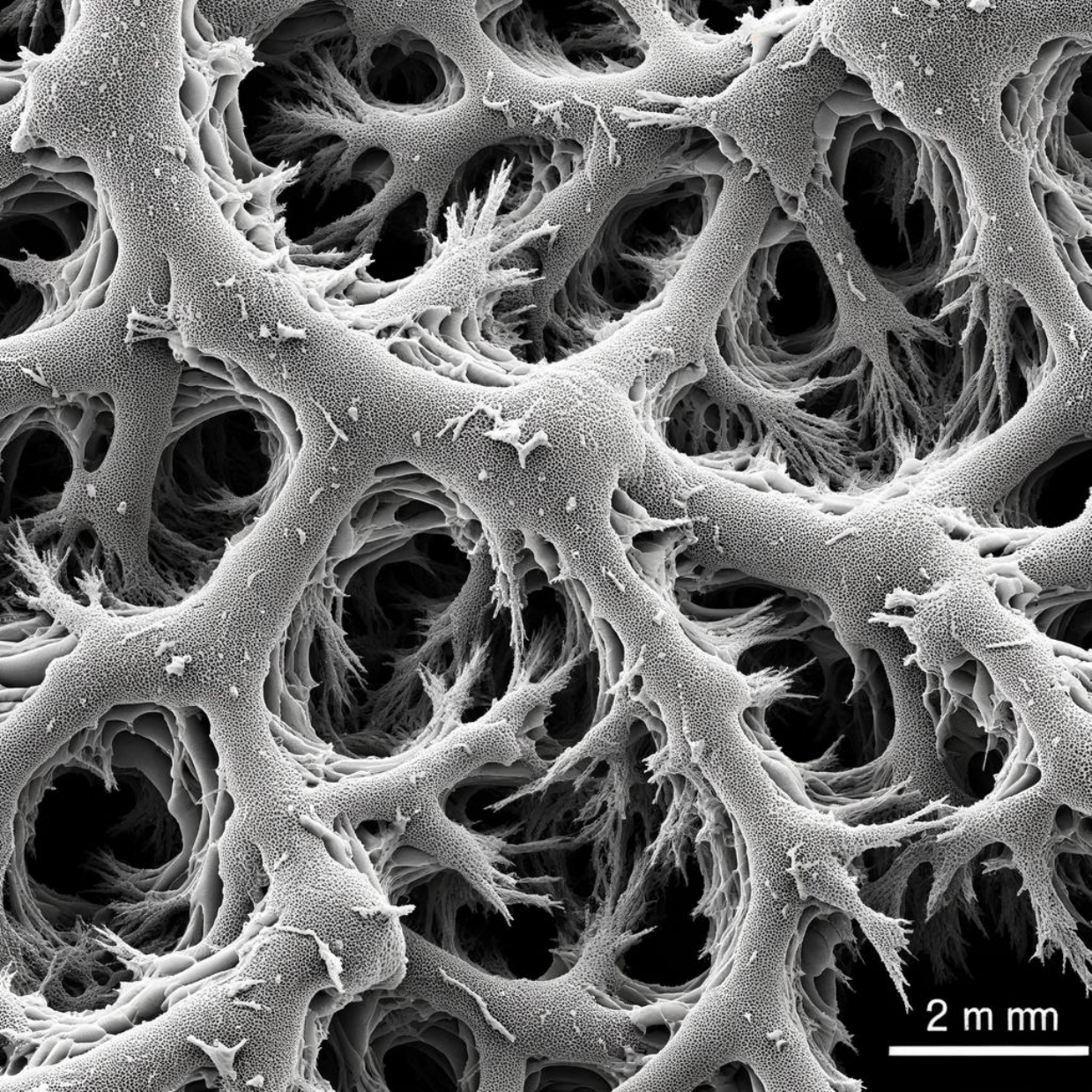} &
\includegraphics[width=0.14\textwidth]{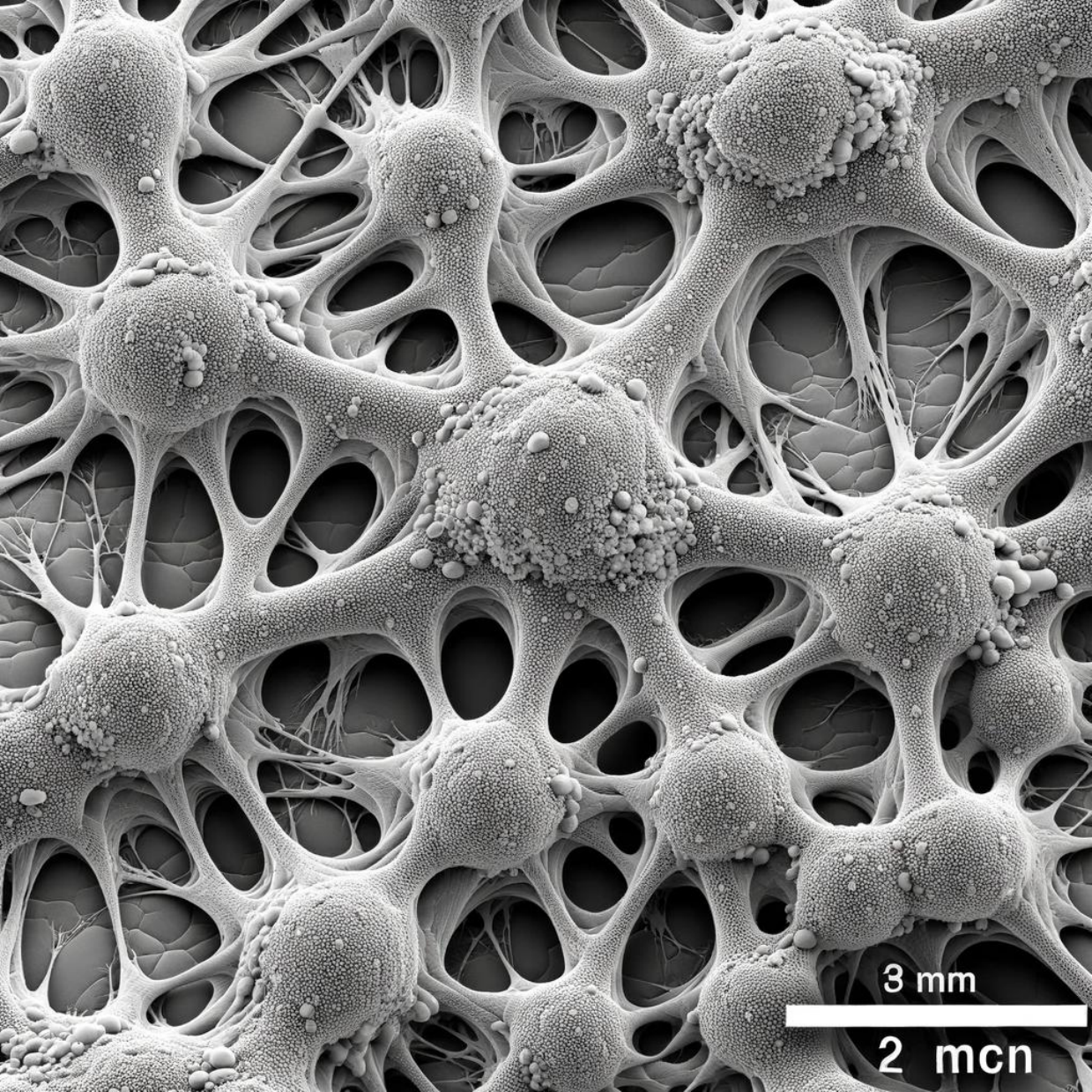} & 
\includegraphics[width=0.14\textwidth]{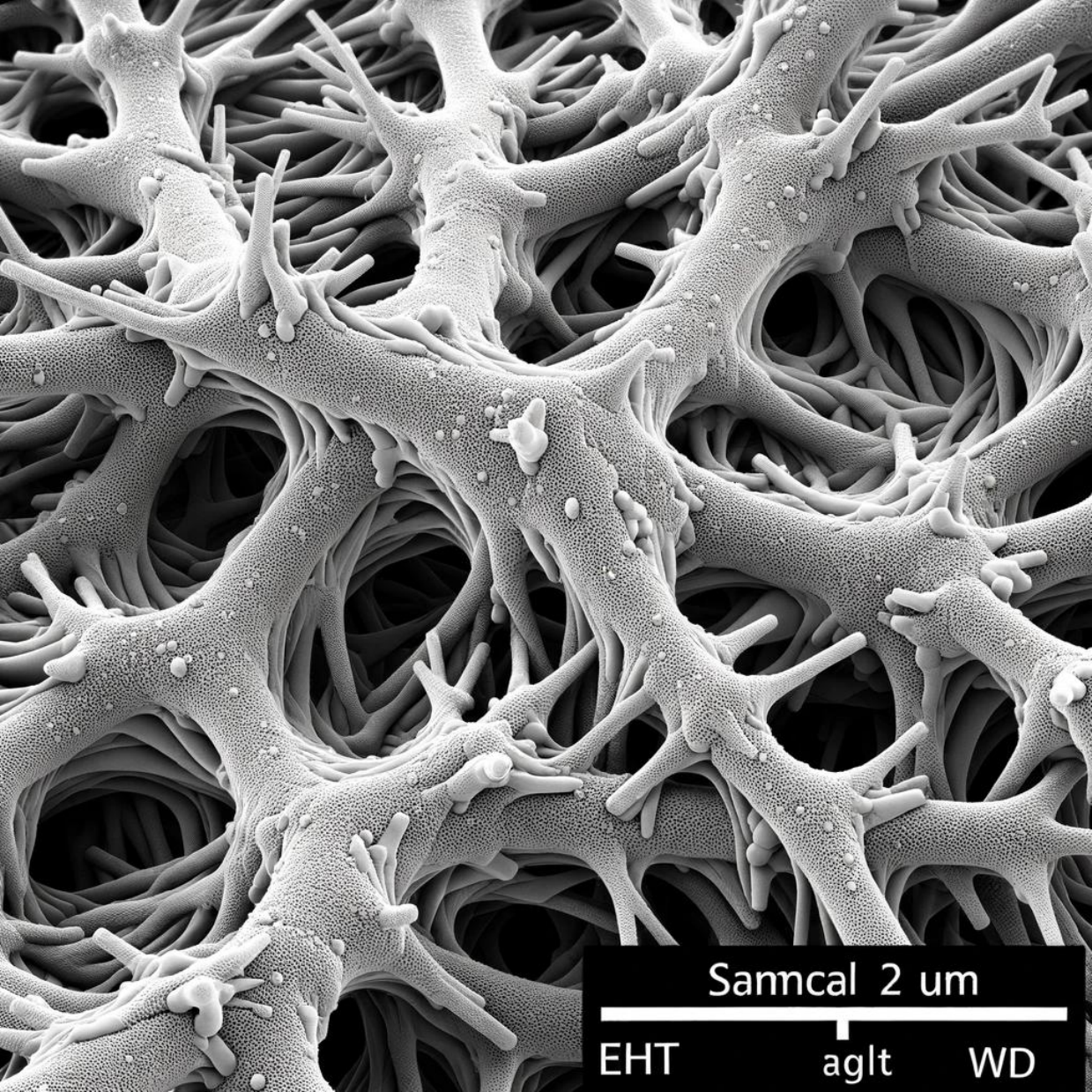} & 
\includegraphics[width=0.14\textwidth]{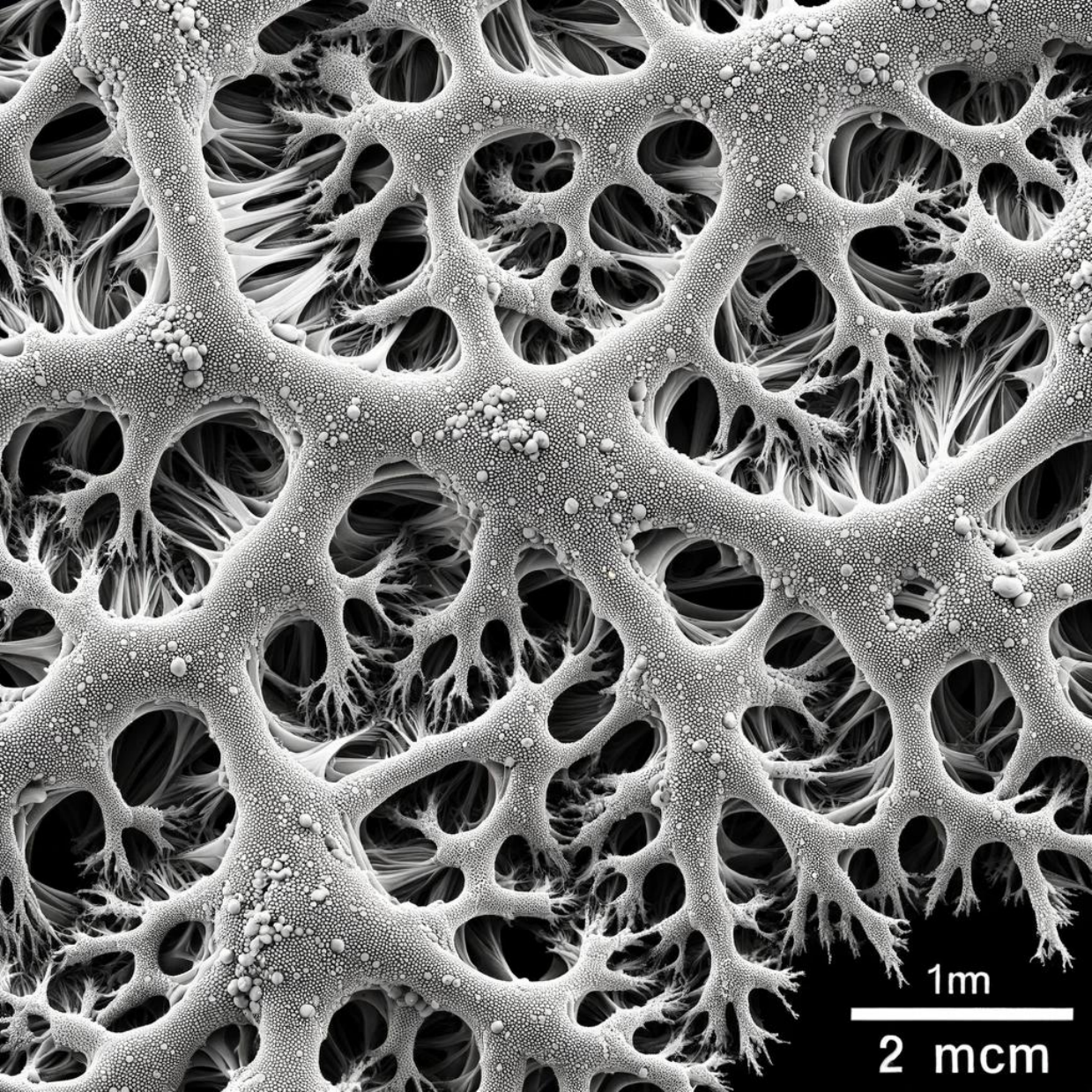} \\ [-0.95em] 
\caption{The table shows sample synthetic images generated by DALLE-3 from textual descriptions provided in a series of question-answer pairs, as shown in Table \ref{tab:tab1}.}
\label{tab:tab2}
\end{tabularx}
\vspace{-3mm}
\end{tcolorbox}

\begin{tcolorbox}[colback=white!2!white,colframe=black!75!black]
\vspace{-5mm}
\begin{tabularx}{1.2\textwidth}{bss}
\caption{The table presents the question-answer pairs generated by the LMM(GPT-4V) in response to natural language questions about the input image of the fibres nanomaterial category. In summary, the question-answer pairs (generated text) provide in-depth information about an image related to fibre nanomaterials.}\label{tab:tab3}\\ [-0.75em]
\toprule 
 \\ [-0.75em]
\begin{overpic}[width=0.135\columnwidth]{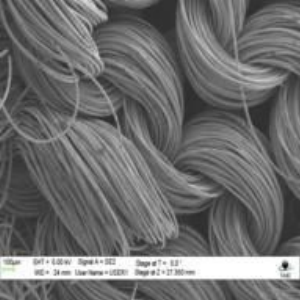}
  \put(475,50){\makebox[0pt][c]{%
    \parbox{0.4\columnwidth}{%
      \centering
      \textbf{Please answer the following questions based on the provided input image belonging to the fibres nanomaterial category.} 
    }%
  }}  
\end{overpic}  \\ [-0.25em]

\midrule
\textbf{Query prompt}  & \textbf{GPT-4 Vision output} \\
\midrule
\endhead
\textcolor{black}{**Basics**} & The image depicts fibrous nanomaterials. The scale of the image is 100µm, meaning each unit of measurement in the scale bar represents 100 micrometers.
\\

\midrule
\textcolor{black}{**Morphology and Structure**} & The nanomaterials in the image have a fibrous or filamentous morphology. There do not appear to be distinct layers or phases, but the fibers are layered upon each other. The fibers seem mostly uniform in terms of their width, but their lengths and curvatures vary.  
\\

\midrule
\textcolor{black}{**Size and Distribution**} & The individual nanostructures, or fibers, appear to be a few micrometers in width, but their lengths are much longer and harder to determine from the image. The nanomaterials are densely packed and intertwined. There is evident bundling of the fibers.
\\

\midrule
\textcolor{black}{**Surface Characteristics**} & The surface of the nanomaterial appears somewhat rough or textured, possibly due to the magnification level and the inherent structure of the fibers. There are visible inconsistencies or impurities on the surface of some fibers.
\\

\midrule
\textcolor{black}{**Composition and Elements**} & There aren't significant compositional variations visible in the grayscale image. The image does not provide specific labels or markers indicating the presence of specific elements or compounds.
\\
 
\midrule
\textcolor{black}{**Interactions and Boundaries**} & The individual nanostructures are closely intertwined and touch each other in many places. Boundaries between individual fibers are evident, but due to their intertwined nature, defining clear separations can be challenging.
\\
 
\midrule
\textcolor{black}{**External Environment**} & There is no evidence in the image of the nanomaterial interacting with an external matrix. All visible structures appear to be fibers; no other objects or structures are discernible.
\\
 
\midrule
\textcolor{black}{**Image Technique and Modifications**} & The imaging technique used is Scanning Electron Microscopy (SEM), as indicated by the ``SE2" signal. There's no explicit indication of post-processing or modifications, but this often isn't mentioned directly in SEM images.
\\
 
\midrule
\textcolor{black}{**Functional Features**} & No specific functional features, like active sites, are discernible from the image. The image seems to be a static representation.
\\

\midrule
\textcolor{black}{**Context and Application**} & The intended application or use of the depicted nanomaterial isn't provided directly in the image, but fibrous nanomaterials can have applications ranging from reinforcement in composite materials to filtration and sensing applications. The image appears to be of a real, experimental sample given the details and characteristics shown.
\\

\bottomrule
\end{tabularx}
\end{tcolorbox}

\vspace{-7mm}

\begin{tcolorbox}[colback=white!2!white,colframe=black!75!black]
\vspace{-5mm}
\begin{tabularx}{1.2\textwidth}{bsssss}
\\ [-0.75em]
\includegraphics[width=0.145\textwidth]{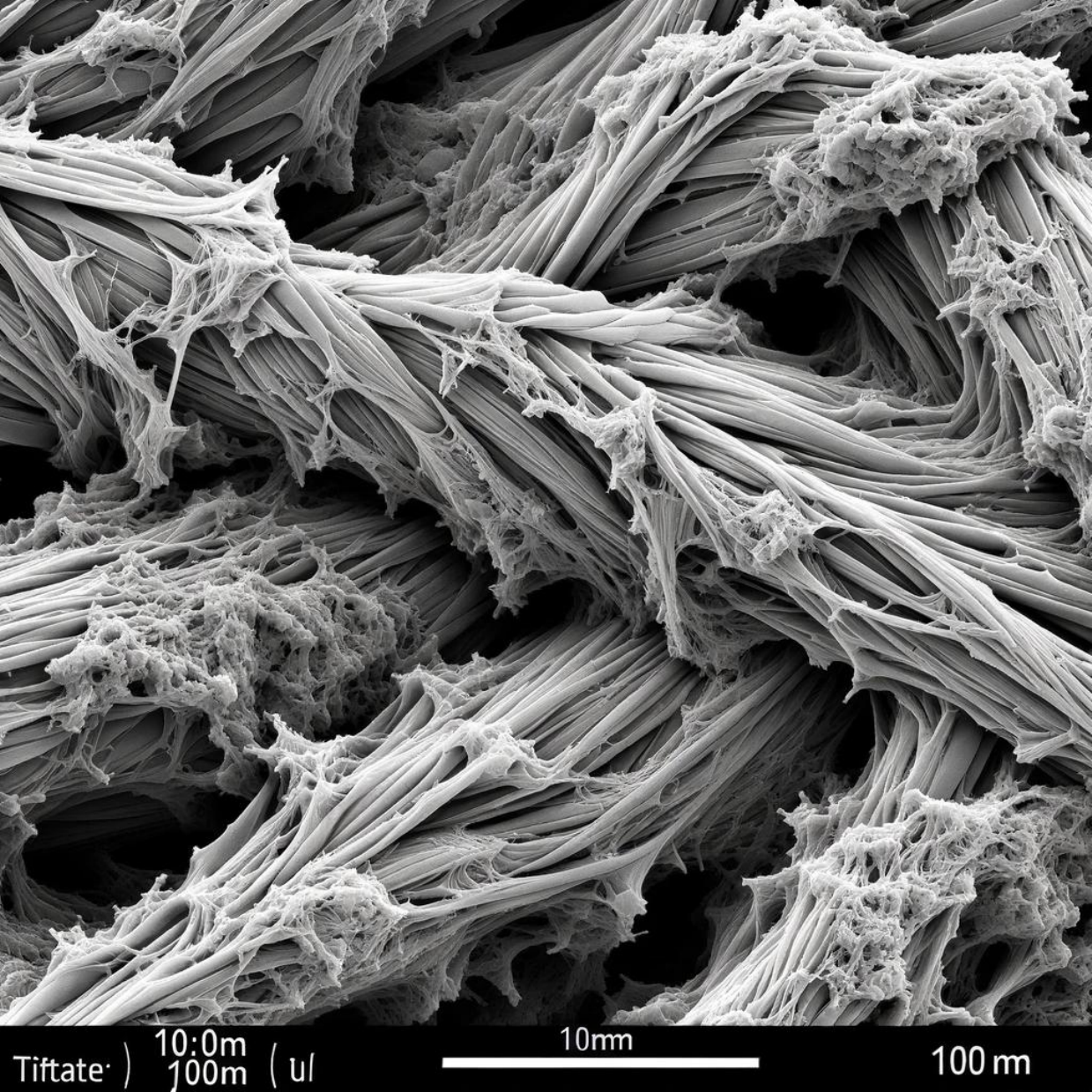} & 
\includegraphics[width=0.145\textwidth]{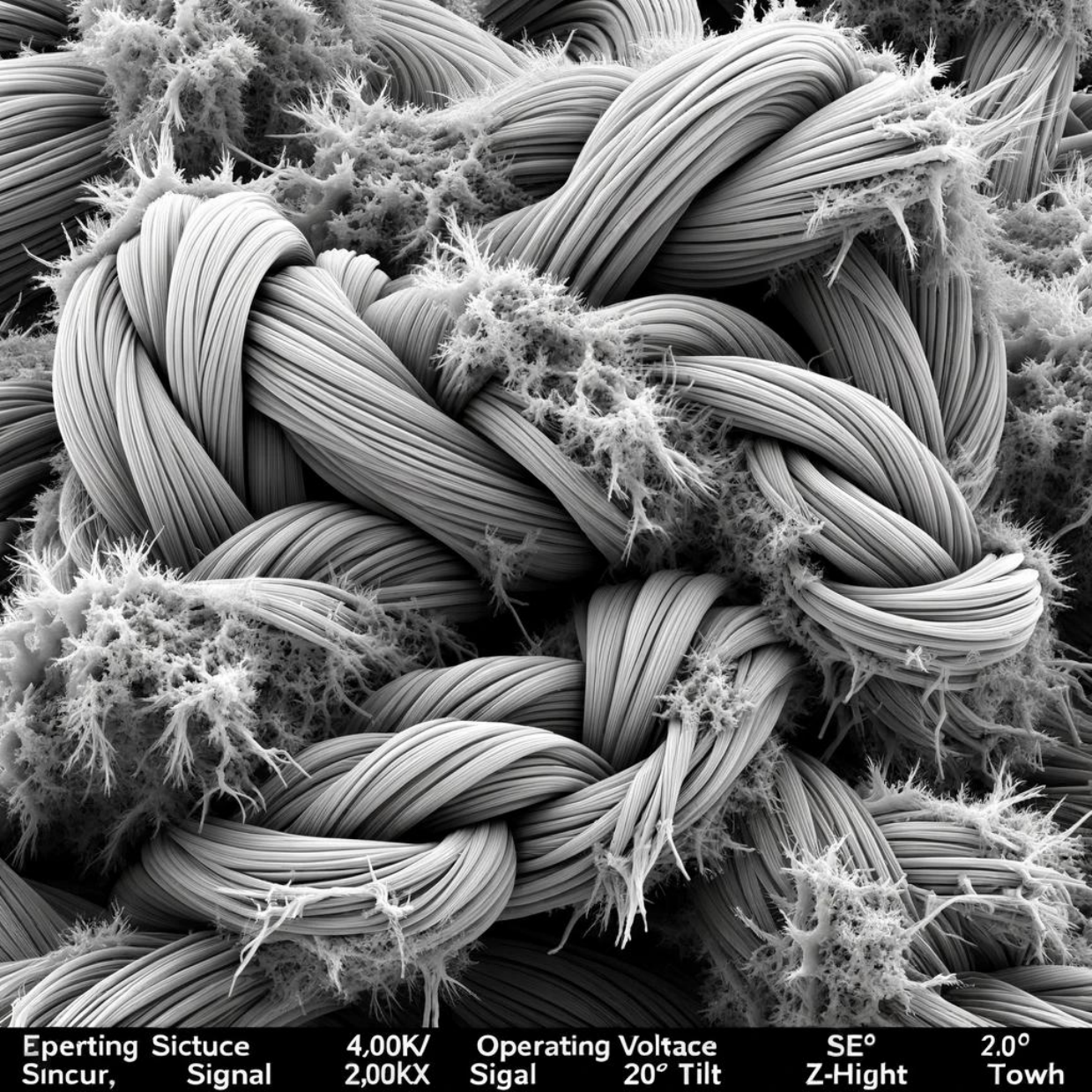} & 
\includegraphics[width=0.145\textwidth]{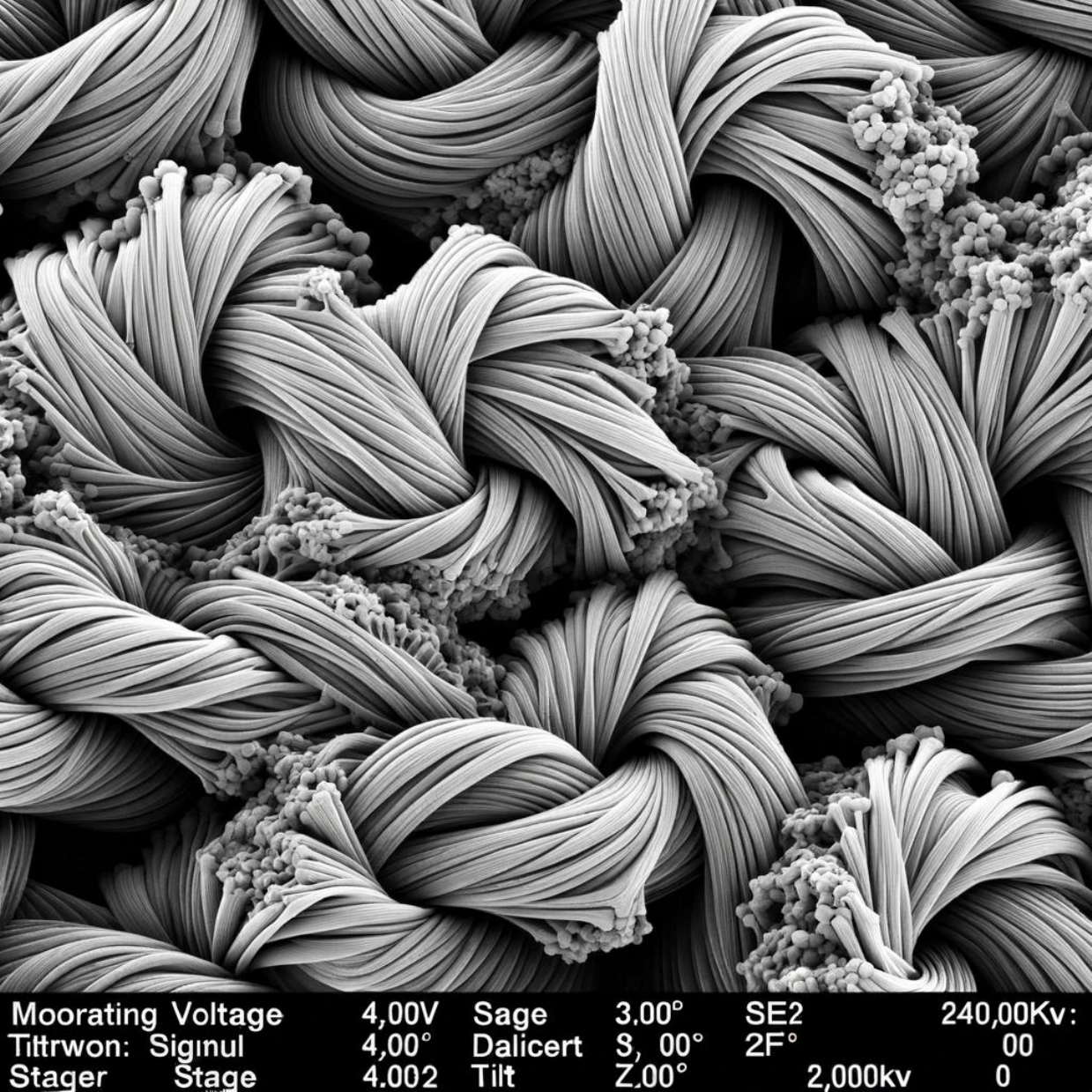} &
\includegraphics[width=0.145\textwidth]{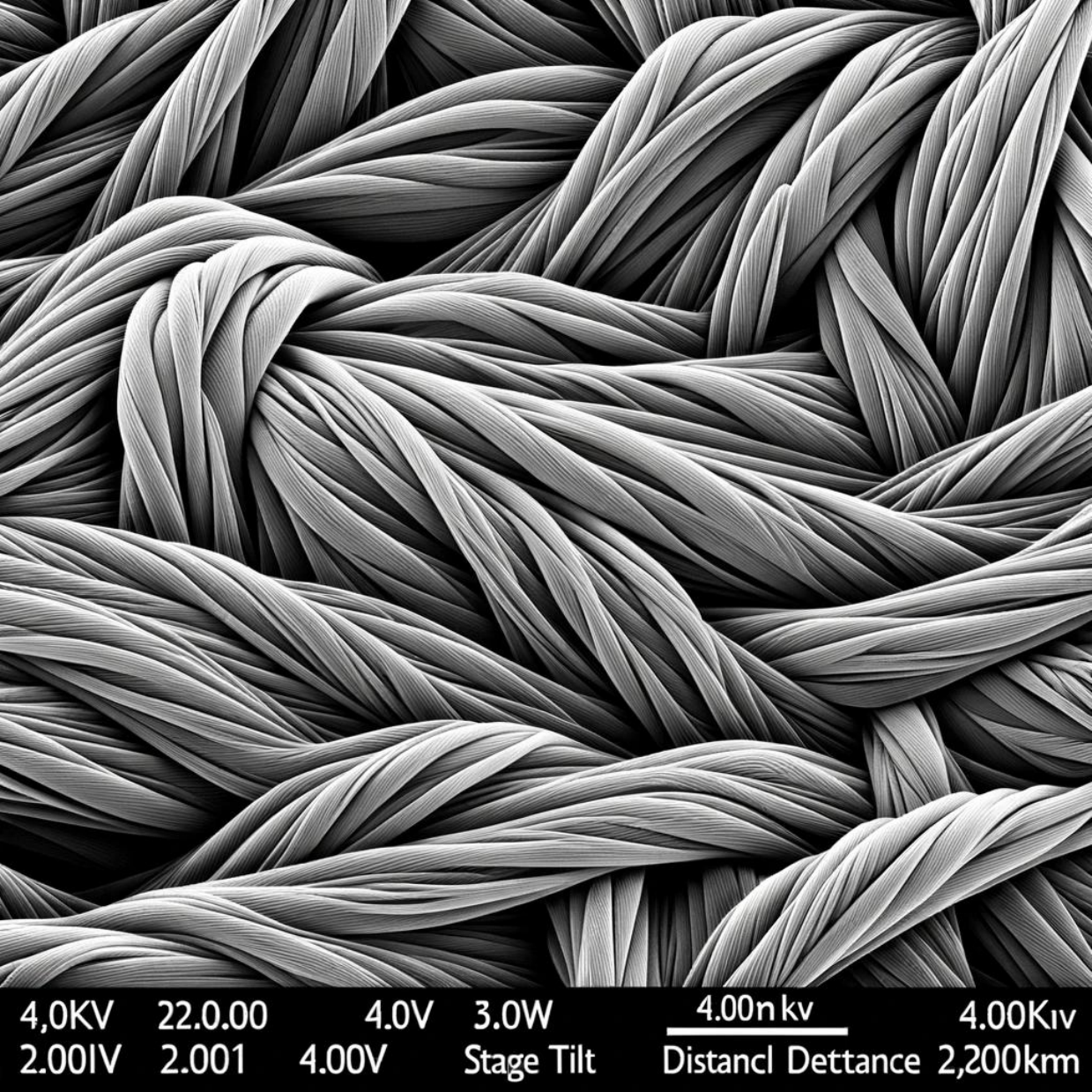} & 
\includegraphics[width=0.145\textwidth]{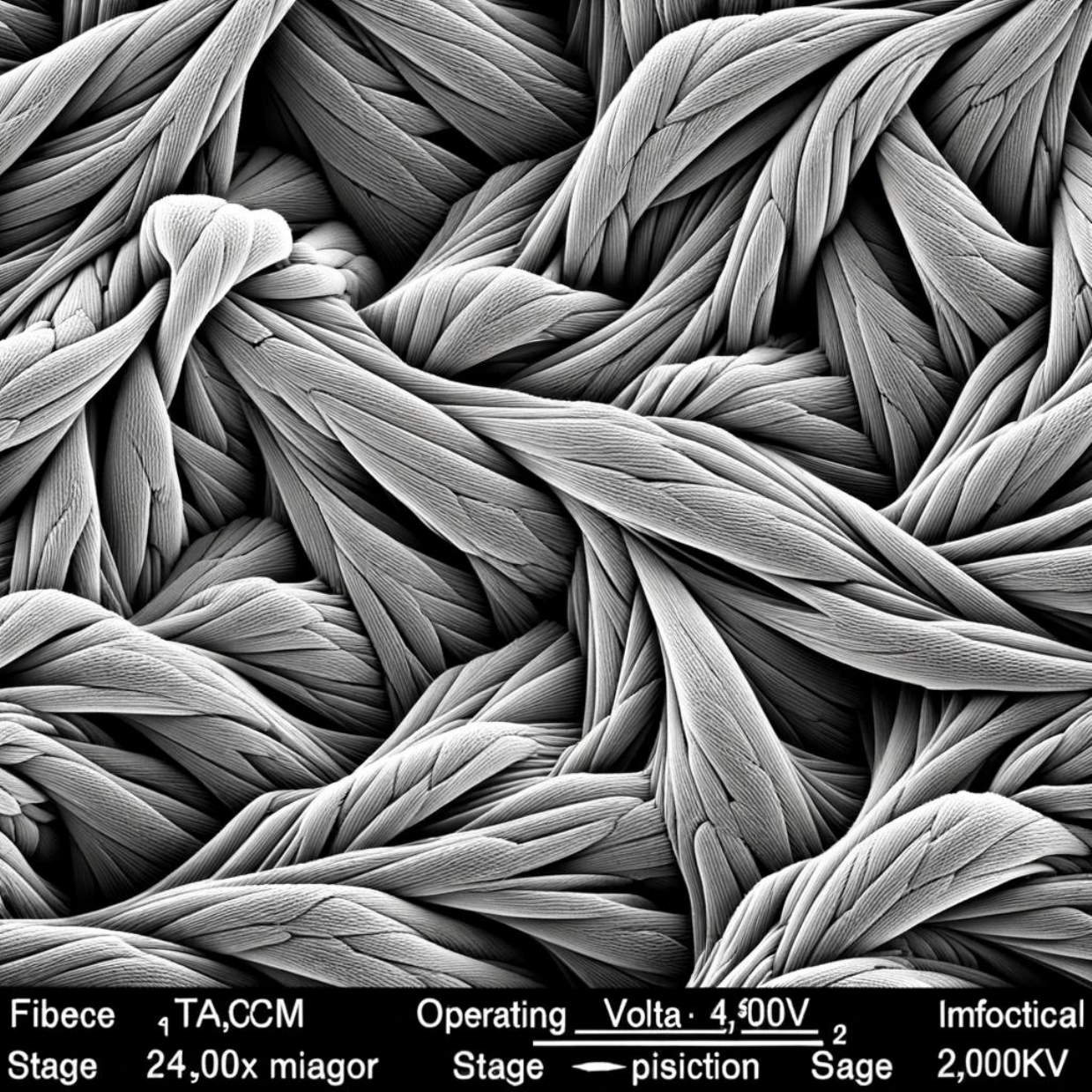} & 
\includegraphics[width=0.145\textwidth]{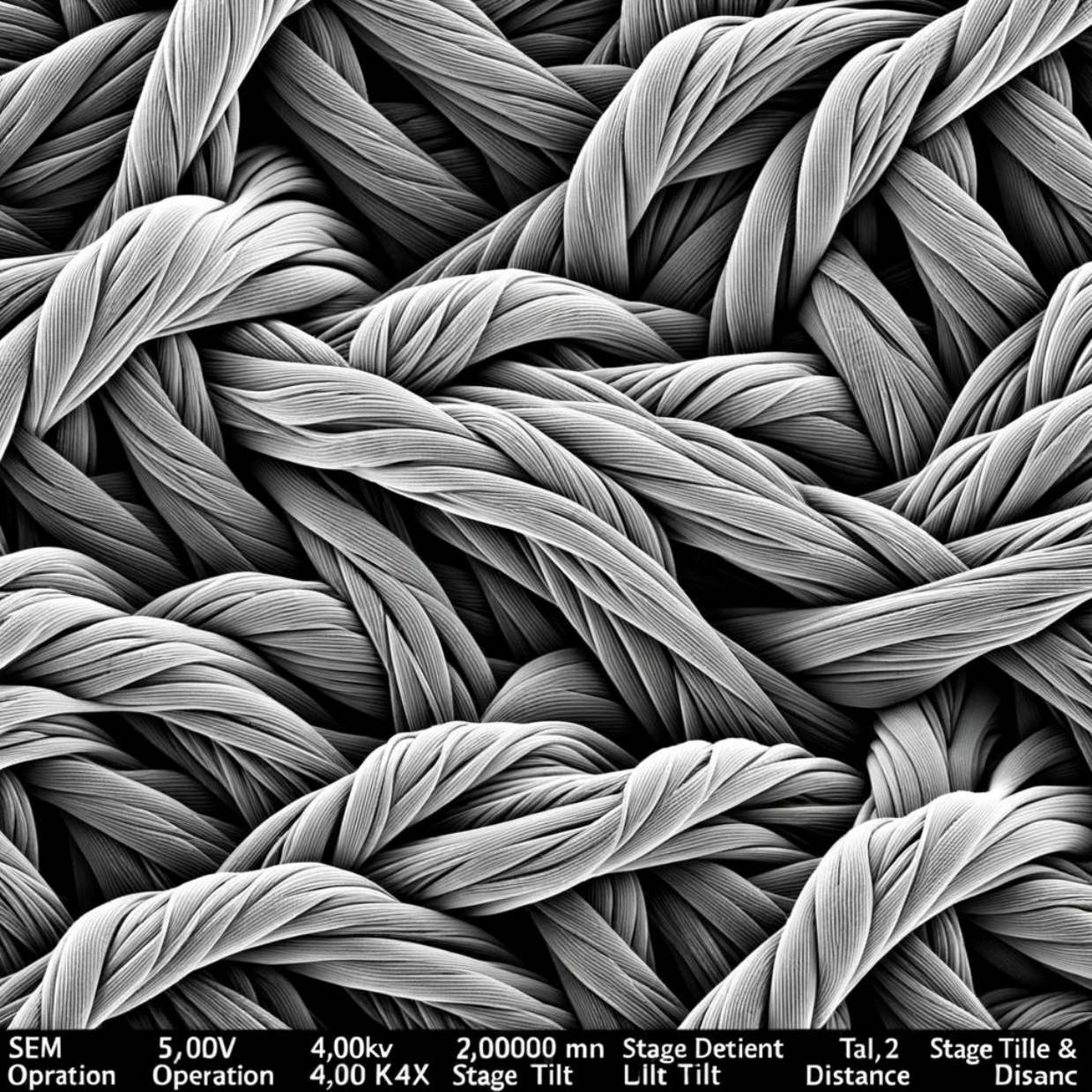} \\ [-0.45em] 
\caption{The table shows sample synthetic images generated by DALLE-3 from textual descriptions provided in a series of question-answer pairs, as shown in Table \ref{tab:tab3}.}
\label{tab:tab4}
\end{tabularx}
\vspace{-3mm}
\end{tcolorbox}

\begin{tcolorbox}[colback=white!2!white,colframe=black!75!black]
\vspace{-5mm}
\begin{tabularx}{1.2\textwidth}{bss}
\caption{The table presents the question-answer pairs generated by the LMM(GPT-4V) in response to natural language questions about the input image of the films and coated surface nanomaterial category. In summary, the question-answer pairs (generated text) provide in-depth information about an image related to films and coated surface nanomaterials.}\label{tab:tab5}\\ [-0.75em]
\toprule 
 \\ [-0.75em]
\begin{overpic}[width=0.175\columnwidth]{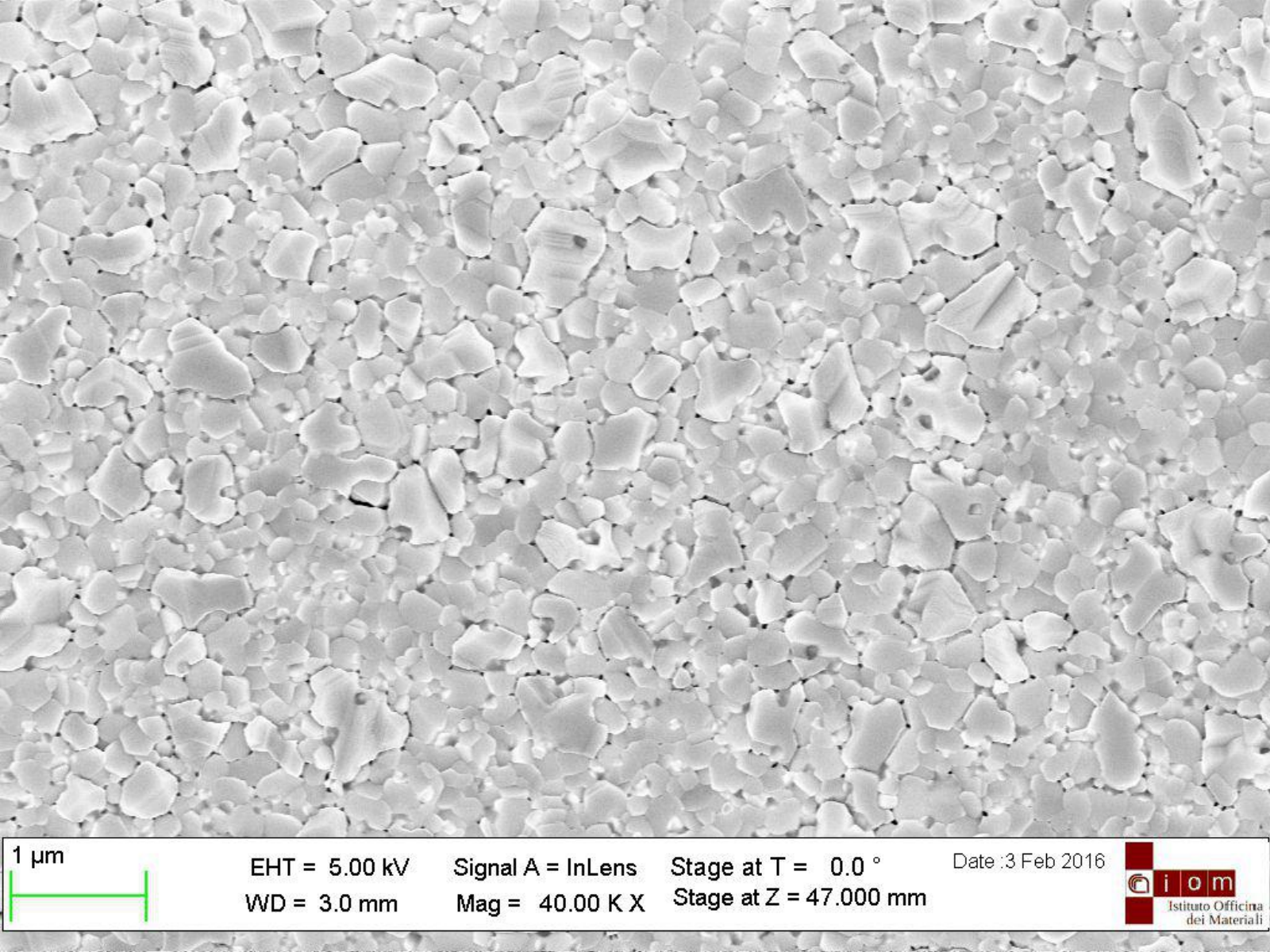}
  \put(375,40){\makebox[0pt][c]{%
    \parbox{0.4\columnwidth}{%
      \centering
      \textbf{Please answer the following questions based on the provided input image belonging to the Films and Coated Surface nanomaterial category.} 
    }%
  }}  
\end{overpic}  \\ [-0.25em]

\midrule
\textbf{Query prompt}  & \textbf{GPT-4 Vision output} \\
\midrule
\endhead
\textcolor{black}{**Basics**} & The image depicts a nanomaterial surface, possibly a thin film or coated surface. The scale bar represents 1 micrometer ($\mu m$).
\\

\midrule
\textcolor{black}{**Morphology and Structure**} & The nanomaterials in the image have a flake or shard-like morphology. There are no distinct layers, phases, or domains visible in the image. The nanomaterials appear somewhat varied in size and shape but have a general uniformity in their shard-like appearance.
\\

\midrule
\textcolor{black}{**Size and Distribution**} & The approximate size of the individual nanostructures ranges from a few nanometers to several hundred nanometers. The nanomaterials are densely packed and distributed uniformly across the image. There is no apparent evidence of aggregation or bundling.
\\

\midrule
\textcolor{black}{**Surface Characteristics**} & The nanomaterial surface appears rough with distinct shard-like structures. There do not appear to be any significant defects, pores, or impurities on the surface.
\\

\midrule
\textcolor{black}{**Composition and Elements**} & There's no evidence of compositional variations based on color, brightness, or contrast in this grayscale image. There are no labels or markers indicating specific elements or compounds present in the image.
\\
 
\midrule
\textcolor{black}{**Interactions and Boundaries**} & The individual nanostructures seem to be touching and closely packed, with some overlapping visible. The boundaries between different structures are clear due to the distinct shapes of the nanostructures.
\\
 
\midrule
\textcolor{black}{**External Environment**} & There's no evident interaction of the nanomaterial with a surrounding matrix, solvent, or polymer. The entire image seems to consist of the nanomaterial; no other distinct structures or objects are observed.
\\
 
\midrule
\textcolor{black}{**Image Technique and Modifications**} & The image appears to be captured using Scanning Electron Microscopy (SEM) based on the information provided and the visual characteristics. There is no clear indication in the image about post-processing or modifications.
\\
 
\midrule
\textcolor{black}{**Functional Features**} & From the image, no specific functional features such as active sites or regions with distinct properties are visible. The image seems to be a static representation, with no dynamic processes evident.
\\

\midrule
\textcolor{black}{**Context and Application**} & The intended application or use of the nanomaterial is not provided in the image. Based on the information and the appearance, it seems like a real, experimental sample rather than a theoretical or simulation-based representation.
\\

\bottomrule
\end{tabularx}
\end{tcolorbox}

\vspace{-8mm}

\begin{tcolorbox}[colback=white!2!white,colframe=black!75!black]
\vspace{-5mm}
\begin{tabularx}{1.5\textwidth}{bsssss}
\\ [-1.05em]
\includegraphics[height=0.105\textwidth]{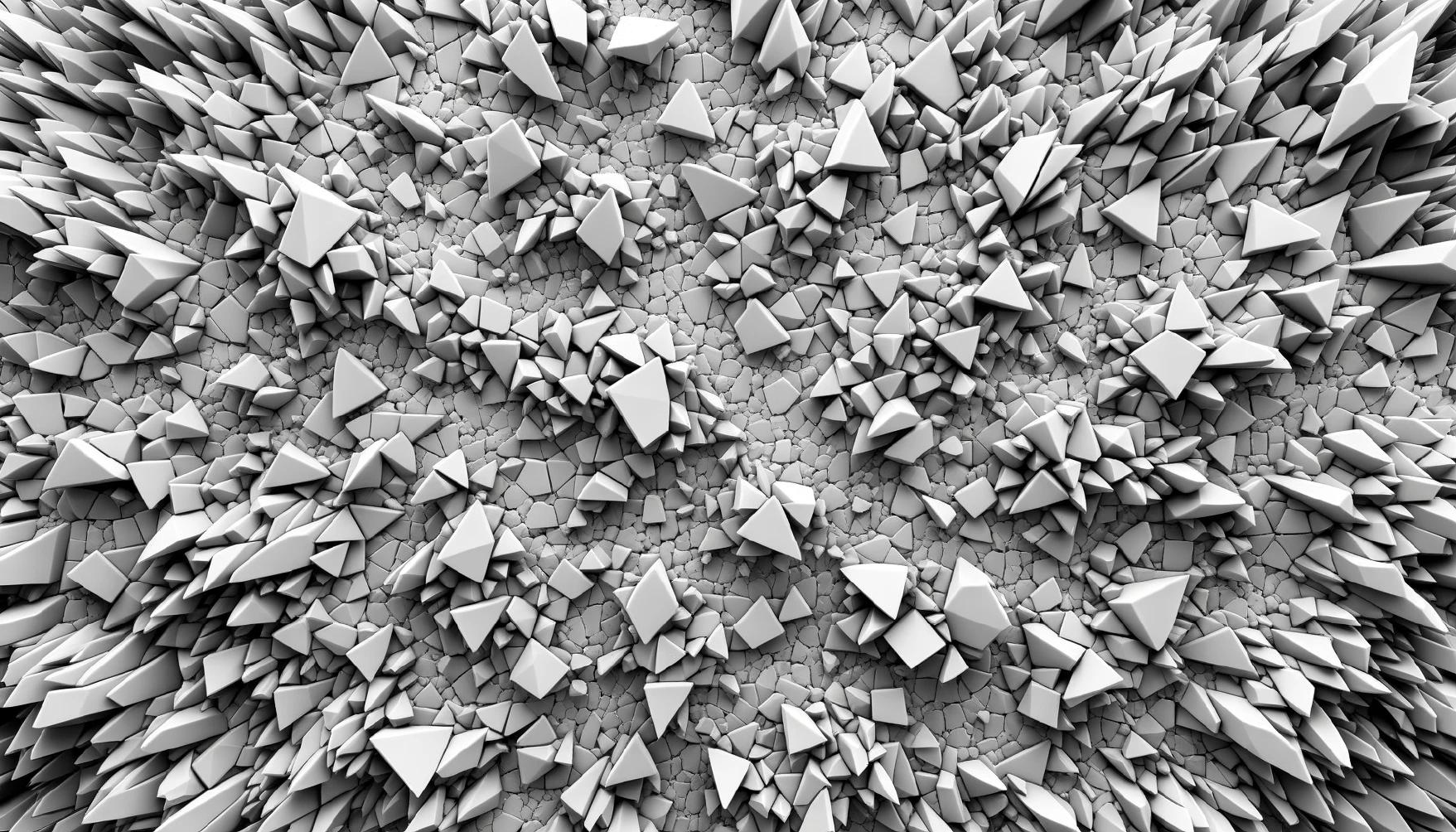} & 
\includegraphics[height=0.105\textwidth]{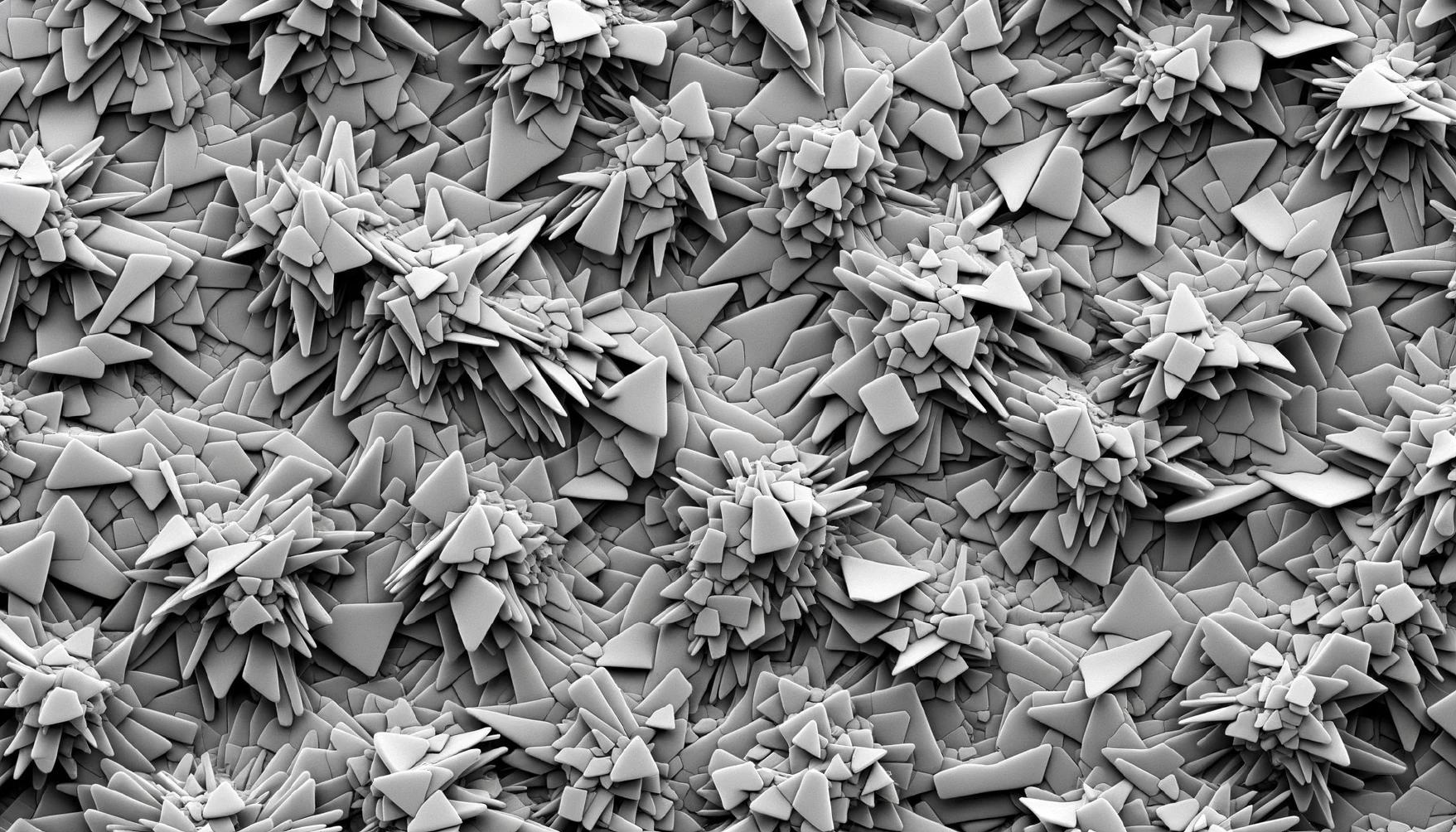} & 
\includegraphics[height=0.105\textwidth]{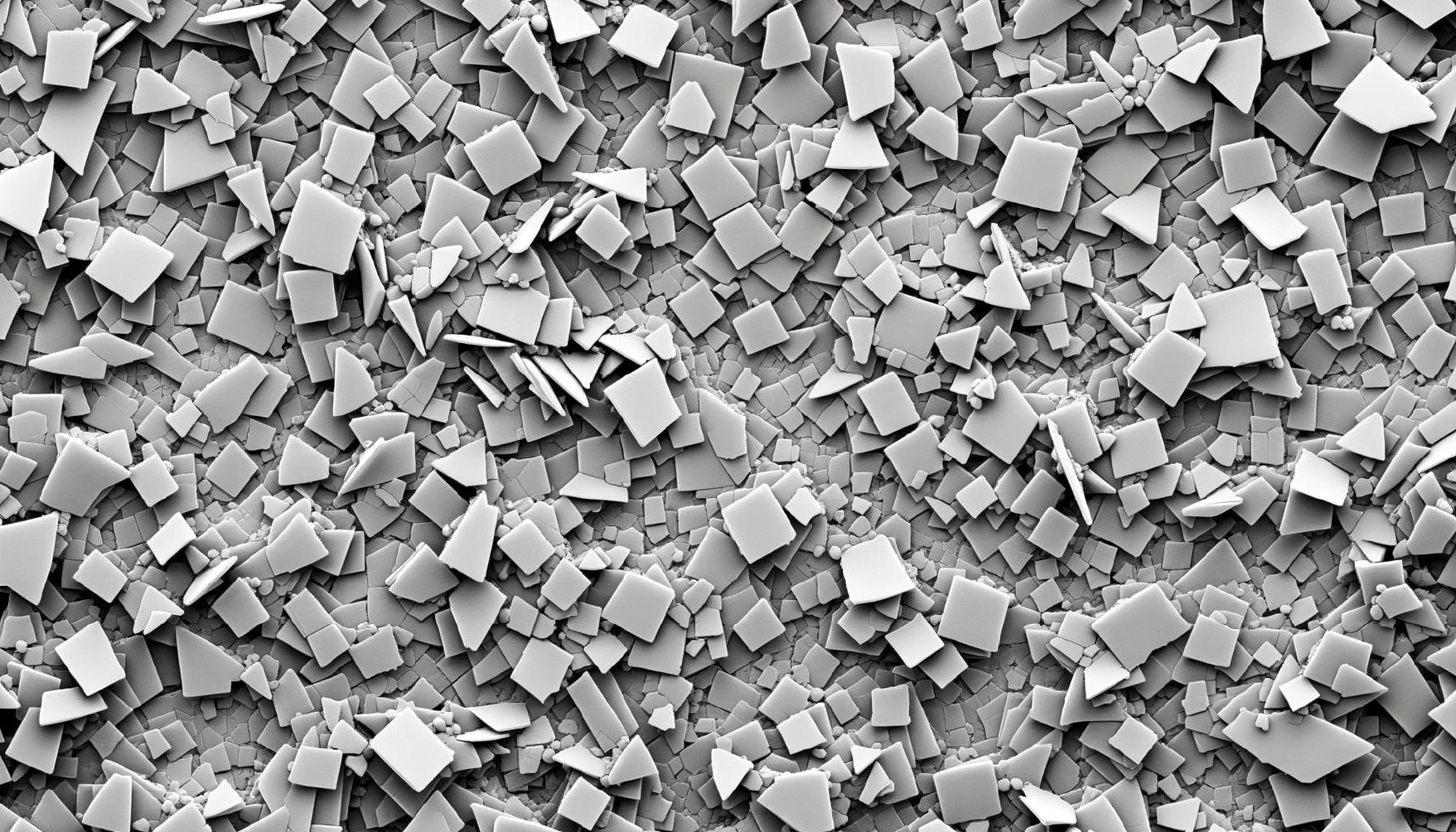} & 
\includegraphics[height=0.105\textwidth]{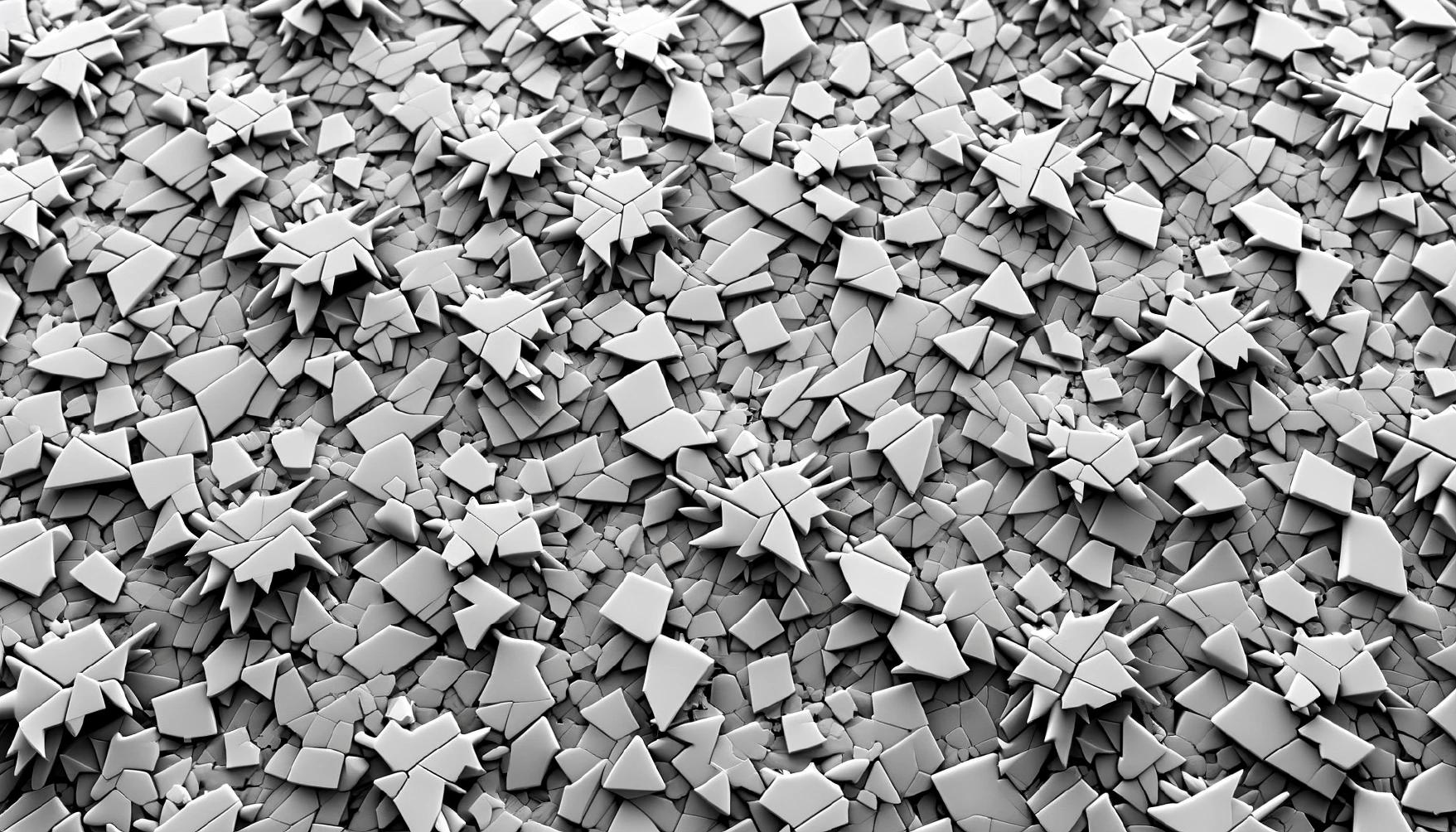} \\ [-0.25em] 
\includegraphics[height=0.105\textwidth]{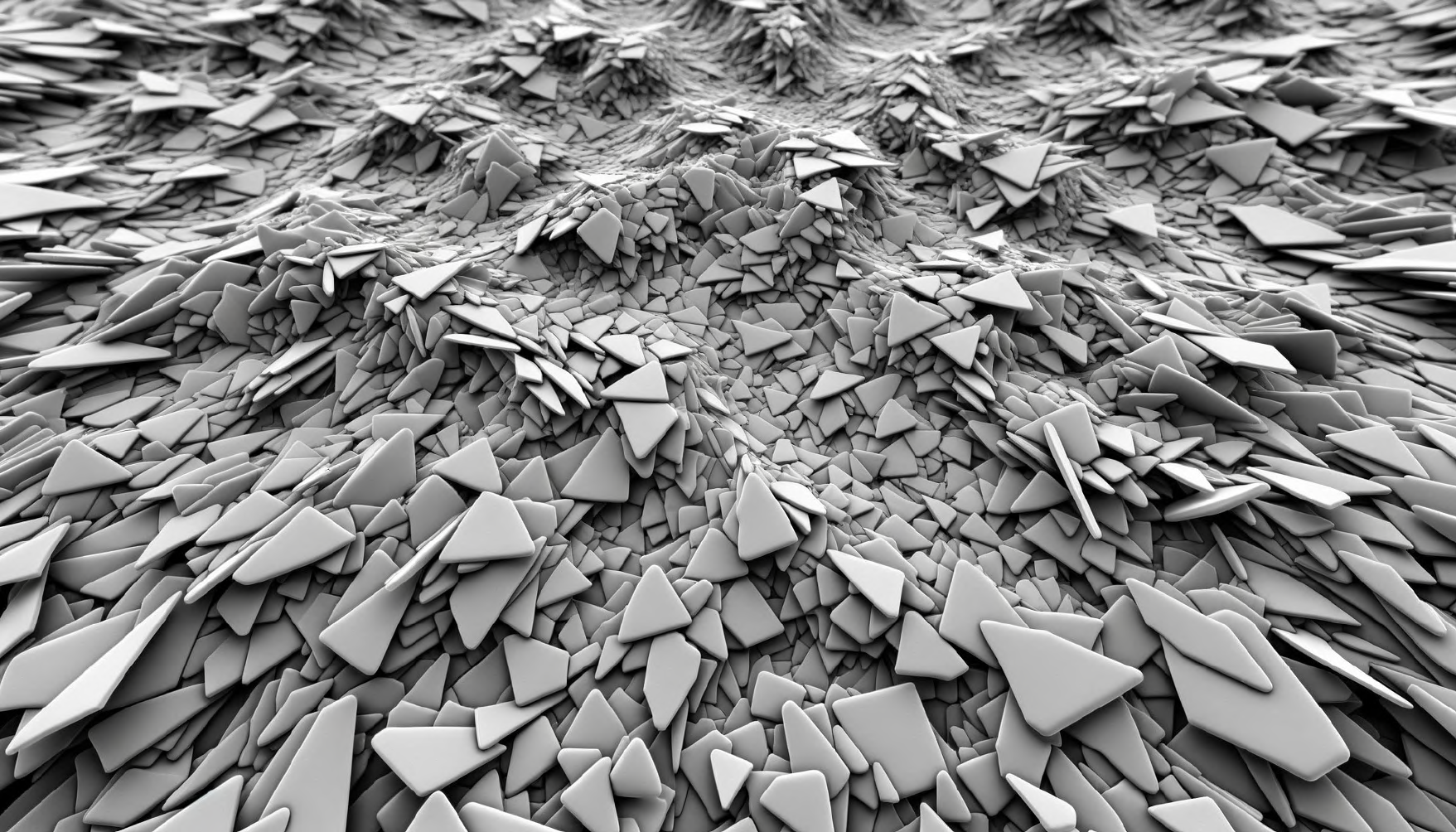} & 
\includegraphics[height=0.105\textwidth]{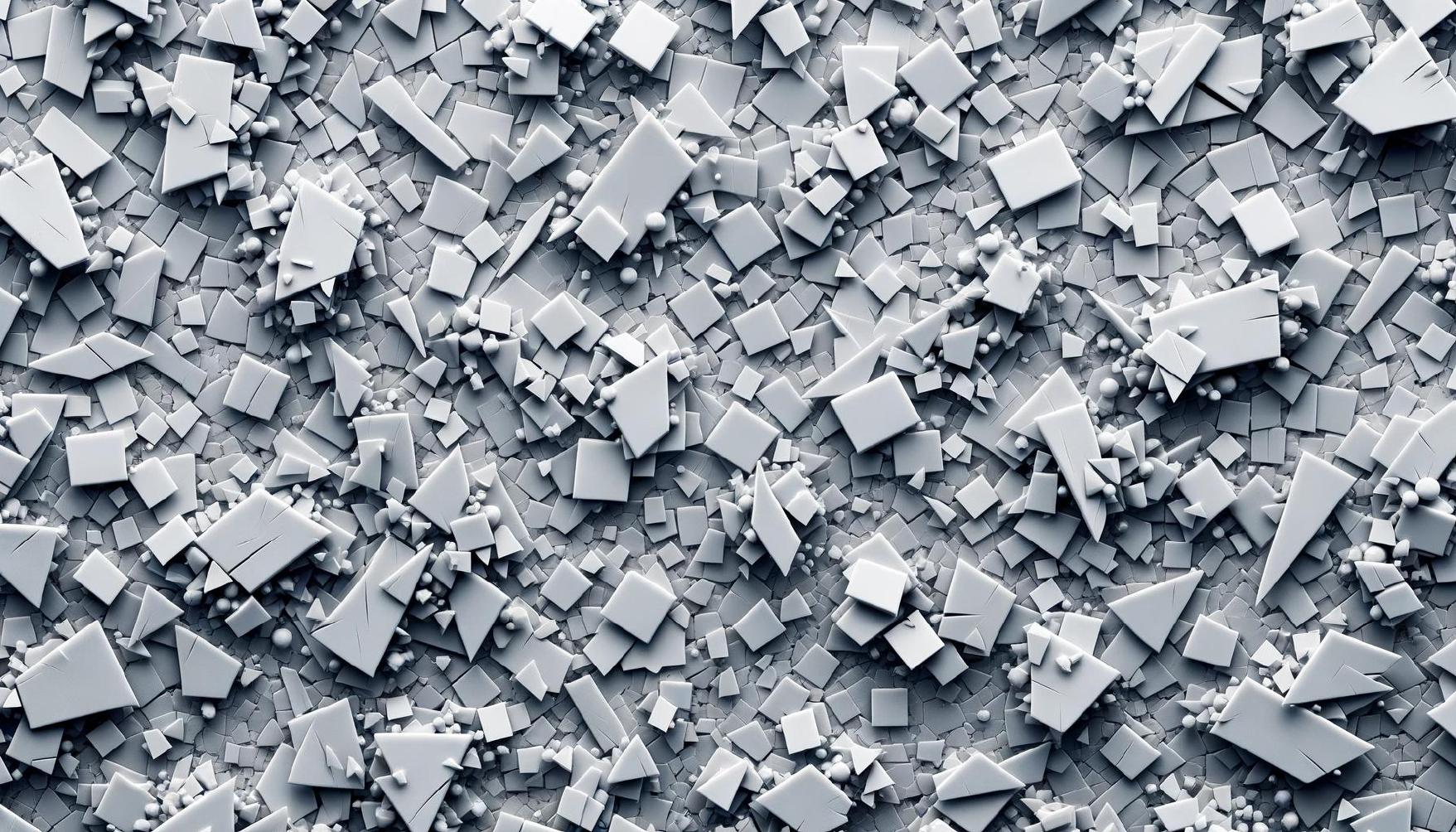} \\ [-0.95em]
\caption{The table shows sample synthetic images generated by DALLE-3 from textual descriptions provided in a series of question-answer pairs, as shown in Table \ref{tab:tab5}.}
\label{tab:tab6}
\end{tabularx}
\vspace{-3mm}
\end{tcolorbox}

\begin{tcolorbox}[colback=white!2!white,colframe=black!75!black]
\vspace{-5mm}
\begin{tabularx}{1.2\textwidth}{bss}
\caption{The table presents the question-answer pairs generated by the LMM(GPT-4V) in response to natural language questions about the input image of the nanowires nanomaterial category. In summary, the question-answer pairs (generated text) provide in-depth information about an image related to nanowire nanomaterials.}\label{tab:tab7}\\ [-0.75em]
\toprule 
 \\ [-0.85em]
\begin{overpic}[width=0.175\columnwidth]{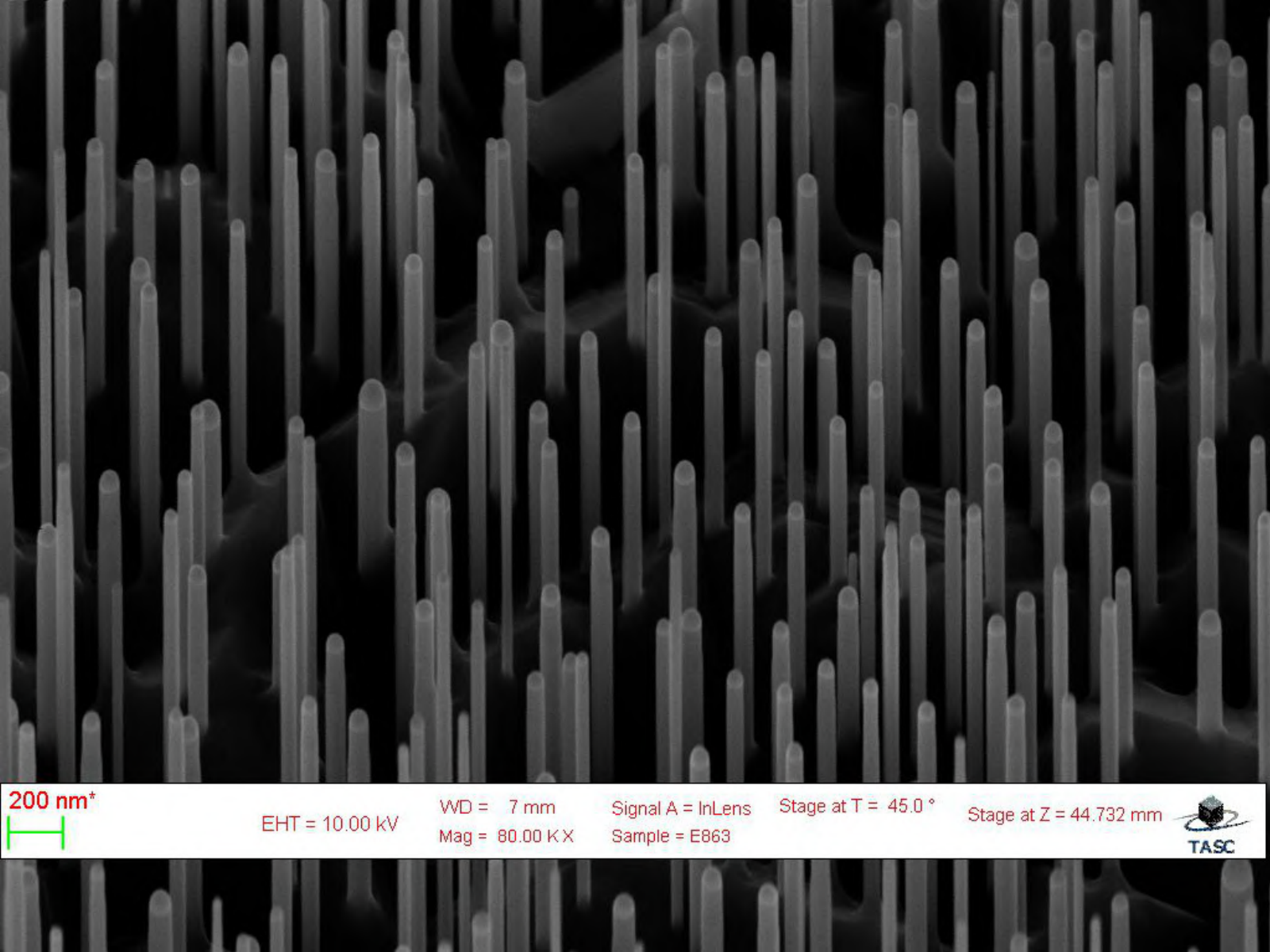}
  \put(400,40){\makebox[0pt][c]{%
    \parbox{0.5\columnwidth}{%
      \centering
      \textbf{Please answer the following questions based on the provided input image belonging to the nanowire nanomaterial category.} 
    }%
  }}  
\end{overpic}  \\ [-0.25em]

\midrule
\textbf{Query prompt}  & \textbf{GPT-4 Vision output} \\
\midrule
\endhead
\textcolor{black}{**Basics**} & The image depicts nanowires. The scale of the image is represented by the bar labeled "200 nm," indicating that the length of the bar corresponds to 200 nanometers.
\\

\midrule
\textcolor{black}{**Morphology and Structure**} & The nanomaterials in the image have a rod-like or wire-like morphology. There are no distinct layers, phases, or domains visible in the image. The nanowires appear relatively uniform in size and shape.
\\

\midrule
\textcolor{black}{**Size and Distribution**} & The approximate size of the individual nanostructures can be deduced to be several hundred nanometers in length and likely tens of nanometers in diameter based on the scale bar. The nanowires are distributed evenly and vertically throughout the image. There is no visible evidence of aggregation or bundling.
\\

\midrule
\textcolor{black}{**Surface Characteristics**} & The nanowires appear smooth. No obvious defects, pores, or impurities are visible on the nanowire surface.
\\

\midrule
\textcolor{black}{**Composition and Elements**} & The image does not show evidence of compositional variations based on the grayscale representation. There are no visible labels or markers indicating specific elements or compounds present.
\\
 
\midrule
\textcolor{black}{**Interactions and Boundaries**} & Individual nanostructures seem separate and do not appear to be touching or fusing with one another. Clear boundaries exist between different nanowires.
\\
 
\midrule
\textcolor{black}{**External Environment**} & There is no evidence of the nanowires interacting with their surrounding environment or matrix in this image. The darker areas between the nanowires do not appear to be nanomaterials, likely representing voids or spaces between the nanowires.

\\
 
\midrule
\textcolor{black}{**Image Technique and Modifications**} & The imaging technique used appears to be Scanning Electron Microscopy (SEM) based on the details provided in the image (e.g., EHT, Mag, Signal). There is no evidence from the image to suggest post-processing or modifications like false coloring or 3D rendering.
\\
 
\midrule
\textcolor{black}{**Functional Features**} & The image does not provide specific information regarding functional features or distinct properties. The image is a static representation and does not capture dynamic processes.
\\

\midrule
\textcolor{black}{**Context and Application**} & The intended application or use of the nanowires is not specified in the image. Based on the details provided, it appears to be a real, experimental sample as opposed to a theoretical or simulation-based representation.
\\

\bottomrule
\end{tabularx}
\end{tcolorbox}

\vspace{-7mm}

\begin{tcolorbox}[colback=white!2!white,colframe=black!75!black]
\vspace{-5mm}
\begin{tabularx}{1.2\textwidth}{bsssss}
\\ [-0.65em]
\includegraphics[width=0.145\textwidth]{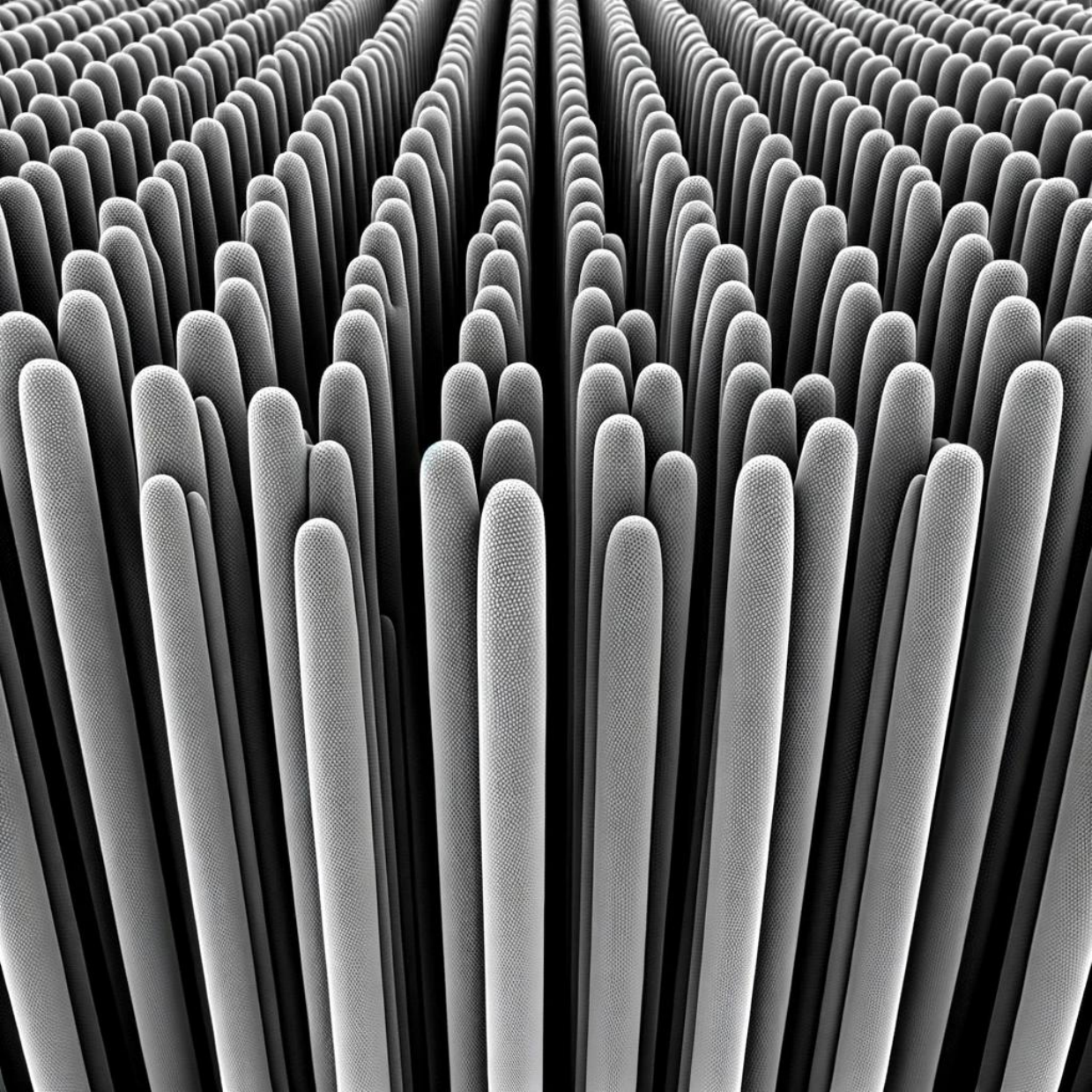} & 
\includegraphics[width=0.145\textwidth]{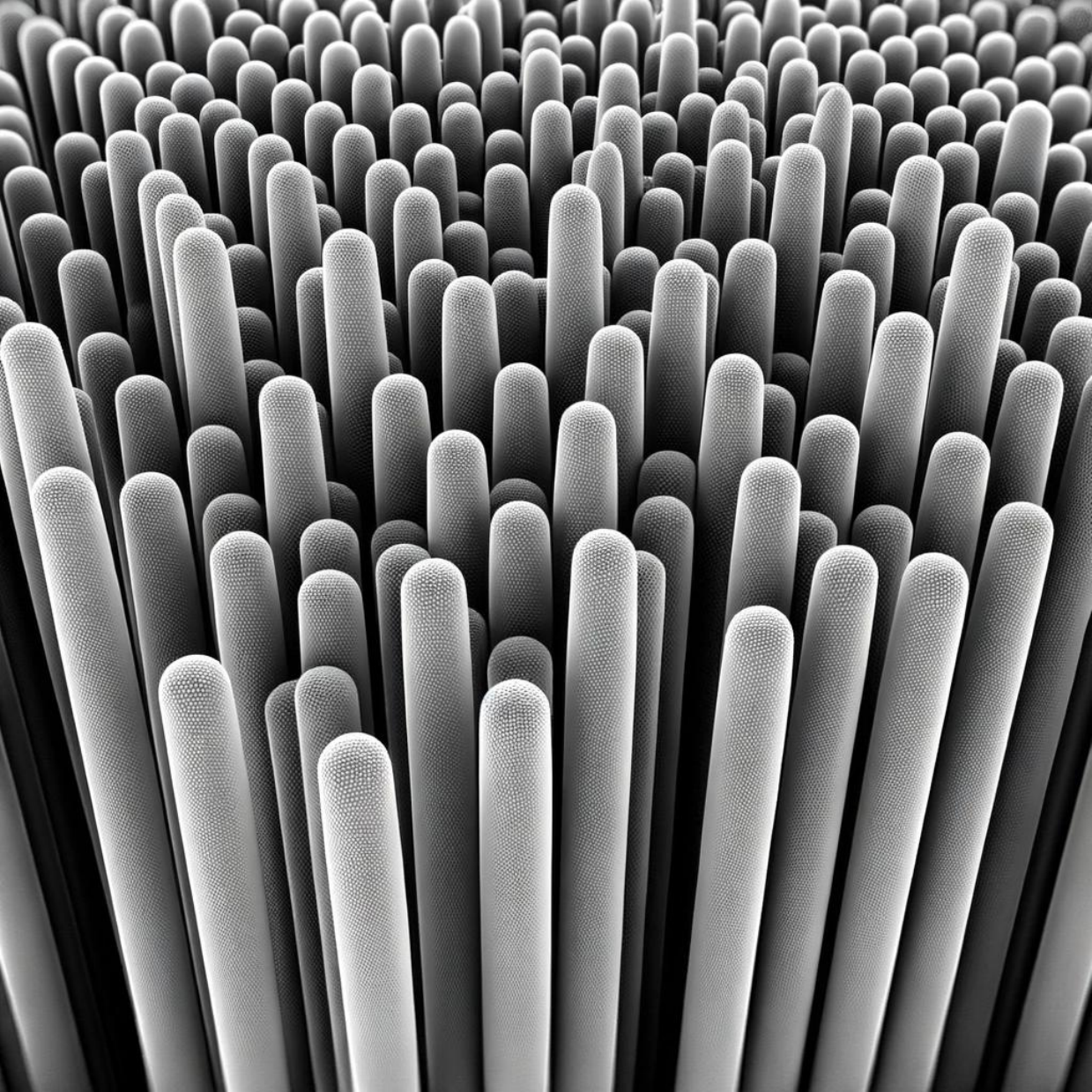} & 
\includegraphics[width=0.145\textwidth]{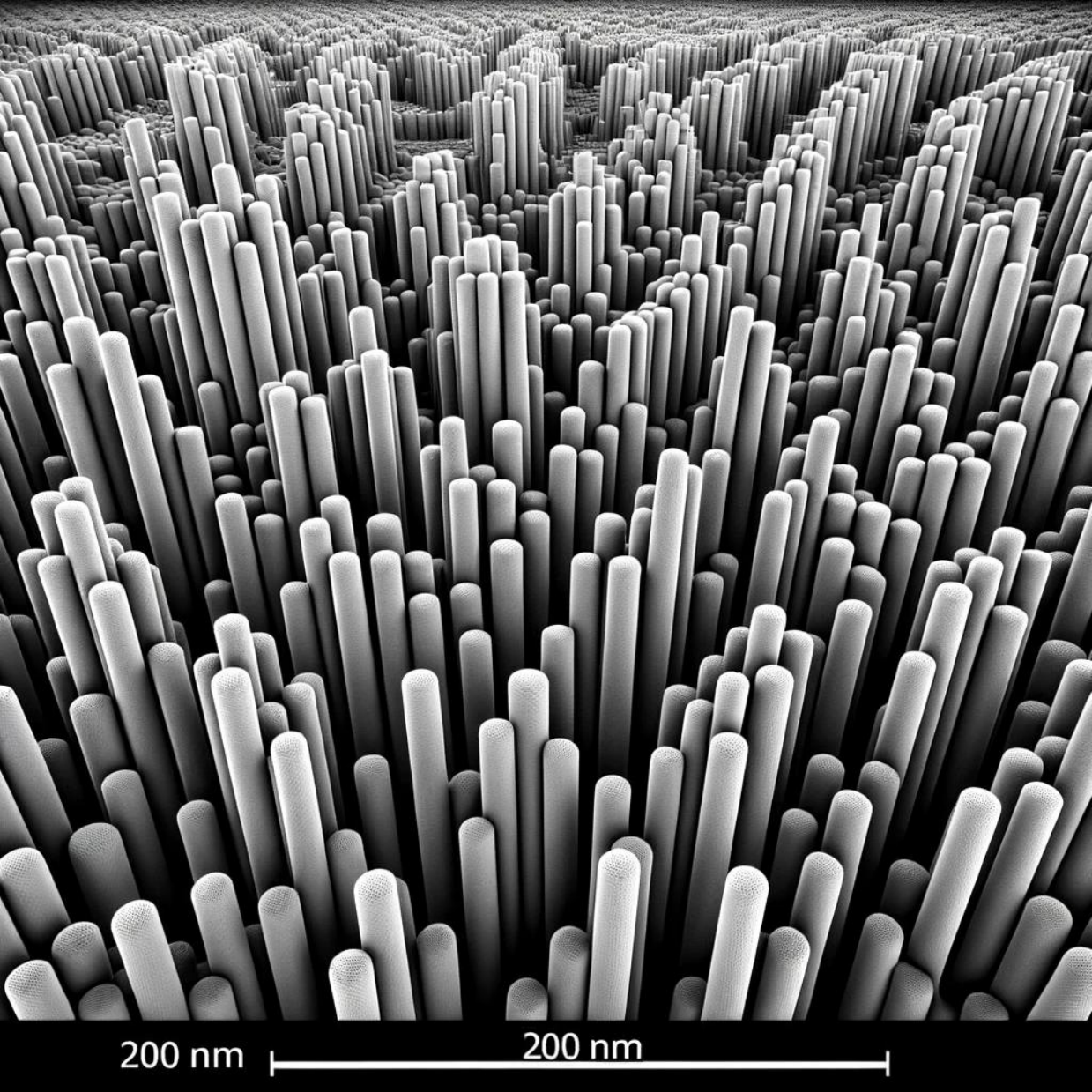} &
\includegraphics[width=0.145\textwidth]{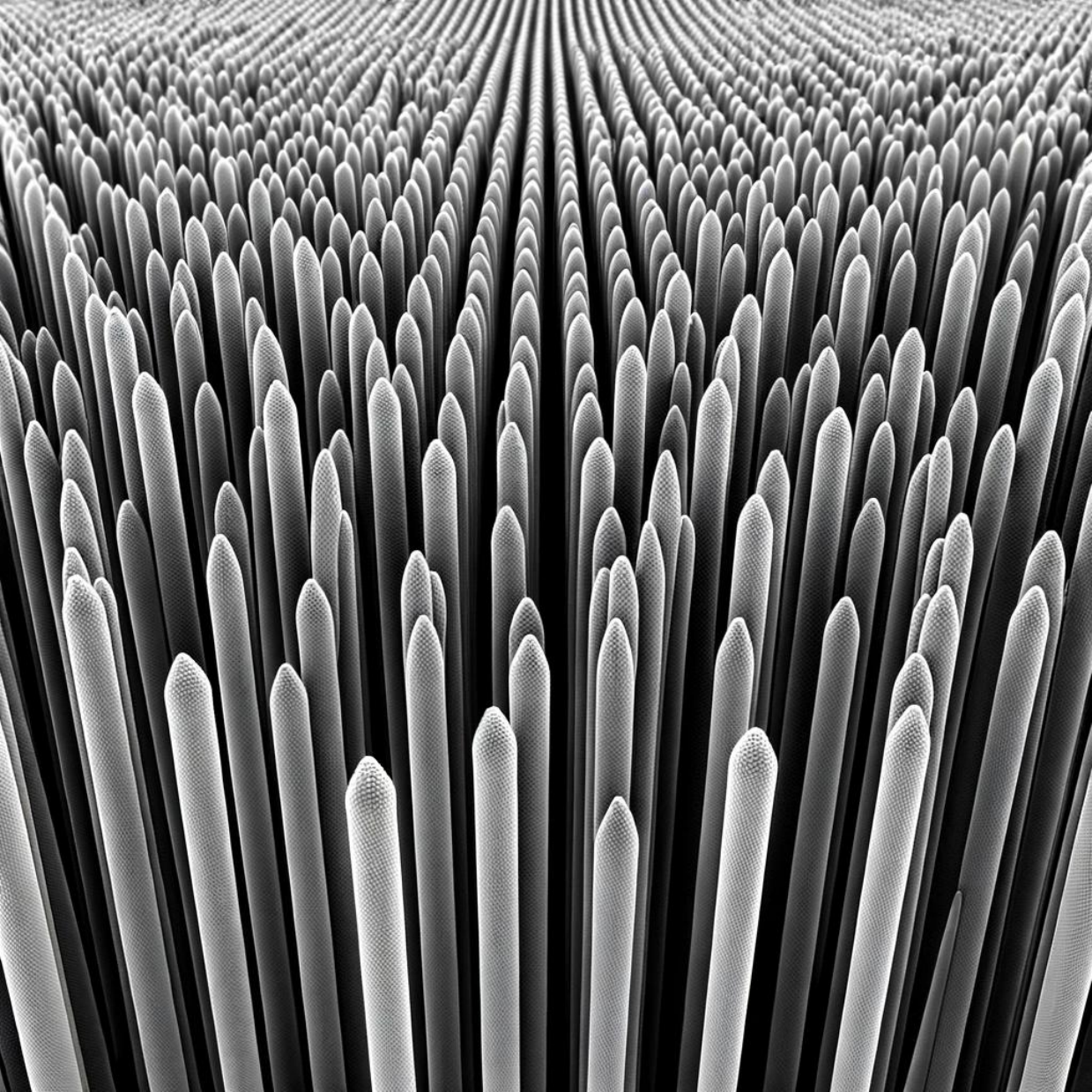} & 
\includegraphics[width=0.145\textwidth]{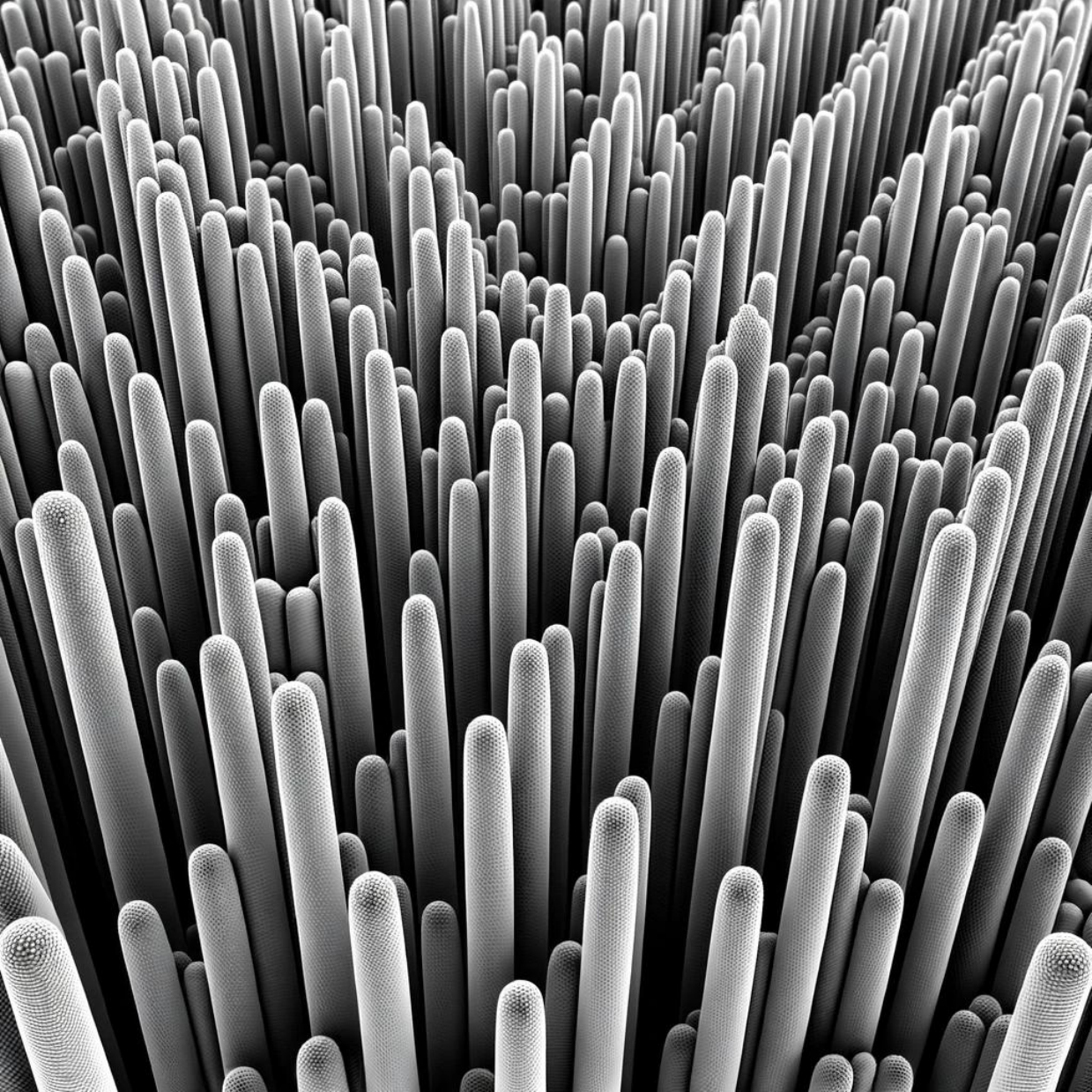} & 
\includegraphics[width=0.145\textwidth]{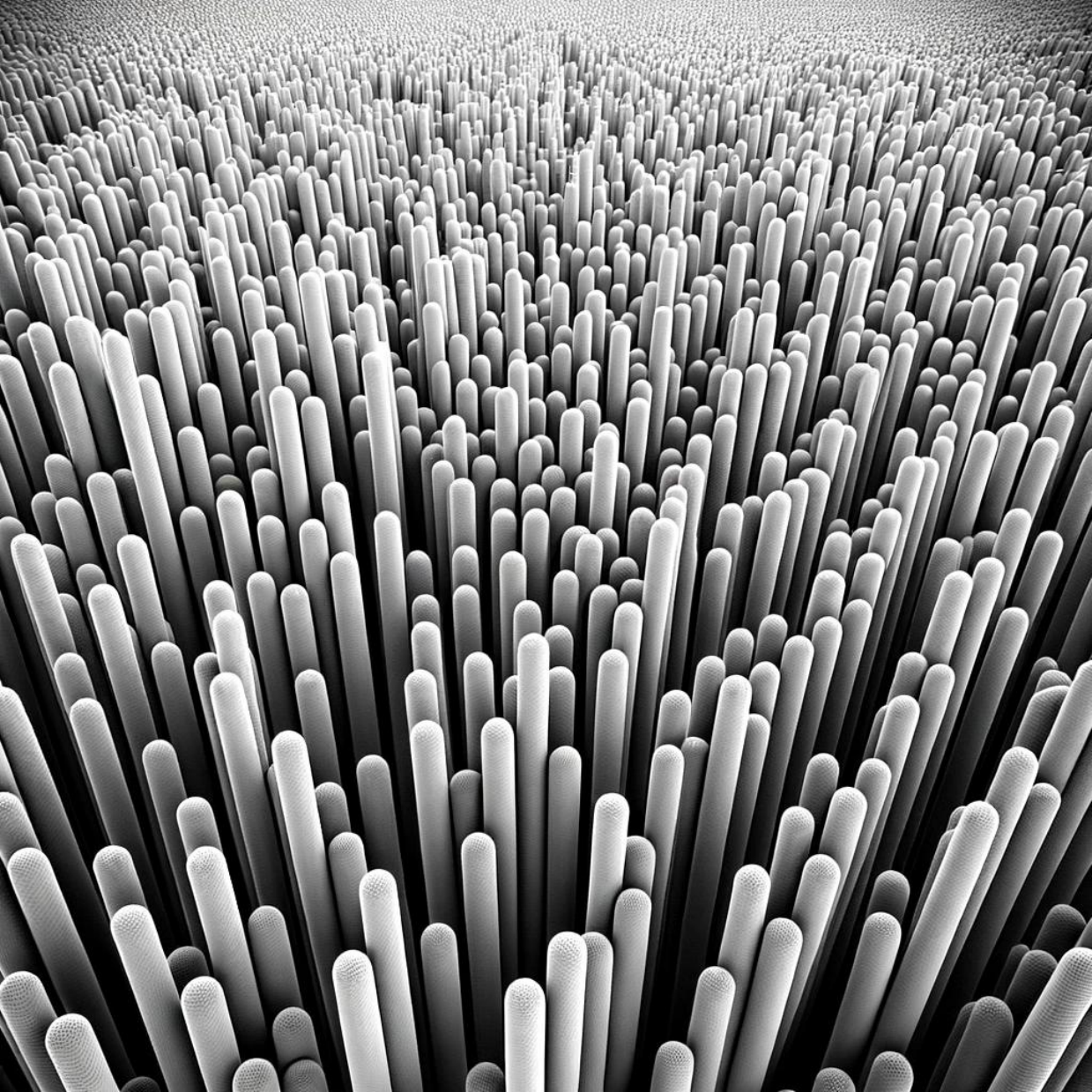} \\ [-0.45em] 
\caption{The table shows sample synthetic images generated by DALLE-3 from textual descriptions provided in a series of question-answer pairs, as shown in Table \ref{tab:tab7}.}
\label{tab:tab8}
\end{tabularx}
\vspace{-3mm}
\end{tcolorbox}

\begin{tcolorbox}[colback=white!2!white,colframe=black!75!black]
\vspace{-5mm}
\begin{tabularx}{1.2\textwidth}{bss}
\caption{The table presents the question-answer pairs generated by the LMM(GPT-4V) in response to natural language questions about the input image of the porous sponge nanomaterial category. In summary, the question-answer pairs (generated text) provide in-depth information about an image related to porous sponge nanomaterials.}\label{tab:tab11}\\ [-0.75em]
\toprule 
 \\ [-0.75em]
\begin{overpic}[width=0.165\columnwidth]{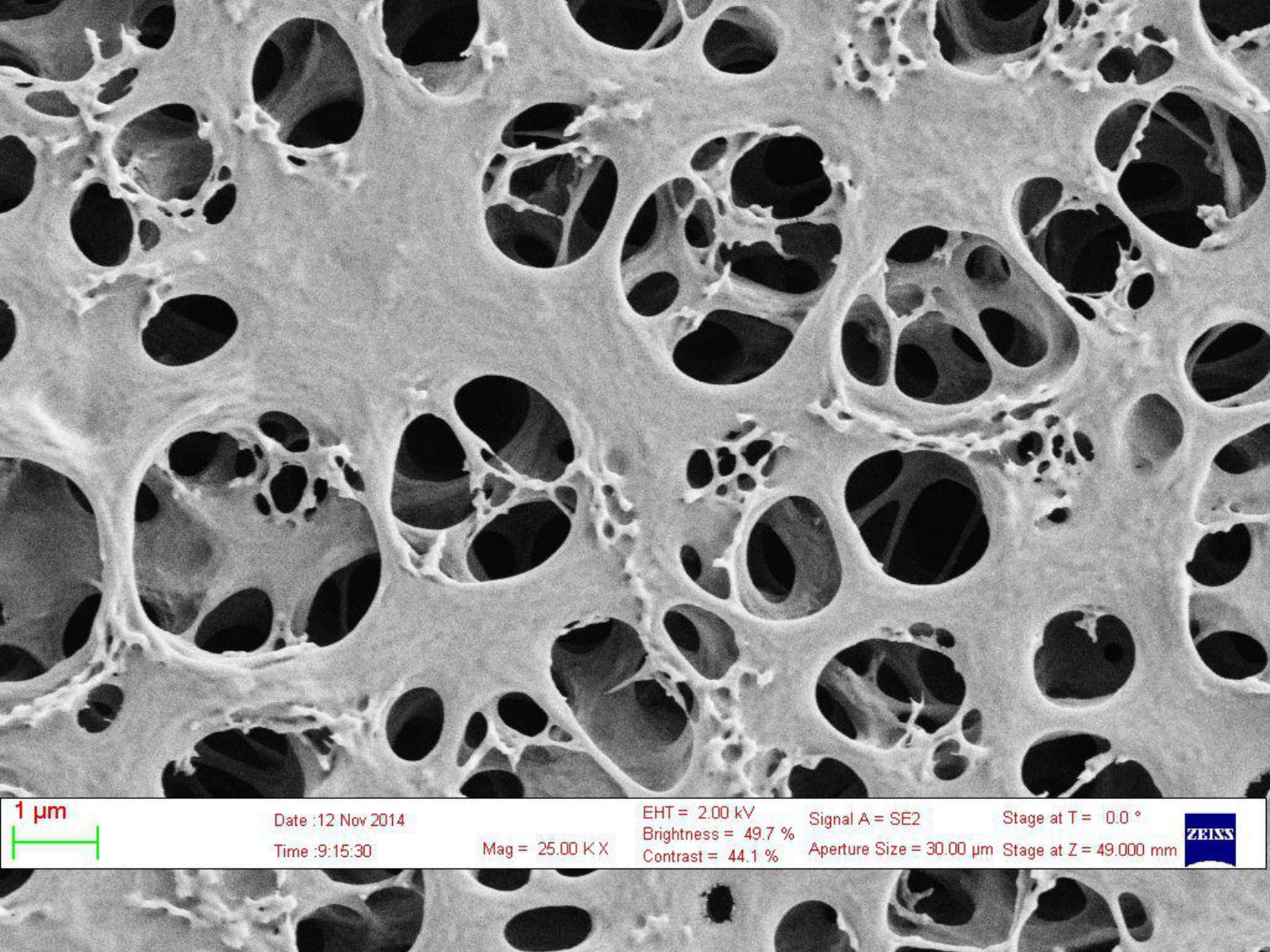}
  \put(400,40){\makebox[0pt][c]{%
    \parbox{0.5\columnwidth}{%
      \centering
      \textbf{Please answer the following questions based on the provided input image belonging to the Porous Sponge nanomaterial category.} 
    }%
  }}  
\end{overpic}  \\ [-0.25em]

\midrule
\textbf{Query prompt}  & \textbf{GPT-4 Vision output} \\
\midrule
\endhead
\textcolor{black}{**Basics**} & The image depicts a porous sponge nanomaterial. The scale of the image indicates that one unit of measurement represents 1 µm (micrometer).
\\

\midrule
\textcolor{black}{**Morphology and Structure**} & The nanomaterials in the image display a sponge-like morphology with interconnected pores. There don't appear to be distinct layers, but the structure is characterized by varied pore sizes and interconnecting channels. The nanomaterials pores do not appear uniform in size and shape.
\\

\midrule
\textcolor{black}{**Size and Distribution**} & The approximate size of the individual pores varies, but many appear to be in the range of several hundred nanometers to a few micrometers. The pores and channels of the nanomaterial are distributed throughout the image in a somewhat random manner. There isn't evidence of aggregation or bundling, but rather of an interconnected porous structure.
\\

\midrule
\textcolor{black}{**Surface Characteristics**} & The surface of the nanomaterial appears rough and textured, consistent with a sponge-like structure. There are visible pores of varying sizes throughout the material, which are intrinsic to its sponge-like nature.
\\

\midrule
\textcolor{black}{**Composition and Elements**} & The image is grayscale, so it's hard to determine compositional variations based on color. However, variations in brightness could suggest differences in material thickness or density. The provided labels do not indicate specific elements.
\\
 
\midrule
\textcolor{black}{**Interactions and Boundaries**} & The nanostructures, or pores, appear interconnected, and many of them are touching or fused with adjacent structures. Boundaries between individual pores are clear, but the interconnected nature makes it a continuous structure
\\
 
\midrule
\textcolor{black}{**External Environment**} & There isn't evidence in the image of the nanomaterial interacting with an external environment or matrix. No other distinct structures or objects that aren't part of the nanomaterial are visible in the image.
\\
 
\midrule
\textcolor{black}{**Image Technique and Modifications**} & The image was captured using Scanning Electron Microscopy (SEM), as suggested by the details in the image corner. There's no indication in the provided data about post-processing or modifications made to the image.
\\
 
\midrule
\textcolor{black}{**Functional Features**} & The interconnected pores and channels can be considered functional features, as they may play a role in substance transport, filtration, or other applications. The image appears to be a static representation of the material.
\\

\midrule
\textcolor{black}{**Context and Application**} & The intended application or use of the nanomaterial is not provided in the image or its accompanying data. Based on the data provided and the appearance of the image, it seems like a real, experimental sample rather than a theoretical or simulation-based representation.
\\

\bottomrule
\end{tabularx}
\end{tcolorbox}

\vspace{-7mm}

\begin{tcolorbox}[colback=white!2!white,colframe=black!75!black]
\vspace{-5mm}
\begin{tabularx}{1.2\textwidth}{bsssss}
\\ [-0.985em]
\includegraphics[width=0.145\textwidth]{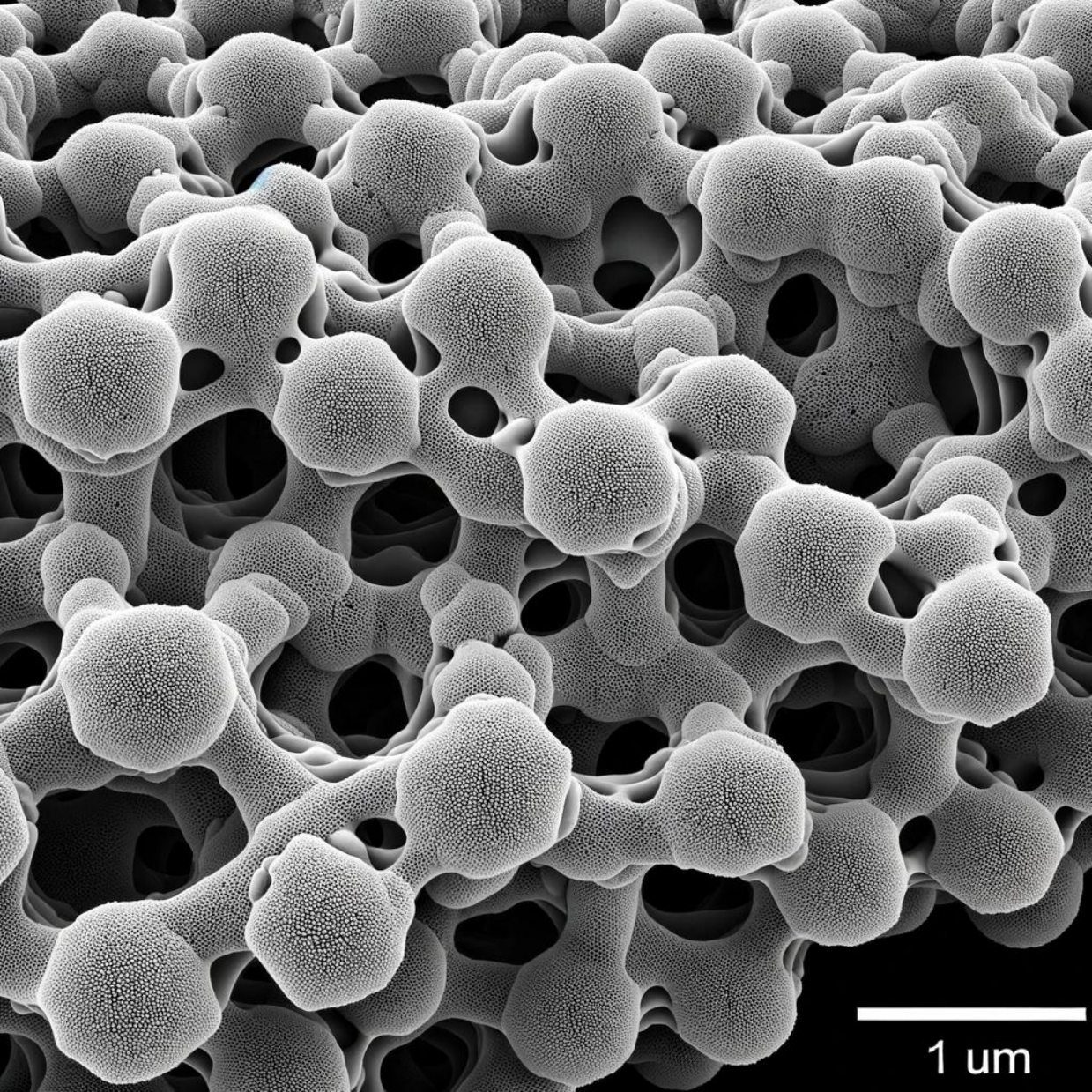} & 
\includegraphics[width=0.145\textwidth]{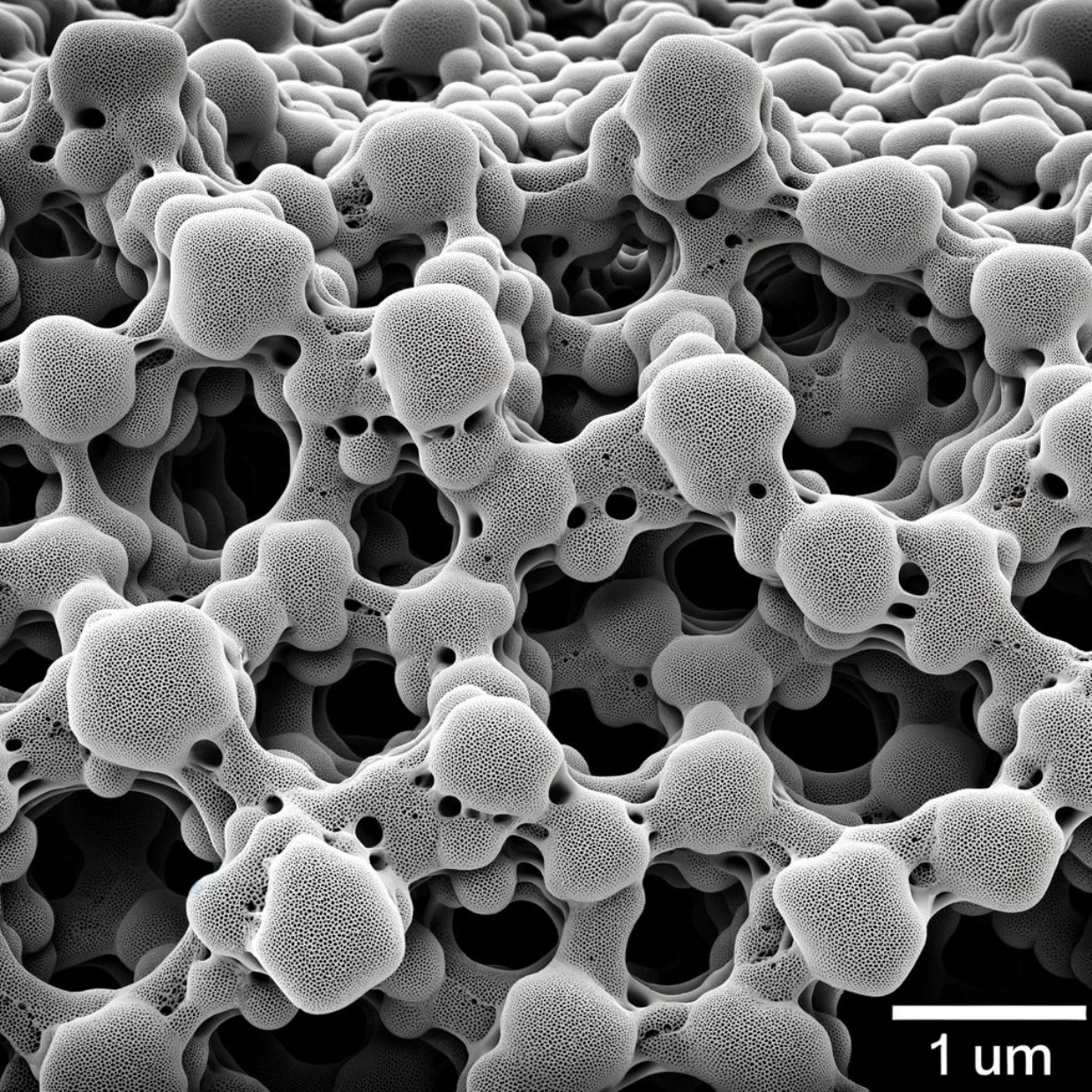} & 
\includegraphics[width=0.145\textwidth]{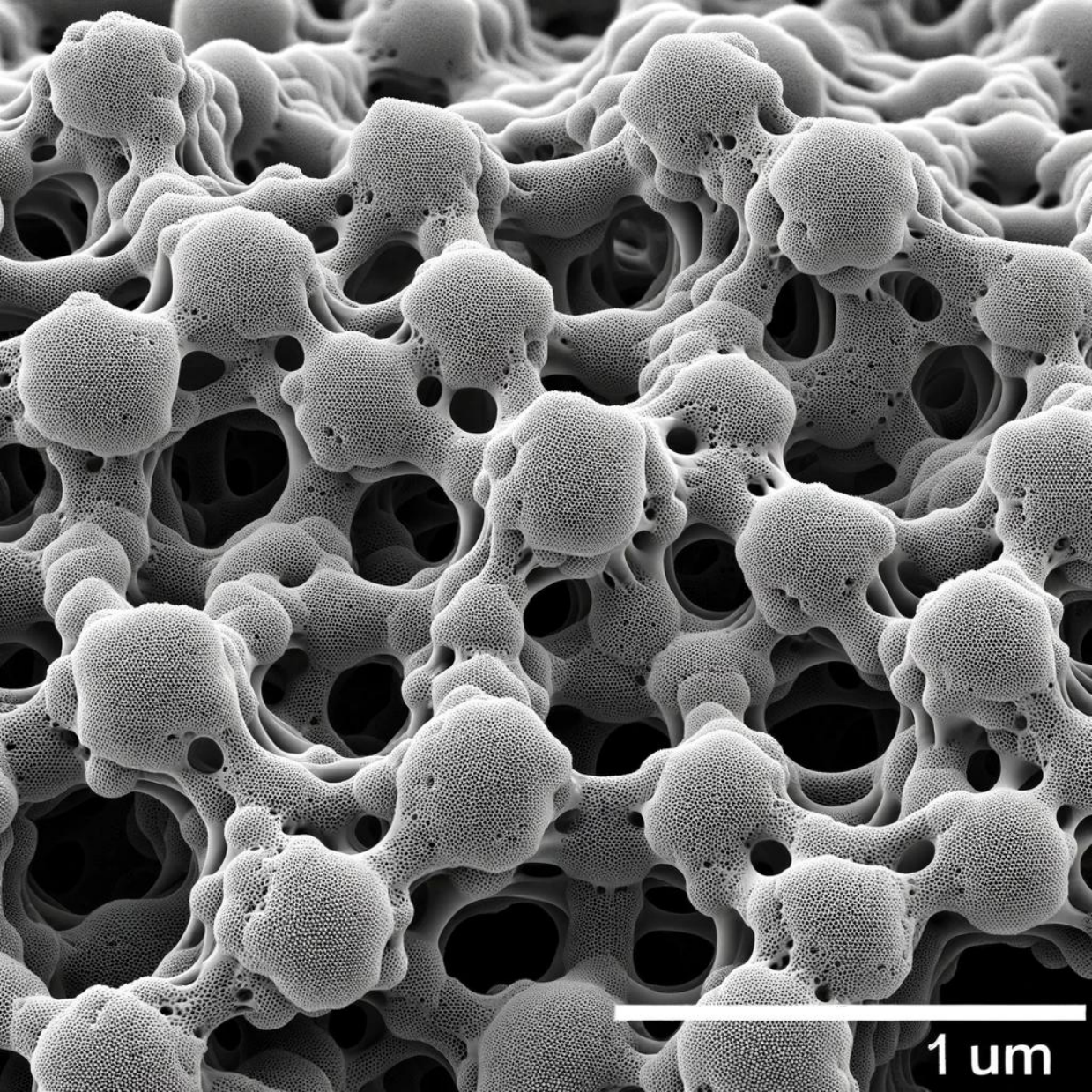} &
\includegraphics[width=0.145\textwidth]{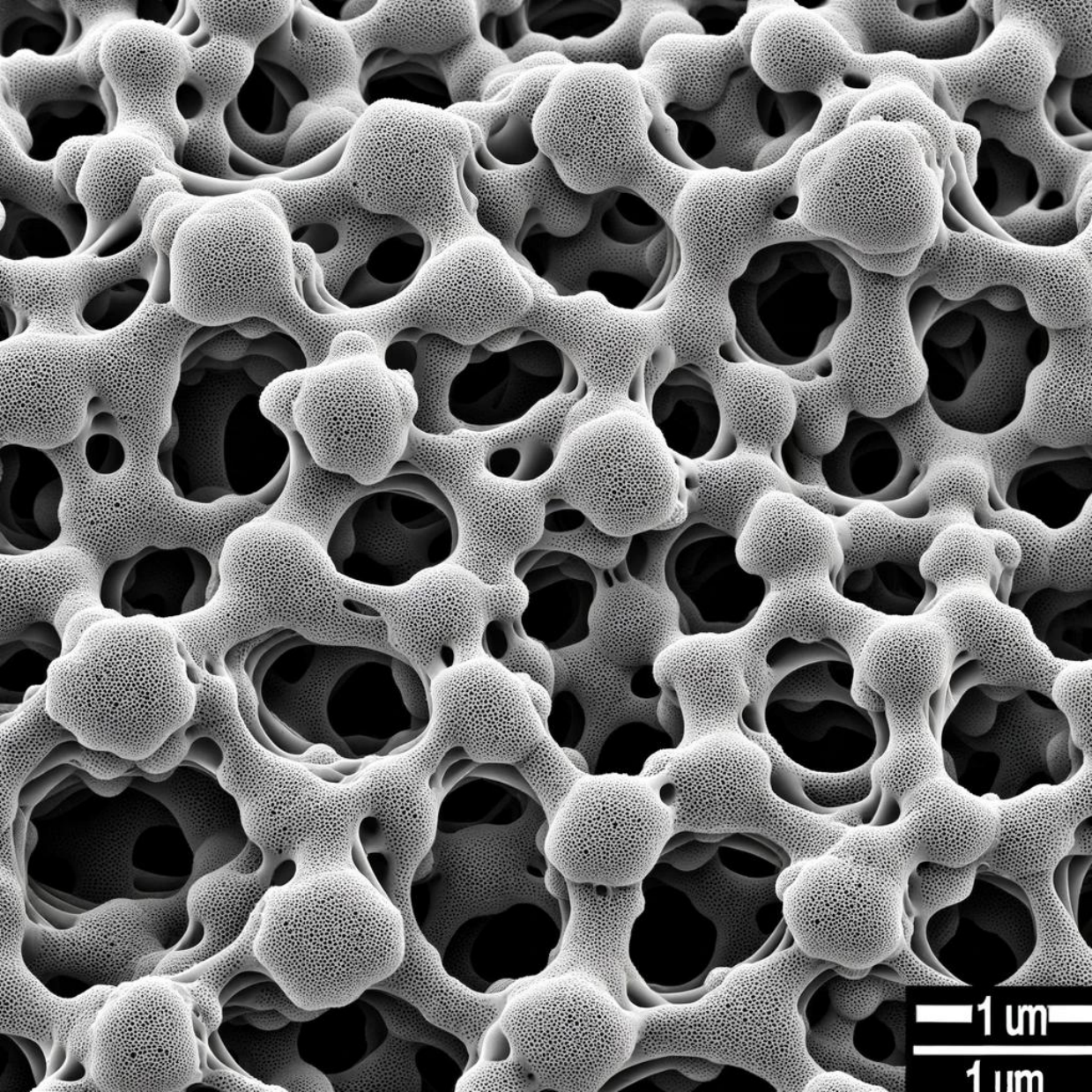} & 
\includegraphics[width=0.145\textwidth]{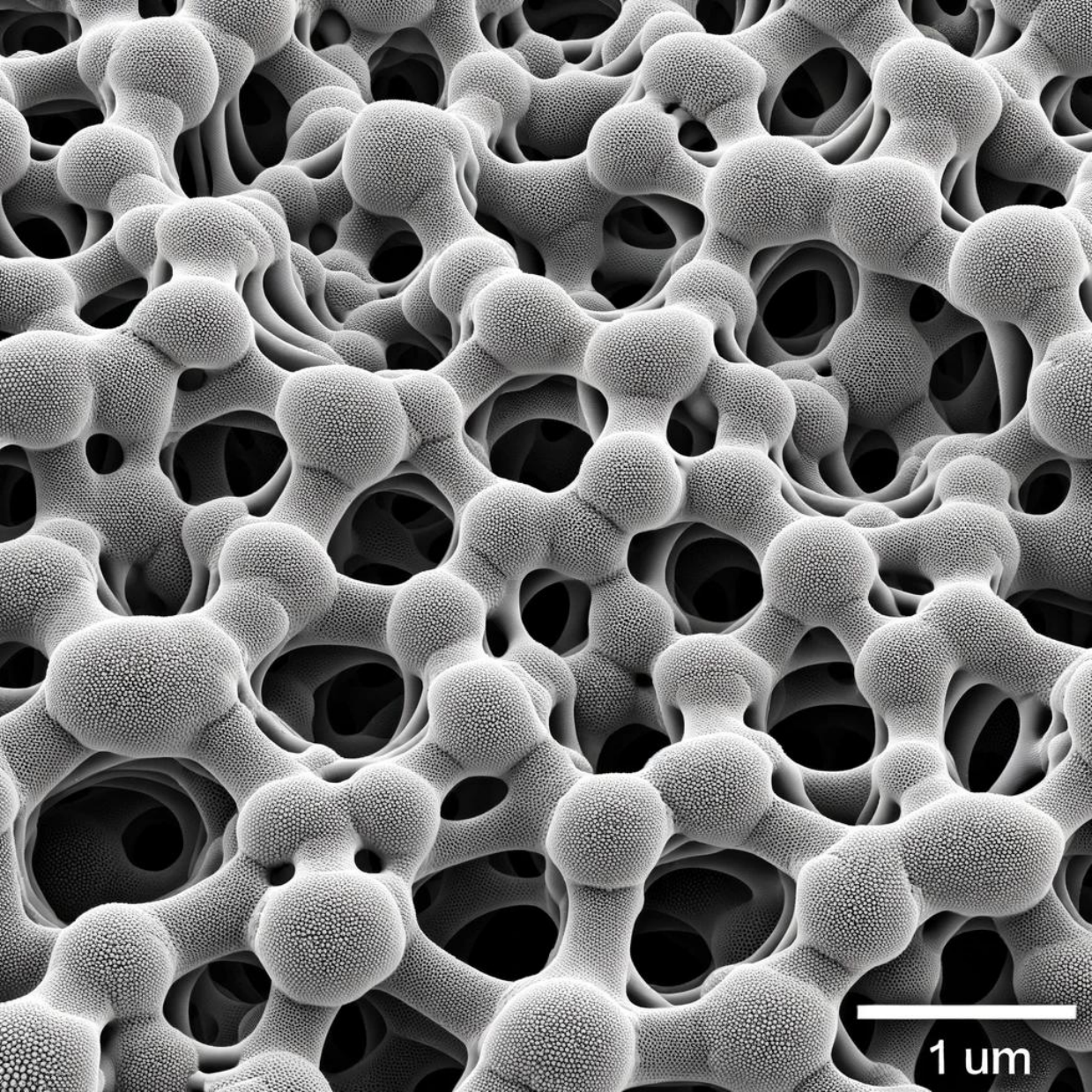} & 
\includegraphics[width=0.145\textwidth]{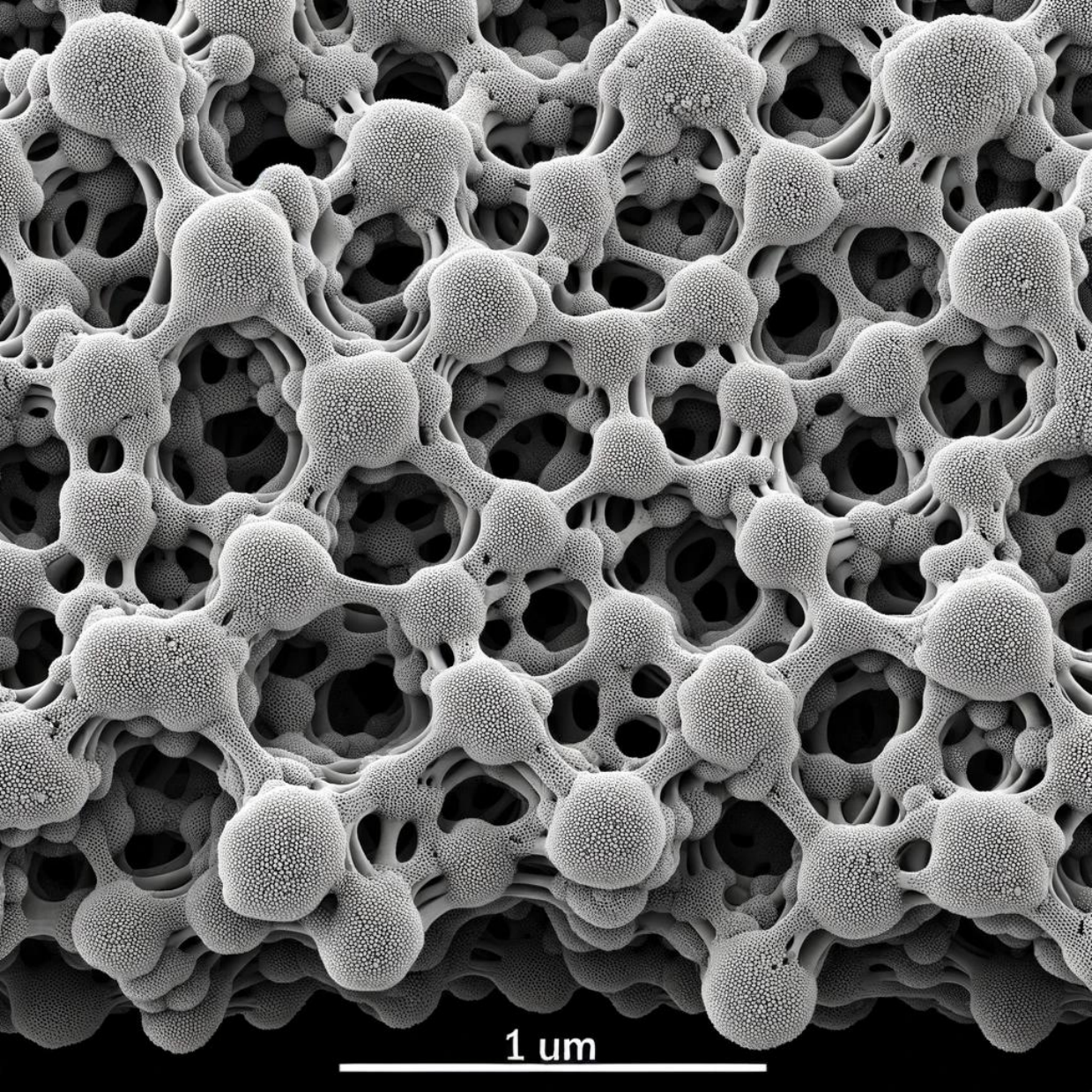} \\ [-0.75em] 
\caption{The table shows sample synthetic images generated by DALLE-3 from textual descriptions provided in a series of question-answer pairs, as shown in Table \ref{tab:tab11}.}
\label{tab:tab12}
\end{tabularx}
\vspace{-3mm}
\end{tcolorbox}

\begin{tcolorbox}[colback=white!2!white,colframe=black!75!black]
\vspace{-5mm}
\begin{tabularx}{1.2\textwidth}{bss}
\caption{The table presents the question-answer pairs generated by the LMM(GPT-4V) in response to natural language questions about the input image of the tips nanomaterial category. In summary, the question-answer pairs (generated text) provide in-depth information about an image related to tips nanomaterials.}\label{tab:tab13}\\ [-0.75em]
\toprule 
 \\ [-3.5em]
\begin{overpic}[width=0.165\columnwidth]{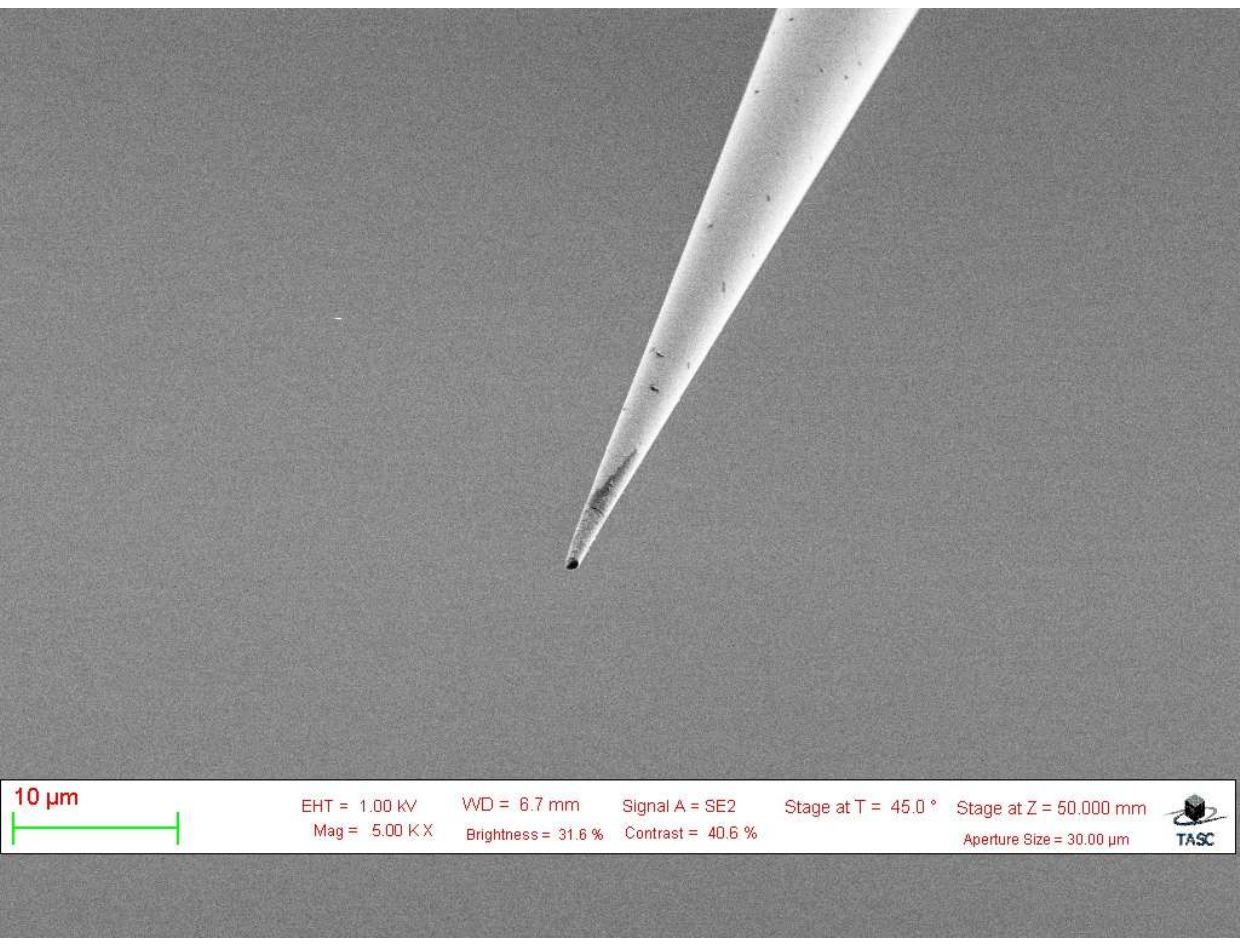}
  \put(300,50){\makebox[0pt][c]{%
    \parbox{0.5\columnwidth}{%
      \centering
      \textbf{Please answer the following questions based on the provided input image belonging to the tips nanomaterial category.} 
    }%
  }}  
\end{overpic}  \\ [-2.5em]

\midrule
\textbf{Query prompt}  & \textbf{GPT-4 Vision output} \\
\midrule
\endhead
\textcolor{black}{**Basics**} & The image appears to depict a sharp nanotip, possibly used in probe microscopy or other related applications. The scale bar in the image is labeled ``10 µm", indicating that one unit of measurement represents 10 micrometers. 
\\

\midrule
\textcolor{black}{**Morphology and Structure**} & The general shape or morphology of the nanomaterial is a sharp, elongated tip. There do not appear to be distinct layers, phases, or domains visible. The nanomaterial seems to be uniform in shape.  
\\

\midrule
\textcolor{black}{**Size and Distribution**} &  The approximate size of the individual nanostructure can be gauged using the scale bar. The tip appears to be several micrometers in length. The image focuses on one nanomaterial, so there is no distribution to assess. There is no evidence of aggregation or bundling.
\\

\midrule
\textcolor{black}{**Surface Characteristics**} & The nanomaterial appears relatively smooth with minor roughness. There seem to be some small dark spots which could be defects, pores, or impurities, but it's challenging to say for sure without more information.
\\

\midrule
\textcolor{black}{**Composition and Elements**} &  The image is in grayscale, so there are no evident compositional variations based on color. However, the variations in brightness and contrast could suggest differences in material composition or thickness. The image has no explicit labels indicating specific elements or compounds.
\\
 
\midrule
\textcolor{black}{**Interactions and Boundaries**} &  Since the image primarily shows a single nanotip, there's no interaction with other nanostructures visible. The boundaries of the nanotip are clear and distinct from the background.
\\
 
\midrule
\textcolor{black}{**External Environment**} &  There's no evidence of interaction with surrounding materials or matrix. The background appears uniform. Other than the nanotip, there are no other distinct structures or objects evident.
\\
 
\midrule
\textcolor{black}{**Image Technique and Modifications**} & The image likely comes from a Scanning Electron Microscope (SEM) based on the data present on the image, such as ``EHT", ``Mag," and ``Signal A", There's no evidence in the image to suggest post-processing or modifications, but the grayscale nature of SEM images is typical, and there's no indication of false coloring.
\\
 
\midrule
\textcolor{black}{**Functional Features**} &  There aren't any distinct functional features visible. The sharpness of the tip suggests it might be used for precise interactions at the nanoscale. The image seems to be a static representation.
\\

\midrule
\textcolor{black}{**Context and Application**} &  The nanotip could be intended for applications like scanning probe microscopy, electron field emission, or other nanoscale interactions. The image appears to be of a real, experimental sample given the details and imperfections visible.
\\ [-0.05em]

\bottomrule
\end{tabularx}
\end{tcolorbox}

\vspace{-7mm}

\begin{tcolorbox}[colback=white!2!white,colframe=black!75!black]
\vspace{-6mm}
\begin{tabularx}{1.2\textwidth}{bsssss}
\\ [-2.65em]
\includegraphics[width=0.135\textwidth]{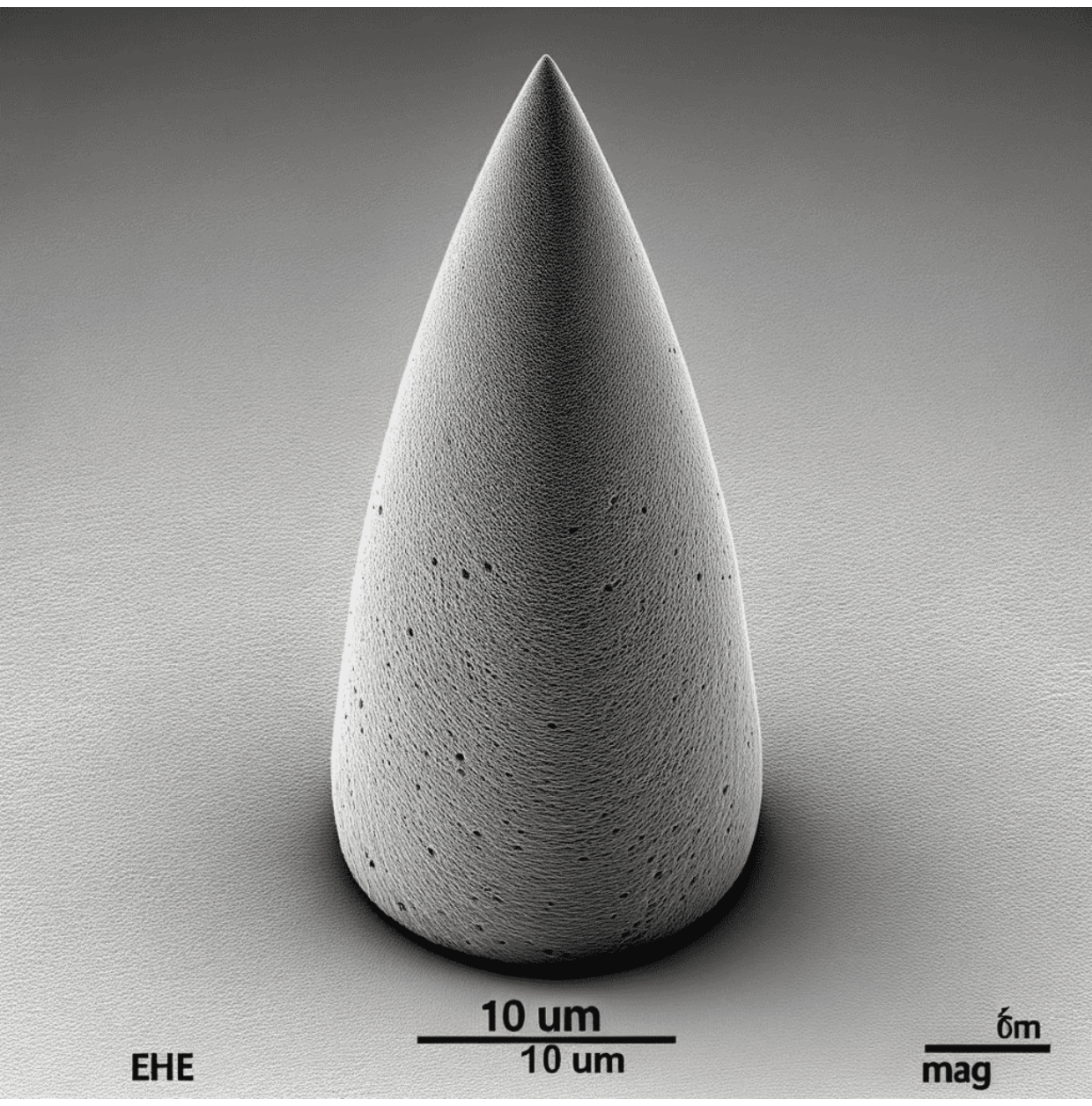} & 
\includegraphics[width=0.135\textwidth]{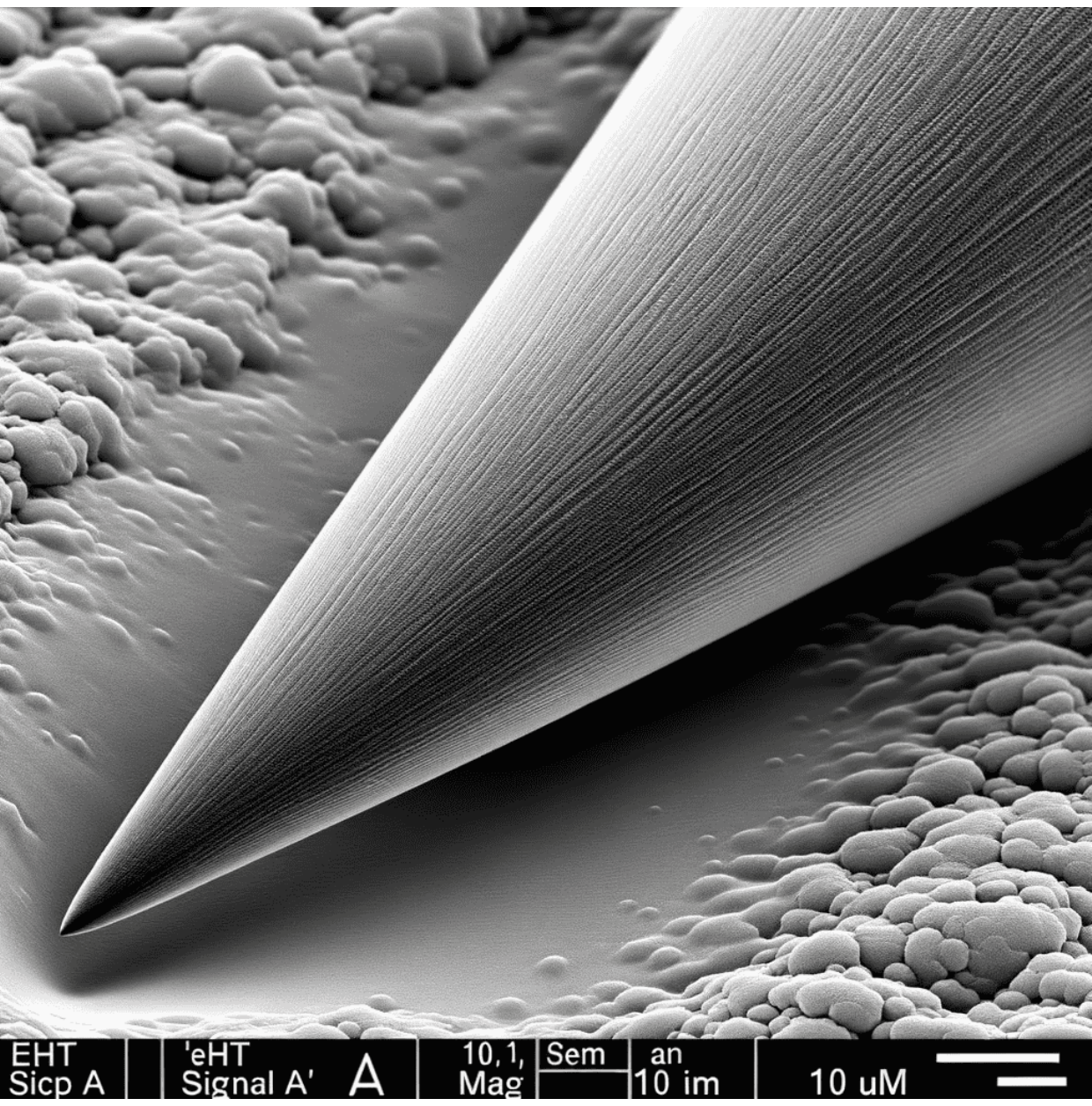} & 
\includegraphics[width=0.135\textwidth]{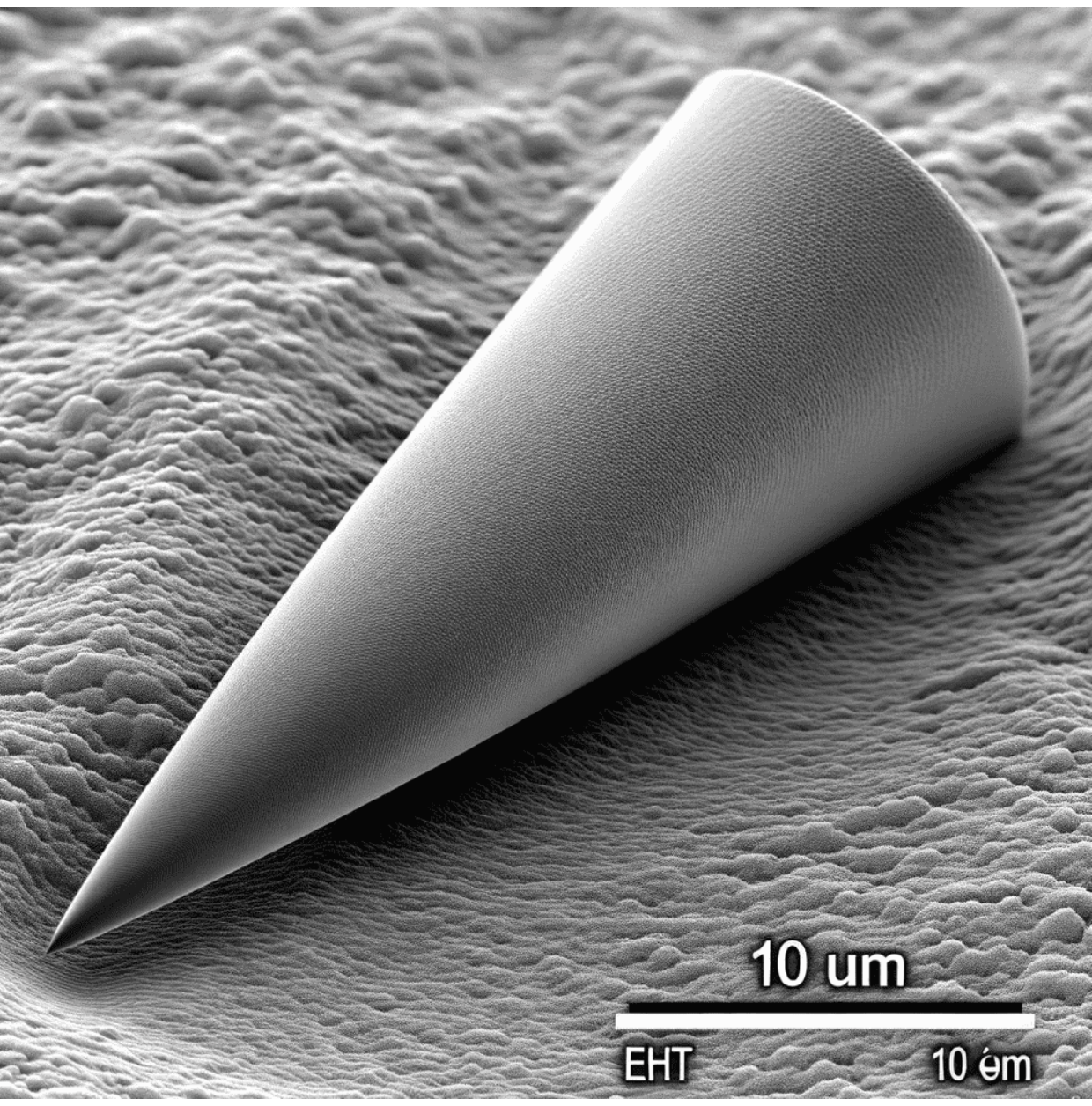} &
\includegraphics[width=0.135\textwidth]{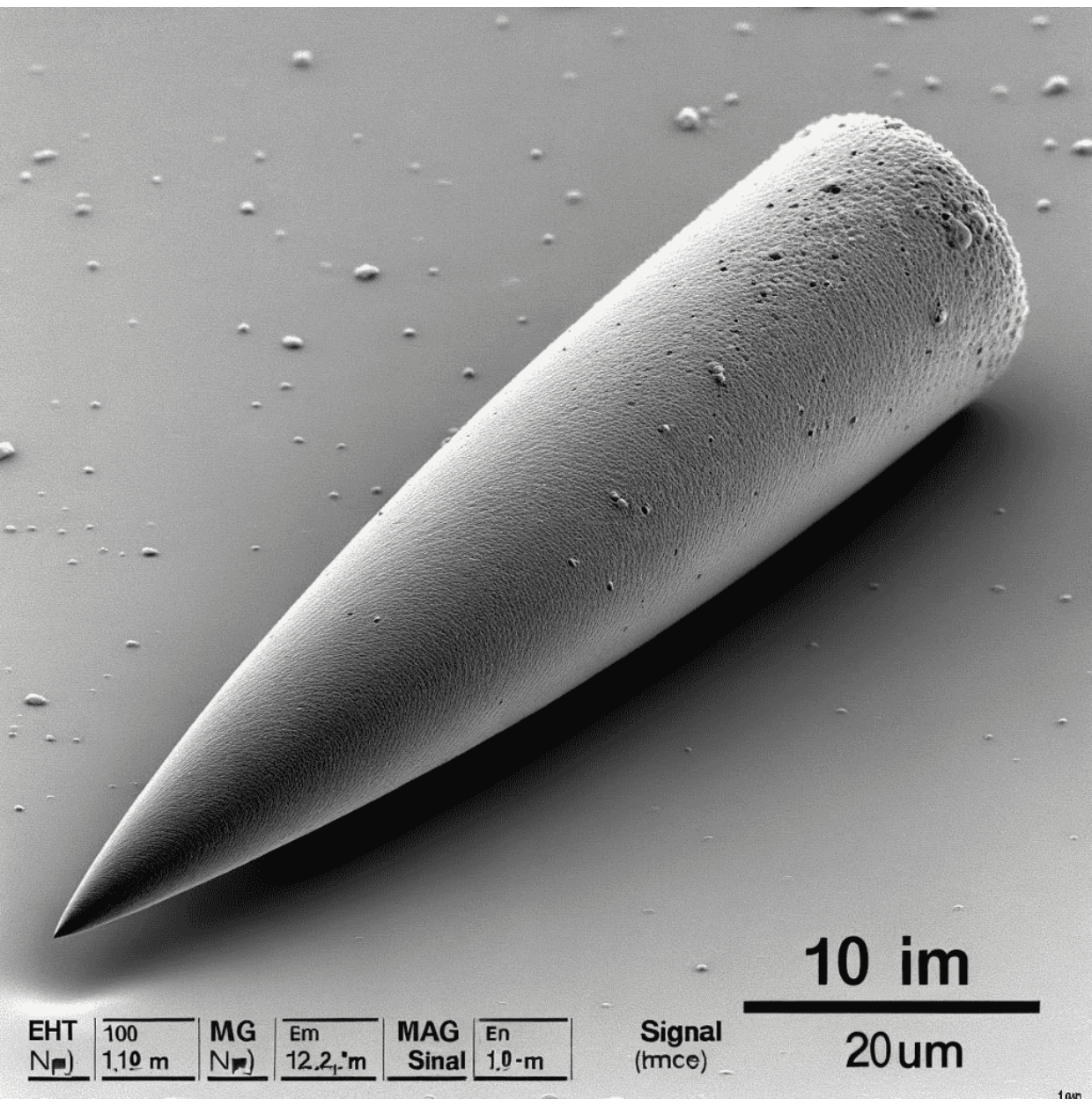} & 
\includegraphics[width=0.135\textwidth]{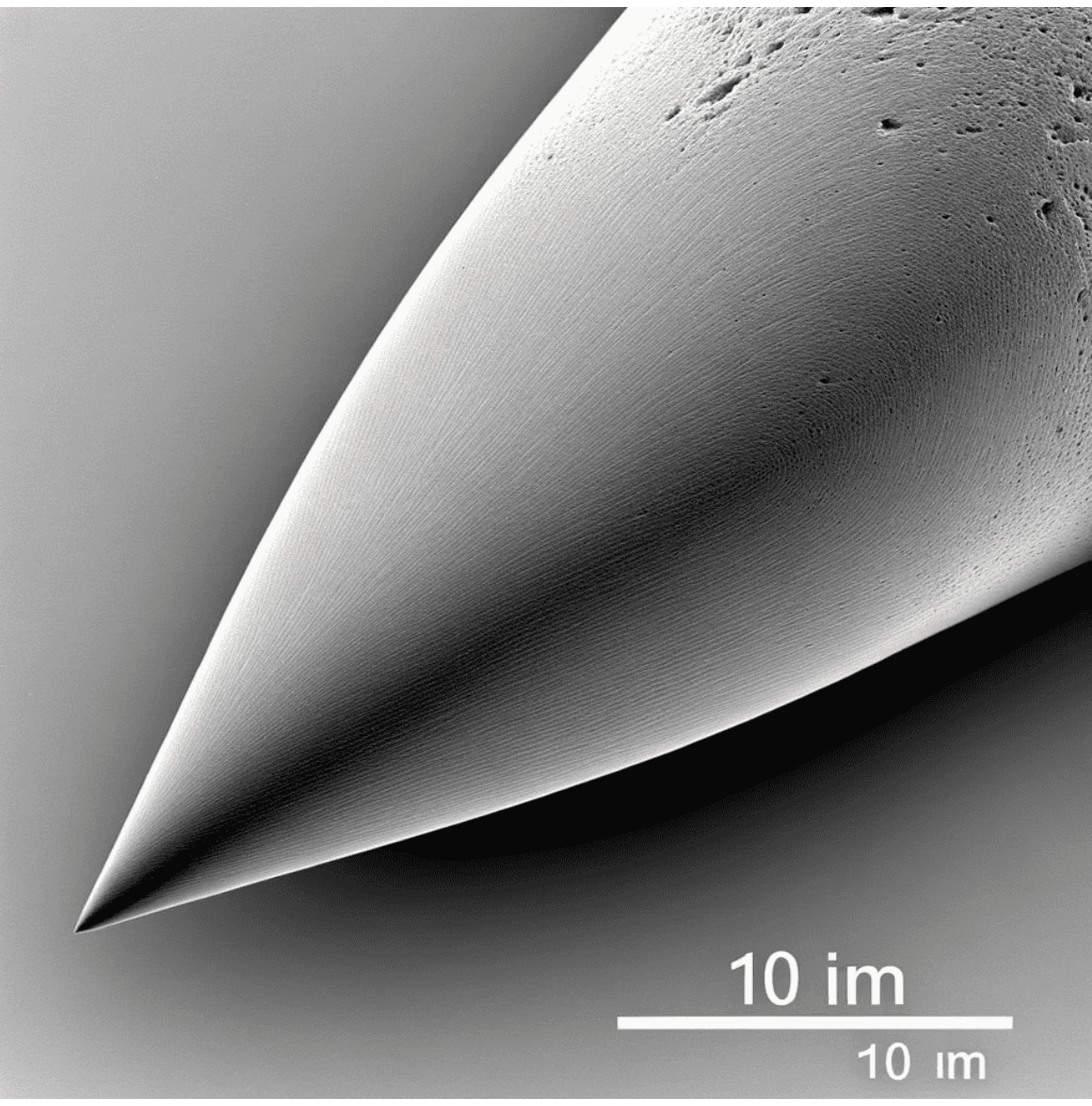} & 
\includegraphics[width=0.135\textwidth]{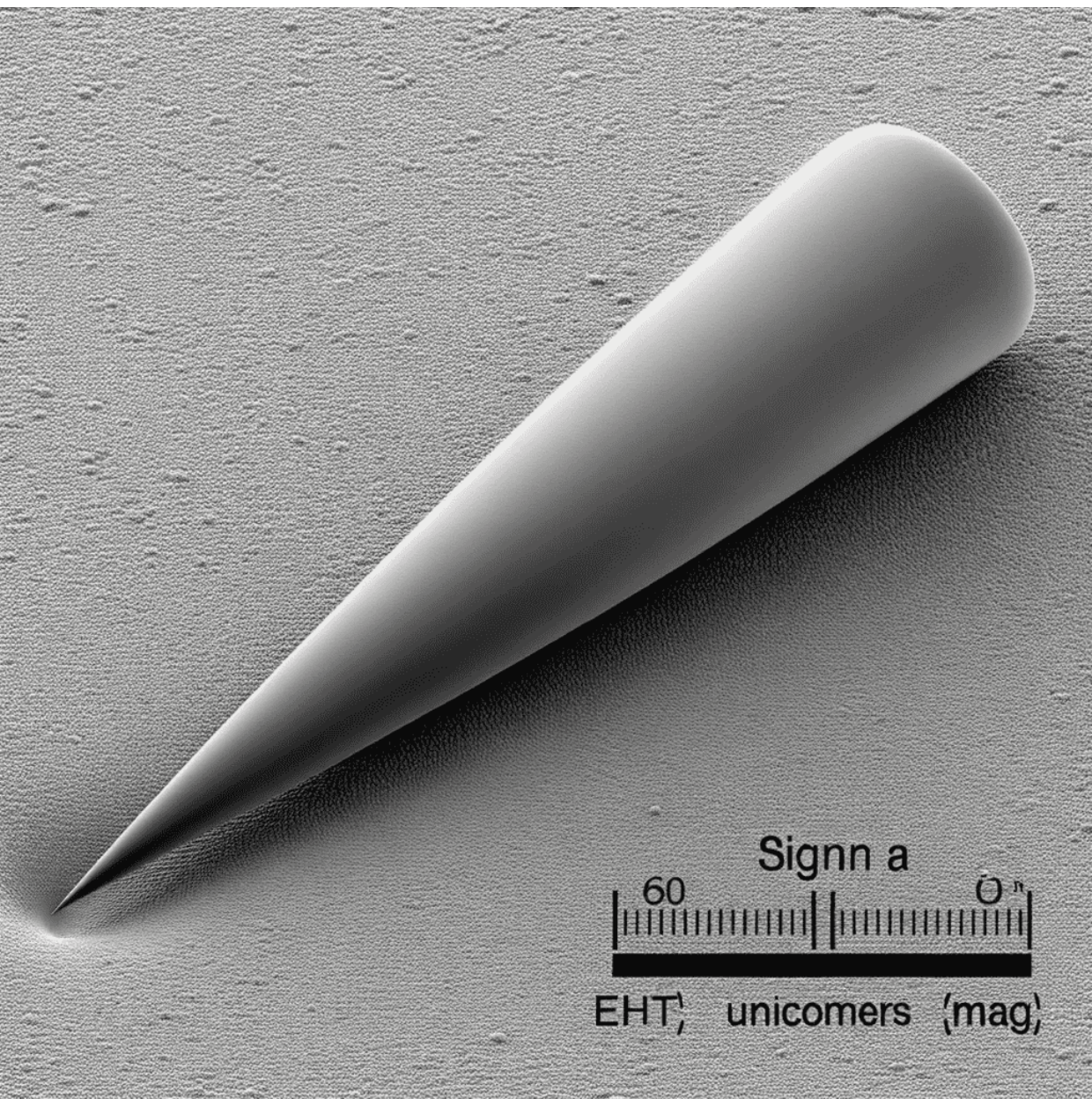} \\ [-2.25em] 
\caption{The table shows sample synthetic images generated by DALLE-3 from textual descriptions provided in a series of question-answer pairs, as shown in Table \ref{tab:tab13}.}
\label{tab:tab14}
\end{tabularx}
\vspace{-3mm}
\end{tcolorbox}

\begin{tcolorbox}[colback=white!2!white,colframe=black!75!black]
\vspace{-5mm}
\begin{tabularx}{1.2\textwidth}{bss}
\caption{The table presents the question-answer pairs generated by the LMM(GPT-4V) in response to natural language questions about the input image of the particles nanomaterial category. In summary, the question-answer pairs (generated text) provide in-depth information about an image related to particles nanomaterials.}\label{tab:tab15}\\ [-0.75em]
\toprule 
 \\ [-0.75em]
\begin{overpic}[width=0.165\columnwidth]{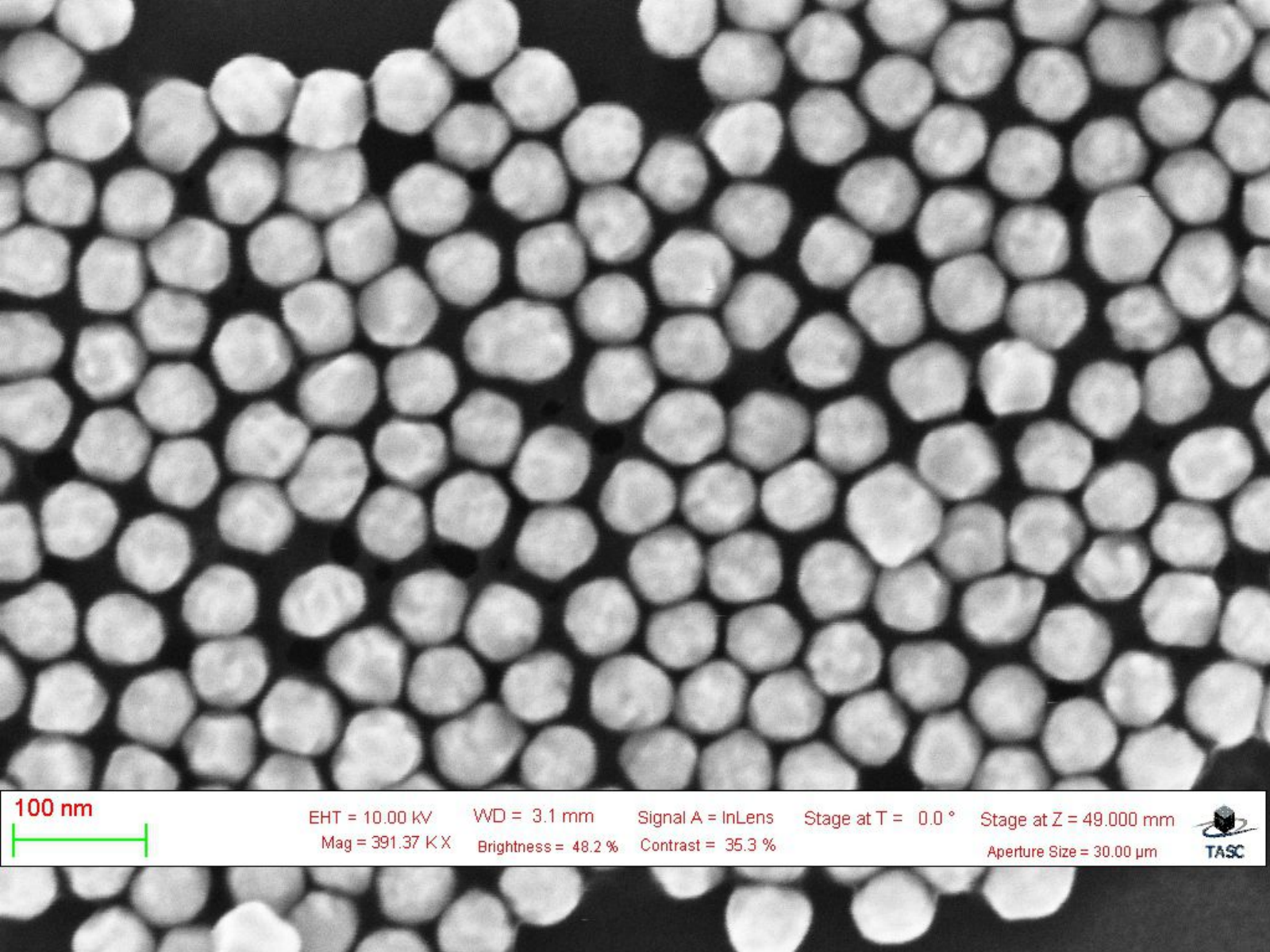}
  \put(400,40){\makebox[0pt][c]{%
    \parbox{0.5\columnwidth}{%
      \centering
      \textbf{Please answer the following questions based on the provided input image belonging to the particles nanomaterial category.} 
    }%
  }}  
\end{overpic}  \\ [-0.25em]

\midrule
\textbf{Query prompt}  & \textbf{GPT-4 Vision output} \\
\midrule
\endhead
\textcolor{black}{**Basics**} & The image appears to depict spherical nanoparticles.
   - The scale of the image is indicated by the "100 nm" scale bar, meaning one unit of measurement represents 100 nanometers.

\\

\midrule
\textcolor{black}{**Morphology and Structure**} & The nanoparticles have a spherical or nearly spherical shape. There are no distinct layers, phases, or domains visible in the image. The nanoparticles appear to be quite uniform in size and shape.
\\

\midrule
\textcolor{black}{**Size and Distribution**} & The approximate size of the individual nanostructures can be gauged from the 100 nm scale bar, suggesting they are likely slightly below 100 nm in diameter. The nanoparticles are distributed in a closely packed arrangement, resembling an organized pattern. There doesn't appear to be evidence of aggregation or bundling.
\\

\midrule
\textcolor{black}{**Surface Characteristics**} & The nanomaterial appears to have a smooth surface. No visible defects, pores, or impurities on the surface are discernible from this image.
\\

\midrule
\textcolor{black}{**Composition and Elements**} & There isn't evidence of compositional variations in the image based on the grayscale nature of the image. No labels or markers indicating specific elements or compounds are present in the image.
\\
 
\midrule
\textcolor{black}{**Interactions and Boundaries**} & The individual nanostructures appear to be in close proximity but not necessarily fused. They are separate from one another. Clear boundaries between the different nanoparticles are visible.
\\
 
\midrule
\textcolor{black}{**External Environment**} & There is no evidence of the nanomaterial interacting with its surrounding environment or matrix. The entire image appears to only showcase the nanoparticles.

\\
 
\midrule
\textcolor{black}{**Image Technique and Modifications**} & The image was captured using Scanning Electron Microscopy (SEM) as indicated by the image details. There's no evidence in the image to suggest post-processing or modifications such as false coloring or 3D rendering.
\\
 
\midrule
\textcolor{black}{**Functional Features**} & No specific functional features, such as active sites or regions with distinct properties, are visible in this image. The image appears to be a static representation.
\\

\midrule
\textcolor{black}{**Context and Application**} & The intended application or use of the nanomaterial isn't provided in the image or context. It appears to be a real, experimental sample given the details and nature of the image.
\\

\bottomrule
\end{tabularx}
\end{tcolorbox}

\vspace{-8mm}

\begin{tcolorbox}[colback=white!2!white,colframe=black!75!black]
\vspace{-5mm}
\begin{tabularx}{1.2\textwidth}{bsssss}
\\ [-0.45em]
\includegraphics[width=0.145\textwidth]{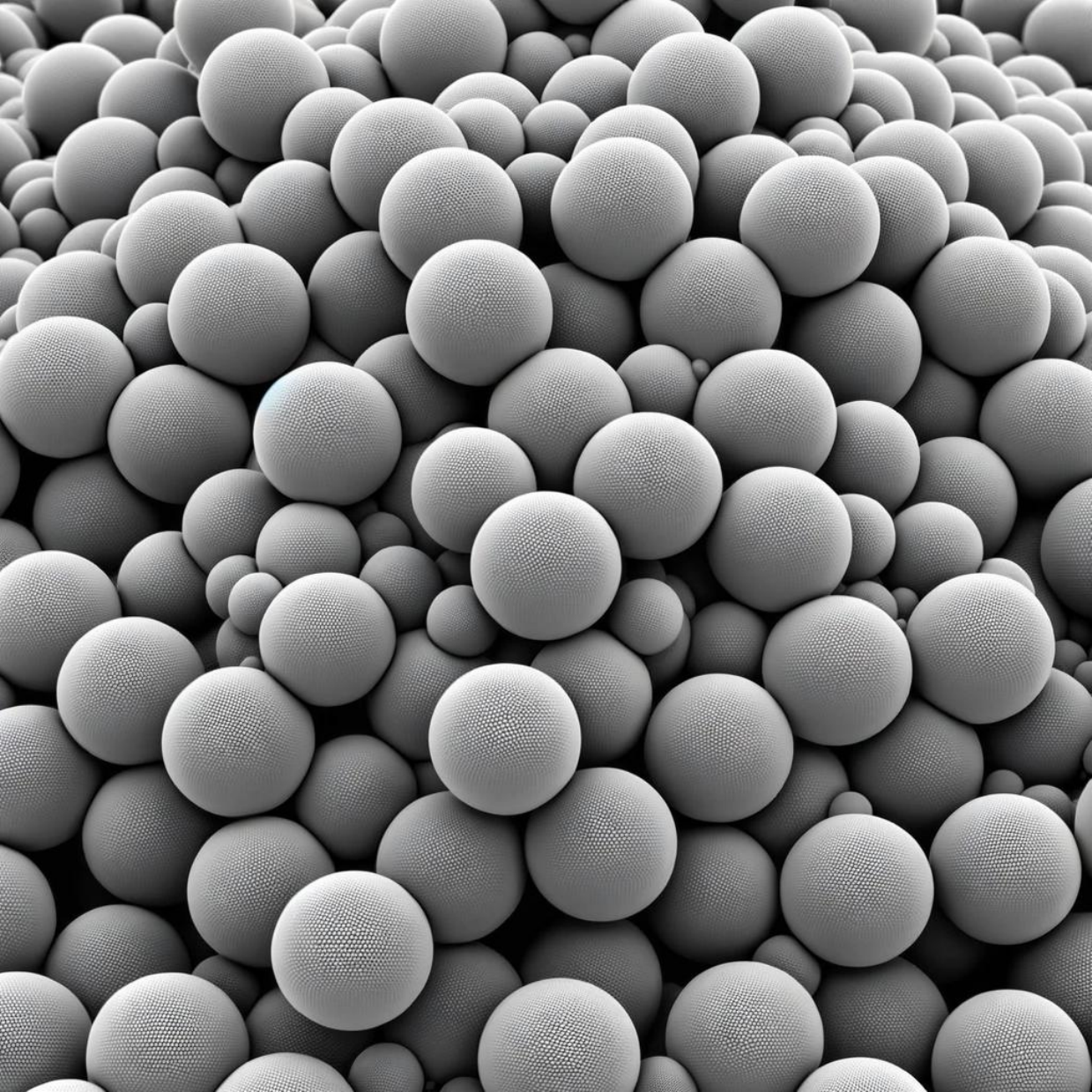} & 
\includegraphics[width=0.145\textwidth]{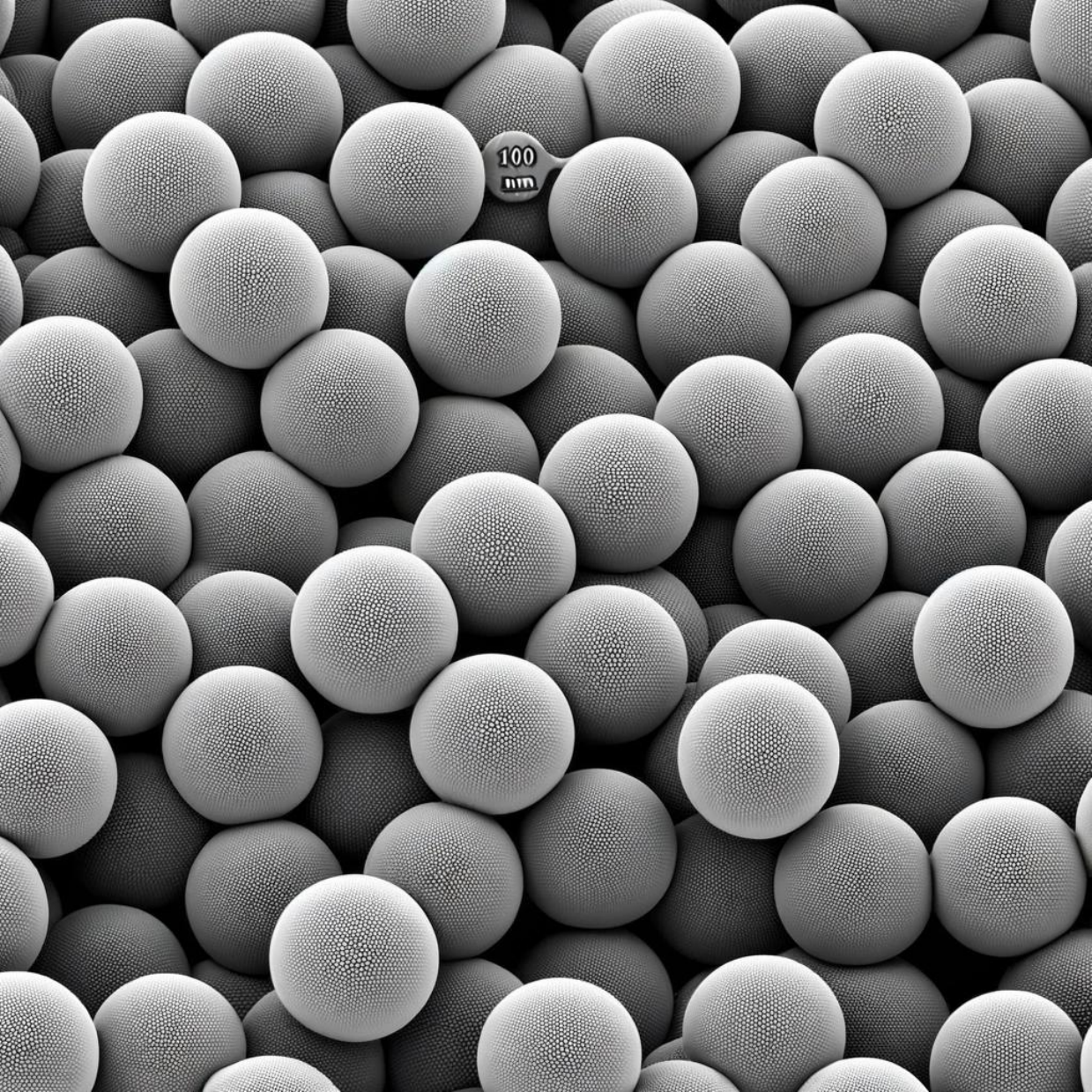} & 
\includegraphics[width=0.145\textwidth]{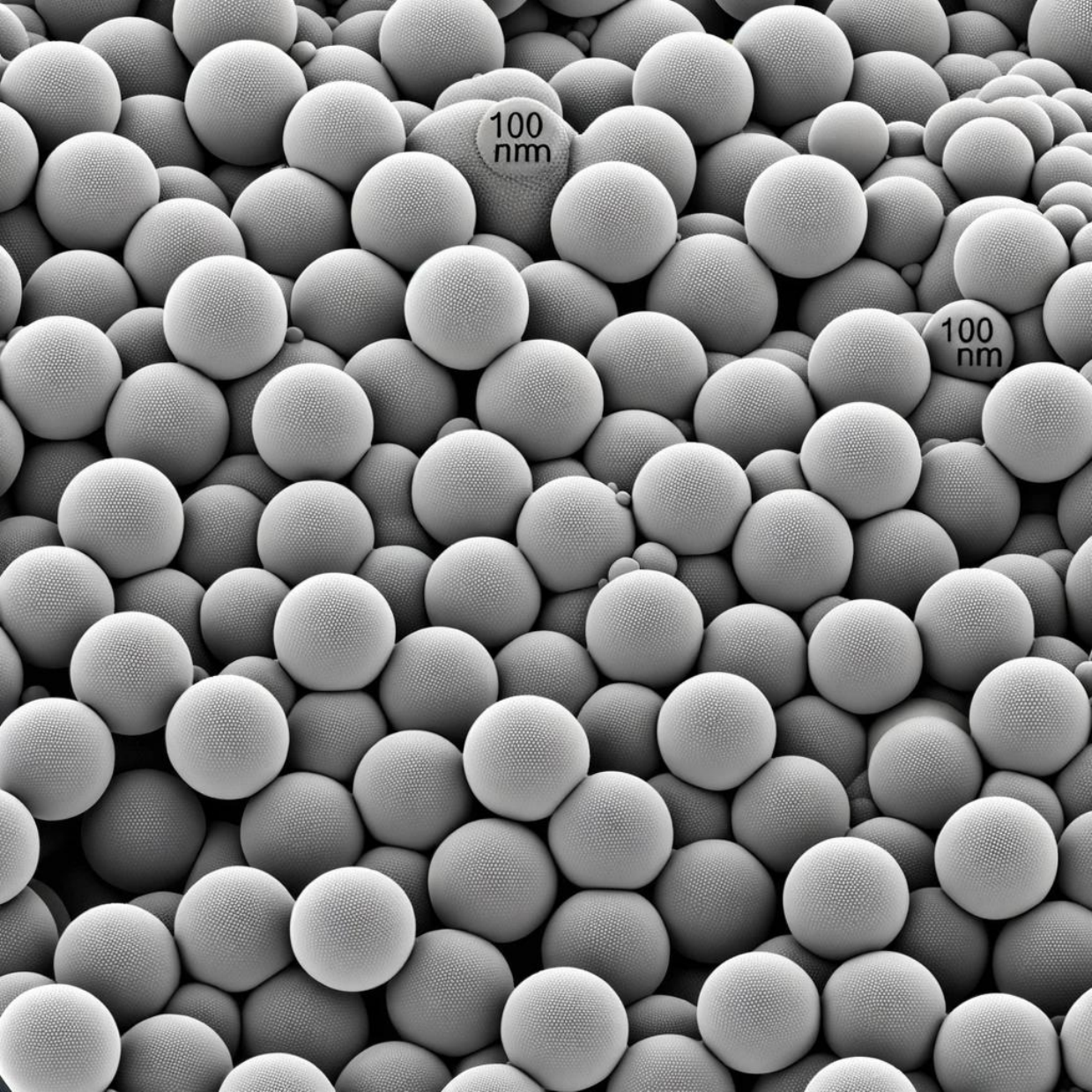} &
\includegraphics[width=0.145\textwidth]{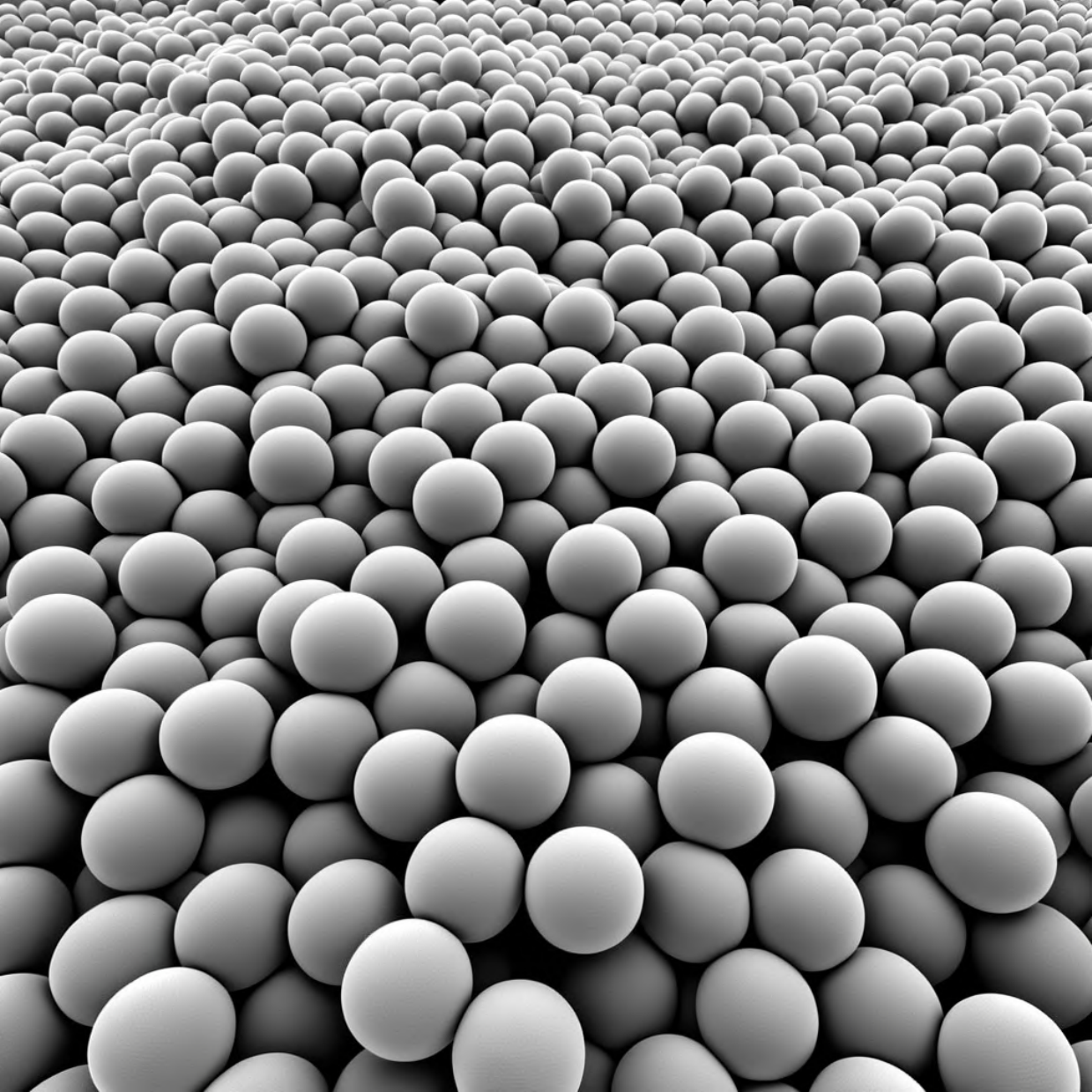} & 
\includegraphics[width=0.145\textwidth]{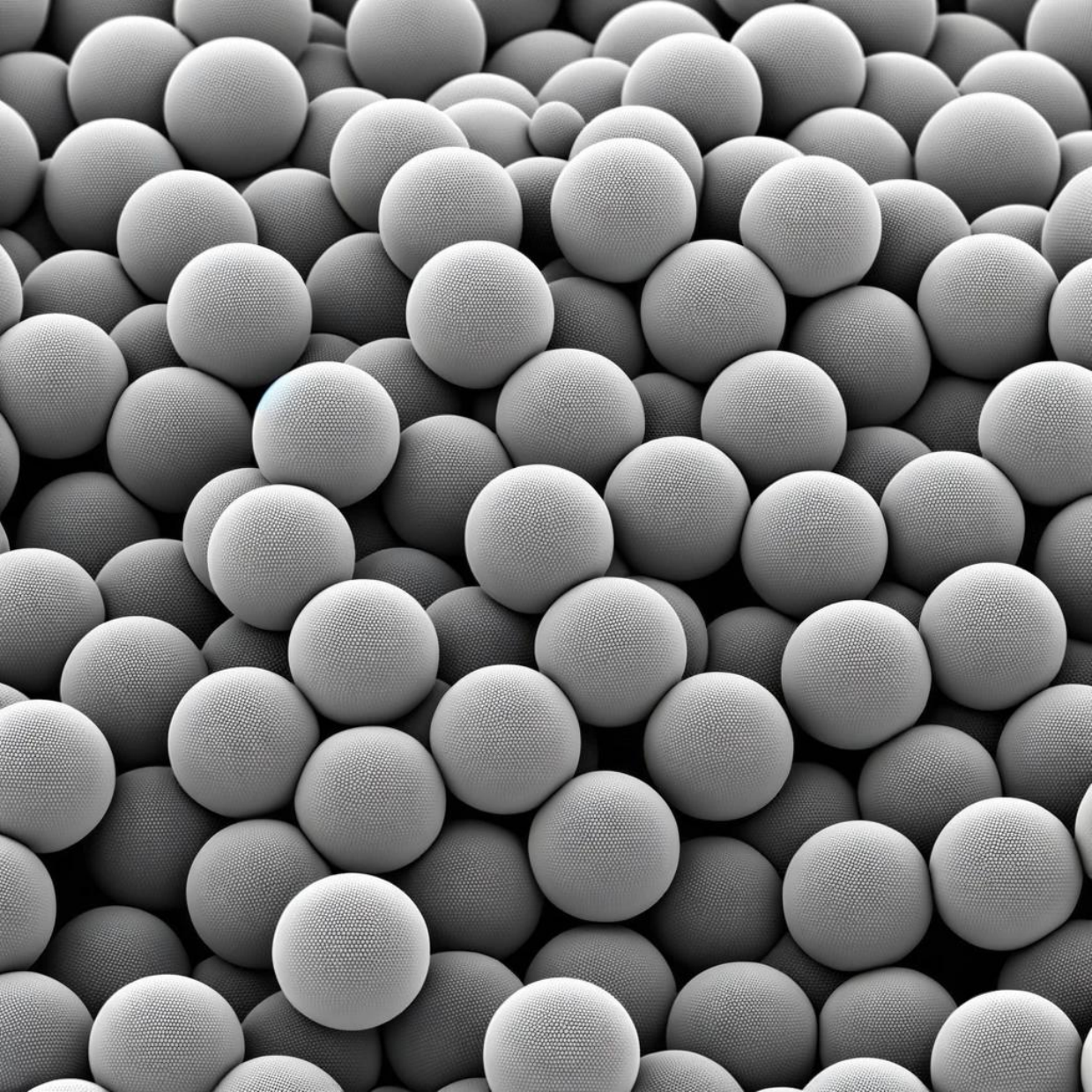} & 
\includegraphics[width=0.145\textwidth]{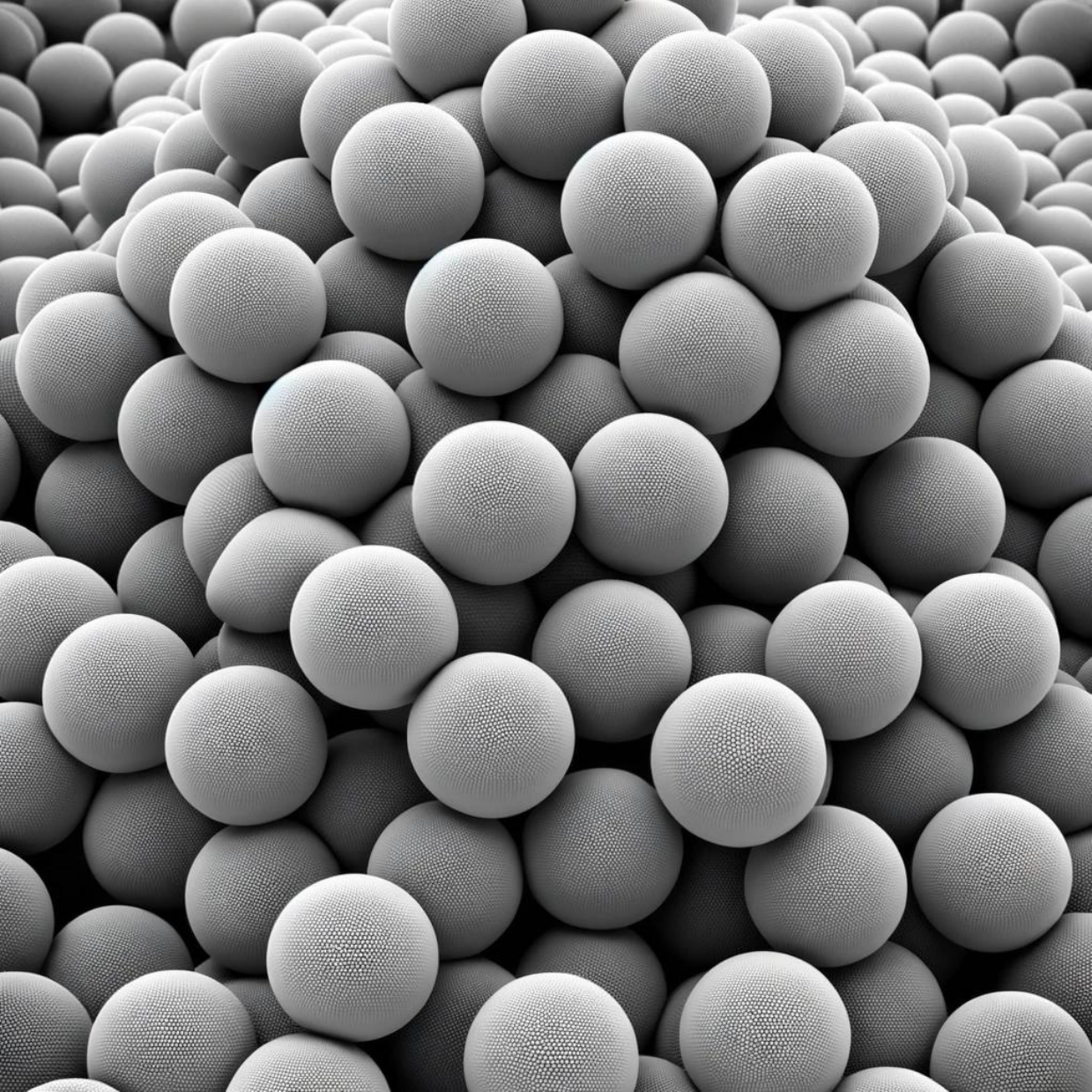} \\ [-0.35em] 
\caption{The table shows sample synthetic images generated by DALLE-3 from textual descriptions provided in a series of question-answer pairs, as shown in Table \ref{tab:tab15}.}
\label{tab:tab16}
\end{tabularx}
\vspace{-3mm}
\end{tcolorbox}

\begin{tcolorbox}[colback=white!2!white,colframe=black!75!black]
\vspace{-5mm}
\begin{tabularx}{1.2\textwidth}{bss}
\caption{The table presents the question-answer pairs generated by the LMM(GPT-4V) in response to natural language questions about the input image of the powder nanomaterial category. In summary, the question-answer pairs (generated text) provide in-depth information about an image related to powder nanomaterials.}\label{tab:tab17}\\ [-0.85em]
\toprule 
\begin{overpic}[width=0.165\columnwidth]{Images/GPT4V-Images/Powder/Image.pdf}
  \put(400,40){\makebox[0pt][c]{%
    \parbox{0.4\columnwidth}{%
      \centering
      \textbf{Please answer the following questions based on the provided input image belonging to the powder nanomaterial category.} 
    }%
  }}  
\end{overpic}  \\ [-0.25em]

\midrule
\textbf{Query prompt}  & \textbf{GPT-4 Vision output} \\
\midrule
\endhead
\textcolor{black}{**Basics**} & The image depicts a nanomaterial in the form of a powder. The scale of the image is denoted by ``2$\mu$m", meaning one unit of the scale bar represents 2 micrometers. 
\\

\midrule
\textcolor{black}{**Morphology and Structure**} & The nanomaterials have a somewhat irregular, clumped morphology resembling that of aggregated particles. There are no distinct layers, phases, or domains visible. The nanomaterials do not appear uniform in size and shape; there are varied sizes and shapes present.  
\\

\midrule
\textcolor{black}{**Size and Distribution**} &  The size of the individual nanostructures varies, with some smaller granules and some larger aggregates. It is hard to give an exact size without further information, but they are definitely within the nanoscale to microscale range. The nanomaterials are clustered together. There is evidence of aggregation as many particles are clumped together.
\\

\midrule
\textcolor{black}{**Surface Characteristics**} & The nanomaterial appears rough in texture. There are no visible defects, pores, or impurities distinguishable at this resolution.
\\

\midrule
\textcolor{black}{**Composition and Elements**} &  The image is monochrome, so there is no direct evidence of compositional variations based on color. The varied brightness might suggest different materials or thicknesses, but it's not definitive. There are no labels or markers indicating specific elements or compounds.
\\
 
\midrule
\textcolor{black}{**Interactions and Boundaries**} & Many of the individual nanostructures are touching or fused together. There are clear boundaries between some structures, but many are aggregated together.
\\
 
\midrule
\textcolor{black}{**External Environment**} & There's no evidence of the nanomaterial interacting with any surrounding environment or matrix in this image. All structures in the image appear to be part of the nanomaterial powder.
\\
 
\midrule
\textcolor{black}{**Image Technique and Modifications**} & The image was captured using Scanning Electron Microscopy (SEM), as suggested by the information provided at the bottom of the image. There's no evidence in the image itself of post-processing or modifications such as false coloring or 3D rendering.
\\
 
\midrule
\textcolor{black}{**Functional Features**} & No specific functional features are visible in the image. The image appears to be a static representation.
\\

\midrule
\textcolor{black}{**Context and Application**} & The intended application or use of the nanomaterial is not provided in the image. Given the detailed nature and the labels present, this appears to be a real, experimental sample rather than a theoretical or simulation-based representation.
\\

\bottomrule
\end{tabularx}
\end{tcolorbox}

\vspace{-8mm}

\begin{tcolorbox}[colback=white!2!white,colframe=black!75!black]
\vspace{-5mm}
\begin{tabularx}{1.5\textwidth}{bsssss}
\\ [-1.1em]
\includegraphics[height=0.09\textwidth]{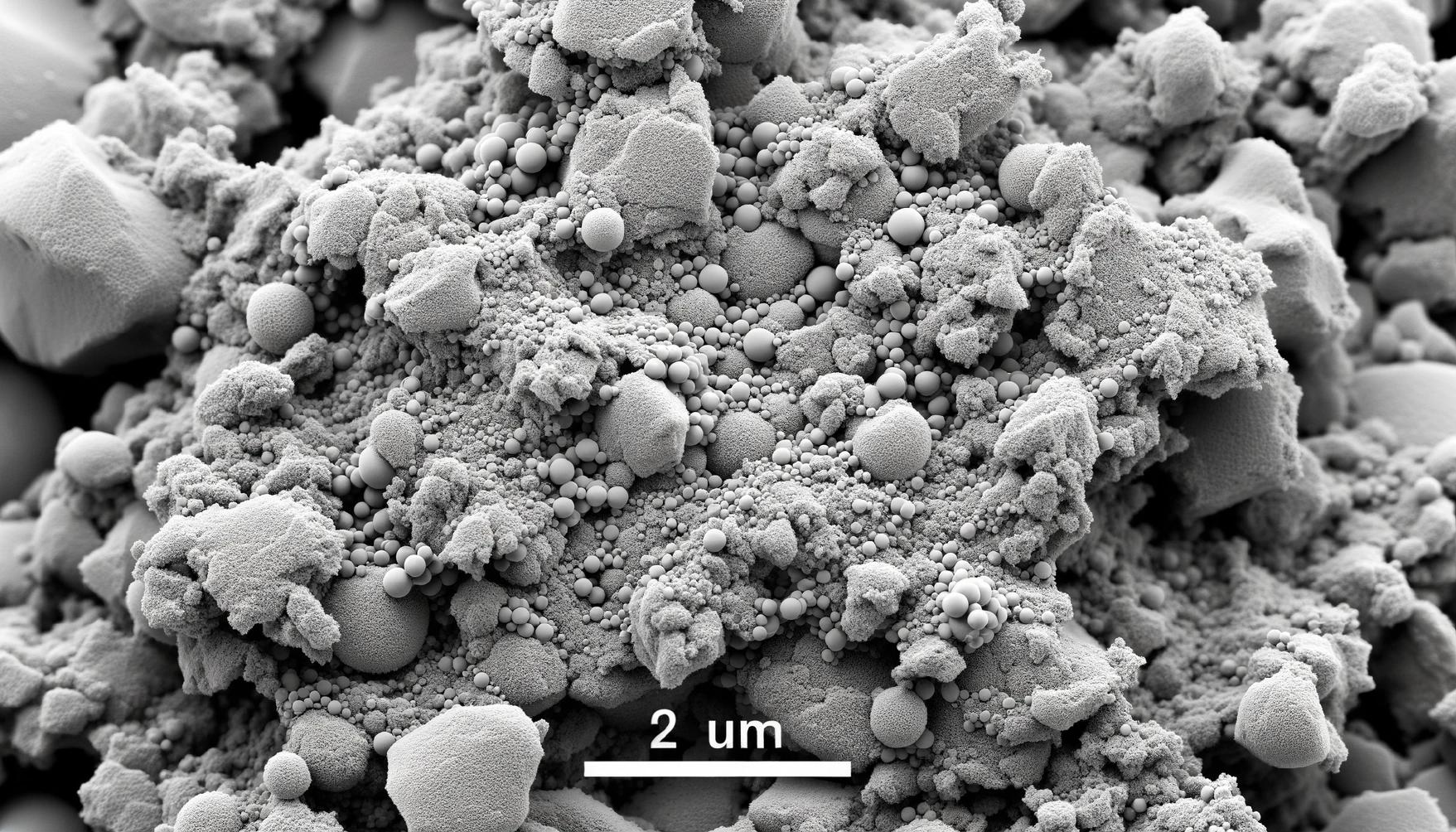} & 
\includegraphics[height=0.09\textwidth]{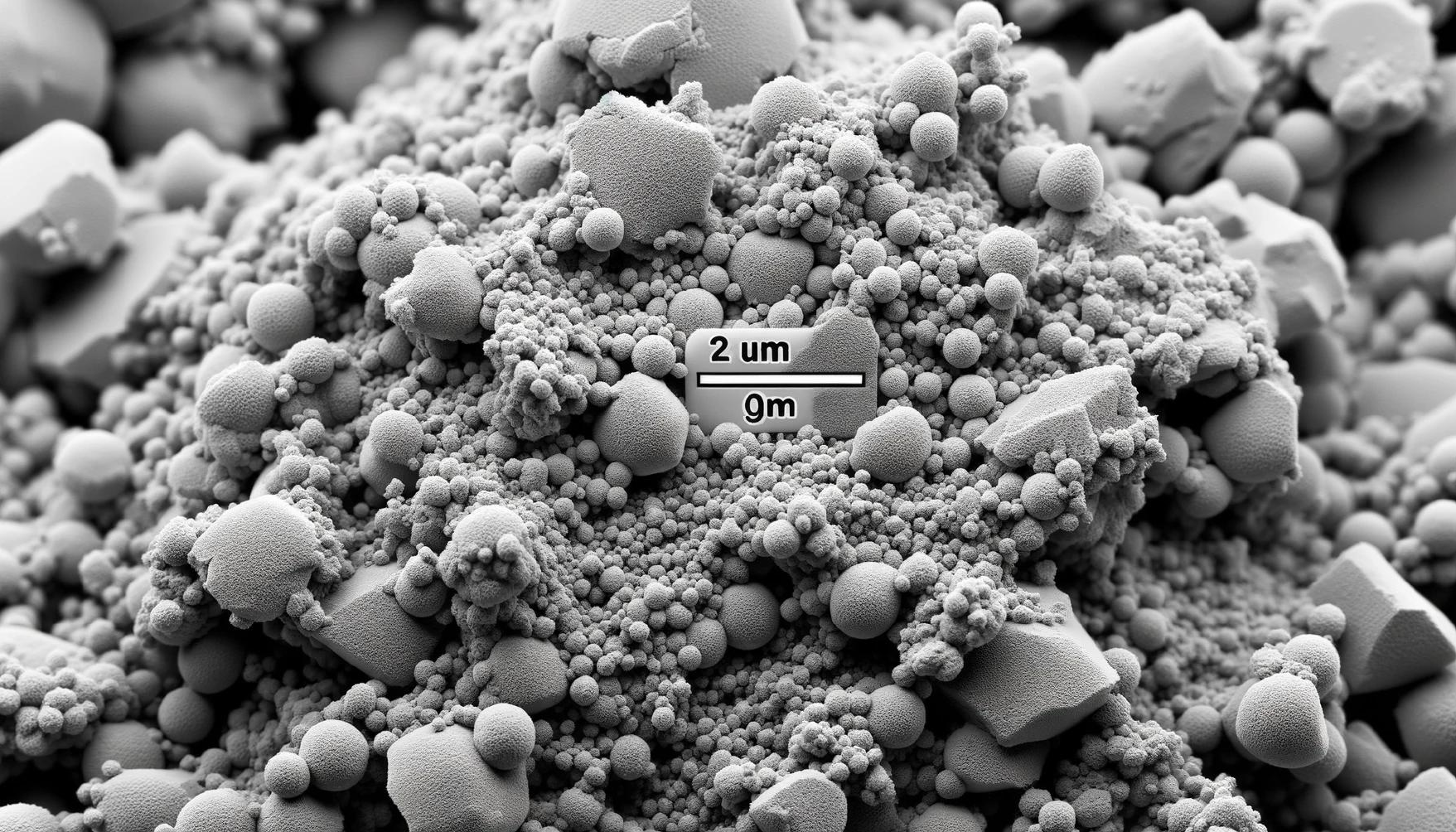} & 
\includegraphics[height=0.09\textwidth]{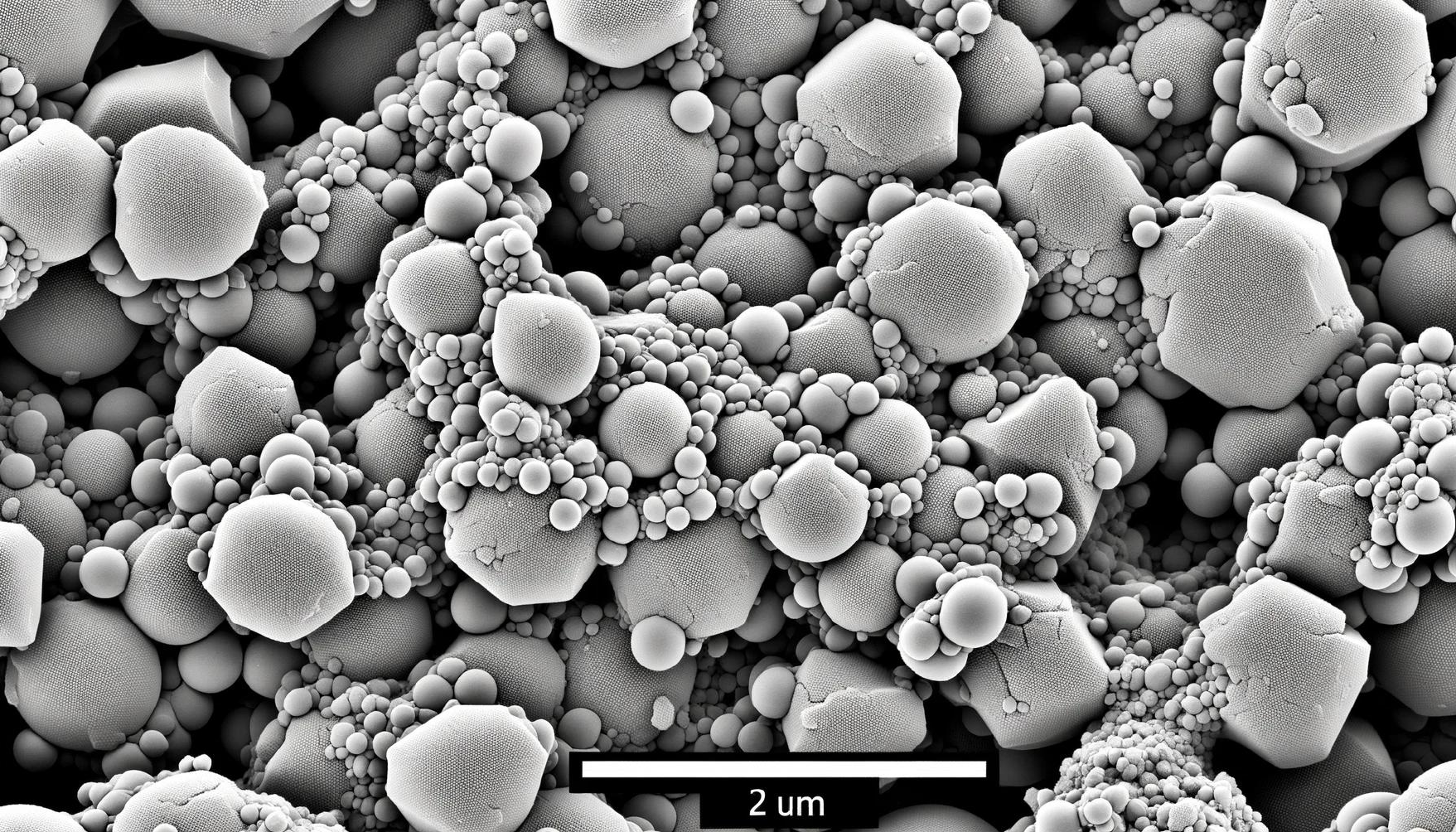} & 
\includegraphics[height=0.09\textwidth]{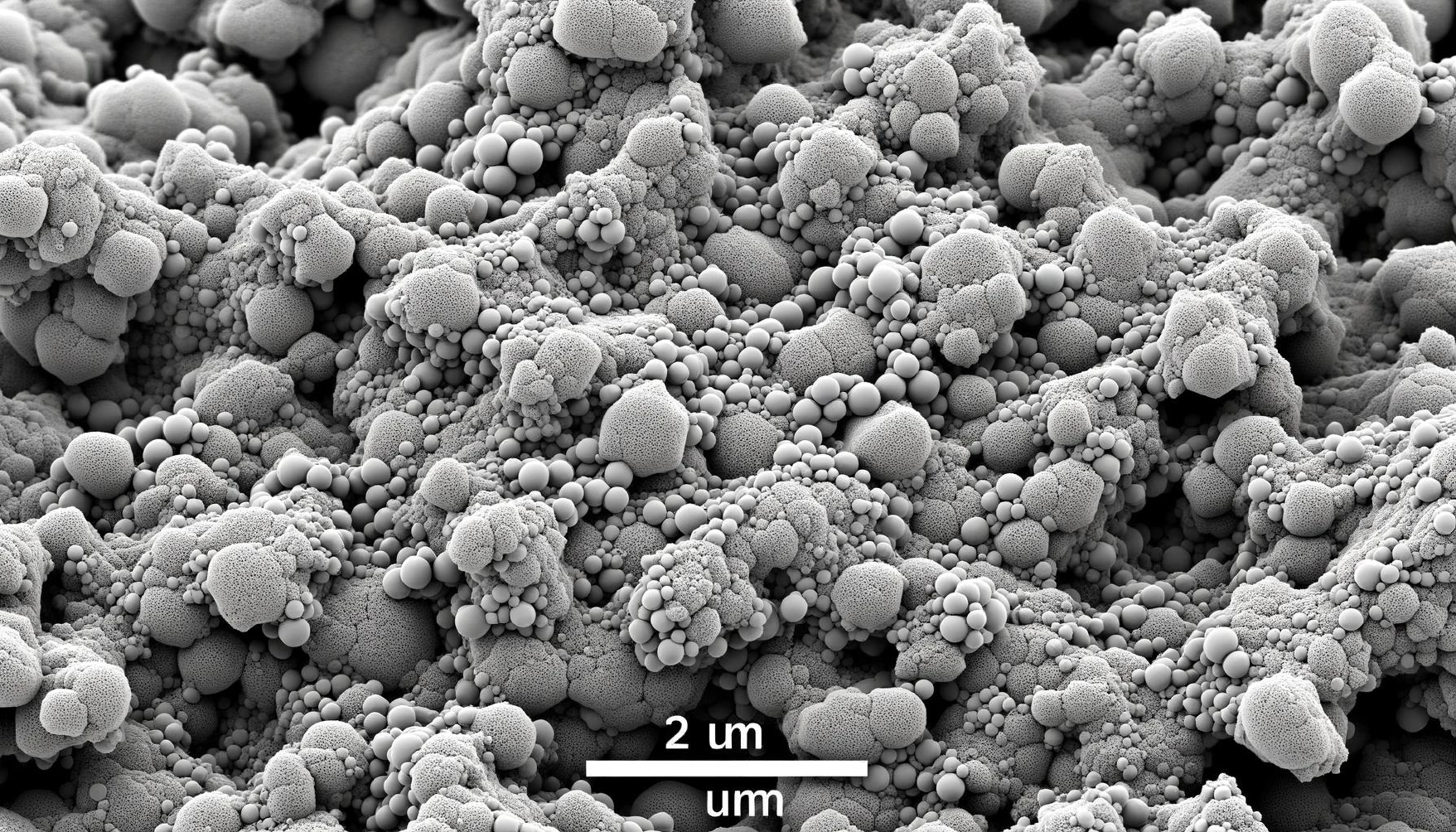} &
\includegraphics[height=0.09\textwidth]{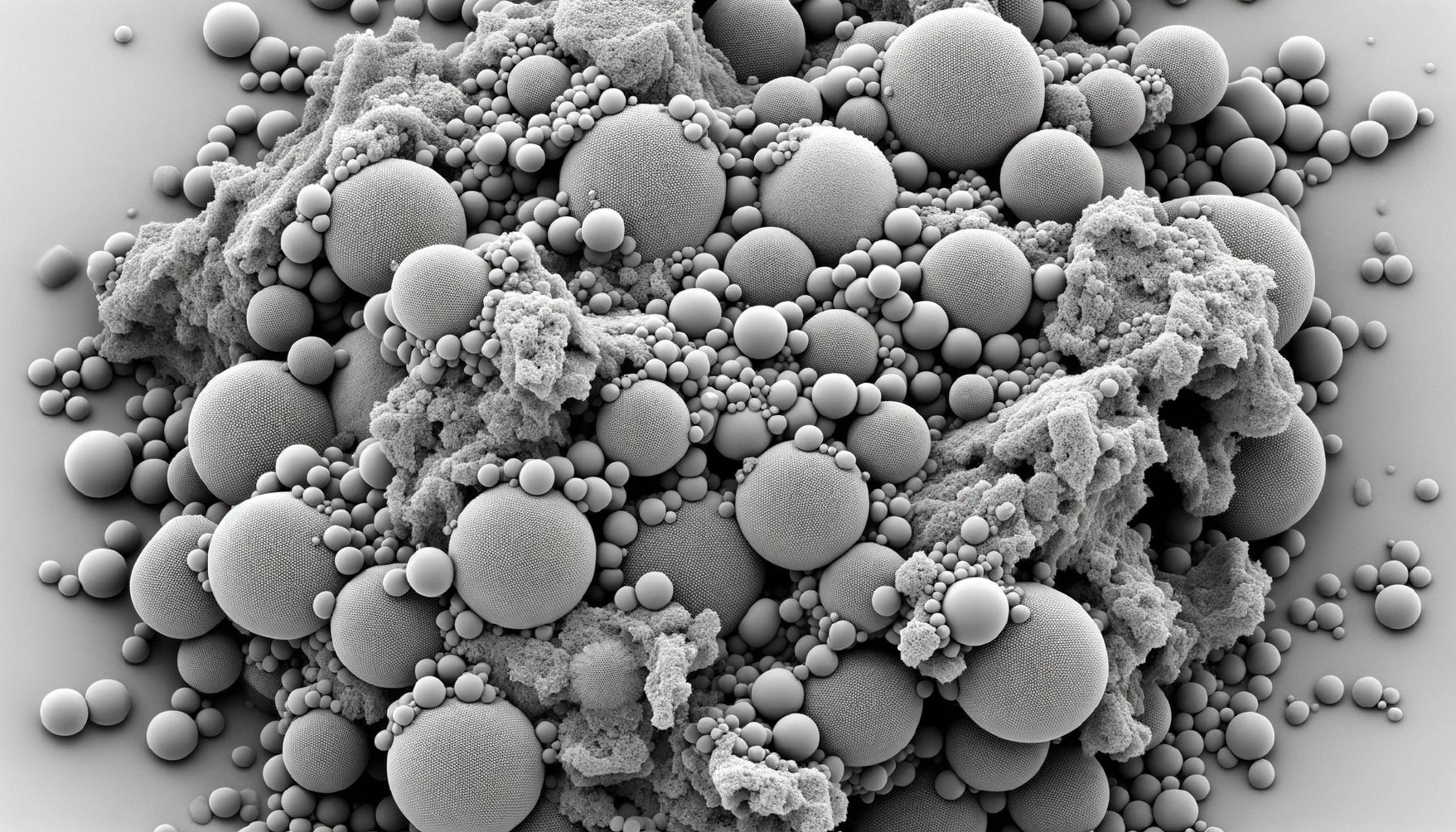} \\ [-0.25em] 
\includegraphics[height=0.09\textwidth]{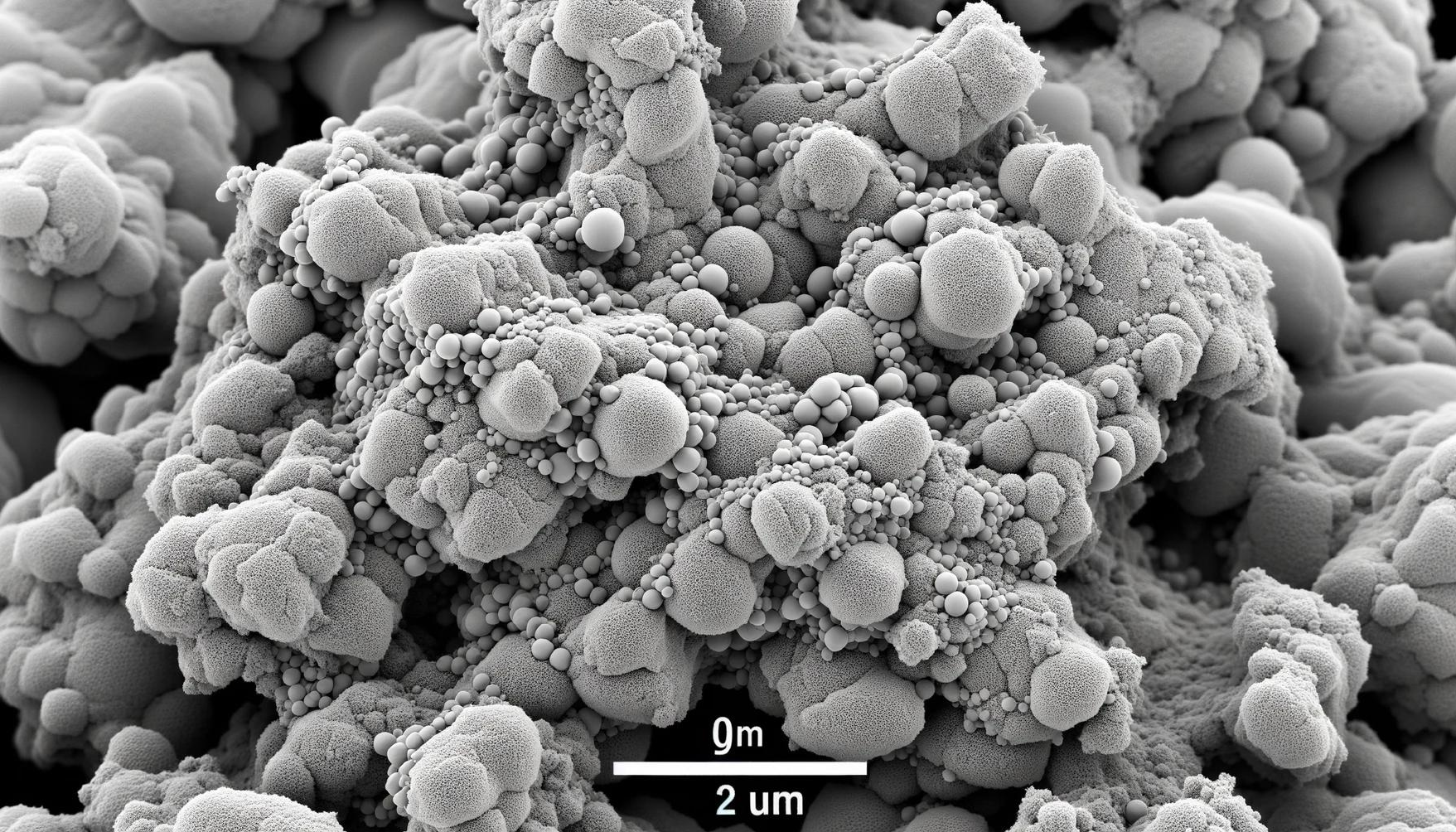} \\ [-0.95em]
\caption{The table shows sample synthetic images generated by DALLE-3 from textual descriptions provided in a series of question-answer pairs, as shown in Table \ref{tab:tab17}.}
\label{tab:tab18}
\end{tabularx}
\vspace{-3mm}
\end{tcolorbox}

\begin{tcolorbox}[colback=white!2!white,colframe=black!75!black]
\vspace{-5mm}
\begin{tabularx}{1.2\textwidth}{bss}
\caption{The table presents the question-answer pairs generated by the LMM(GPT-4V) in response to natural language questions about the input image of the MEMS devices and electrodes nanomaterial category. In summary, the question-answer pairs (generated text) provide in-depth information about an image related to MEMS devices nanomaterial category.}\label{tab:tab19}\\ [-0.85em]
\toprule 
\begin{overpic}[width=0.15\columnwidth]{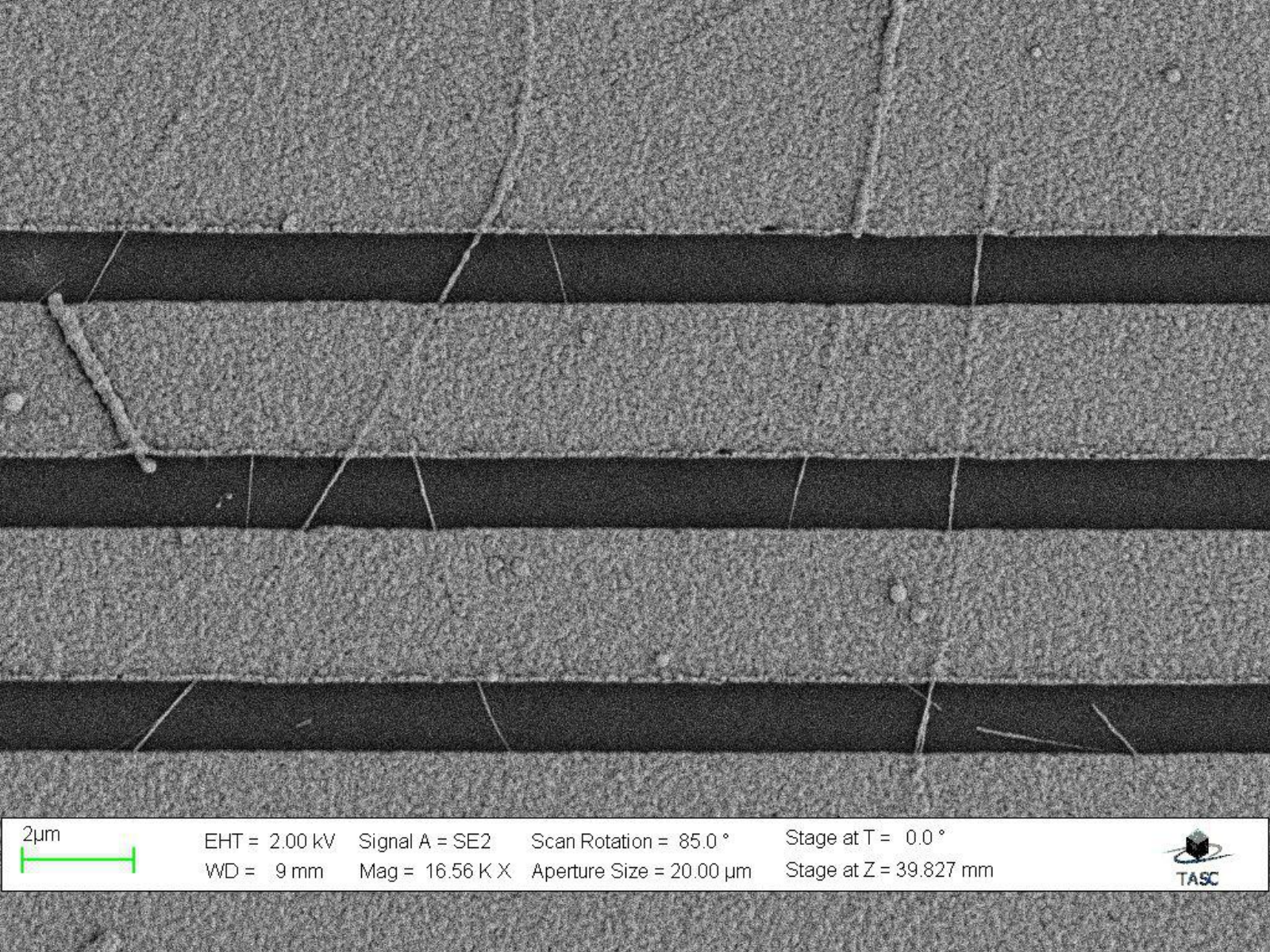}
  \put(400,40){\makebox[0pt][c]{%
    \parbox{0.5\columnwidth}{%
      \centering
      \textbf{Please answer the following questions based on the provided input image belonging to the MEMS devices and electrodes nanomaterial category.} 
    }%
  }}  
\end{overpic}  \\ [-0.25em]

\midrule
\textbf{Query prompt}  & \textbf{GPT-4 Vision output} \\
\midrule
\endhead
\textcolor{black}{**Basics**} & The image appears to depict nanoscale lines or strips which could be electrodes or other conductive paths. The scale at the bottom left indicates 2$\mu$m, so one unit of measurement represents 2 micrometers.
\\

\midrule
\textcolor{black}{**Morphology and Structure**} & The nanomaterials in the image are linear and strip-like in shape. There seem to be two distinct phases: the darker lines or strips and the lighter background. The nanomaterial lines appear uniform in width, but there is variability in the length and positioning of some lines.
\\

\midrule
\textcolor{black}{**Size and Distribution**} & The width of the individual nanostructures (lines or strips) is smaller than the provided scale, so they are less than 2$\mu$m in width. The nanomaterial lines are parallel to one another and appear evenly spaced. No evident aggregation.
\\

\midrule
\textcolor{black}{**Surface Characteristics**} & The nanomaterial lines appear smooth, while the background has a textured appearance. There are a few visible defects or imperfections, particularly breaks in some of the lines.
\\

\midrule
\textcolor{black}{**Composition and Elements**} & The image is in grayscale, so it's challenging to identify compositional variations solely based on color. However, the differences in brightness might indicate different compositions. No explicit labels or markers are indicating specific elements or compounds.
\\
 
\midrule
\textcolor{black}{**Interactions and Boundaries**} & Individual nanostructures (lines) are separate from each other with clear spaces in between. There are clear boundaries between the lines and the surrounding matrix.
\\
 
\midrule
\textcolor{black}{**External Environment**} & The background or matrix in which these nanostructures are embedded appears to be a different material, possibly an insulating layer. There are other structures like breaks and dots, which might not be nanomaterials. Their nature is not clear from the image.
\\
 
\midrule
\textcolor{black}{**Image Technique and Modifications**} & The imaging technique used is Scanning Electron Microscopy (SEM) as indicated by the ``Signal A = SE2" in the bottom right corner. There is no clear indication of any post-processing or modifications made to the image in the provided information.
\\
 
\midrule
\textcolor{black}{**Functional Features**} & There aren't any explicitly visible functional features like active sites. The linear nature suggests they could serve as conductive paths or channels. The image appears to be a static representation.
\\ [-0.15em]

\bottomrule
\end{tabularx}
\end{tcolorbox}

\vspace{-6mm}

\begin{tcolorbox}[colback=white!2!white,colframe=black!75!black]
\vspace{-6mm}
\begin{tabularx}{1.2\textwidth}{bss}
\includegraphics[width=0.10\textwidth]{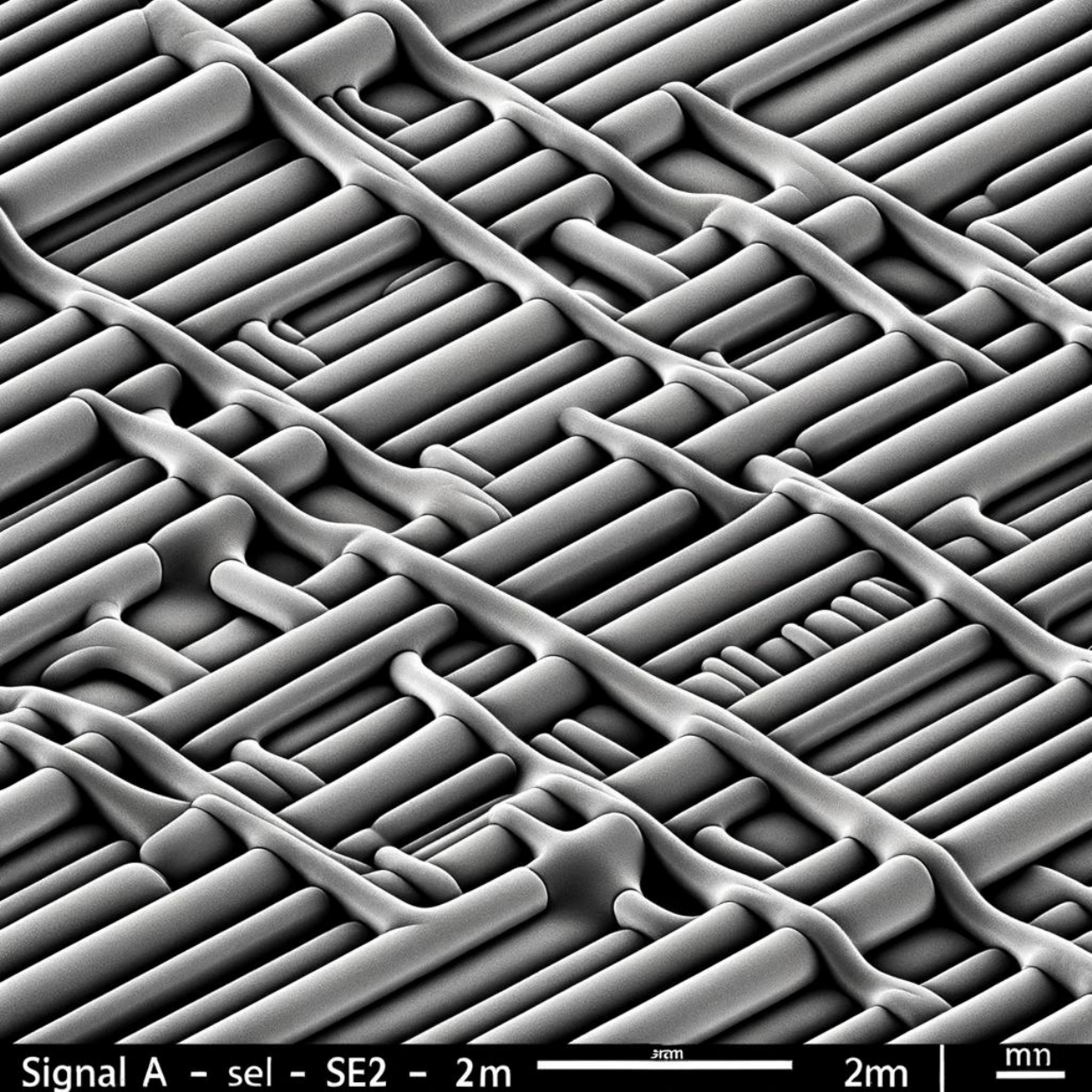} & 
\includegraphics[width=0.10\textwidth]{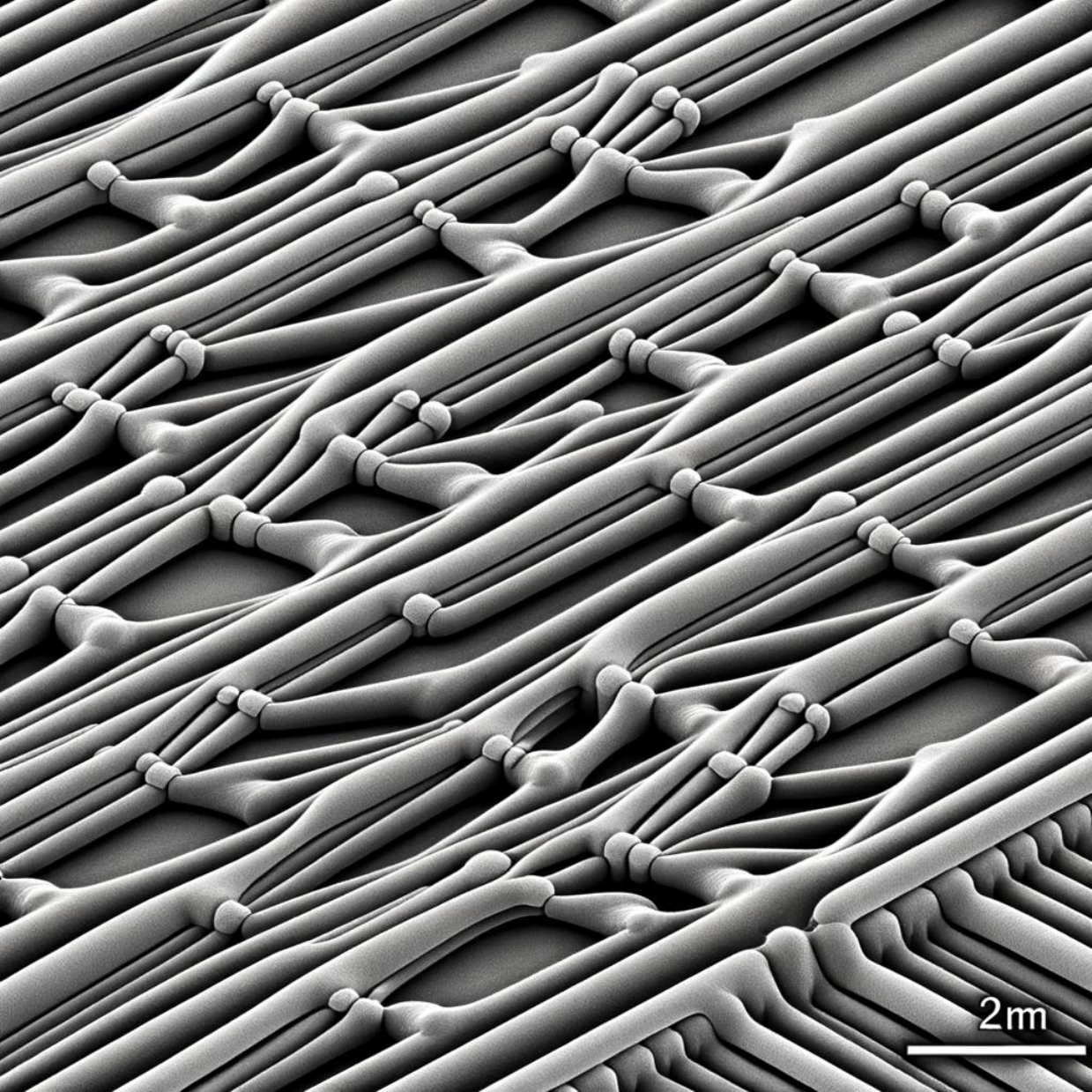} & 
\includegraphics[width=0.10\textwidth]{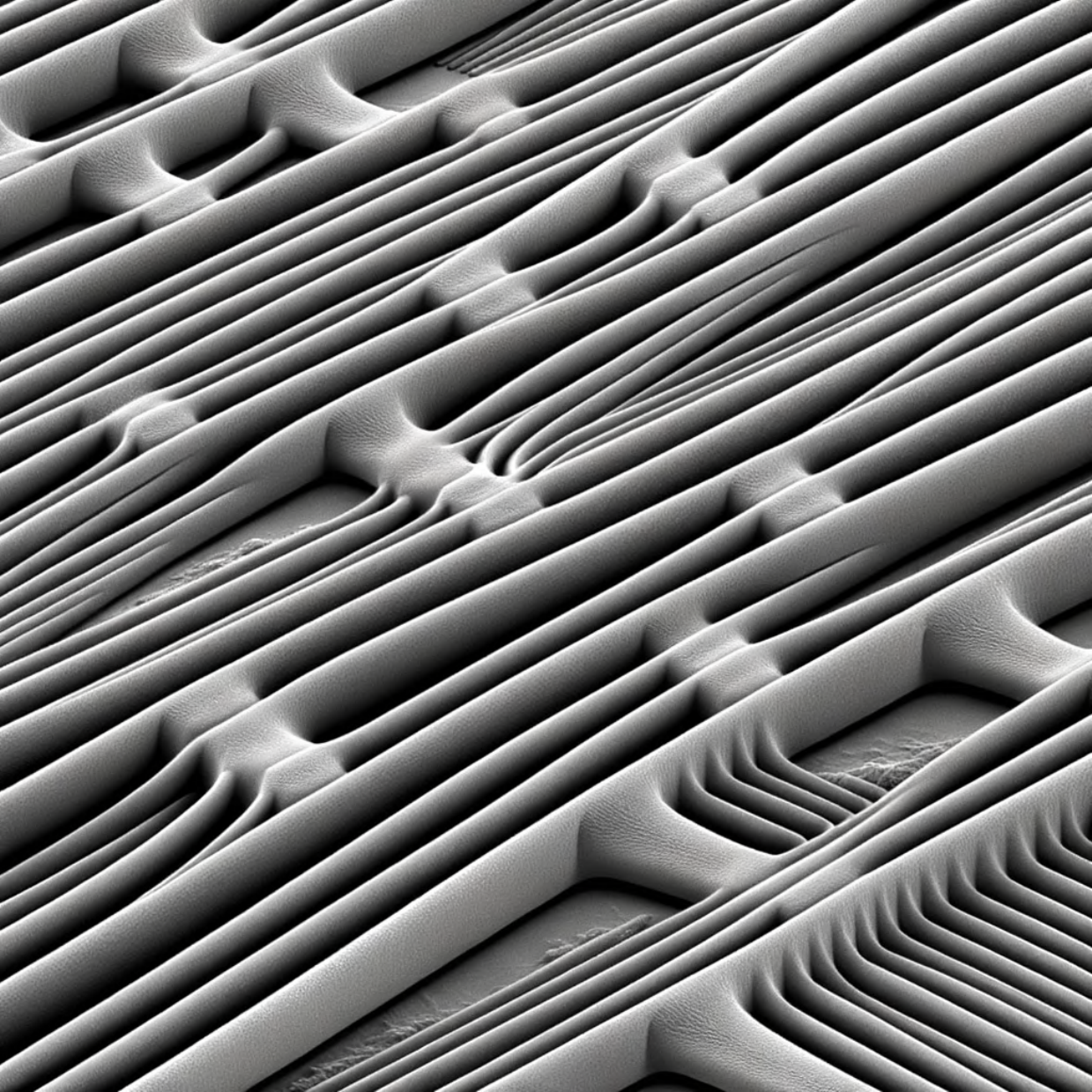} \\
\includegraphics[width=0.10\textwidth]{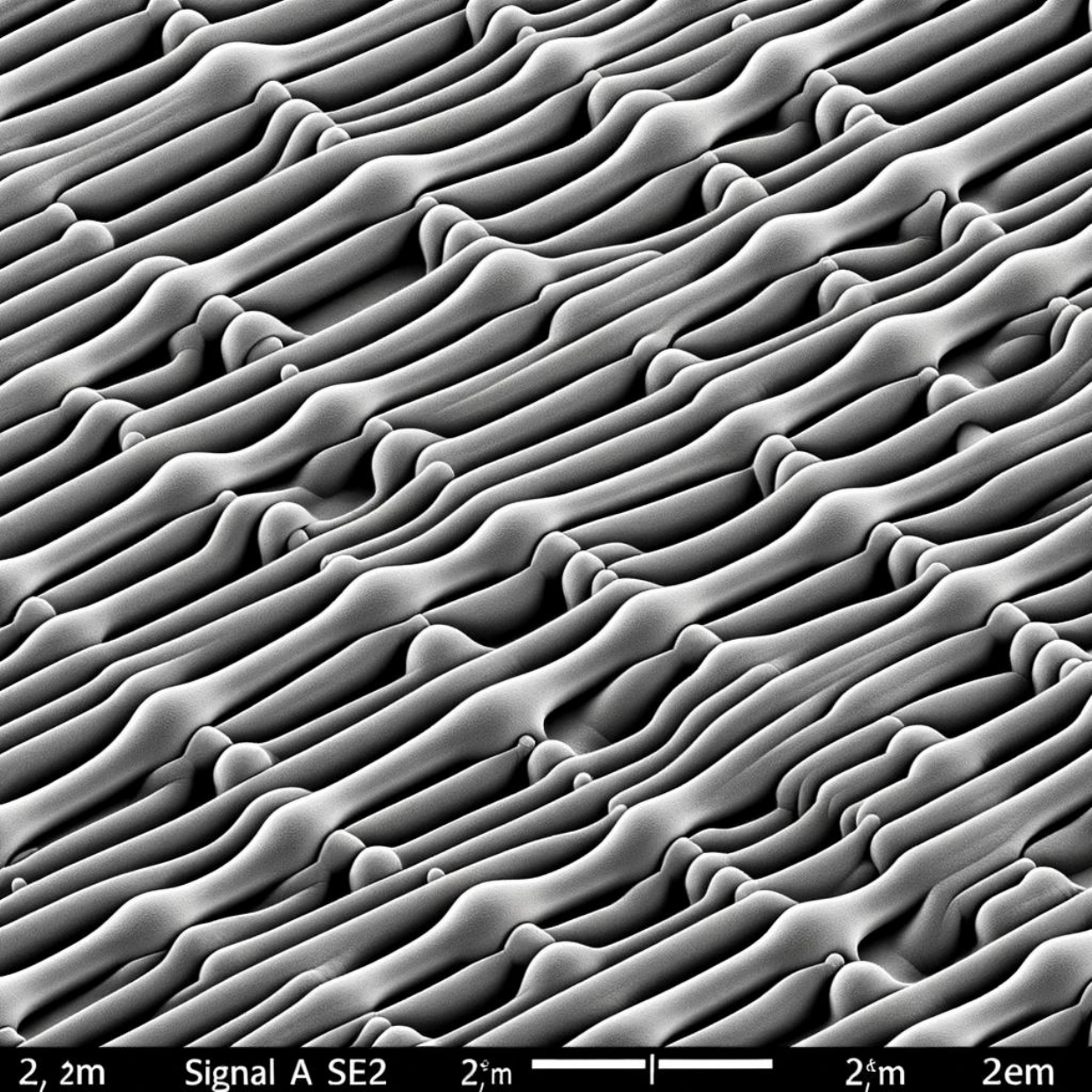} & 
\includegraphics[width=0.10\textwidth]{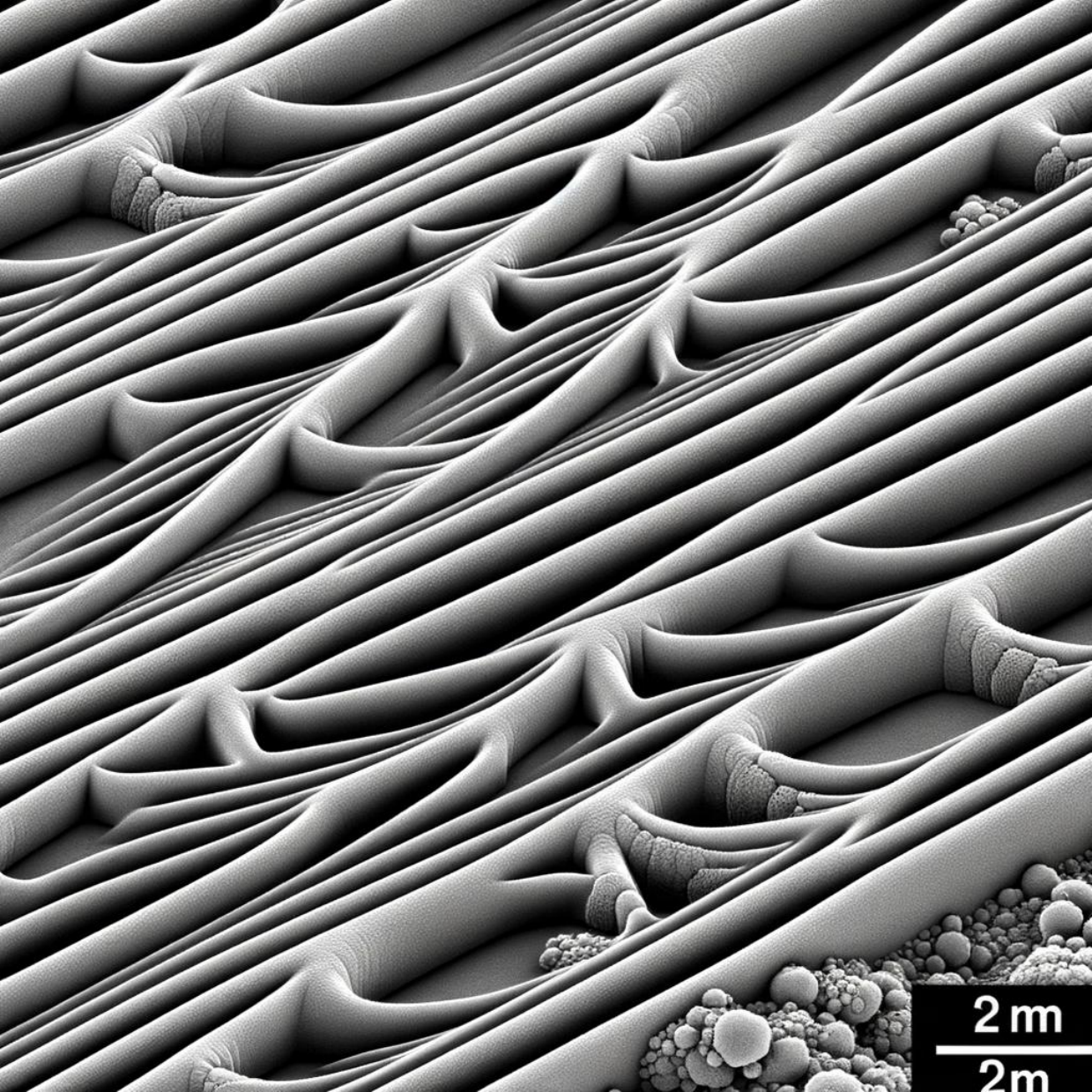} & 
\includegraphics[width=0.10\textwidth]{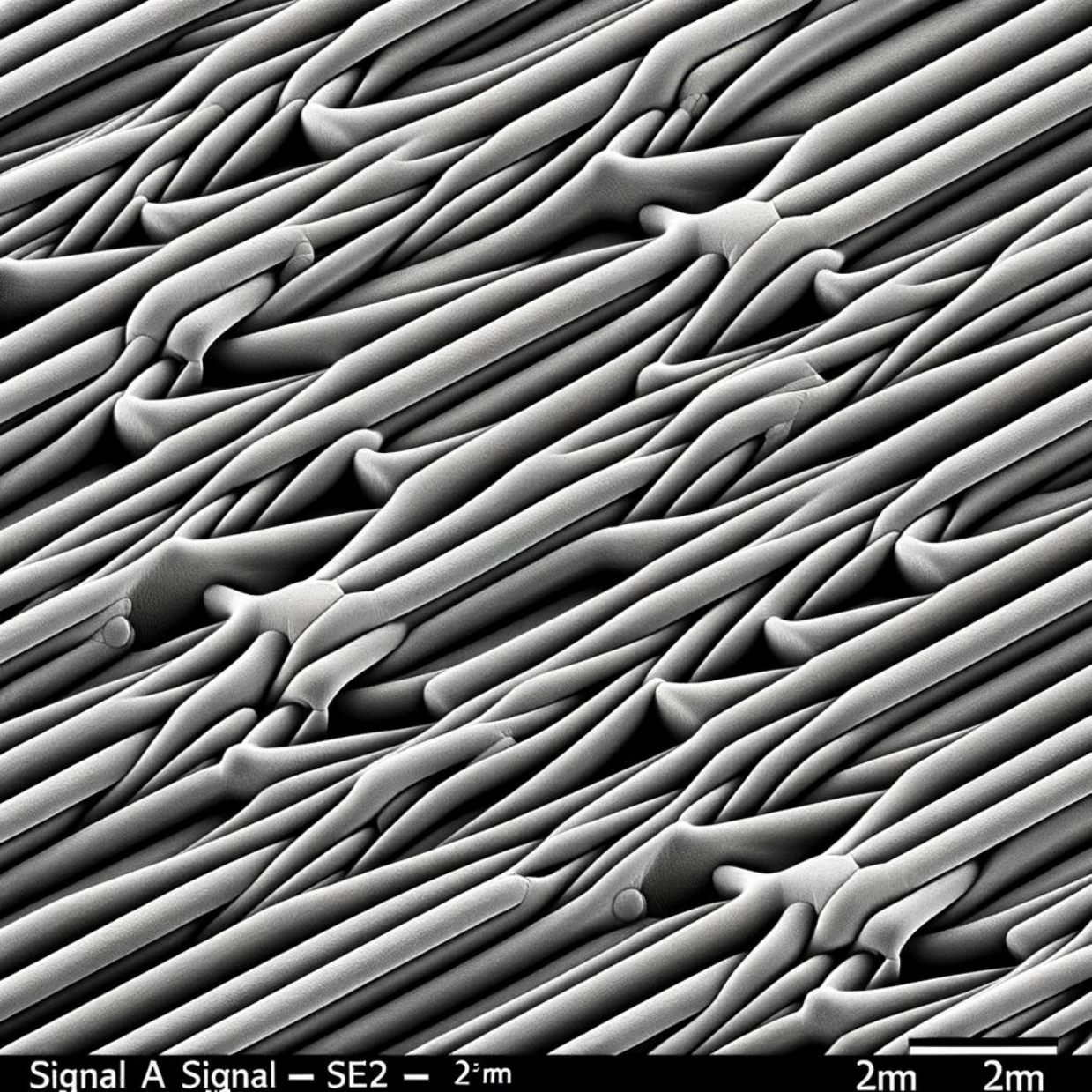} \\ [-0.8em]
\caption{The table shows the sample images generated by DALLE-3 with hints provided from a series of question-answer pairs based on the provided MEMS devices and electrodes nano image as shown in the above Table \ref{tab:tab19}.}
\vspace{-3mm}
\label{tab:tab20}
\end{tabularx}
\end{tcolorbox}

\clearpage
\twocolumn

\vspace{-2mm}
\subsection{Additional datasets and Experimental results}
\vspace{1mm}
To enhance the robustness and validity of our framework, we conducted evaluations using multiple open-source benchmark datasets that are pertinent to our research area and cover a variety of applications. This approach enabled us to confirm the effectiveness of our framework and establish its suitability for a wider array of datasets beyond merely the SEM dataset.

\subsubsection{NEU-SDD(\cite{deshpande2020one}):} 
\begin{figure}[ht!]
    \centering
    \includegraphics[width=8.5cm]{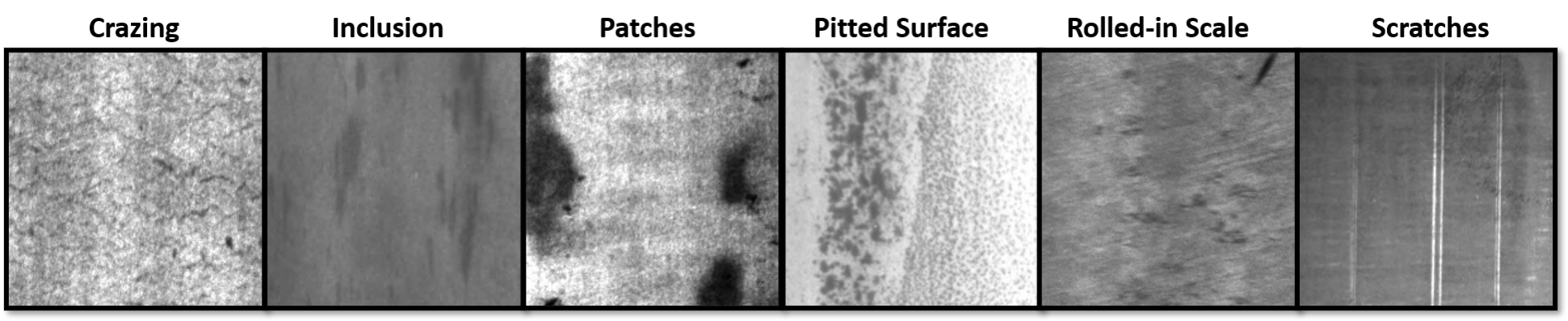}
    \vspace{-6mm}
    \caption{Representative electron microscopy images from the NEU-SDD dataset, showcasing six types of surface defects on hot-rolled steel strips: \textit{pitted surfaces, scratches, rolled-in scale, crazing, patches, and inclusions}.}
    \vspace{-1mm}
    \label{fig:addataset3}
    \vspace{-3mm}
\end{figure}

\vspace{-1mm}
The NEU-SDD dataset\footnote{Datasource: \url{http://faculty.neu.edu.cn/yunhyan/NEU_surface_defect_database.html}\label{note1}} comprises 1800 electron microscopy images of surface irregularities on hot-rolled steel strips. These grayscale images are 200$\times$200 pixels each and are categorized into six distinct types of defects, with each category containing 300 representative micrographs. The categories include \textit{pitted surfaces, scratches, rolled-in scale, crazing, patches, and inclusion defects}. Figure \ref{fig:addataset3} provides illustrative images from each of these categories. To evaluate the performance of our proposed methodology, particularly for multi-category defect detection tasks, we conducted a comparative analysis using several benchmark algorithms.

\vspace{-3mm}

\vspace{-3mm}
\subsubsection{CMI:}
\begin{figure}[ht!]
    \centering
    \includegraphics[width=7cm]{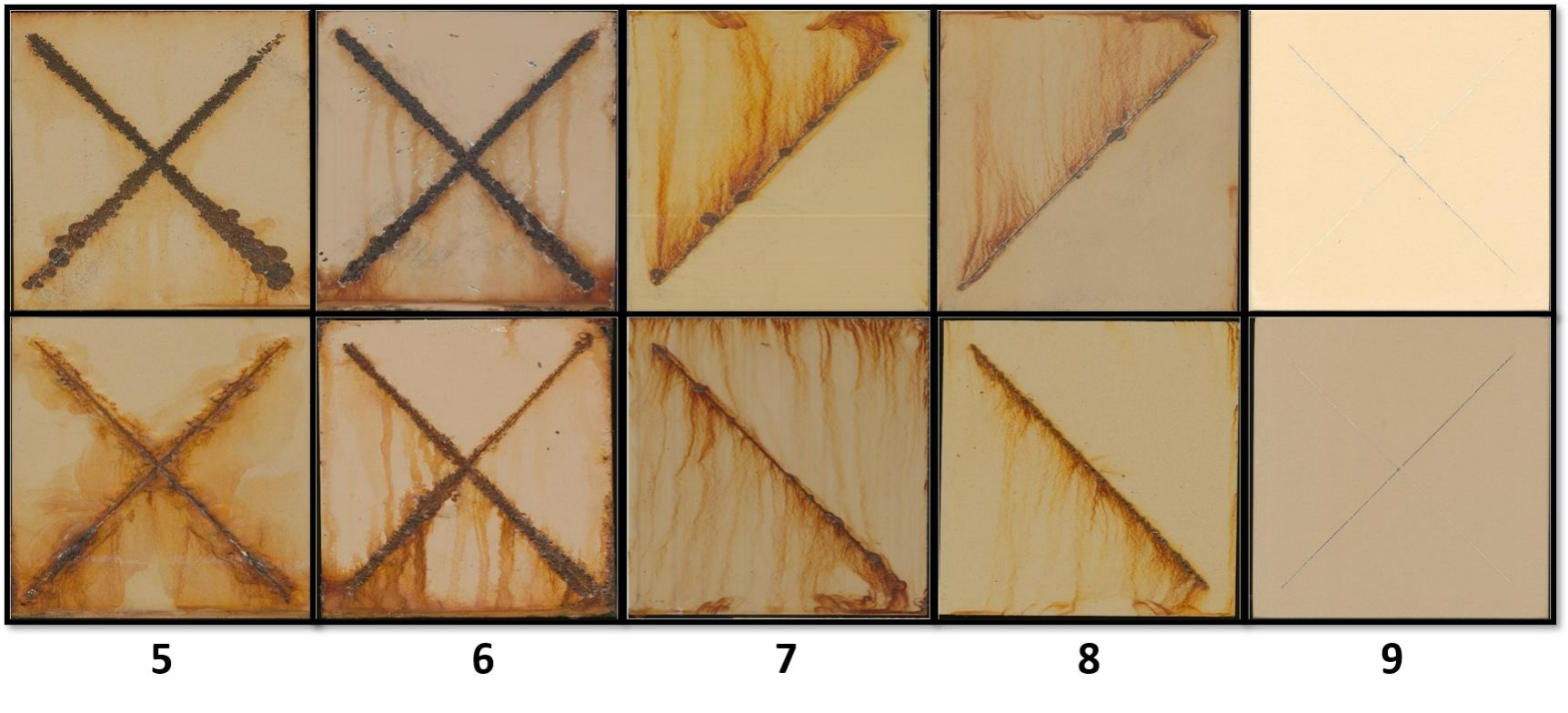}
    \vspace{-3mm}
    \caption{Samples of electron micrographs from the CMI dataset, categorized by corrosion severity scores ranging from 5 to 9 according to ASTM-D1654 standards.}
    \label{fig:addataset1}
    \vspace{-3mm}
\end{figure}

\vspace{1mm}
The CMI dataset\footnote{\url{https://arl.wpi.edu/corrosion_dataset}\label{note2}} consists of 600 detailed electron micrographs that display corroded panels. These images have been annotated by corrosion experts according to the ASTM-D1654 standards, with individual scores ranging from 5 to 9. Each score corresponds to a set of 120 unique micrographs, each with a resolution of 512$\times$512 pixels. Figure \ref{fig:addataset1} presents examples from each scoring category. We evaluated the effectiveness of our proposed technique for multi-category classification by comparing it with various established algorithms.

\vspace{-1mm}

\vspace{-3mm}
\subsubsection{KTH-Tips:}
\begin{figure}[ht!]
    \centering
    \includegraphics[width=8cm]{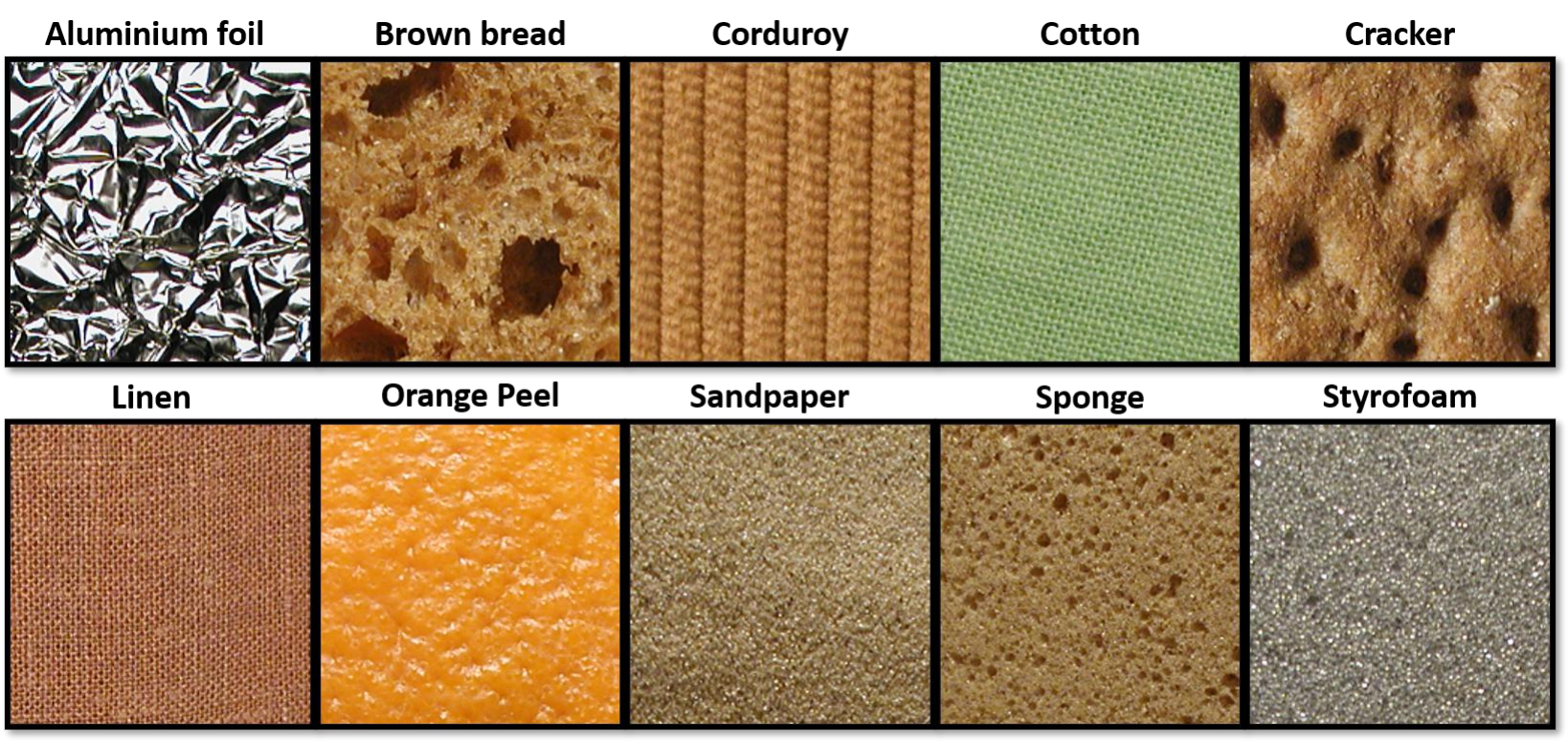}
    \vspace{-2mm}
    \caption{Representative electron micrographs from the KTH-TIPS texture dataset, depicting the ten material classes including \textit{sponge, orange peel, styrofoam, cotton, cracker, linen, crust, sandpaper, aluminum foil, and corduroy}}
    \label{fig:addataset2}
    \vspace{-3mm}
\end{figure}

\vspace{1mm}
KTH-TIPS\footnote{\url{https://www.csc.kth.se/cvap/databases/kth-tips/index.html}\label{note3}} is a comprehensive texture dataset that contains 810 electron micrographs, each representing one of ten specific material classes. Each image, with a resolution of 200$\times$200 pixels, captures a wide range of materials under varying lighting conditions, orientations, and scales. The diverse textures include \textit{sponge, orange peel, styrofoam, cotton, cracker, linen, crust, sandpaper, aluminum foil, and corduroy}. Figure \ref{fig:addataset2} showcases representative images from each material class. We assessed the performance of our proposed method by comparing it with results from several benchmark algorithms, specifically for multi-category texture recognition tasks.

\vspace{-2mm}

Table \ref{tab:auxexp} provides an extensive comparative analysis, detailing the performance of our method in relation to a range of established techniques, across various datasets. 
Our findings indicate that our method not only performs with remarkable consistency but also achieves superior results when compared to the standard approaches, thereby underscoring its high effectiveness and dependability in different experimental scenarios.

\vspace{-4mm}
\begin{table}[ht!]
\footnotesize
\centering
\resizebox{0.4\textwidth}{!}{%
\subfloat{%
\setlength{\tabcolsep}{3pt}
\begin{tabular}{cc|cccc}
\hline
\multicolumn{2}{c|}{\textbf{Algorithms}}                                       & \textbf{NEU-SDD} & \textbf{CMI} & \textbf{KTH-TIPS}  \\ \hline
\multicolumn{1}{c|}{\multirow{4}{*}{\rotatebox[origin=c]{90}{\textbf{Baselines}}}} & ResNet                   & 0.906	& 0.928	& 0.941 &             \\
\multicolumn{1}{c|}{}                                          & GoogleNet                & 0.936	& 0.928	& 0.929
              \\
\multicolumn{1}{c|}{}                                          & SqueezeNet                & 0.955	& 0.943	& 0.963
              \\ 
\multicolumn{1}{c|}{}                                          & VanillaViT               & 0.962	& 0.968	& 0.972
 \\ 
\hline
\multicolumn{1}{c|}{}                                          & \textbf{MultiFusion-LLM}                  &     \textbf{1.0}              &      \textbf{1.0}             &    \textbf{1.0}               &                     \\ \hline
\end{tabular}}}
\vspace{0mm}
\caption{The table presents an in-depth comparison of the performance metrics of our proposed framework against a selection of benchmark algorithms, showcasing the results obtained from the evaluations conducted on a diverse collection of datasets.}
\label{tab:auxexp}
\vspace{-3mm}
\end{table}

However, when it comes to generating question-answer pairs or providing in-depth technical descriptions for the analysis of material images from different datasets using GPT-4(V), the idea that a single, universal prompt can cater to all these varied tasks is a misconception. There is no one-size-fits-all solution; each task demands a uniquely crafted prompt, created with intention and understanding. In the AI landscape, a diversity of prompting strategies is not just beneficial—it’s essential. By tailoring our prompts to our specific needs, we unlock the full potential of these advanced AI models, ensuring that they serve us in the most effective way possible. In these additional experiments, we utilize GPT-4V to generate question-and-answer pairs for the material category based on the input microscopy image, rather than custom prompts generated using GPT-4(language-only).

\clearpage
\onecolumn

\section{CMI}
\begin{tcolorbox}[colback=white!2!white,colframe=black!75!black]
\vspace{-5mm}
\begin{tabularx}{1.2\textwidth}{bss}
\caption{The table presents the question-answer pairs generated by the LMM(GPT-4V) in response to natural language questions about the input image of the corrosion 5 category from the CMI dataset. In summary, the question-answer pairs (generated text) provide in-depth information about an image related to the corrosion 5 category.}\label{tab:tab31}\\
\toprule
\begin{overpic}[width=0.185\columnwidth]{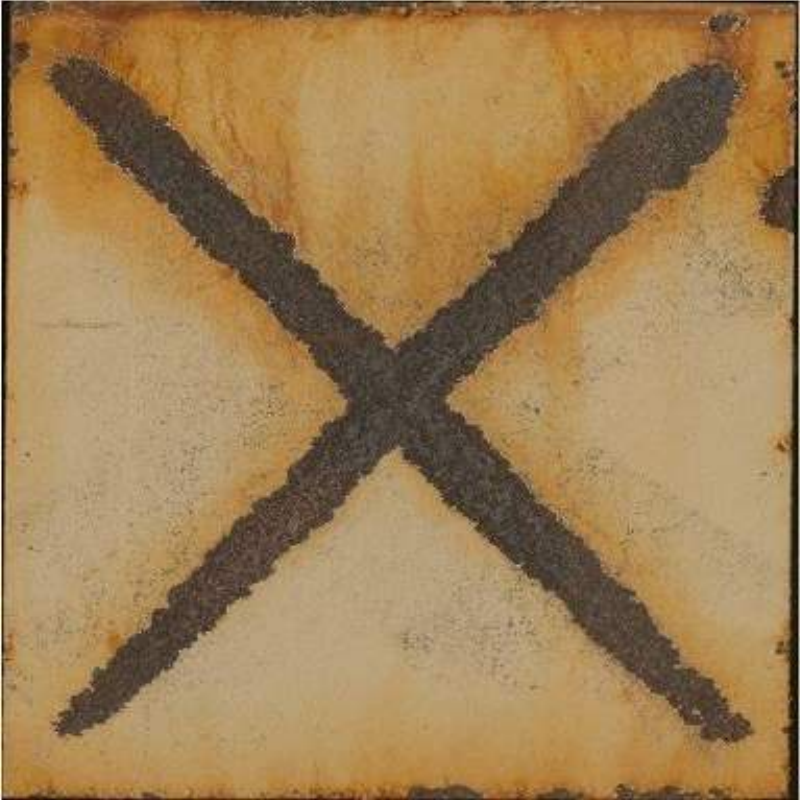}
\vspace{-5mm}
  \put(350,50){\makebox[0pt][c]{%
    \parbox{0.5\columnwidth}{%
      \centering
      \textbf{Please answer the following questions based on the provided input image belonging to the corrosion 5 category.} 
    }%
  }}  
\end{overpic}  \\

\midrule
\textbf{Query prompt}  & \textbf{GPT-4 Vision response} \\
\midrule
\endhead
\textcolor{black}{What is the primary purpose of the image?} & The image is a representation of a corroded panel from the Corrosion Image Data Set for Automating Scientific Assessment of Materials, specifically with a Corrosion Rating of 5.
\\

\midrule
\textcolor{black}{How is the severity of corrosion visually represented in this rating?} & In the context of a rating of 5, there are visible signs of corrosion, but they are not overwhelmingly dominant. The corrosion is evident but still moderate.
\\

\midrule
\textcolor{black}{What is the dominant feature in the image?} & The dominant feature is a large X-shape, which appears to be scribed onto the panel, displaying signs of corrosion emanating from it.
\\

\midrule
\textcolor{black}{Describe the color and texture of the X-shape.} & The X-shape is dark brown to black, with a rough and corroded texture. There are patches of uneven coloration, with some areas appearing lighter, indicating variations in the severity of corrosion.
\\

\midrule
\textcolor{black}{What's notable about the background surrounding the X-shape?} & The background is a beige hue, with a gradient that transitions from a lighter shade at the top to a slightly darker shade at the bottom. It shows signs of aging and subtle, uneven streaks and patches, indicating initial stages of corrosion.
\\
 
\midrule
\textcolor{black}{Are there visible boundaries or divisions between the X-shape and the background?} & The X-shape and the background are distinct, but the boundaries between them are not sharply defined. The edges of the X appear corroded, making them blend unevenly with the background.
\\
 
\midrule
\textcolor{black}{Is there a pattern to the corrosion emanating from the X-shape?} & The corrosion seems to emanate outward from the scribed X-shape, spreading in a manner that suggests it's creeping away from the scribe lines. 
\\
 
\midrule
\textcolor{black}{How do the dark speckles on the background contribute to the overall appearance of corrosion?} & The sporadic, small dark speckles scattered throughout the beige background add to the visual cues of wear and corrosion, indicating that the panel is undergoing a consistent, albeit moderate, corrosive process.
\\
 
\midrule
\textcolor{black}{Given the context of corrosion ratings, what does a rating of 5 signify in terms of material integrity and longevity?} & A rating of 5 suggests that while there are evident signs of corrosion, the material still retains a significant portion of its original integrity. It's midway between pristine condition (rating 10) and severe corrosion (rating 0), indicating that the material has undergone wear but is not entirely compromised.
\\ 

\bottomrule
\end{tabularx}
\end{tcolorbox}

\vspace{-5mm}

\begin{tcolorbox}[colback=white!2!white,colframe=black!75!black]
\vspace{-4mm}
\begin{tabularx}{1.2\textwidth}{bss}
\includegraphics[width=0.185\textwidth]{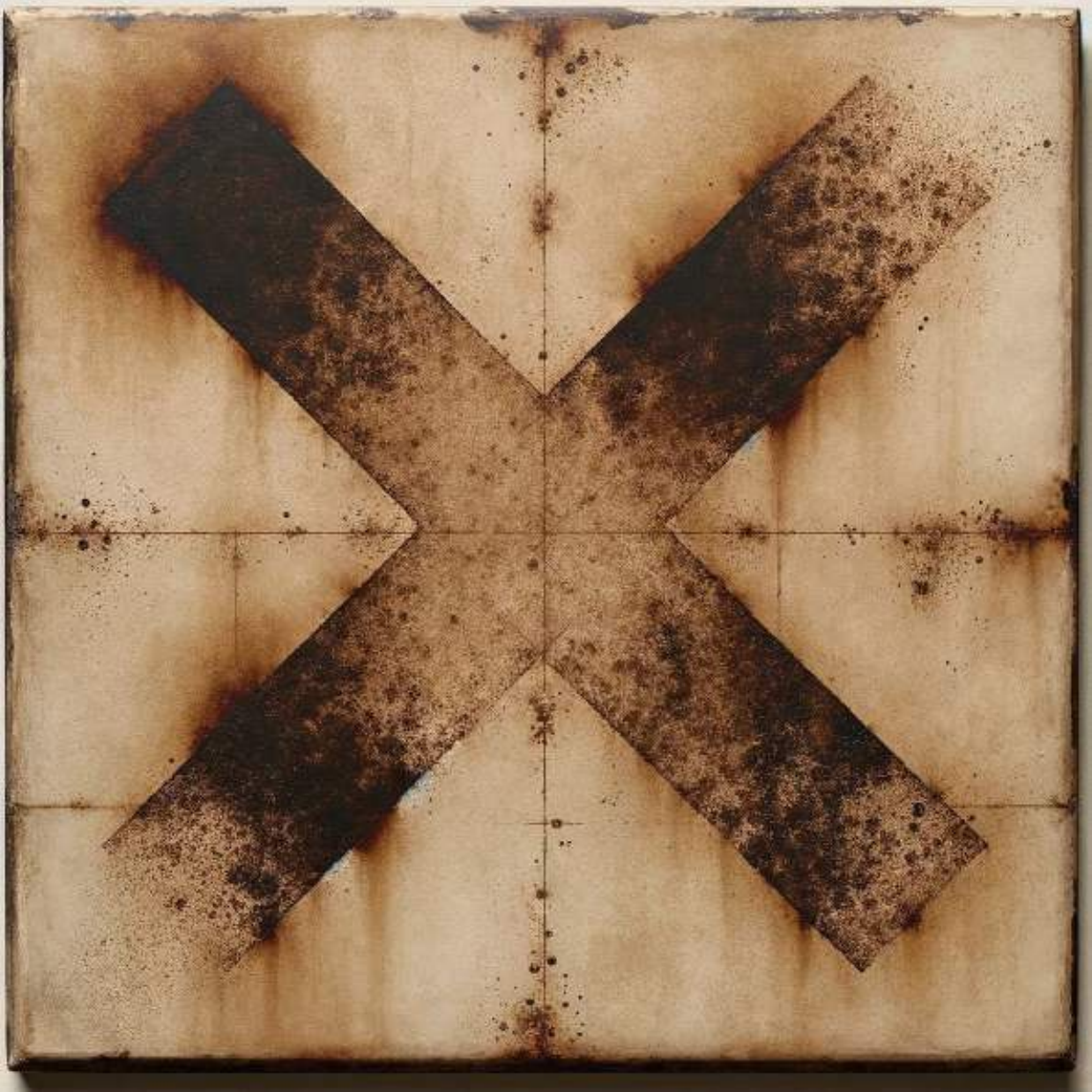} & 
\includegraphics[width=0.185\textwidth]{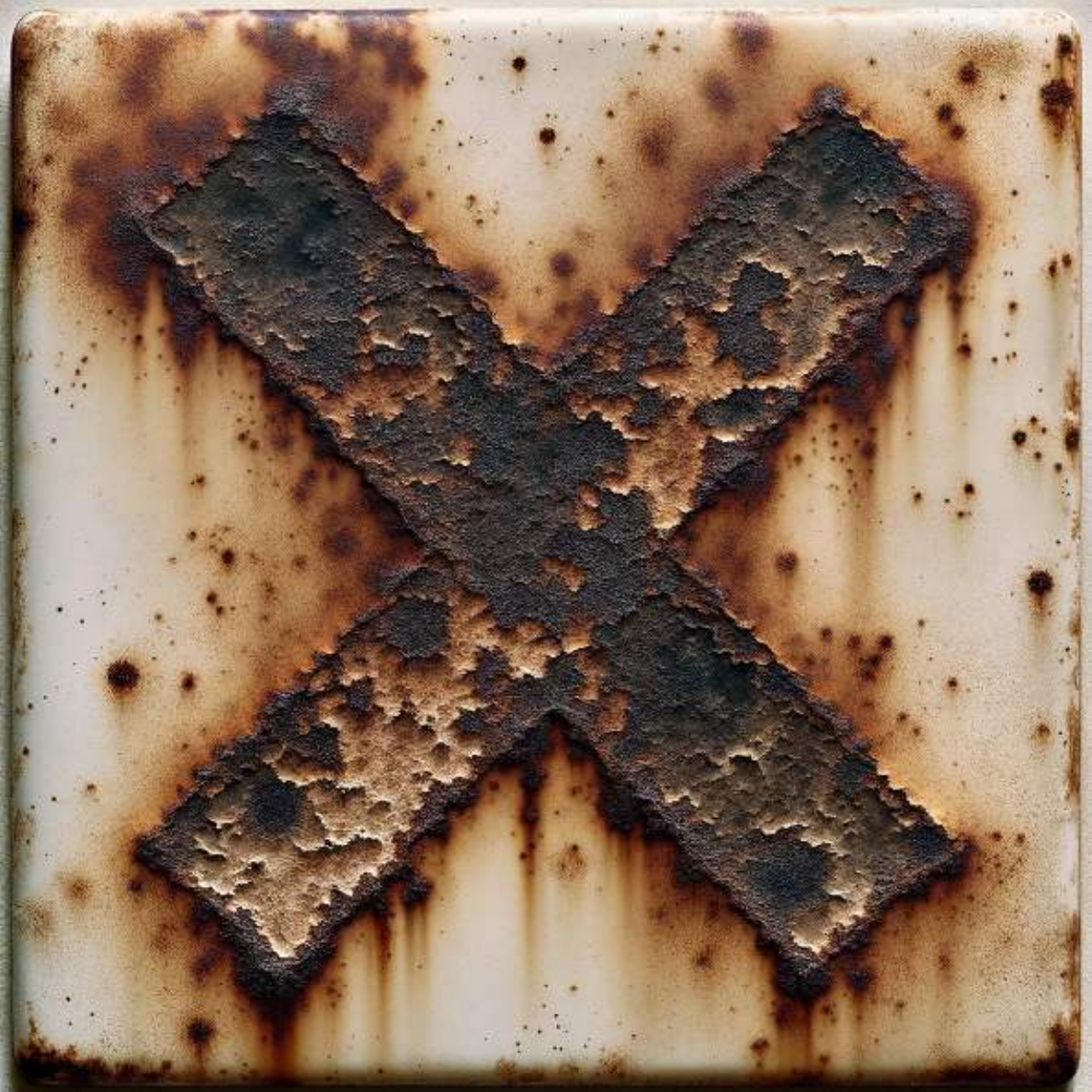} & 
\includegraphics[width=0.185\textwidth]{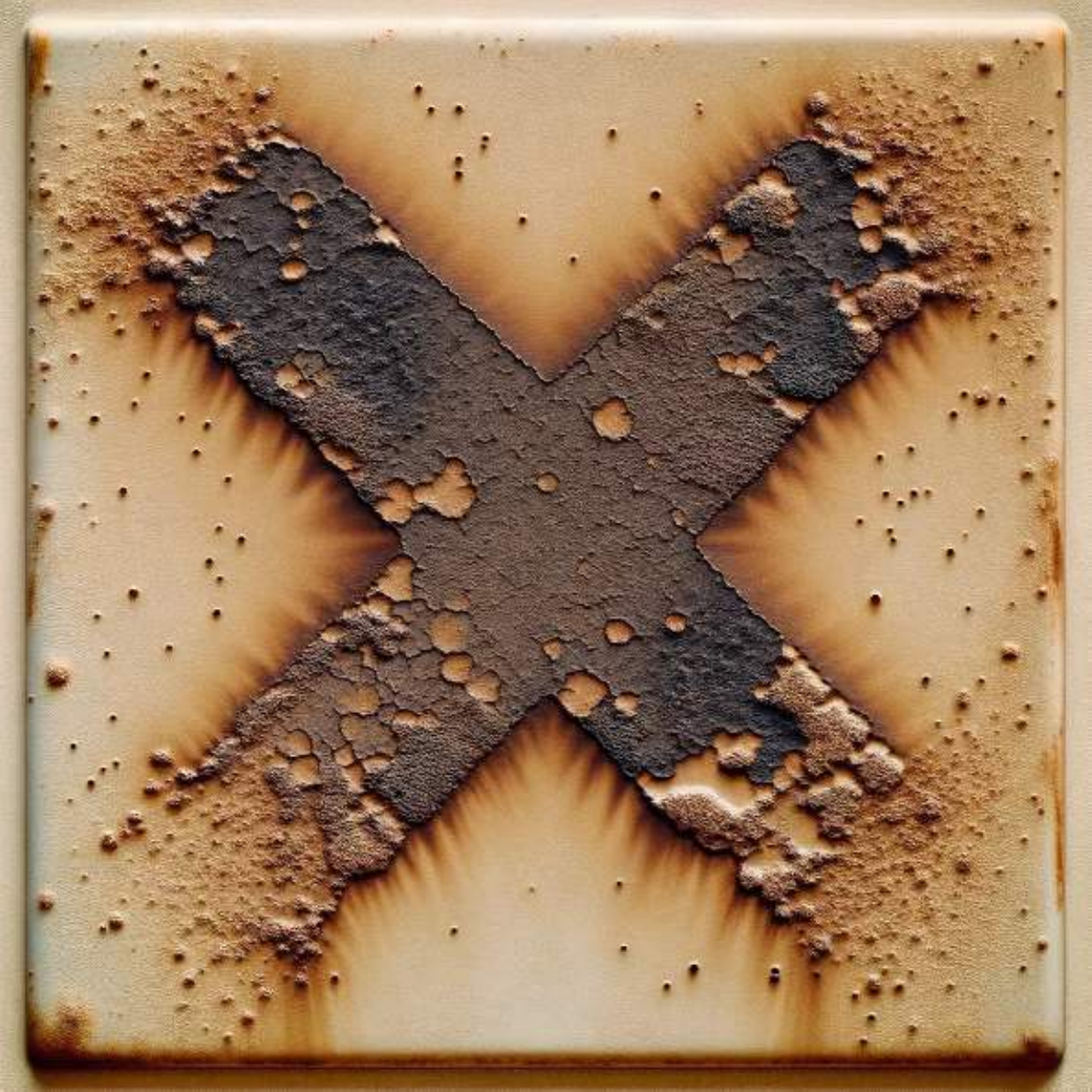} \\ [-0.65em]
\caption{The table shows the sample images generated by DALLE-3 with hints provided from a series of question-answer pairs based on the provided corrosion 5 category image as shown in the above Table \ref{tab:tab31}.}
\label{tab:tab32}
\end{tabularx}
\end{tcolorbox}

\clearpage
\onecolumn

\begin{tcolorbox}[colback=white!2!white,colframe=black!75!black]
\vspace{-5mm}
\begin{tabularx}{1.2\textwidth}{bss}
\caption{The table presents the question-answer pairs generated by the LMM(GPT-4V) in response to natural language questions about the input image of the corrosion 9 category from the CMI dataset. In summary, the question-answer pairs (generated text) provide in-depth information about an image related to the corrosion 9 category.}\label{tab:tab33}\\
\toprule
\begin{overpic}[width=0.185\columnwidth]{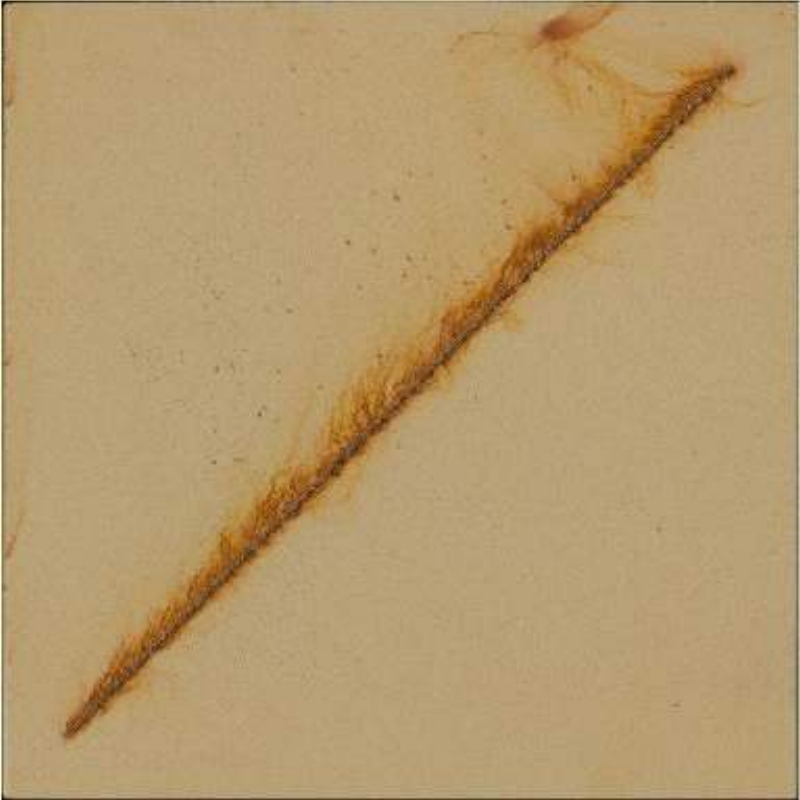}
\vspace{-5mm}
  \put(350,50){\makebox[0pt][c]{%
    \parbox{0.5\columnwidth}{%
      \centering
      \textbf{Please answer the following questions based on the provided input image belonging to the corrosion 9 category.} 
    }%
  }}  
\end{overpic}  \\

\midrule
\textbf{Query prompt}  & \textbf{GPT-4 Vision response} \\
\midrule
\endhead
\textcolor{black}{What is the primary intent of the image?} & The image is a representation of a corroded panel from the Corrosion Image Data Set for Automating Scientific Assessment of Materials, specifically with a Corrosion Rating of 9.
\\

\midrule
\textcolor{black}{Can you describe the overall color scheme of the image?} & The image predominantly features a light beige background, with the corrosion markings appearing in shades of brown, ranging from light to somewhat dark.
\\

\midrule
\textcolor{black}{What is the main corroded feature present in the image?} & The principal corroded feature is a single, diagonal scribe or line that runs across the panel, emanating signs of corrosion.
\\

\midrule
\textcolor{black}{How does the corrosion manifest around this scribed line?} & The corrosion around the scribe appears as a gradient, with the most intense corrosion closely hugging the scribe and gradually fading out as it spreads outward. It has an uneven, feathery appearance, with some areas showing darker brown patches.
\\

\midrule
\textcolor{black}{Given the context of corrosion ratings, how does a rating of 9 compare to other ratings in terms of corrosion severity?} & A rating of 9 indicates very minimal corrosion. The material is close to its pristine state, with only slight, localized signs of wear or corrosion. It's significantly better preserved than materials with lower ratings.
\\
 
\midrule
\textcolor{black}{Is the corrosion consistent throughout the scribed line?} & No, the corrosion varies in intensity. Some parts of the scribe show denser, more pronounced corrosion, while other parts are lighter and less distinct.
\\
 
\midrule
\textcolor{black}{How does the background of the image appear in comparison to the corroded area?} & The background is mostly uniform and retains its original beige color, with the exception of the area immediately surrounding the scribed line. This contrast emphasizes the localized nature of the corrosion in a rating 9 image. 
\\
 
\midrule
\textcolor{black}{If one were to run their fingers across the image, what differences in texture might they expect between the corroded scribe and the background?} & While the background would feel smooth, the corroded scribe would offer a rougher, uneven texture, reminiscent of early-stage rust or wear on a metal surface.
\\

\bottomrule
\end{tabularx}
\end{tcolorbox}

\vspace{-5mm}

\begin{tcolorbox}[colback=white!2!white,colframe=black!75!black]
\vspace{-4mm}
\begin{tabularx}{1.2\textwidth}{bss}
\includegraphics[width=0.185\textwidth]{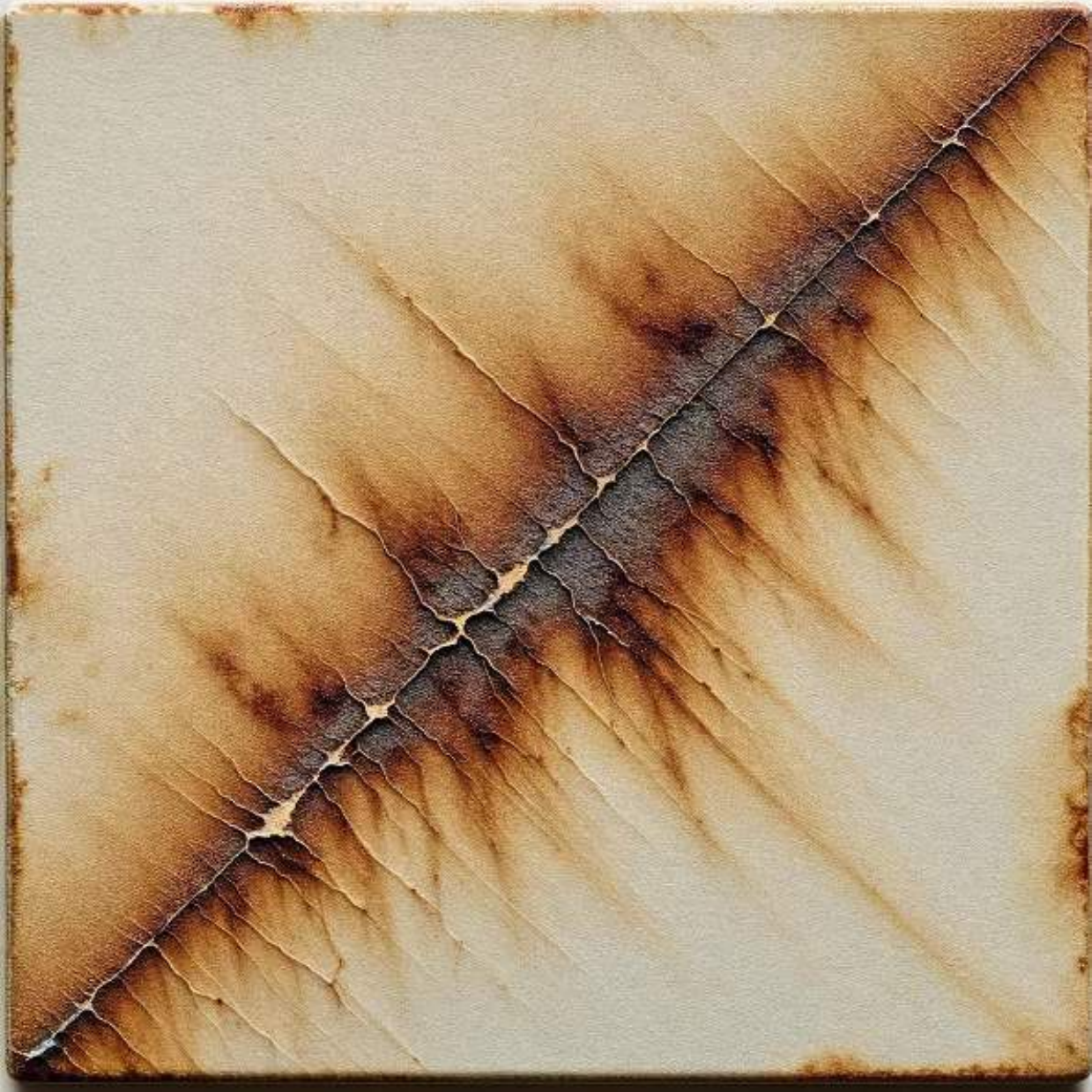} & 
\includegraphics[width=0.185\textwidth]{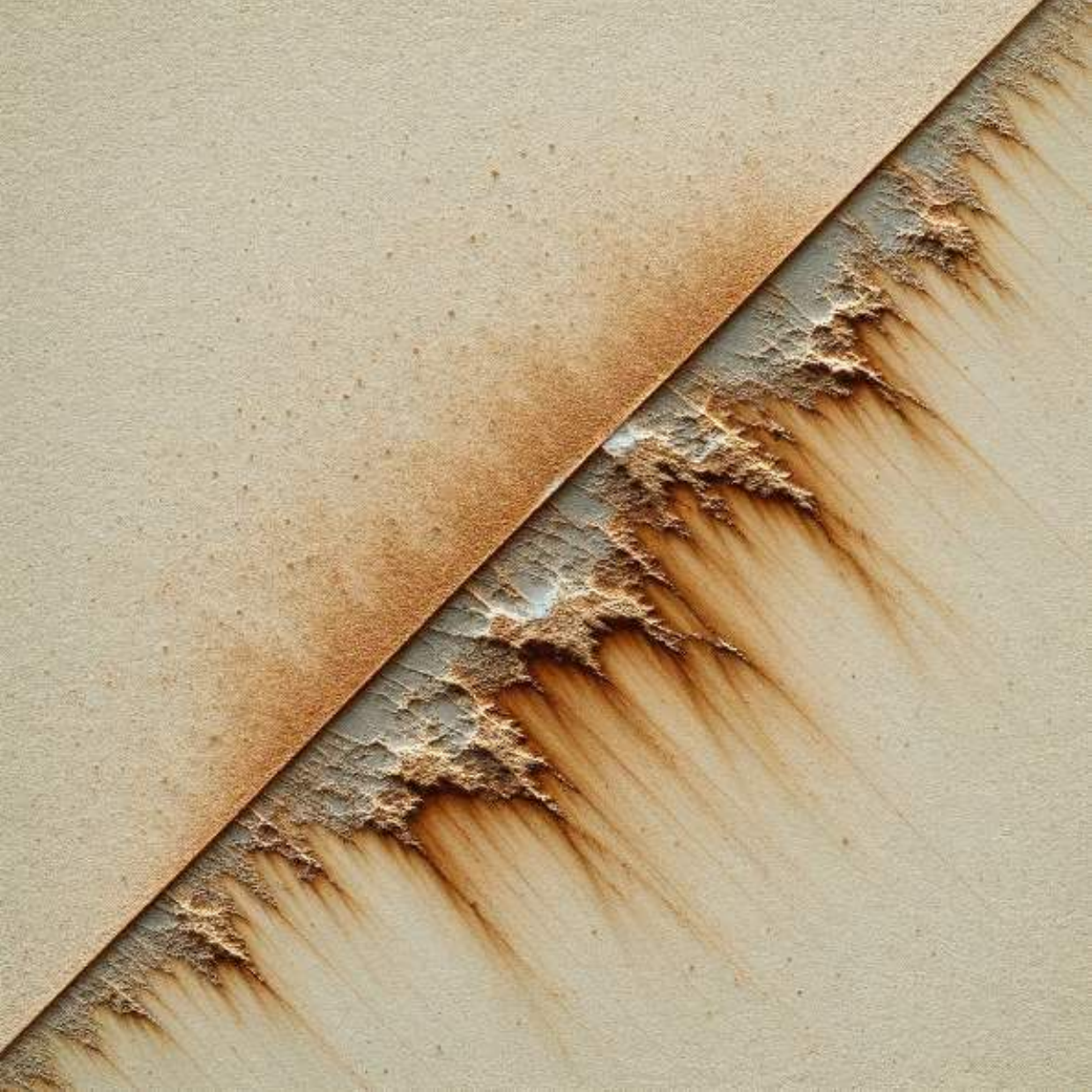} & 
\includegraphics[width=0.185\textwidth]{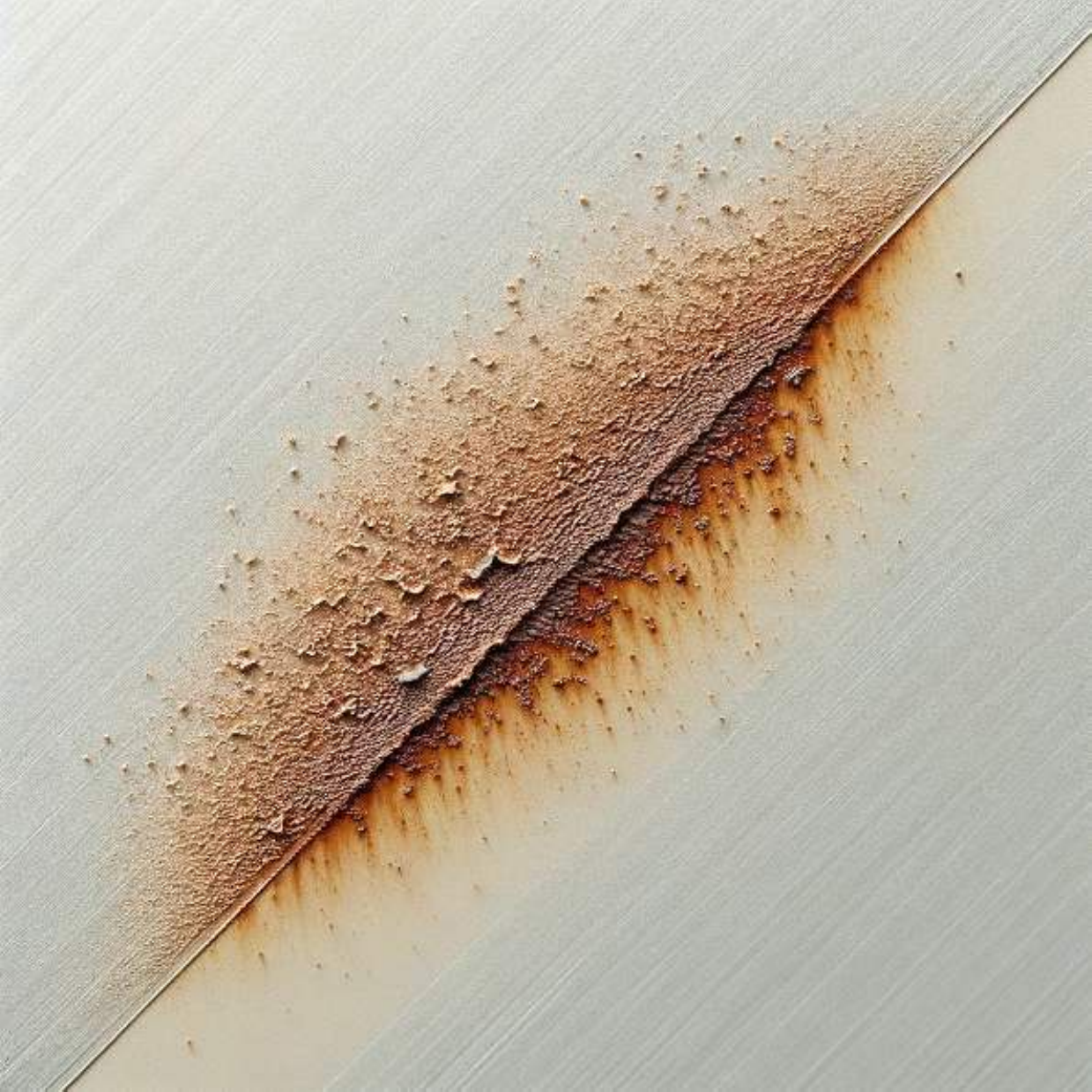} \\ [-0.65em]
\caption{The table shows the sample images generated by DALLE-3 with hints provided from a series of question-answer pairs based on the provided corrosion 9 category image as shown in the above Table \ref{tab:tab33}.}
\label{tab:tab34}
\end{tabularx}
\end{tcolorbox}

\clearpage
\onecolumn

\section{KTH-TIPS}
\begin{tcolorbox}[colback=white!2!white,colframe=black!75!black]
\vspace{-5mm}
\begin{tabularx}{1.2\textwidth}{bss}
\caption{The table presents the question-answer pairs generated by the LMM(GPT-4V) in response to natural language questions about the input image of the aluminium foil category from the KTH-Tips dataset. In summary, the question-answer pairs (retrieved text) provide in-depth information about an image related to aluminium foil category.}\label{tab:tab21}\\
\toprule
\begin{overpic}[width=0.185\columnwidth]{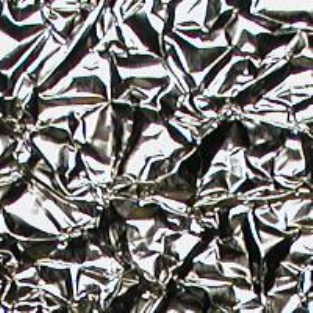}
\vspace{-5mm}
  \put(350,50){\makebox[0pt][c]{%
    \parbox{0.5\columnwidth}{%
      \centering
      \textbf{Please answer the following questions based on the provided input image belonging to the aluminium foil category.} 
    }%
  }}  
\end{overpic}  \\

\midrule
\textbf{Query prompt}  & \textbf{GPT-4 Vision response} \\
\midrule
\endhead
\textcolor{black}{What is the primary material depicted in the image?} & The image primarily shows aluminium foil.
\\

\midrule
\textcolor{black}{How is the surface texture of the material?} & The surface texture appears crumpled with multiple creases and folds. 
\\

\midrule
\textcolor{black}{What color predominates in the image?} &  The predominant color is metallic silver.
\\

\midrule
\textcolor{black}{Are there any reflections or shines visible on the material?} &  Yes, there are several shiny reflections visible due to the metallic nature of the foil.
\\

\midrule
\textcolor{black}{How would you describe the overall pattern on the foil?} & The foil presents a chaotic pattern with random creases and crinkles throughout.
\\
 
\midrule
\textcolor{black}{Is the foil flat or does it appear to have depth and volume?} &  The foil appears to have depth and volume due to its crumpled nature.
\\
 
\midrule
\textcolor{black}{Does the foil cover the entirety of the image, or are there any other elements present?} &  The foil covers the entire image, with no other elements present.
\\
 
\midrule
\textcolor{black}{How would you describe the lighting on the foil?} &  The lighting appears to be coming from multiple directions, causing multiple bright reflections on the foil's surface.
\\
 
\midrule
\textcolor{black}{Is there any uniformity in the crinkles and folds, or are they random?} &  The crinkles and folds seem to be random with no uniform pattern.
\\ 

\midrule
\textcolor{black}{Is the background distinguishable, or does the foil cover it completely?} & The foil covers the entire image, making the background indistinguishable.
\\
\bottomrule
\end{tabularx}
\end{tcolorbox}

\vspace{-4mm}

\begin{tcolorbox}[colback=white!2!white,colframe=black!75!black]
\vspace{-3mm}
\begin{tabularx}{1.2\textwidth}{bss}
\includegraphics[width=0.185\textwidth]{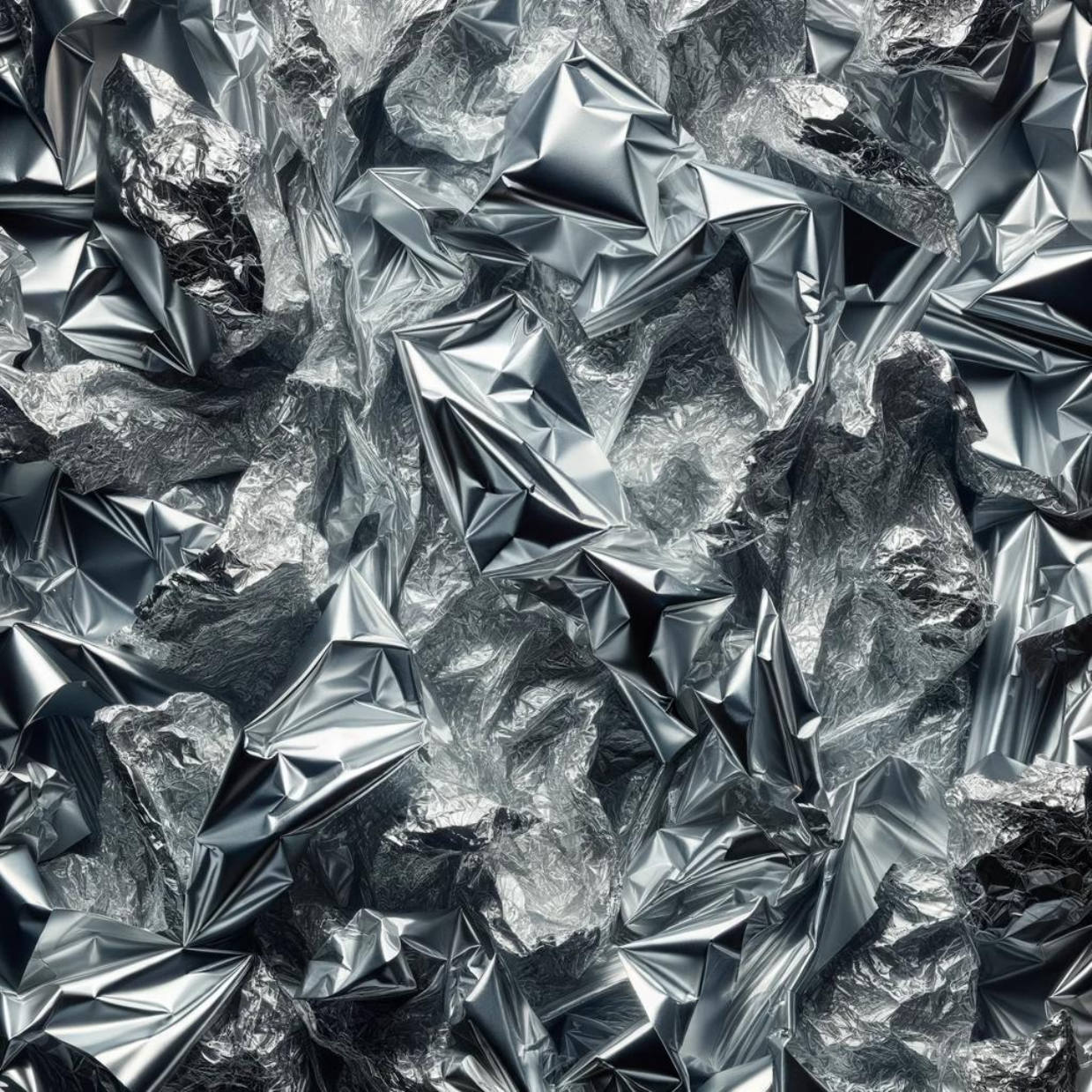} & 
\includegraphics[width=0.185\textwidth]{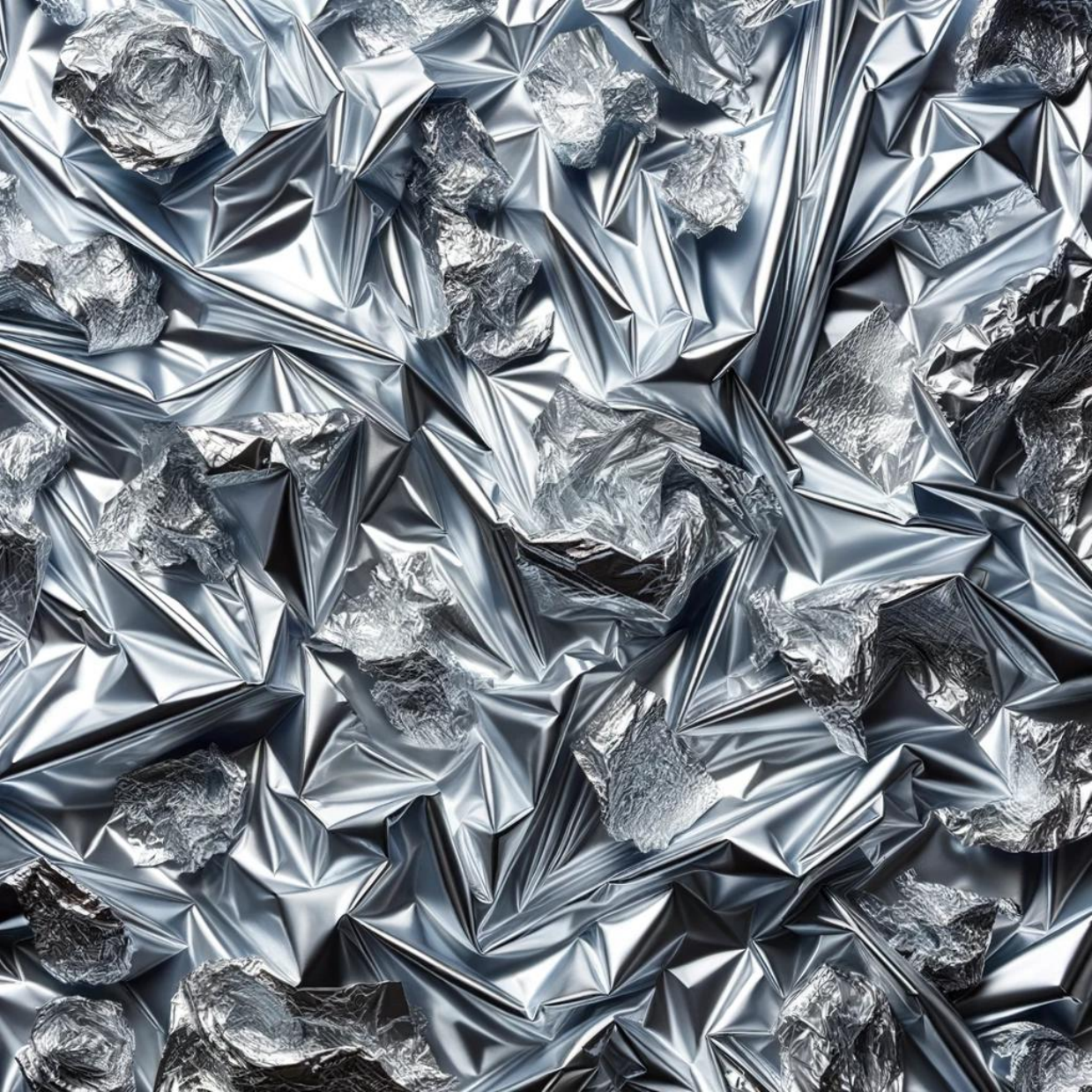} & 
\includegraphics[width=0.185\textwidth]{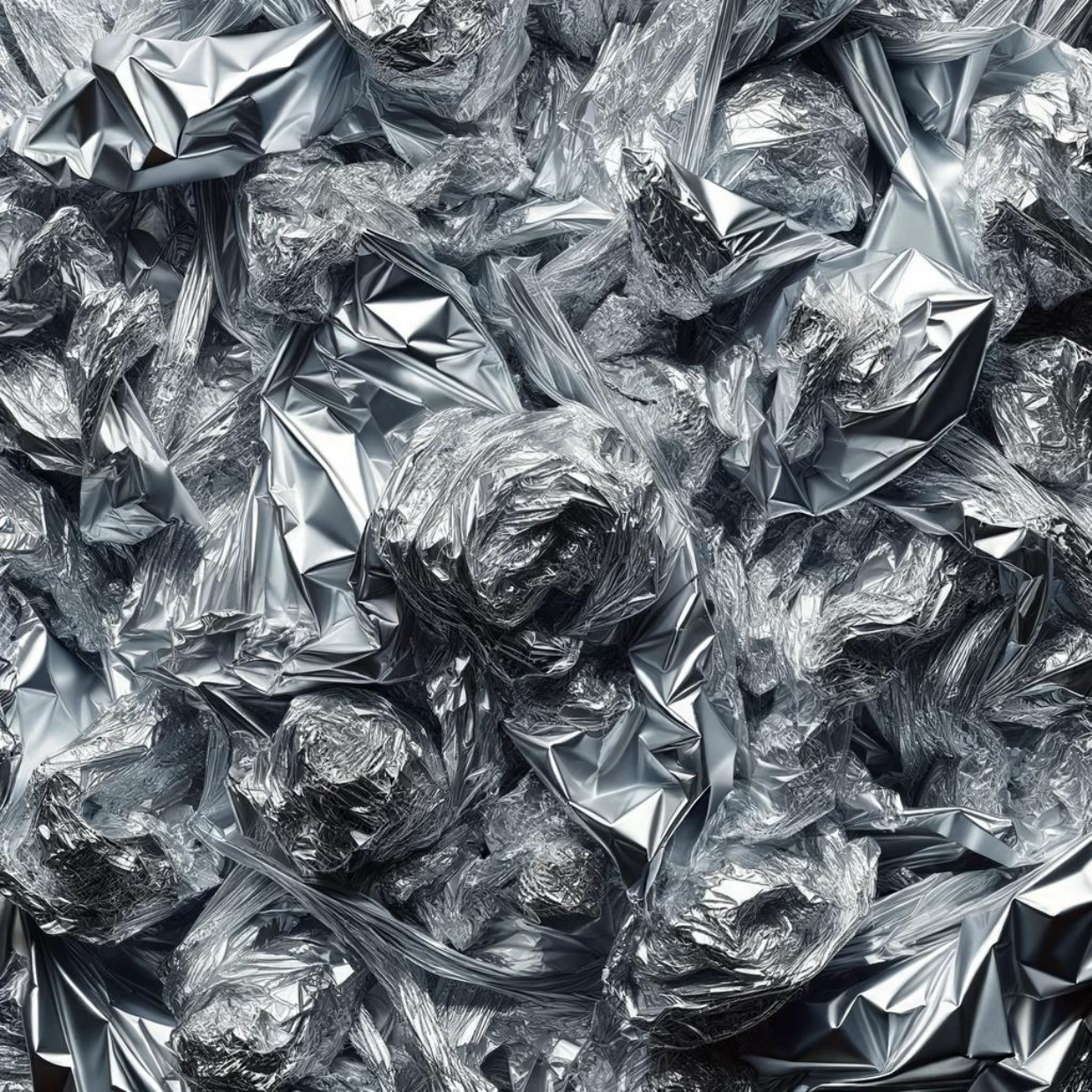} \\

\caption{The table shows the sample images generated by DALLE-3 with hints provided from a series of question-answer pairs based on the provided aluminium foil image as shown in the above Table \ref{tab:tab21}.}
\label{tab:tab22}
\end{tabularx}
\end{tcolorbox}

\clearpage
\onecolumn

\begin{tcolorbox}[colback=white!2!white,colframe=black!75!black]
\vspace{-5mm}
\begin{tabularx}{1.2\textwidth}{bss}
\caption{The table presents the question-answer pairs generated by the LMM(GPT-4V) in response to natural language queries about the input image of the corduroy category from the KTH-Tips dataset. In summary, the question-answer pairs (generated text) provide in-depth information about an image related to corduroy category.}\label{tab:tab23}\\
\toprule
\begin{overpic}[width=0.185\columnwidth]{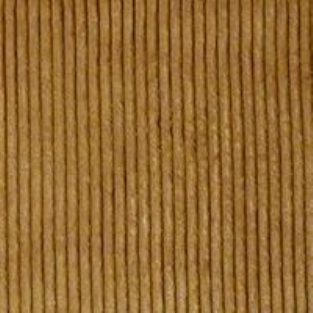}
\vspace{-5mm}
  \put(350,50){\makebox[0pt][c]{%
    \parbox{0.5\columnwidth}{%
      \centering
      \textbf{Please answer the following questions based on the provided input image belonging to the corduroy category.} 
    }%
  }}  
\end{overpic}  \\

\midrule
\textbf{Query prompt}  & \textbf{GPT-4 Vision response} \\
\midrule
\endhead
\textcolor{black}{What type of material is depicted in the image?} & The image illustrates a segment of corduroy fabric.
\\

\midrule
\textcolor{black}{What color does the corduroy fabric primarily exhibit?} & The corduroy fabric primarily presents a warm tan or light brown hue.
\\

\midrule
\textcolor{black}{Describe the texture and pattern noticeable on the corduroy.} & The corduroy exhibits a distinct vertical ribbed texture, characterized by continuous raised lines or wales.
\\

\midrule
\textcolor{black}{Are these raised lines uniformly distributed throughout the fabric?} & Yes, the raised lines appear evenly spaced and maintain a consistent width across the fabric.
\\

\midrule
\textcolor{black}{How does light interact with the fabric's texture?} & The raised lines or wales reflect light more prominently than the recessed sections, giving them a subtle sheen.
\\
 
\midrule
\textcolor{black}{Describe the spacing between the ribs.} &  The ribs are closely spaced, giving the fabric a densely ribbed appearance.
\\
 
\midrule
\textcolor{black}{How does the fabric's texture vary between the raised lines and the gaps in between?} &  The raised lines have a slightly rough and elevated texture, while the intervening gaps are smoother and more recessed.
\\
 
\midrule
\textcolor{black}{Are there any visible imperfections or signs of wear on the corduroy?} &  The fabric appears to be in excellent condition with no apparent signs of wear or imperfections.
\\
 
\midrule
\textcolor{black}{From which angle or perspective is the fabric being observed?} &  The fabric is viewed from a close-up perspective, highlighting its vertical ribbed texture in detail.
\\ 

\midrule
\textcolor{black}{Is the image a full representation of the fabric or just a section?} & The image showcases a section of the corduroy fabric, emphasizing its signature texture and pattern.
\\
\bottomrule
\end{tabularx}
\end{tcolorbox}

\vspace{-4mm}

\begin{tcolorbox}[colback=white!2!white,colframe=black!75!black]
\vspace{-3mm}
\begin{tabularx}{1.2\textwidth}{bss}
\includegraphics[width=0.185\textwidth]{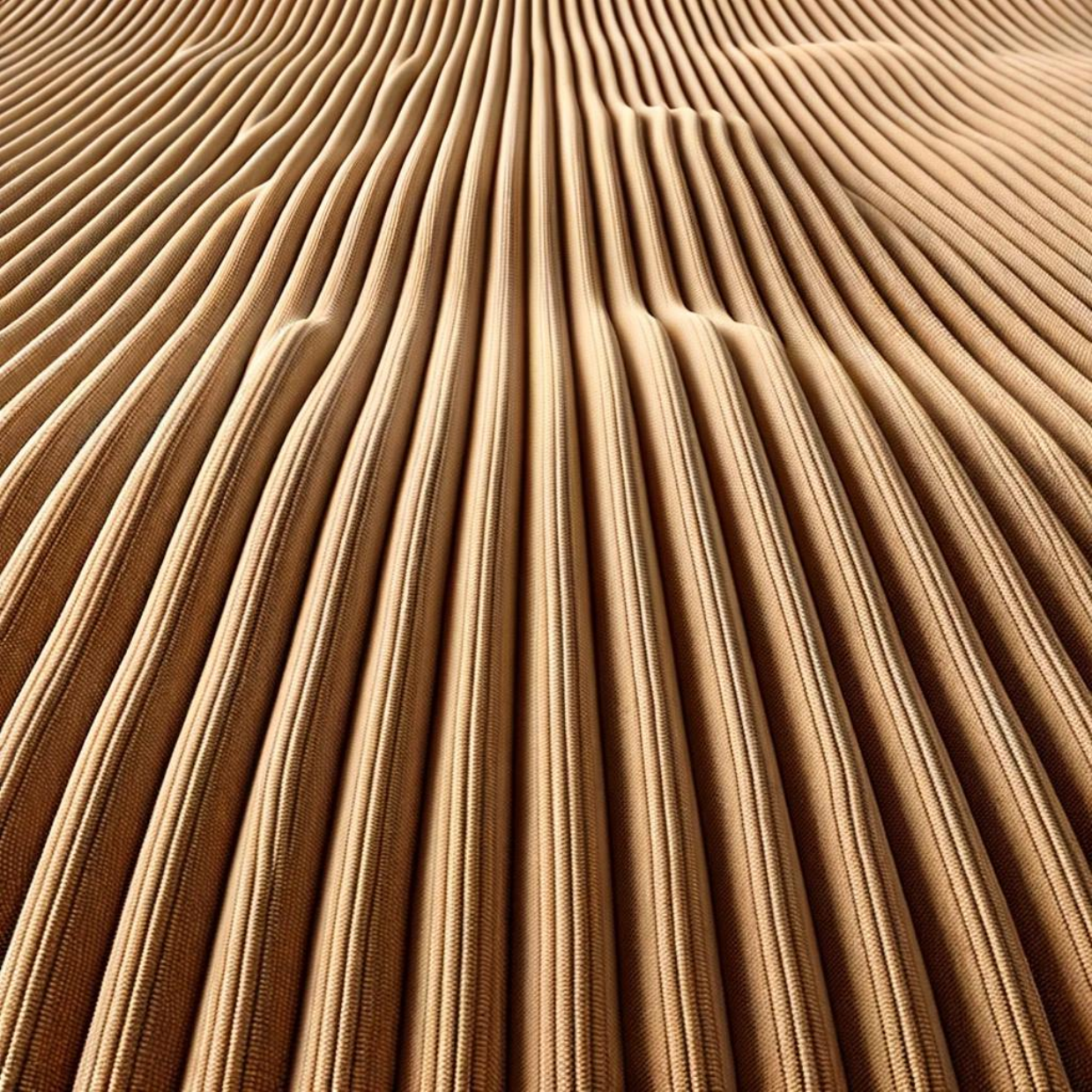} & 
\includegraphics[width=0.185\textwidth]{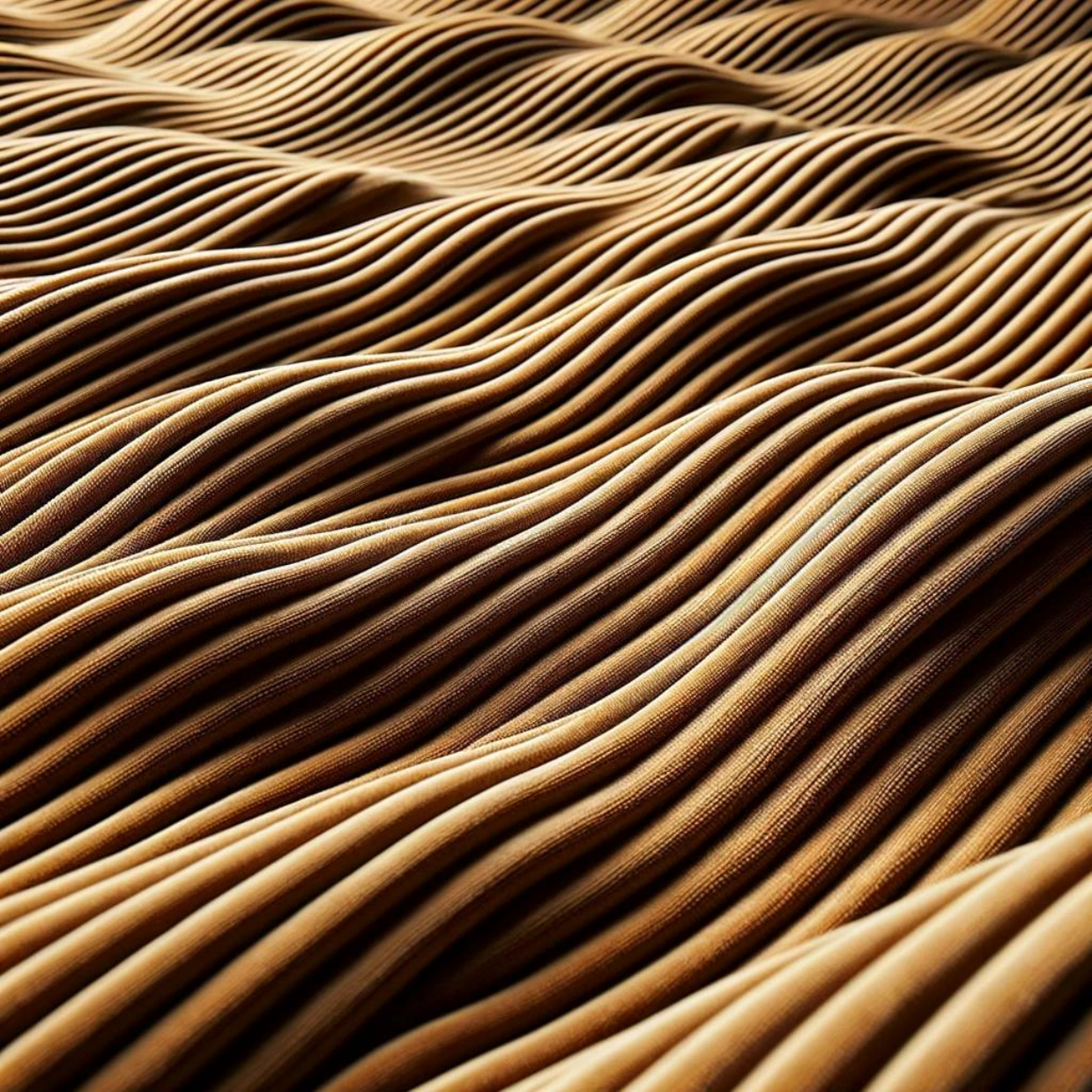} & 
\includegraphics[width=0.185\textwidth]{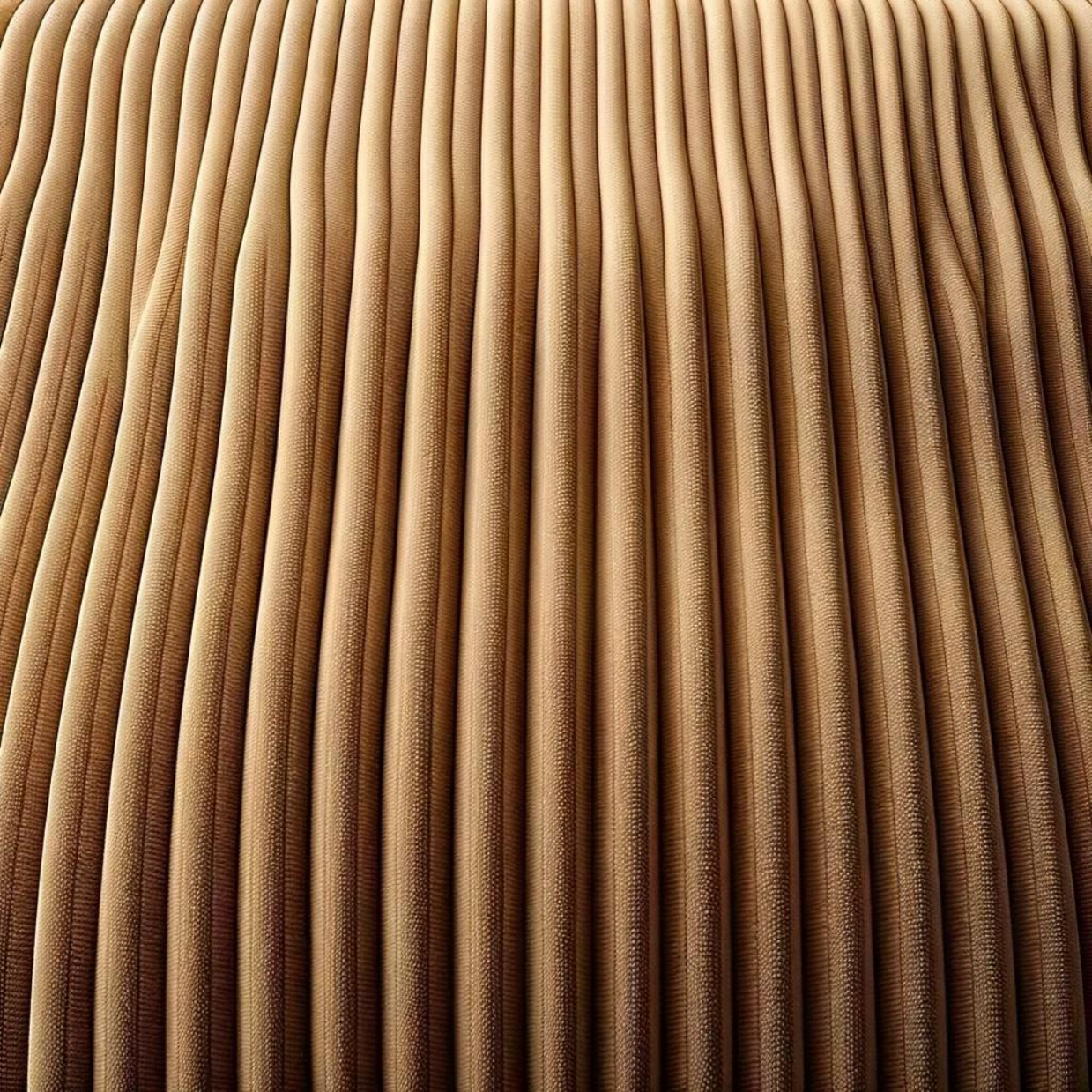} 
\\
\caption{The table shows the sample images generated by DALLE-3 with hints provided from a series of question-answer pairs based on the provided Corduroy image as shown in the above Table \ref{tab:tab23}.}
\label{tab:tab24}
\end{tabularx}
\end{tcolorbox}

\clearpage
\onecolumn

\begin{tcolorbox}[colback=white!2!white,colframe=black!75!black]
\vspace{-5mm}
\begin{tabularx}{1.2\textwidth}{bss}
\caption{The table presents the question-answer pairs generated by the LMM(GPT-4V) in response to natural language questions about the input image of the linen category from the KTH-Tips dataset. In summary, the question-answer pairs (generated text) provide in-depth information about an image related to linen category.}\label{tab:tab25}\\
\toprule
\begin{overpic}[width=0.185\columnwidth]{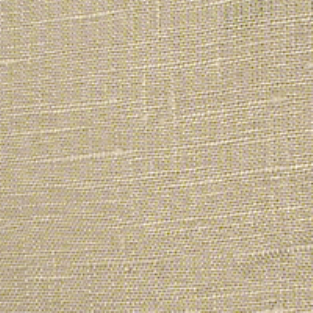}
\vspace{-5mm}
  \put(350,50){\makebox[0pt][c]{%
    \parbox{0.5\columnwidth}{%
      \centering
      \textbf{Please answer the following questions based on the provided input image belonging to the linen category.} 
    }%
  }}  
\end{overpic}  \\

\midrule
\textbf{Query prompt}  & \textbf{GPT-4 Vision response} \\
\midrule
\endhead
\textcolor{black}{What is the primary material displayed in the image?} & The image prominently features a section of linen fabric.
\\

\midrule
\textcolor{black}{Can you identify the primary color of the linen fabric in the image?} & The linen fabric has a light beige or off-white color.
\\

\midrule
\textcolor{black}{Describe the texture or pattern visible on the linen.} & The linen showcases a distinct woven texture, characterized by closely interlaced threads in a crisscross pattern.
\\

\midrule
\textcolor{black}{How would you describe the spacing between the woven threads?} & The threads are finely woven with a relatively tight weave, but individual threads can be distinctly seen due to the inherent coarse nature of linen.
\\

\midrule
\textcolor{black}{Does the linen appear soft or rigid based on the visible texture?} & Based on the texture, the linen seems to have a crisp yet naturally soft appearance, typical of the material.
\\
 
\midrule
\textcolor{black}{Are there any observable sheen or reflective properties on the linen?} &  The linen has a matte finish with no significant sheen, reflecting its natural and breathable quality.

\\
 
\midrule
\textcolor{black}{How consistent is the texture throughout the displayed section?} & The texture is relatively consistent, but there are subtle variations in the thread thickness and weave, imparting a natural and organic feel to the linen.

\\
 
\midrule
\textcolor{black}{ Based on the visible attributes, how would you describe the tactile sensation of the fabric?} &  Given its appearance, the fabric likely feels textured but soft to the touch, cool, and breathable — typical qualities of linen.
\\
 
\midrule
\textcolor{black}{Are there any discernible imperfections or unique characteristics in the fabric?} &  The linen has an inherently uneven texture with some minor variations in the weave, which is characteristic of natural linen and adds to its charm. There don't appear to be any significant imperfections or stains.
\\ 

\bottomrule
\end{tabularx}
\end{tcolorbox}

\vspace{-4mm}

\begin{tcolorbox}[colback=white!2!white,colframe=black!75!black]
\vspace{-3mm}
\begin{tabularx}{1.2\textwidth}{bss}
\includegraphics[width=0.185\textwidth]{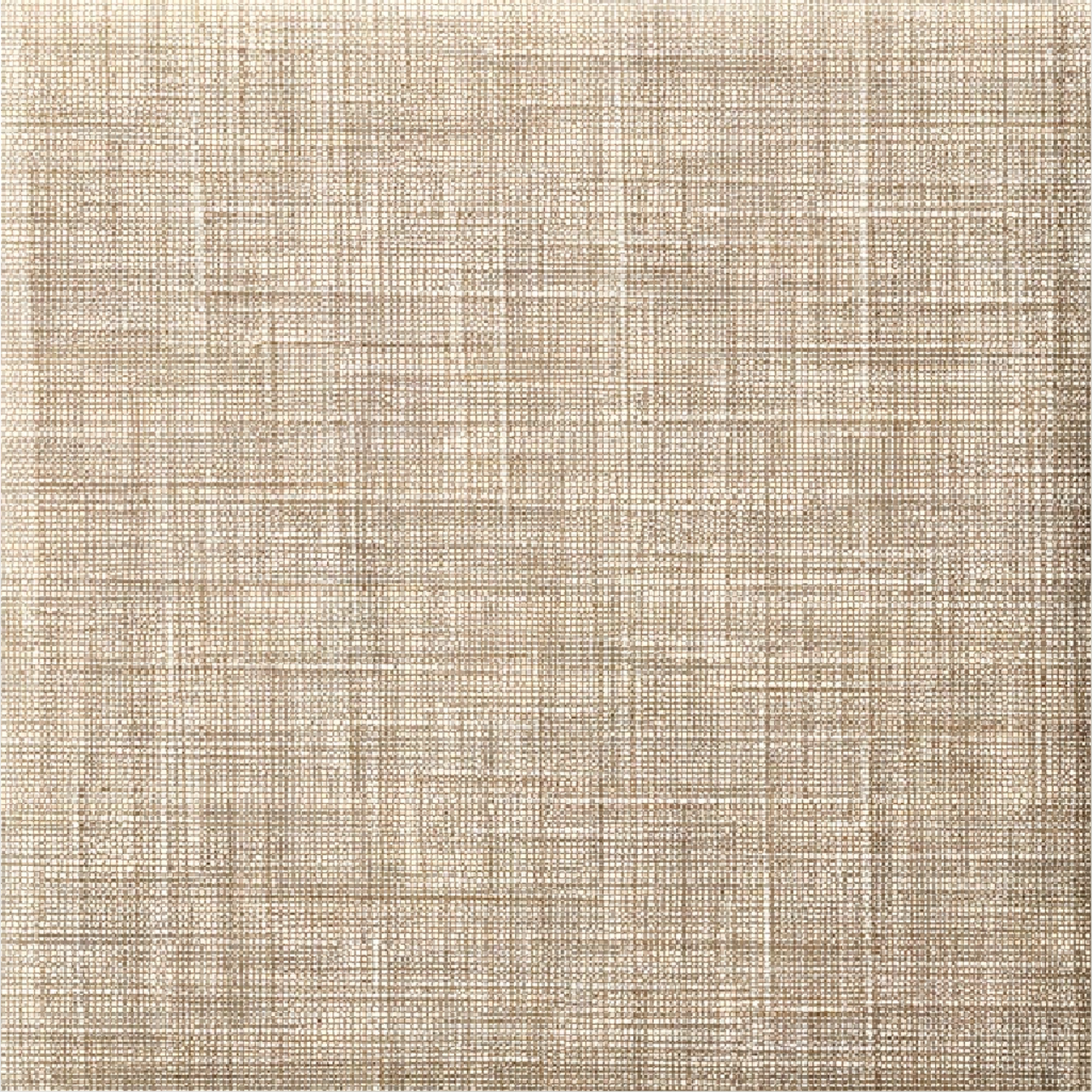} & 
\includegraphics[width=0.185\textwidth]{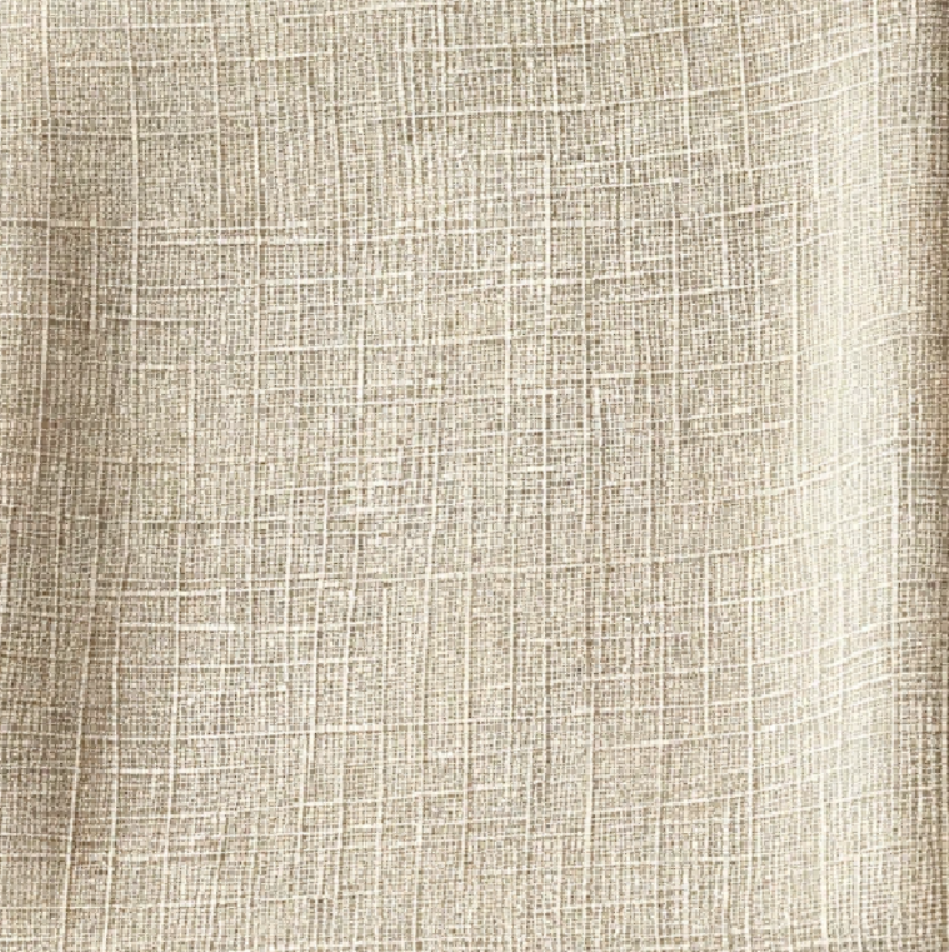} & 
\includegraphics[width=0.185\textwidth]{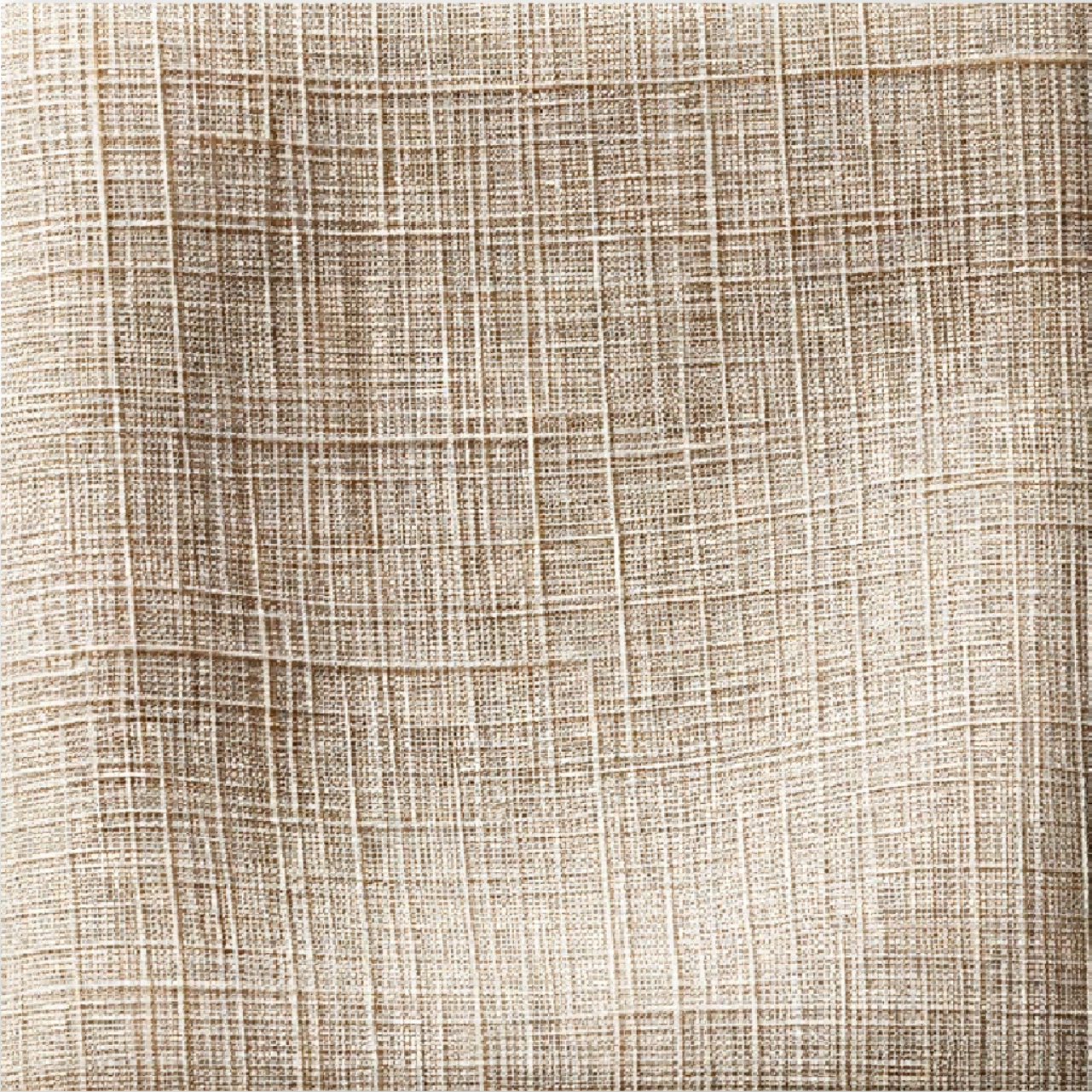} 
\\
\caption{The table shows the sample images generated by DALLE-3 with hints provided from a series of question-answer pairs based on the provided linen image as shown in the above Table \ref{tab:tab25}.}
\label{tab:tab26}
\end{tabularx}
\end{tcolorbox}

\clearpage
\onecolumn

\section{NEU-SDD}
\begin{tcolorbox}[colback=white!2!white,colframe=black!75!black]
\vspace{-5mm}
\begin{tabularx}{1.2\textwidth}{bss}
\caption{The table presents the question-answer pairs generated by the LMM(GPT-4V) in response to natural language questions about the input image of the scratches category from the NEU-SDD dataset. In summary, the question-answer pairs (generated text) provide in-depth information about an image related to the scratches category.}\label{tab:tab27}\\
\toprule
\begin{overpic}[width=0.185\columnwidth]{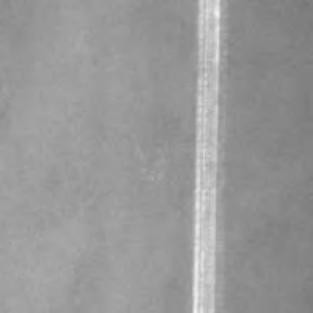}
\vspace{-5mm}
  \put(350,50){\makebox[0pt][c]{%
    \parbox{0.5\columnwidth}{%
      \centering
      \textbf{Please answer the following questions based on the provided input image belonging to the scratches category.} 
    }%
  }}  
\end{overpic}  \\

\midrule
\textbf{Query prompt}  & \textbf{GPT-4 Vision response} \\
\midrule
\endhead
\textcolor{black}{What category does this texture belong to?} & The scratches category of the NEU-SDD dataset.
\\

\midrule
\textcolor{black}{Describe the central feature in the image.} & A pronounced vertical scratch or line that is brighter than the surrounding area.
\\

\midrule
\textcolor{black}{How would you describe the contrast between the scratch and the surrounding texture?} & The scratch is much brighter and distinct, contrasting strongly with the darker, uniform background.
\\

\midrule
\textcolor{black}{What color tones dominate the image?} & Shades of gray dominate the major parts of the image.
\\

\midrule
\textcolor{black}{Can you define the orientation of the scratch?} & The scratch is vertical.
\\
 
\midrule
\textcolor{black}{How would you describe the edges of the scratch?} & The edges of the scratch are somewhat jagged and irregular, with varying brightness.

\\
 
\midrule
\textcolor{black}{Is the scratch in the center or towards the side of the image?} & It's centralized in the image.

\\
 
\midrule
\textcolor{black}{Are there multiple scratches or just a single prominent one?} & There's a single prominent scratch, but it appears to have multiple adjacent lines or striations.
\\
 
\midrule
\textcolor{black}{Is the rest of the image's texture smooth or grainy?} & The surrounding texture is relatively smooth with a subtle graininess.
\\ 

\midrule
\textcolor{black}{How does the brightness of the scratch compare to the overall image?} & The scratch is notably brighter than the overall image.
\\

\midrule
\textcolor{black}{If you were to touch the scratch, would it feel raised or recessed?} & It might feel recessed or indented compared to the surrounding surface.
\\

\bottomrule
\end{tabularx}
\end{tcolorbox}

\vspace{-6mm}

\begin{tcolorbox}[colback=white!2!white,colframe=black!75!black]
\vspace{-5mm}
\begin{tabularx}{1.2\textwidth}{bss}
\includegraphics[width=0.185\textwidth]{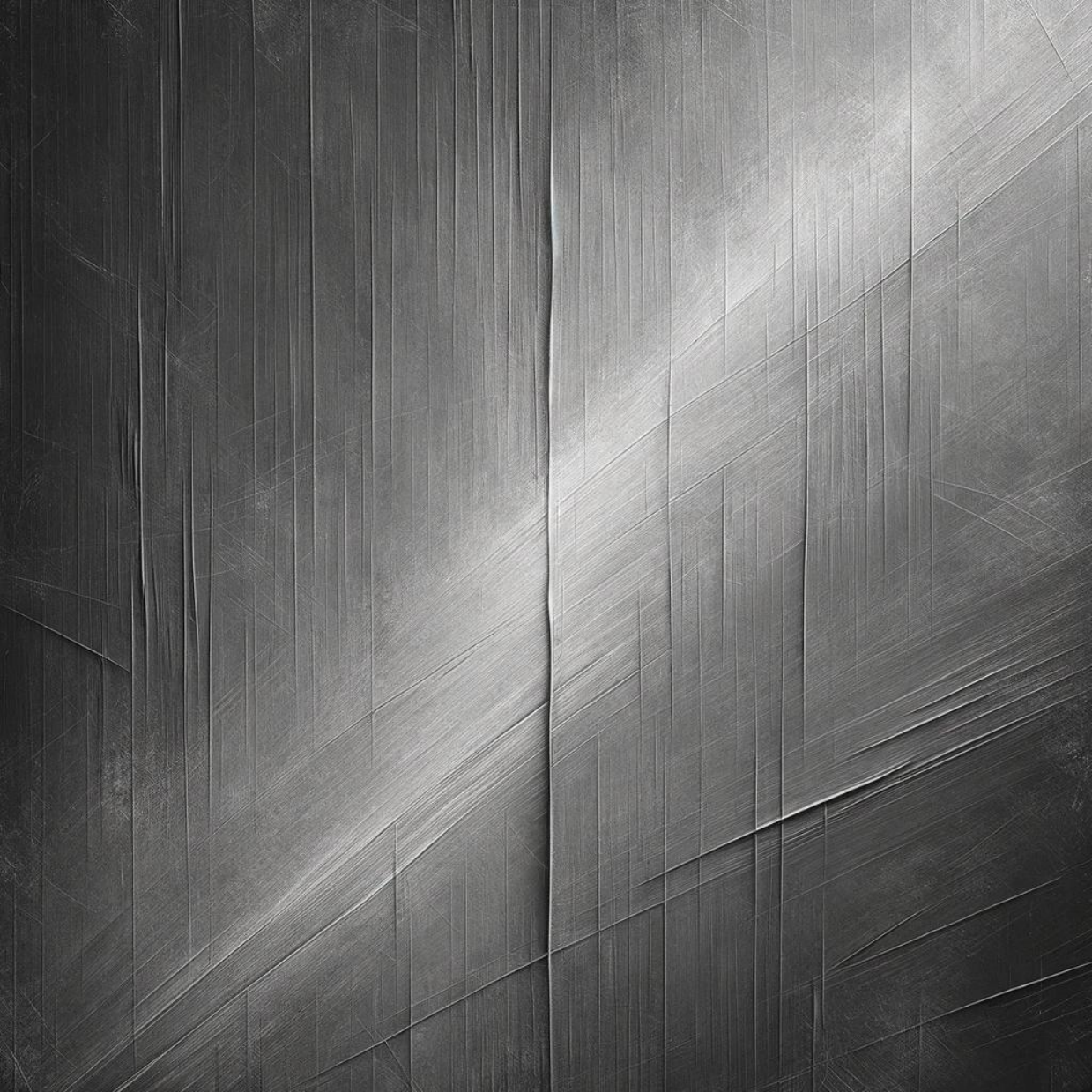} & 
\includegraphics[width=0.185\textwidth]{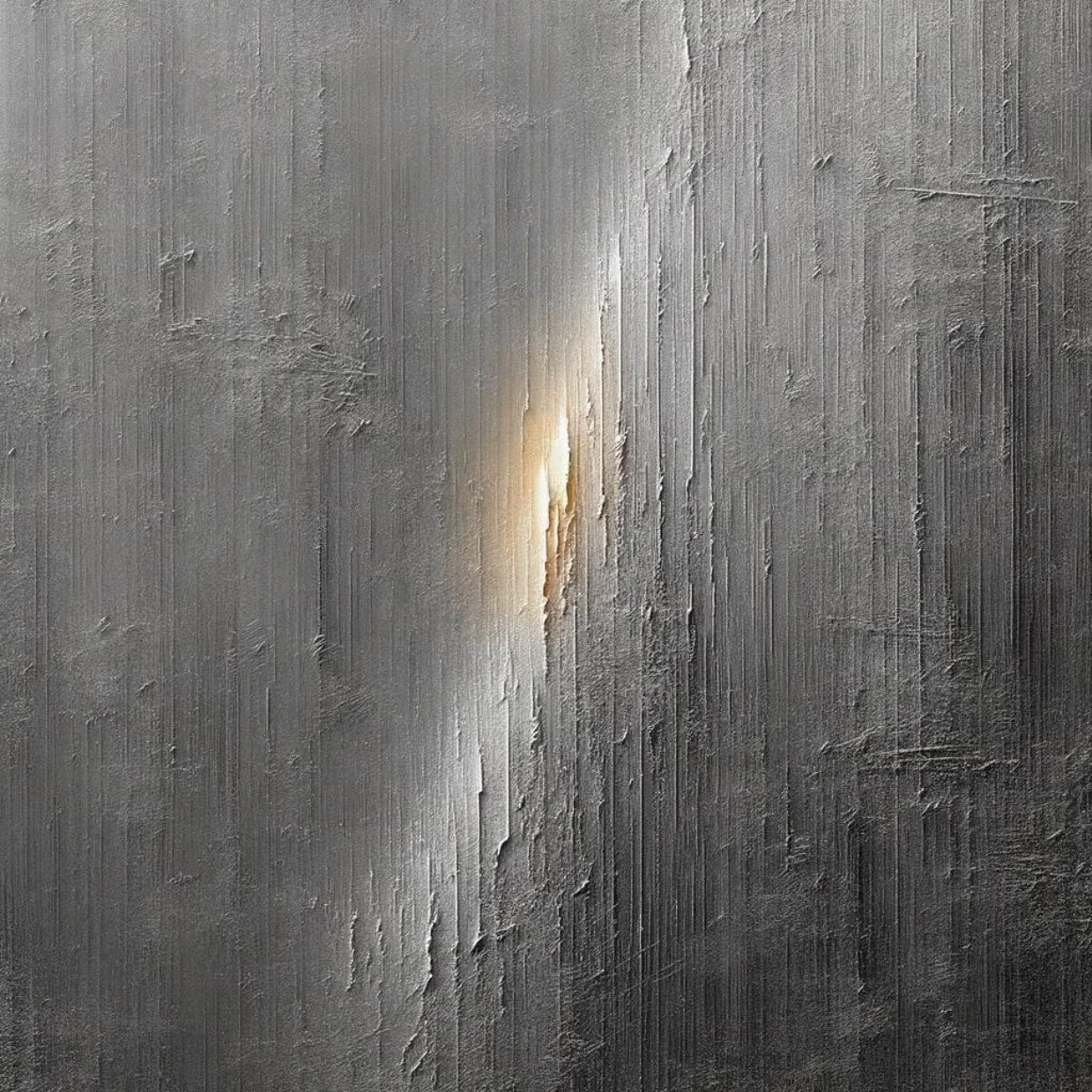} & 
\includegraphics[width=0.185\textwidth]{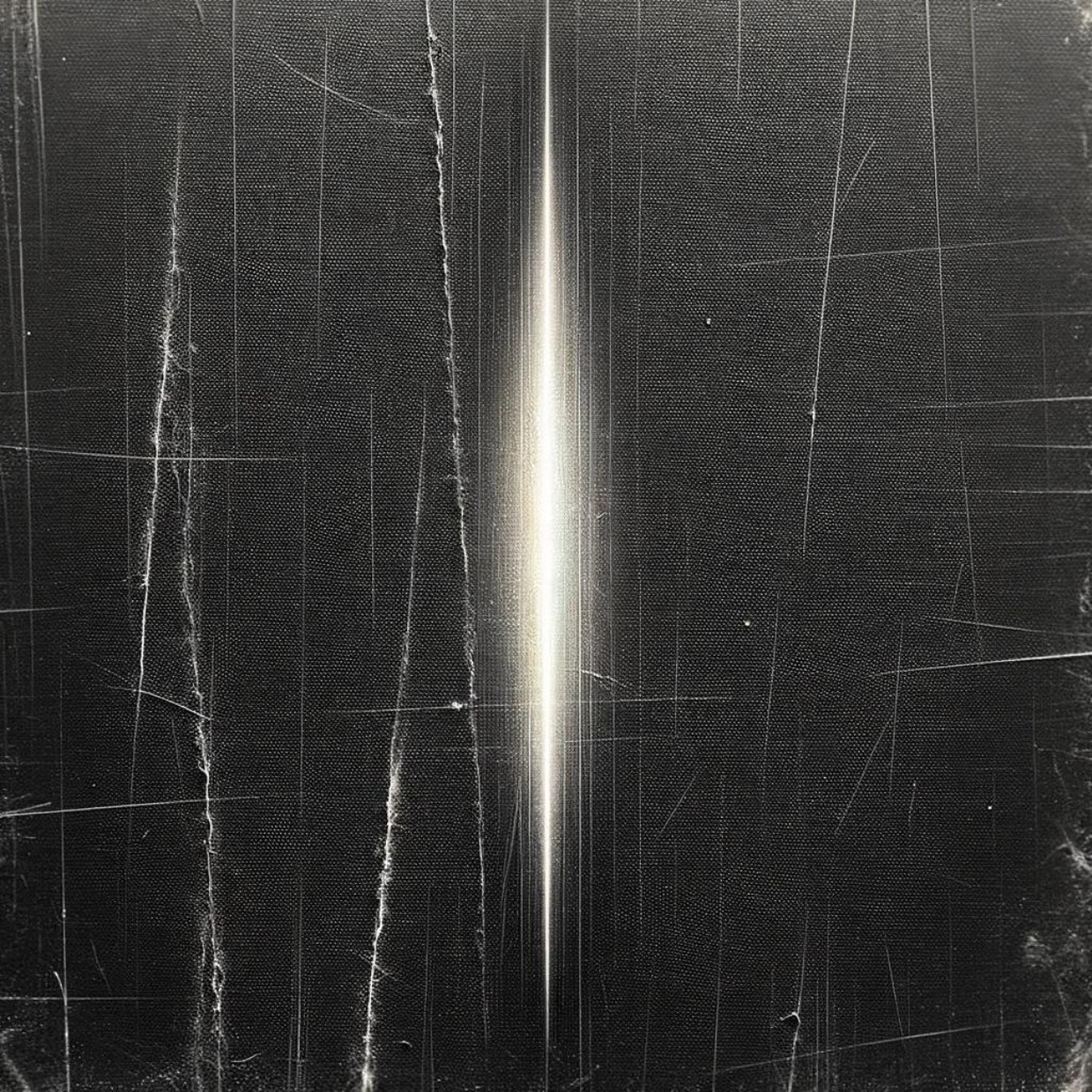} \\  [-0.65em]
\caption{The table shows the sample images generated by DALLE-3 with hints provided from a series of question-answer pairs based on the provided scratches category image as shown in the above Table \ref{tab:tab27}.}
\label{tab:tab28}
\end{tabularx}
\end{tcolorbox}

\clearpage
\onecolumn

\begin{tcolorbox}[colback=white!2!white,colframe=black!75!black]
\vspace{-5mm}
\begin{tabularx}{1.2\textwidth}{bss}
\caption{The table presents the question-answer pairs generated by the LMM(GPT-4V) in response to natural language questions about the input image of the patches category from the NEU-SDD dataset. In summary, the question-answer pairs (generated text) provide in-depth information about an image related to the patches category.}\label{tab:tab29}\\
\toprule
\begin{overpic}[width=0.185\columnwidth]{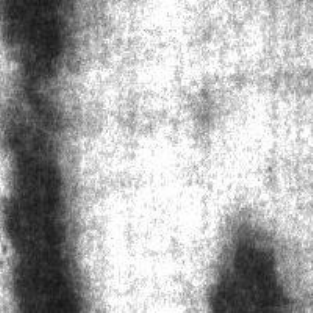}
\vspace{-5mm}
  \put(350,50){\makebox[0pt][c]{%
    \parbox{0.5\columnwidth}{%
      \centering
      \textbf{Please answer the following questions based on the provided input image belonging to the patches category.} 
    }%
  }}  
\end{overpic}  \\

\midrule
\textbf{Query prompt}  & \textbf{GPT-4 Vision response} \\
\midrule
\endhead
\textcolor{black}{What category does this texture belong to?} & The patches category of the NEU-SDD dataset.
\\

\midrule
\textcolor{black}{Describe the dominant features in the image.} & There are darker patches that contrast with the grainy and lighter background.
\\

\midrule
\textcolor{black}{How would you characterize the contrast between the patches and the surrounding area?} & The patches are noticeably darker and stand out against the grainy, lighter background.
\\

\midrule
\textcolor{black}{What color tones are prevalent in the image?} & Various shades of gray.
\\

\midrule
\textcolor{black}{Describe the shape and arrangement of the patches.} & The patches have irregular shapes and are spread out, with one large patch on the left side of the image.
\\
 
\midrule
\textcolor{black}{How does the texture within the patches compare to the rest of the image?} & The patches have a smoother, more uniform texture, while the surrounding area appears noisier and more grainy.

\\
 
\midrule
\textcolor{black}{Would you say the patches are centralized or more dispersed?} & One significant patch is more towards the left, but overall they are dispersed.

\\
 
\midrule
\textcolor{black}{Do the patches have well-defined boundaries?} & The boundaries are somewhat blended and gradient, transitioning subtly into the surrounding texture.
\\
 
\midrule
\textcolor{black}{Describe the granularity of the background texture.} & The background has a fine-grained, noisy appearance with a speckled pattern.
\\ 

\midrule
\textcolor{black}{In a real-world scenario, how might such patches form on materials?} & These patches could result from discoloration, wear, stains, or certain manufacturing processes.
\\

\midrule
\textcolor{black}{Does the image lean more towards being bright, dark, or neutral overall?} & It strikes a balance, with the patches being dark and the surrounding being brighter and neutral.
\\

\bottomrule
\end{tabularx}
\end{tcolorbox}

\vspace{-4mm}

\begin{tcolorbox}[colback=white!2!white,colframe=black!75!black]
\vspace{-3mm}
\begin{tabularx}{1.2\textwidth}{bss}
\includegraphics[width=0.185\textwidth]{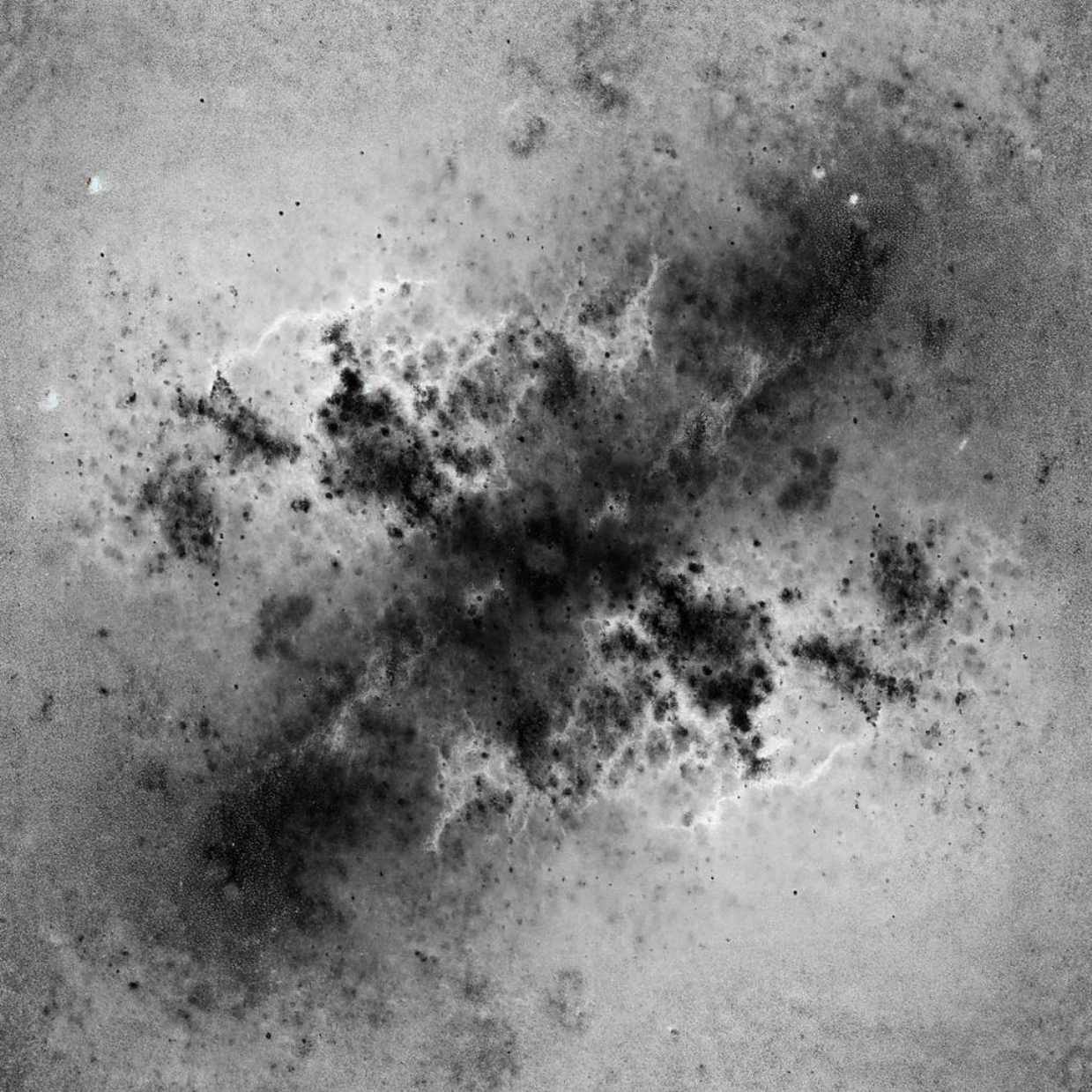} & 
\includegraphics[width=0.185\textwidth]{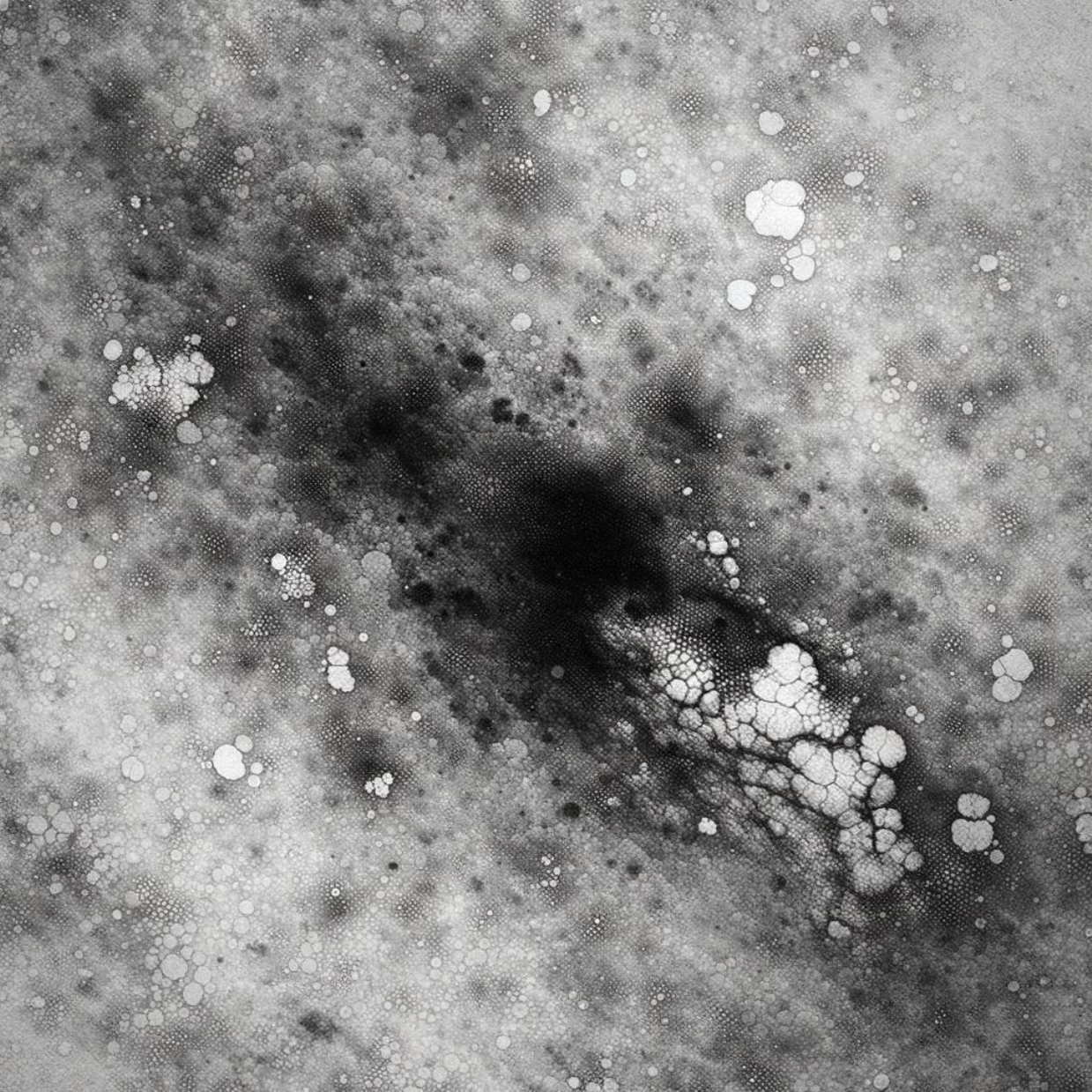} & 
\includegraphics[width=0.185\textwidth]{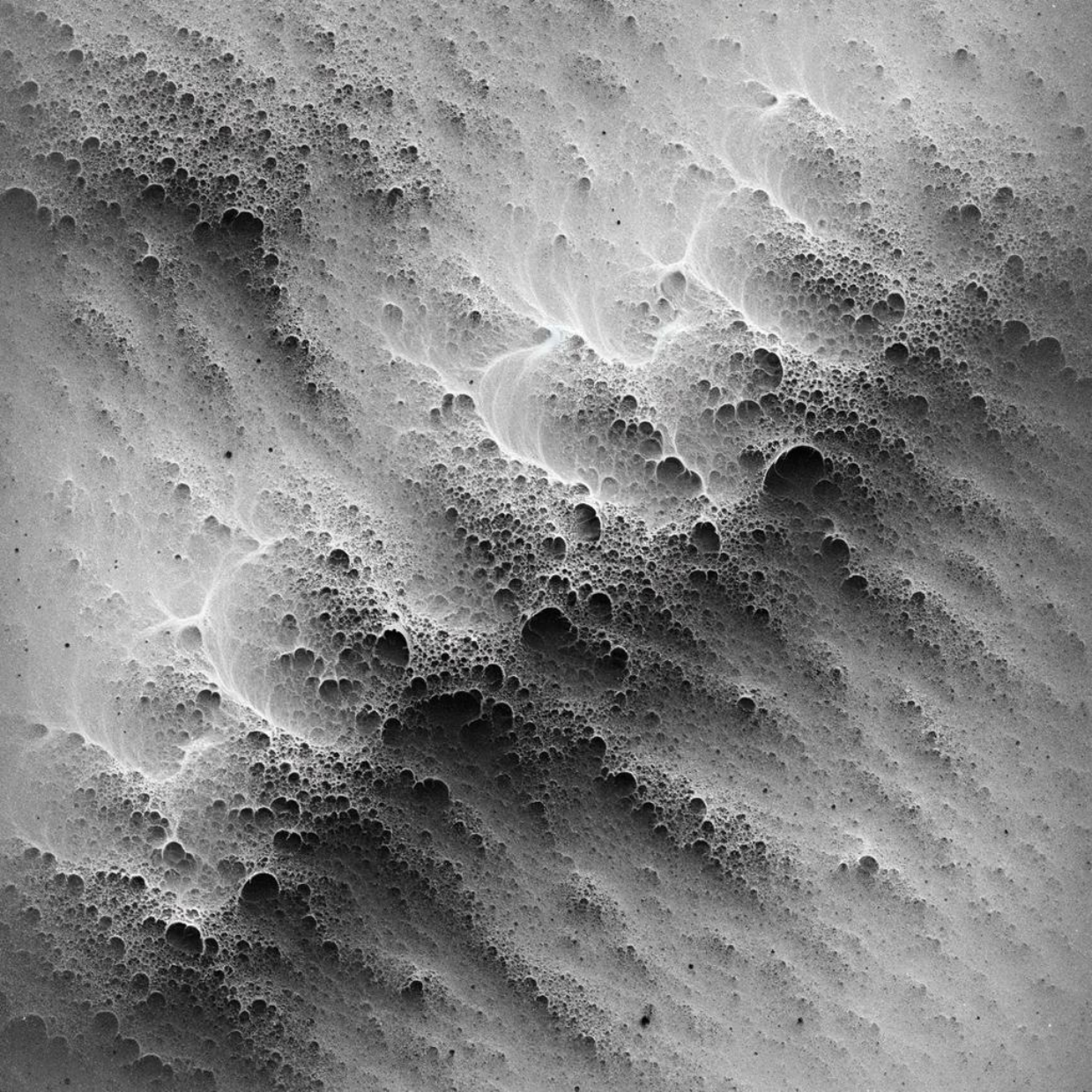} \\ [-0.65em]
\caption{The table shows the sample images generated by DALLE-3 with hints provided from a series of question-answer pairs based on the provided patches category image as shown in the above Table \ref{tab:tab29}.}
\label{tab:tab30}
\end{tabularx}
\end{tcolorbox}

\clearpage

\end{document}